\PassOptionsToPackage{table}{xcolor}
\documentclass[]{fairmeta}

\usepackage{wrapfig}
\usepackage{tabularx}
\usepackage{textcomp}
\usepackage{stfloats}
\usepackage{url}
\usepackage{verbatim}
\usepackage{titlesec}
\usepackage{adjustbox}
\usepackage{tikz}
\usetikzlibrary{backgrounds}
\usepackage{comment}
\usepackage{amsmath,amsfonts,amssymb}
\usepackage{anyfontsize}
\usepackage{colortbl}
\usepackage{color}
\usepackage{hyperref}
\usepackage{graphicx}
\usepackage{subcaption} 
\usepackage{multirow}
\usepackage{booktabs}
\usepackage{xcolor}
\usepackage{pifont}
\usepackage{bbding}
\usepackage{fontawesome}
\makeatletter
\let\caffeFaLightbulbO\faLightbulbO
\DeclareRobustCommand{\faLightbulbO}{\textmd{\caffeFaLightbulbO}}
\makeatother
\usepackage{xspace}
\usepackage{float}
\usepackage{algorithm}
\usepackage{listings}
\usepackage{algcompatible}
\usepackage{algpseudocode}
\usepackage{soul}
\usepackage{array}
\usepackage{tcolorbox}
\usepackage{needspace}
\usepackage{diagbox}
\usepackage{makecell}
\usepackage{rotating}
\usepackage{enumitem}
\usepackage{CJKutf8}
\ifPDFTeX
\newcommand{\figuretwotimes}{\fontfamily{ptm}\selectfont}
\else
\newfontfamily{\figuretwotimes}{Times New Roman}
\fi
\DeclareCaptionFormat{figtwotimes}{{\figuretwotimes\bfseries #1 #2} {\figuretwotimes #3}}

\usepackage{amsmath,amsfonts,bm}

\def\eqref#1{equation~\ref{#1}}

\def\1{\bm{1}}

\DeclareMathAlphabet{\mathsfit}{\encodingdefault}{\sfdefault}{m}{sl}
\SetMathAlphabet{\mathsfit}{bold}{\encodingdefault}{\sfdefault}{bx}{n}

\usepackage{lipsum}
\definecolor{lightgrey}{RGB}{244,244,244}

\RequirePackage{xspace}
\makeatletter
\DeclareRobustCommand\onedot{\futurelet\@let@token\@onedot}
\def\@onedot{\ifx\@let@token.\else.\null\fi\xspace}

\def\ie{\emph{i.e}\onedot}

\makeatother

\definecolor{lightgrey}{RGB}{201,203,209}

\definecolor{token_blue}{RGB}{84, 120, 140}

\newlength\savewidth

\definecolor{tableHeaderBack}{RGB}{225,236,247}
\definecolor{tableStripe}{RGB}{244,248,253}
\definecolor{tableHeadText}{RGB}{28,52,98}
\definecolor{tableGroupBack}{RGB}{210,225,242}
\definecolor{tableGroupText}{RGB}{28,52,98}
\definecolor{benchmarkHead}{RGB}{28,52,98}
\definecolor{benchmarkTagBack}{RGB}{220,234,250}
\definecolor{benchmarkRealBack}{RGB}{220,234,250}
\definecolor{benchmarkRealText}{RGB}{28,52,98}
\definecolor{benchmarkSimBack}{RGB}{220,234,250}
\definecolor{benchmarkSimText}{RGB}{28,52,98}
\colorlet{benchmarkHeaderBack}{tableHeaderBack}
\colorlet{performanceHeaderBack}{tableHeaderBack}
\colorlet{performanceGroupBack}{tableGroupBack}
\colorlet{performanceGroupText}{tableGroupText}
\definecolor{performanceHead}{RGB}{28,52,98}
\newcommand{\surveytable}{%
  \rowcolors{2}{white}{tableStripe}%
}

\let\benchmarktablestyle\surveytablestyle
\let\performancetablestyle\surveytablestyle
\newcommand{\tblhead}[1]{\textcolor{tableHeadText}{{\rmfamily\bfseries #1}}}
\newcommand{\benchhead}{\rowcolor{tableHeaderBack}}
\newcommand{\perfhead}{\rowcolor{tableHeaderBack}}
\newdimen\savedaboverulesep
\newdimen\savedbelowrulesep
\newcommand{\perfheadersep}[1]{%
  \noalign{%
    \global\savedaboverulesep=\aboverulesep
    \global\savedbelowrulesep=\belowrulesep
    \global\aboverulesep=0pt
    \global\belowrulesep=0pt
  }%
  \arrayrulecolor{tableHeadText}\cmidrule{#1}%
  \noalign{%
    \global\aboverulesep=\savedaboverulesep
    \global\belowrulesep=\savedbelowrulesep
  }%
  \arrayrulecolor{black}%
}

\newcommand{\tabletag}[3]{\begingroup\setlength{\fboxsep}{1.3pt}\colorbox{#2}{\textcolor{#3}{\scriptsize\rmfamily\bfseries #1}}\endgroup}
\newcommand{\settingtag}[3]{\begingroup\setlength{\fboxsep}{1.3pt}\colorbox{#2}{\textcolor{#3}{\scriptsize\rmfamily\bfseries\strut\makebox[2.75em][c]{#1}}}\endgroup}
\newcommand{\realtag}{\settingtag{Real}{benchmarkRealBack}{benchmarkRealText}}
\newcommand{\simtag}{\settingtag{Sim}{benchmarkSimBack}{benchmarkSimText}}
\newcommand{\settingtagcell}[1]{\makebox[6.35em][c]{#1}}
\newcommand{\realcell}{\settingtagcell{\realtag}}
\newcommand{\simcell}{\settingtagcell{\simtag}}
\newcommand{\realsimcell}{\settingtagcell{\realtag\hspace{0.45em}\simtag}}
\newcommand{\audtag}[1]{\tabletag{#1}{benchmarkTagBack}{benchmarkHead}}
\newcommand{\perfgroup}[2]{\multicolumn{#1}{l}{\textcolor{tableGroupText}{\rmfamily\bfseries\itshape #2}}\\}
\newsavebox{\surveytablebox}
\newdimen\surveytableextra
\newdimen\surveytabletargetsep
\newcount\surveytablepads
\NewDocumentEnvironment{fullwidthtabular}{m m +b}{%
  \begingroup
  \surveytablepads=#1\relax
  \ifnum\surveytablepads<1\relax
    \surveytablepads=1\relax
  \fi
  \sbox{\surveytablebox}{\begin{tabular}{#2}#3\end{tabular}}%
  \ifdim\wd\surveytablebox<\linewidth
    \surveytableextra=\dimexpr\linewidth-\wd\surveytablebox\relax
    \divide\surveytableextra by \surveytablepads
    \tabcolsep=\dimexpr\tabcolsep+\surveytableextra\relax
  \else
    \surveytableextra=\dimexpr\wd\surveytablebox-\linewidth\relax
    \divide\surveytableextra by \surveytablepads
    \surveytabletargetsep=\dimexpr\tabcolsep-\surveytableextra\relax
    \ifdim\surveytabletargetsep<0pt
      \tabcolsep=0pt
    \else
      \tabcolsep=\surveytabletargetsep
    \fi
  \fi
  \begin{tabular}{#2}#3\end{tabular}%
  \endgroup
}{}

\makeatletter
\AtBeginDocument{%
  \ifdefined\rownum
    \pretocmd{\@array}{\global\rownum\z@}{}{}%
  \fi
}%
\makeatother

\newcolumntype{x}[1]{>{\centering\arraybackslash}p{#1pt}}
\newcolumntype{y}[1]{>{\raggedright\arraybackslash}p{#1pt}}
\newcolumntype{z}[1]{>{\raggedleft\arraybackslash}p{#1pt}}

\renewcommand{\paragraph}[1]{\vspace{1mm}\noindent\textbf{#1}}

\renewcommand{\paragraph}[1]{\vspace{1.25mm}\noindent\textbf{#1}}

\definecolor{codeblue}{rgb}{0.25, 0.5, 0.5}
\definecolor{codekw}{rgb}{0.35, 0.35, 0.75}
\lstdefinestyle{Pytorch}{
    language = Python,
    backgroundcolor = \color{white},
    basicstyle = \fontsize{9pt}{8pt}\selectfont\ttfamily\bfseries,
    columns = fullflexible,
    aboveskip=1pt,
    belowskip=1pt,
    breaklines = true,
    captionpos = b,
    commentstyle = \color{codeblue},
    keywordstyle = \color{codekw},
}

\definecolor{green}{HTML}{009000}
\definecolor{red}{HTML}{ea4335}

\newcommand{\eqcontrib}{\clubsuit}
\newcommand{\correspond}{\spadesuit}

\newtcolorbox{promptblock}{
    colback=gray!5,
    colframe=gray!15,
    boxrule=0.5pt,
    arc=3pt,
    left=12pt,
    right=12pt,
    top=8pt,
    bottom=8pt,
    boxsep=8pt,
    breakable
}

\definecolor{insightTrendBack}{RGB}{240,249,243}
\definecolor{insightTrendFrame}{RGB}{48,119,81}
\definecolor{insightDiffBack}{RGB}{237,248,253}
\definecolor{insightDiffFrame}{RGB}{42,105,152}
\definecolor{insightProblemBack}{RGB}{254,248,235}
\definecolor{insightProblemFrame}{RGB}{169,104,27}
\colorlet{insight_blue}{insightDiffFrame}
\colorlet{insight_olive}{insightTrendFrame}
\tcbset{
    insight common/.style={
        enhanced,
        boxrule=1.5pt,
        arc=3pt,
        left=10pt,
        right=10pt,
        top=12pt,
        bottom=6pt,
        boxsep=6pt,
        fonttitle=\bfseries,
        coltitle=white,
        attach boxed title to top left={xshift=10pt,yshift=-2mm},
        varwidth boxed title=0.92\linewidth,
        before={\par\Needspace{0.26\textheight}},
        after={\par\medskip}
    }
}
\newtcolorbox{insightbox}[1][]{
    insight common,
    colback=insightTrendBack,
    colframe=insightTrendFrame,
    colbacktitle=insightTrendFrame,
    title={\small #1},
    boxed title style={
        colback=insightTrendFrame,
        colframe=insightTrendFrame,
        boxrule=0pt,
        arc=2pt,
        outer arc=2pt,
        left=6pt,
        right=6pt,
        top=2pt,
        bottom=2pt,
    }
}
\newtcolorbox{differencebox}[1][]{
    insight common,
    colback=insightDiffBack,
    colframe=insightDiffFrame,
    colbacktitle=insightDiffFrame,
    title={\small #1},
    boxed title style={
        colback=insightDiffFrame,
        colframe=insightDiffFrame,
        boxrule=0pt,
        arc=2pt,
        outer arc=2pt,
        left=6pt,
        right=6pt,
        top=2pt,
        bottom=2pt,
    }
}
\newtcolorbox{frontierbox}[1][]{
    insight common,
    colback=insightProblemBack,
    colframe=insightProblemFrame,
    colbacktitle=insightProblemFrame,
    title={\small #1},
    boxed title style={
        colback=insightProblemFrame,
        colframe=insightProblemFrame,
        boxrule=0pt,
        arc=2pt,
        outer arc=2pt,
        left=6pt,
        right=6pt,
        top=2pt,
        bottom=2pt,
    }
}

\definecolor{my_green}{RGB}{51,102,0}
\definecolor{my_red}{RGB}{204, 0, 0}
\renewcommand{\checkmark}{\textcolor{my_green}{\ding{51}}} 
\newcommand{\crossmark}{\textcolor{my_red}{\ding{55}}} 

\newcommand{\paratitle}[1]{\vspace{1.2ex}\noindent $\bullet$ \textbf{#1}}

\newcommand{\javisheaderlogo}{\includegraphics[height=0.8cm,keepaspectratio]{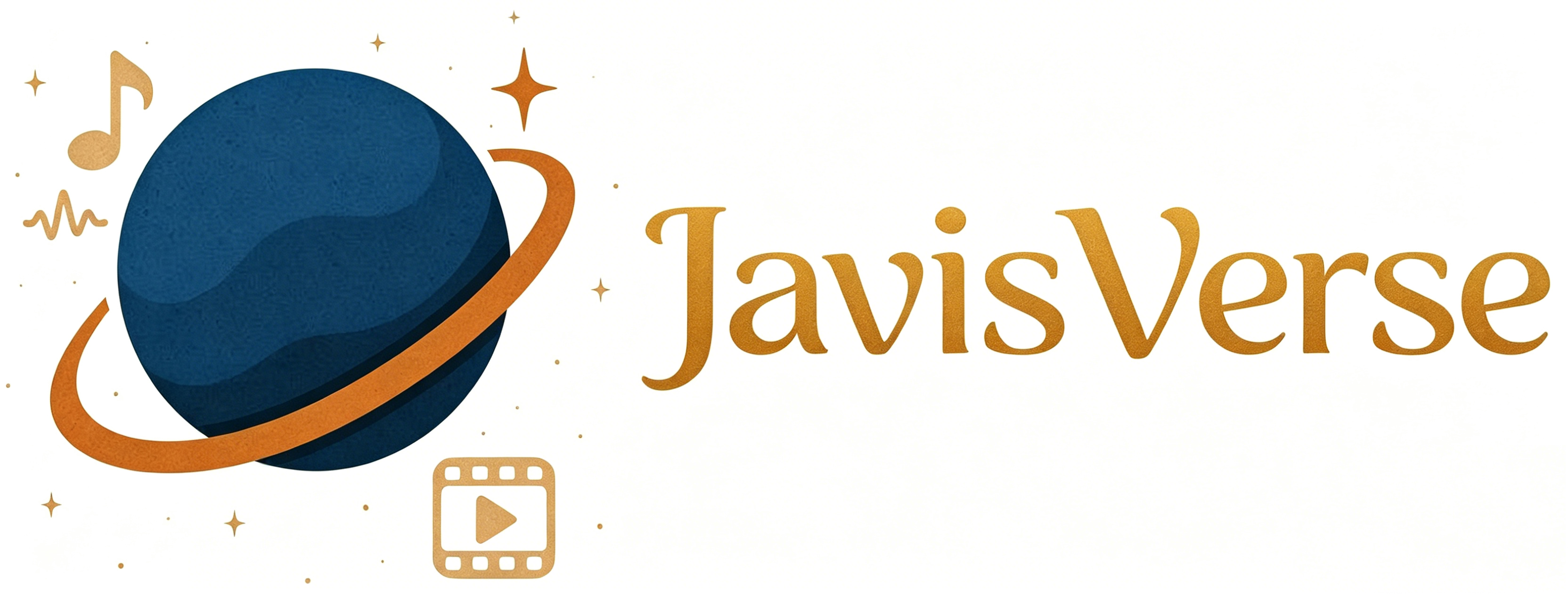}}
\fancypagestyle{javisheader}{
    \fancyhf{}
    \fancyhead[L]{\javisheaderlogo}
    \fancyfoot[C]{\thepage}
    
}
\fancypagestyle{firstheader}{
    \fancyhf{}
    \fancyhead[C]{
        \includegraphics[height=0.8cm]{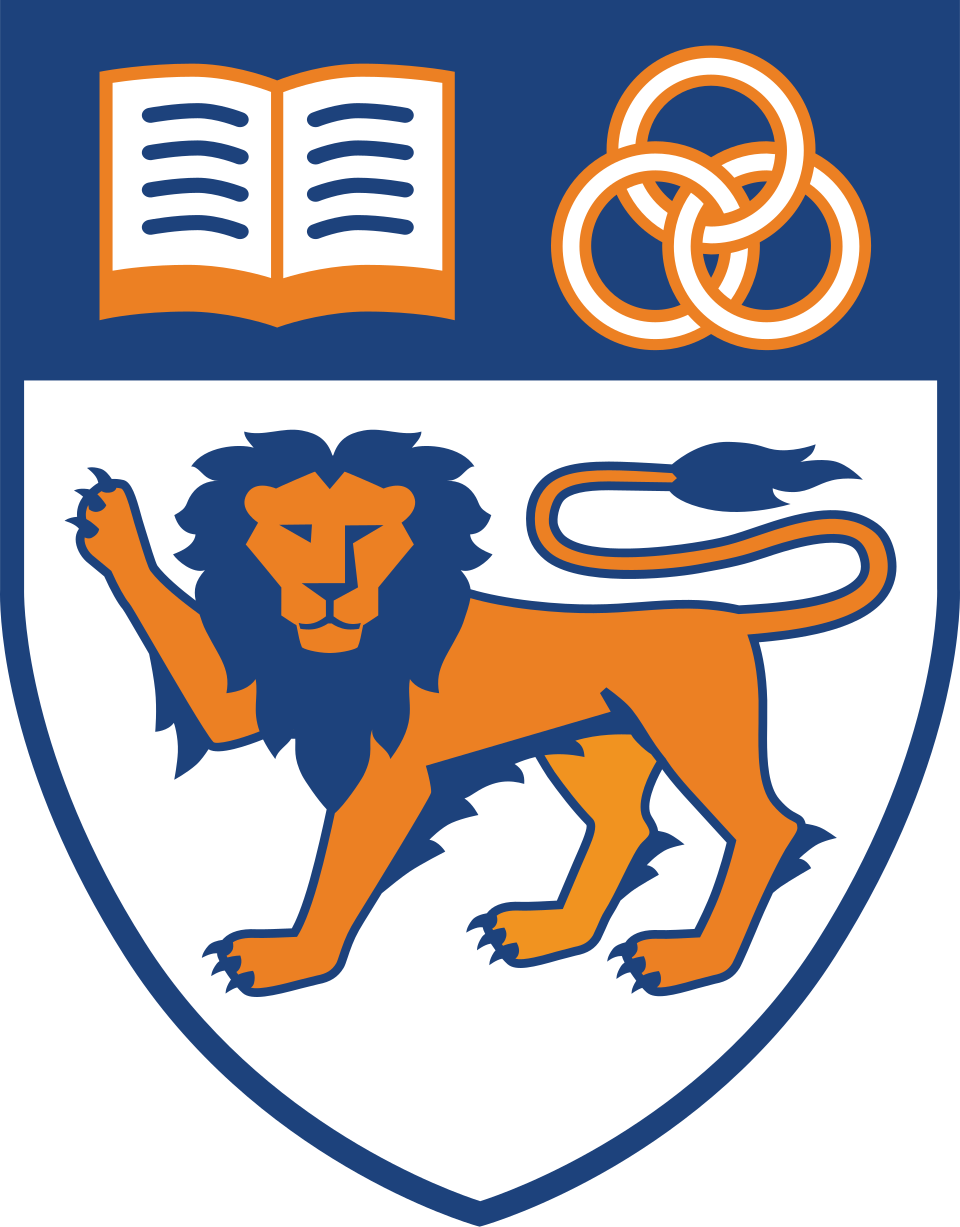}\hspace{3em}
        \includegraphics[height=0.8cm]{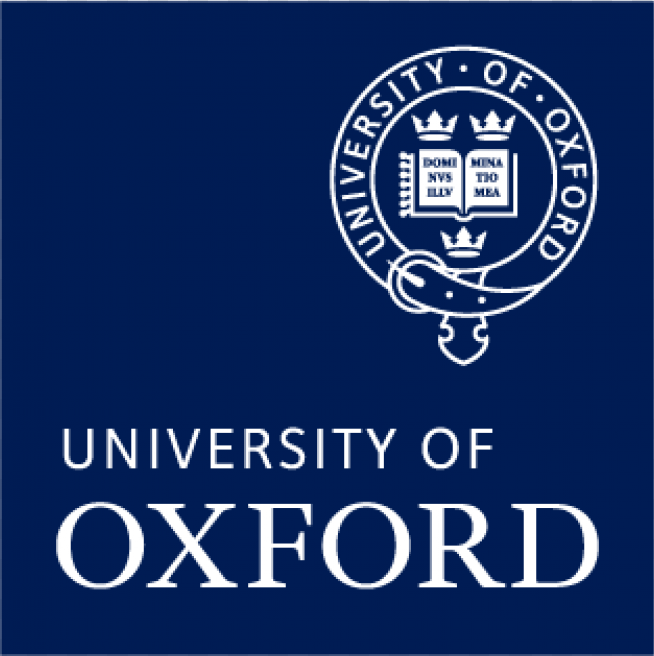}\hspace{3em}
        \includegraphics[height=0.8cm]{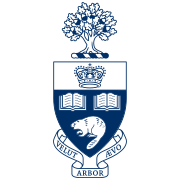}\hspace{3em}
        \includegraphics[height=0.8cm]{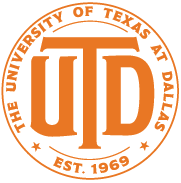}\hspace{3em}
        \includegraphics[height=0.8cm]{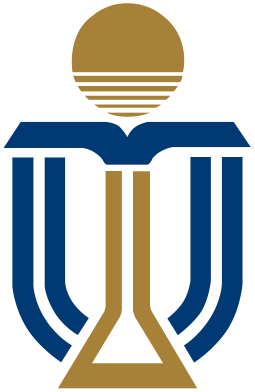}\hspace{3em}
        \includegraphics[height=0.8cm]{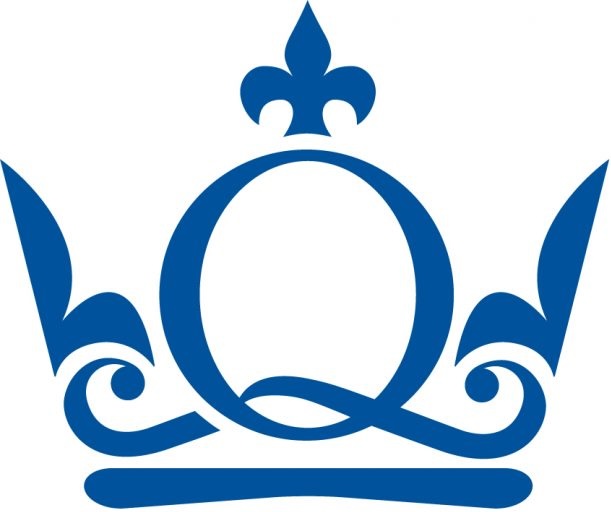}\hspace{3em}
        \includegraphics[height=0.8cm]{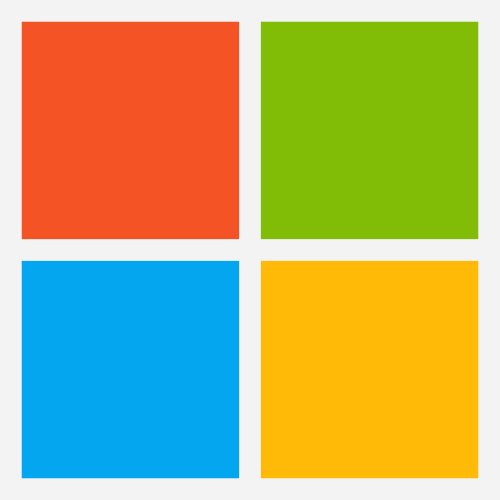}\hspace{3em}
        \includegraphics[height=0.8cm]{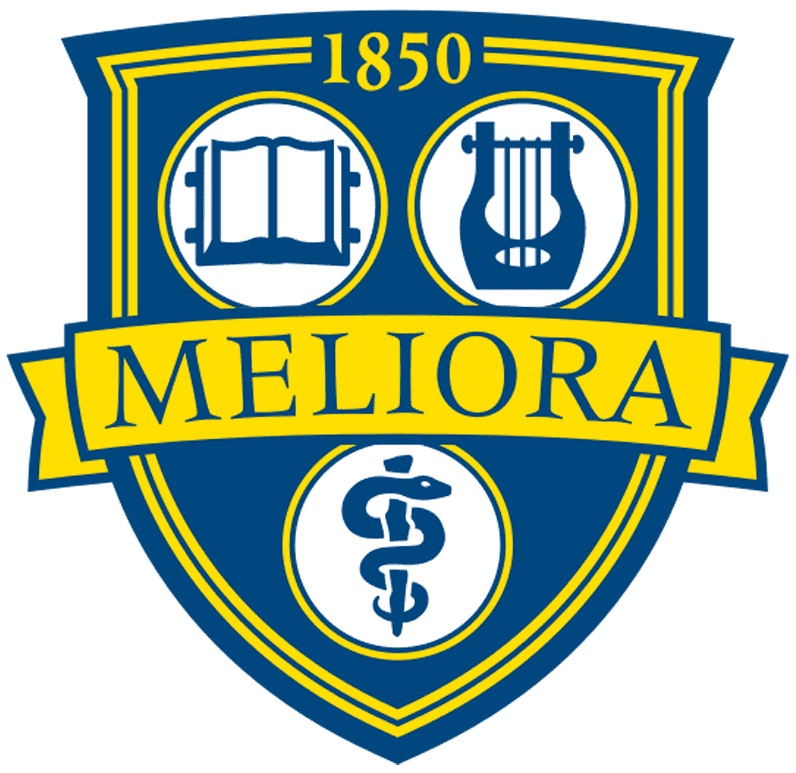}
    }
    
}

\title{Audio-Visual Intelligence in Large Foundation Models: A Comprehensive Survey}

\author[1,\eqcontrib]{You Qin}
\author[1,\eqcontrib]{Kai Liu}
\author[2]{Shengqiong Wu}
\author[3]{Kai Wang}
\author[4]{Shijian Deng}
\author[4]{Yapeng Tian}
\author[1]{Junbin Xiao}
\author[5]{Yazhou Xing}
\author[6]{Yinghao Ma}
\author[1]{Bobo Li}
\author[1]{Roger Zimmermann}
\author[7]{Lei Cui}
\author[7]{Furu Wei}
\author[8]{Jiebo Luo}
\author[2,\correspond]{Hao Fei}

\affiliation{$^1$National University of Singapore}
\affiliation{$^2$University of Oxford}
\affiliation{$^3$University of Toronto}  
\affiliation{$^4$The University of Texas at Dallas}
\affiliation{$^5$The Hong Kong University of Science and Technology}
\affiliation{$^6$Queen Mary University of London}
\affiliation{$^7$Microsoft Research}
\affiliation{$^8$University of Rochester}

\contribution[\eqcontrib]{Equal Contribution}
\contribution[\correspond]{Correspondence (\email{haofei7419@gmail.com})}

\abstract{
Audio-Visual Intelligence (AVI) has emerged as a central frontier in artificial intelligence, bridging auditory and visual modalities to enable machines that can perceive, generate, and interact in the multimodal real world.
In the era of large foundation models, joint modeling of audio and vision has become increasingly crucial, i.e., not only for understanding but also for controllable generation and reasoning across dynamic, temporally grounded signals.
Recent advances, such as Meta MovieGen and Google Veo-3, highlight the growing industrial and academic focus on unified audio–vision architectures that learn from massive multimodal data.
However, despite rapid progress, the literature remains fragmented, spanning diverse tasks, inconsistent taxonomies, and heterogeneous evaluation practices that impede systematic comparison and knowledge integration.
This survey provides the first comprehensive review of AVI through the lens of large foundation models.
We establish a unified taxonomy covering the broad landscape of AVI tasks, ranging from understanding (e.g., speech recognition, sound localization) to generation (e.g., audio-driven video synthesis, video-to-audio) and interaction (e.g., dialogue, embodied, or agentic interfaces).
We synthesize methodological foundations, including modality tokenization, cross-modal fusion, autoregressive and diffusion-based generation, large-scale pretraining, instruction alignment, and preference optimization.
Furthermore, we curate representative datasets, benchmarks, and evaluation metrics, offering a structured comparison across task families and identifying open challenges in synchronization, spatial reasoning, controllability, and safety.
By consolidating this rapidly expanding field into a coherent framework, this survey aims to serve as a foundational reference for future research on large-scale AVI.
}

\date{\today}
\metadata[Homepage]{\url{https://github.com/JavisVerse/Awesome-AVI}}

\begin{document}
\thispagestyle{firstheader}
\maketitle
\pagestyle{javisheader}

\begin{figure}[H] 
\centering
\vspace{-2mm}
\includegraphics[width=0.89\linewidth]{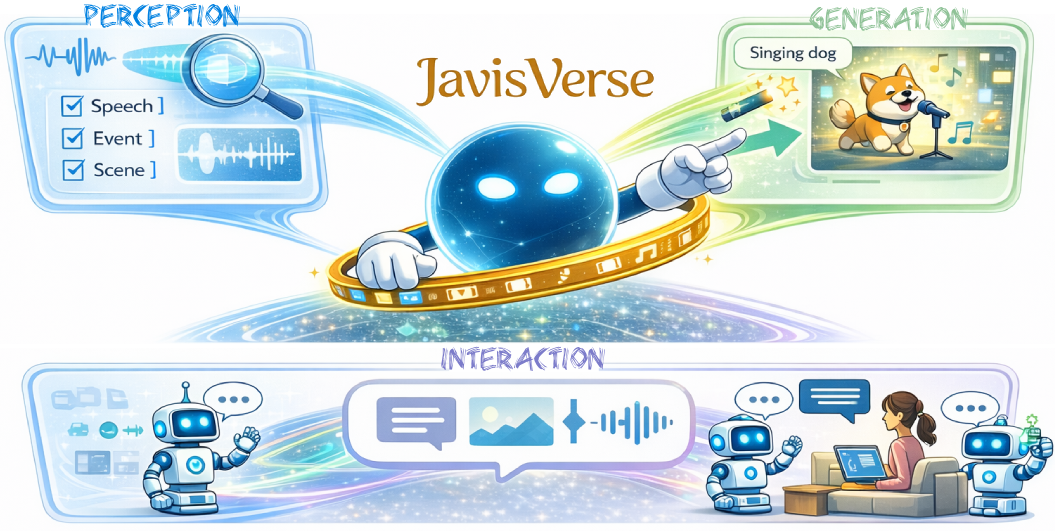}
\vspace{-4mm}
\end{figure}

\newpage

{
  \hypersetup{linktoc=page}
  \tableofcontents
}

\clearpage

\begin{figure}[H]
    \centering
    \vspace{-6mm}
    \includegraphics[width=\textwidth, height=0.97\textheight, keepaspectratio]{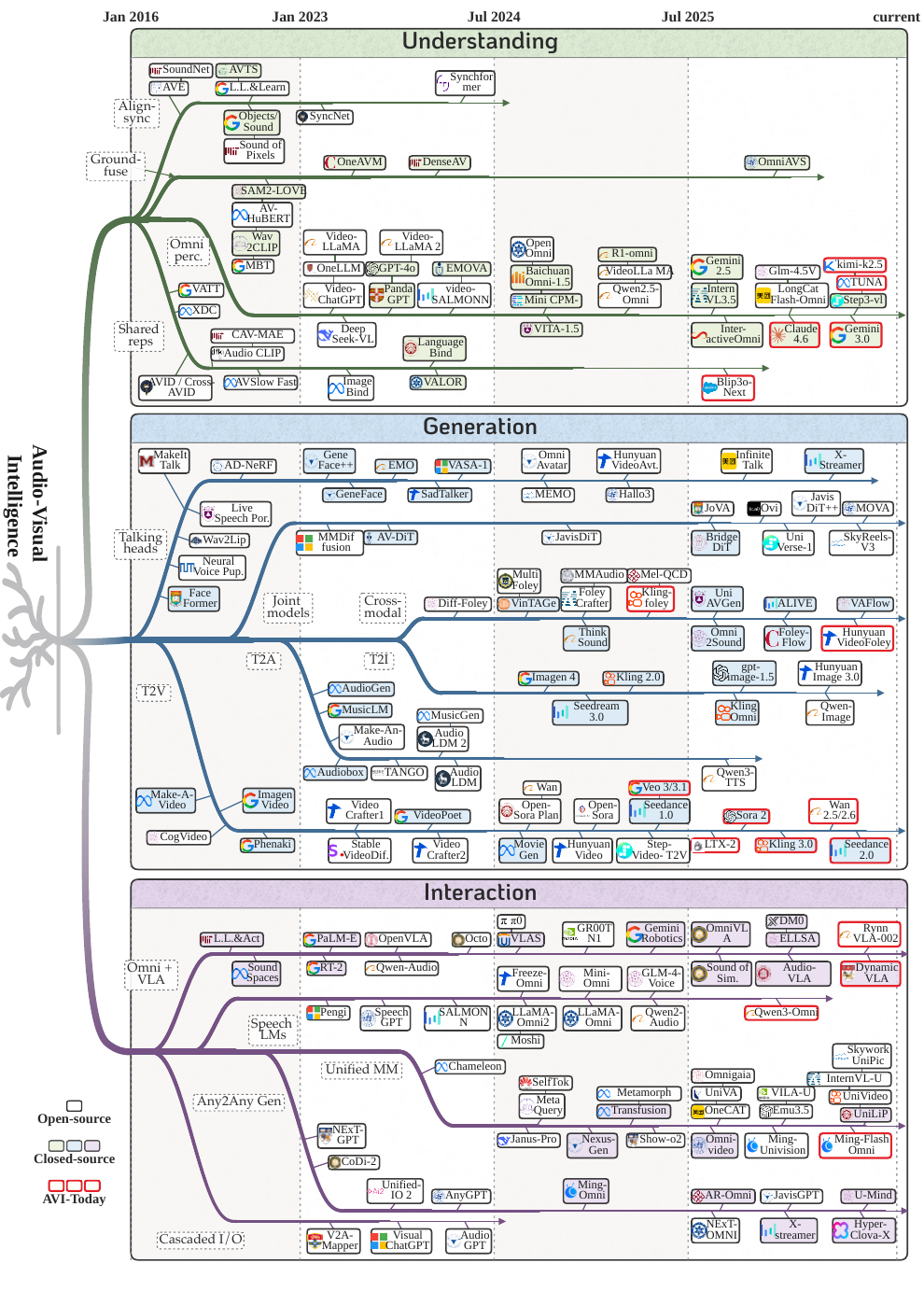}
    \caption{The evolutionary tree of Audio-Visual Intelligence from 2016 to 2026.
  The four columns track representative methods across understanding the world (audio-visual perception), creating the world (audio-visual generation), interacting with the world (unified perception and generation), and the converged AVI systems of today.}
    \label{fig:intro_timeline}
\end{figure}

\section{Introduction}
\label{sec:intro}

Large foundation models have transformed artificial intelligence by scaling data, compute, and model capacity to unlock broad generalization and emergent capabilities \citep{hurst2024gpt,guo2025deepseek,yang2025qwen3}.
Human perception is inherently multimodal, with audio and vision forming the most pervasive and complementary pair for understanding, prediction, and control in real-world environments \citep{xu2025qwen3omni, kong2024hunyuanvideo, kimopenvla,000100L0LWWZMSZ25}.
Across society, this audio-visual pairing underpins assistive technologies, education, robotics, entertainment, and creative tooling, where perception and generation increasingly co-exist rather than appear in isolation.
In the large-model era, progress therefore depends on unifying audio-visual intelligence (AVI) to support robust understanding, controllable generation, and interactive reasoning under temporal and spatial constraints.
Industry systems such as Meta MovieGen \citep{polyak2025moviegen} and Google Veo-3 \citep{google2025veo3} exemplify this strategic shift toward end-to-end audio-vision modeling and coordinated synthesis, signaling both technical maturity and growing application demand.
Against this backdrop, we position our work to consolidate concepts, methods, and trends for AVI at foundation-model scale, laying a coherent basis for research and deployment.

As outlined in \cref{fig:intro_timeline},
AVI spans a wide and rapidly evolving task spectrum that reflects how sound and sight co-occur in the wild.
On the perception side, representative problems include audio-visual speech recognition \citep{malik2021automatic}, lip reading \citep{son2017lip}, active speaker detection \citep{roth2020ava}, sound source localization and separation \citep{tzinis2020two}, event understanding \citep{monfort2019moments}, cross-modal retrieval \citep{mark2024denseav}, and AV question answering \citep{yang2022avqa}, all of which rely on synchronization and grounding across modalities.
On the generative side, research explores audio-driven talking heads \citep{wang2021audio2head}, video-conditioned speech or Foley \citep{cheng2025mmaudio}, music generation aligned to visual rhythm \citep{hong2025musicinfuser}, dubbing and alignment \citep{comunita2024diff}, cross-modal editing and stylization \citep{fu2025objectavedit}, and controllable multimodal storytelling with long-horizon coherence \citep{zhang2025dialogue}.
Interactive settings further introduce streaming inference \citep{xie2025x}, tool use \citep{yao2023react}, and embodiment \citep{kimopenvla}, where systems must reason over temporally entangled signals while respecting latency and user intent.
Collectively, these task families motivate unified formalisms for inputs, objectives, and evaluations that can compare methods fairly across heterogeneous settings.

Methodologically, AVI builds on modality-specific encoders or tokenizers for audio and vision, cross-modal fusion and alignment, and powerful generative decoders \citep{xu2025qwen3omni, chen2025blip3o}.
Autoregressive transformers \citep{brown2020language} drive sequence modeling for speech, music, and video tokens, enabling conditional decoding and instruction following at scale.
Diffusion models \citep{ho2020denoising,liu2023flow} provide high-fidelity synthesis and flexible editing, increasingly adapted to multimodal control via cross-attention and guidance mechanisms.
Self-supervised objectives such as contrastive alignment and masked/denoising modeling remain central to representation learning, while instruction tuning and preference optimization tailor behaviors for interactive AV use \citep{girdhar2023imagebind, xu2025qwen3omni}.
Scaling laws, data mixtures, and curation strategies, i.e., spanning speech corpora, music datasets, video-audio pairs, and synthetic pipelines, jointly determine capability and robustness, raising new questions about coverage, bias, and licensing \citep{yin2024survey}.

Despite impressive progress, the literature remains fragmented across subcommunities with overlapping definitions, inconsistent terminology, and divergent taxonomies that hinder cumulative understanding.
Evaluation practices vary widely in datasets, metrics, and protocols, especially for open-ended generation, alignment quality, temporal coherence, and human-centric judgments, which complicates reproducibility and benchmarking \citep{yang2022avqa,cao2025t2av}.
Safety and governance concerns, e.g., privacy in audio/video, consent for speech and music, watermarking and provenance, and the energy footprint of foundation-scale training, are increasingly consequential yet unevenly addressed \citep{chen2025seedance,google2025veo3}.
These gaps might underscore the need for a comprehensive, taxonomy-driven survey that unifies the area and establishes actionable, comparable standards for AVI research and practice.

\subsection{Contribution Summary}

\begin{itemize}
    \item \textbf{First Comprehensive Survey:}  
    This work provides the first systematic and in-depth survey of \textit{Audio-Visual Intelligence} within the paradigm of large foundation models, unifying perception, generation, and interaction research under a coherent framework.

    \item \textbf{Unified Taxonomy:}  
    We establish a principled taxonomy that organizes the diverse audio-visual tasks (covering speech, music, sound events, video, and open-world understanding, generation, and interaction) while clarifying task scope, assumptions, and relationships among subproblems.

    \item \textbf{Core Methodological Synthesis:}  
    We consolidate the methodological foundations of AVI, including audio/visual tokenization, cross-modal fusion, autoregressive and diffusion-based generation, instruction alignment, and large-scale pretraining strategies.

    \item \textbf{Benchmark and Evaluation Overview:}  
    We curate and summarize datasets, benchmarks, and evaluation metrics across task families, identify key gaps in assessment protocols, and propose practical guidelines to promote fair and reproducible comparisons.

    \item \textbf{Frontiers and Future Directions:}  
    We highlight emerging research challenges, such as temporal synchronization, spatial audio reasoning, multimodal controllability, safety, watermarking, and governance, and outline promising avenues for the next generation of audio-visual foundation models.

    \item \textbf{Resource Sharing:}  
    All summarized resources, references, and organizational structures will be publicly released to support transparency and accelerate progress within the research community.
\end{itemize}

\subsection{Scope and Organization}
\begin{figure*}[p]
    \centering
    \captionsetup{format=figtwotimes}
    \resizebox{\textwidth}{0.94\textheight}{\input{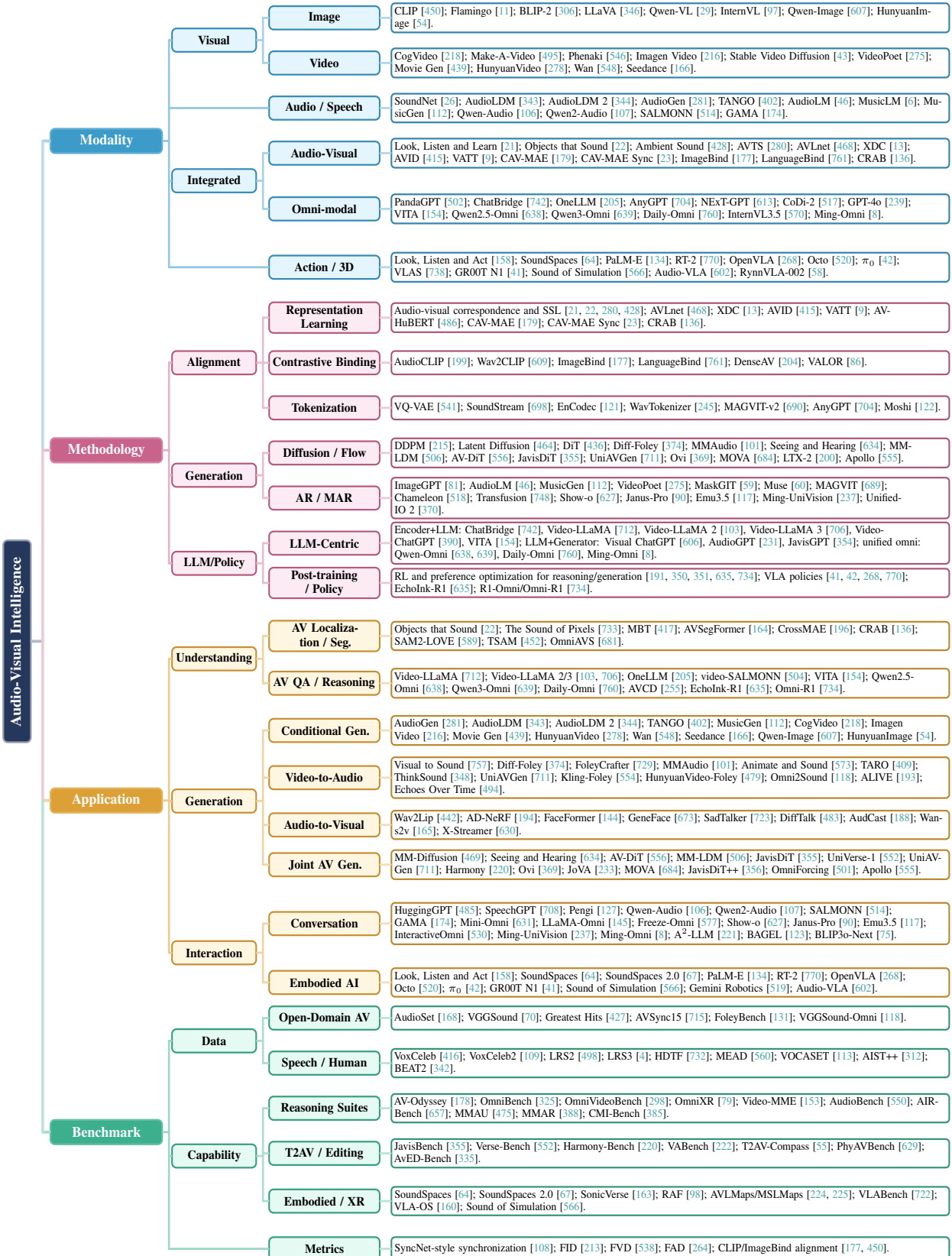}}
    \caption{Taxonomy of Audio-Visual Intelligence.}
    \label{fig:avi_taxonomy}
\end{figure*}

As depicted in \cref{fig:avi_taxonomy}, after laying a formal definition of the data modalities of audio and vision representation (\cref{sec:prel}), we first present a principled taxonomy of Audio-Visual Intelligence (\cref{sec:task}) that clarifies scope, assumptions, supervision signals, and relationships among task families, providing a global map before delving into specifics.
We then synthesize the foundation techniques (\cref{sec:tech}) for AVI, covering tokenization and representations, cross-modal fusion and alignment, autoregressive and diffusion generation, training objectives, data scaling and curation, and instruction alignment for interactive use.
Building on this base, we organize the literature along three pillars: perception  (\cref{sec:und}), generation (\cref{sec:gen}), and interaction (\cref{sec:inter}), where for each pillar, we review representative methods, summarize datasets and metrics, and provide comparative analyses with distilled takeaways.
Afterwards, we walk through the representative applications (\cref{sec:app}) under AVI, such as digital humans and immersive experience, during which we also highlight the relevant methods used in the pipeline.
Finally, we discuss open challenges and forward directions (\cref{sec:future}), including synchronization and temporal reasoning, spatial audio and 3D/4D grounding, controllability and editing, streaming efficiency and latency, evaluation for open-ended AV generation, and safety, watermarking, and data governance at scale.

\section{Preliminary}
\label{sec:prel}

\begin{figure}[!t]
    \centering
    \includegraphics[width=0.8\linewidth]{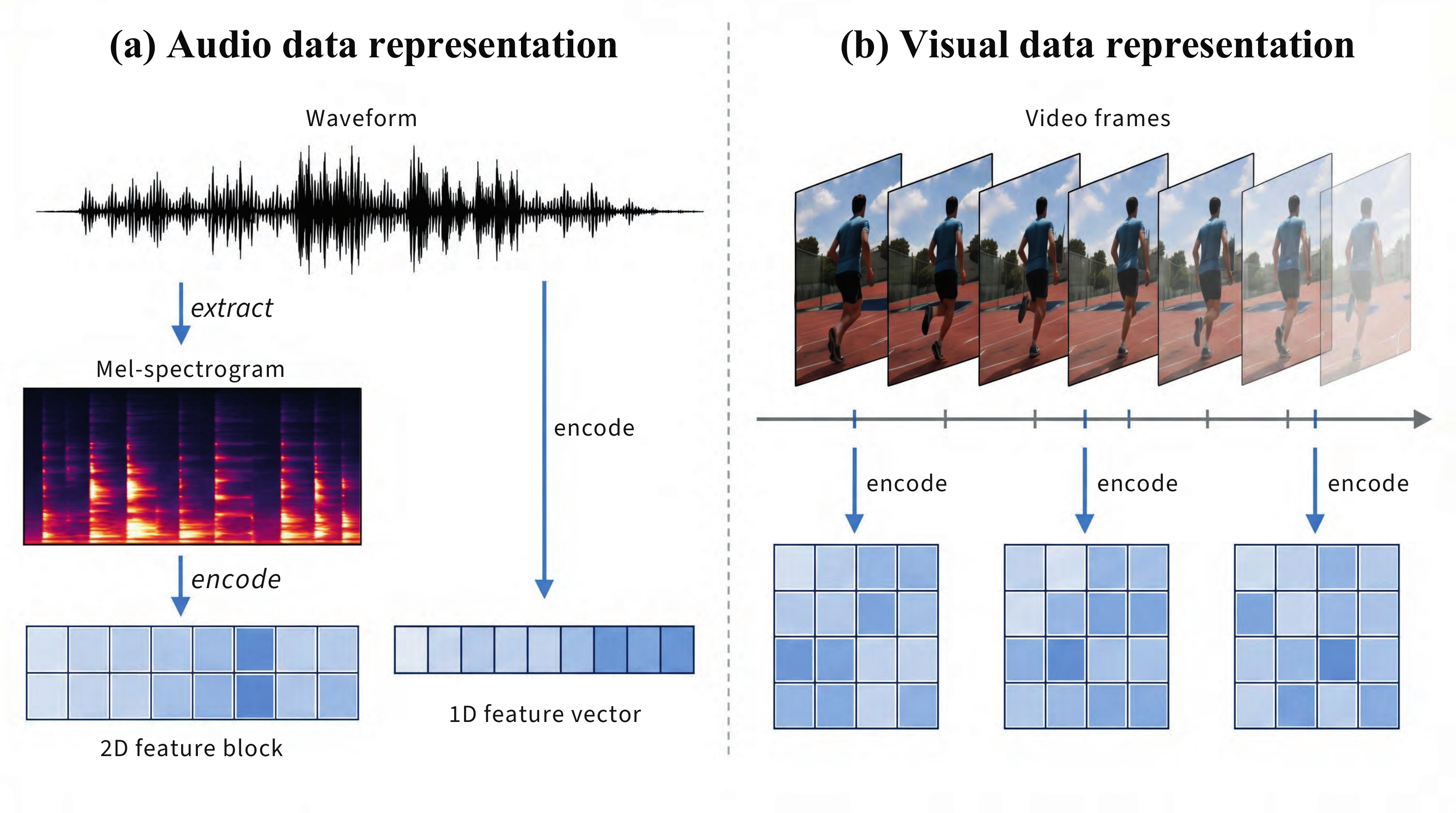}
    \caption{The overview of Audio-Visual data representation.}
    \label{fig:prel_data}
\end{figure}

The remainder of the survey presupposes a clear picture of what ``audio'' and ``vision'' mean in digital form: how physical signals are sampled, how they are turned into tensors or tokens for neural models, and what unimodal representations feed cross-modal and generative systems later on.
We therefore begin by defining each modality in isolation before any fusion, alignment, or task-specific processing is introduced in later sections.
The overview in \cref{fig:prel_data} summarizes the main data shapes and representation families; the subsections that follow make these notions precise for sound and for images and video, respectively, including both continuous embeddings and discrete tokenization.

\subsection{The Audio Modality: Data and Representation}
\label{sec:prel:audio}
The audio modality originates from physical vibrations propagating as sound waves, which can be categorized into speech, music (including instrumental and vocal), and general sound events, which encompass acoustic scenes such as environmental, urban, domestic sounds, or sound effects. 
For digital processing, these analog signals are captured by vibrations of sensors in microphones, and further sampled to produce 1D time-series digital signals known as the waveform. This is typically formatted as mono (1-channel) or stereo (2-channel), though multi-channel spatial audio formats also exist. 

As shown in \cref{fig:prel_data}, audio signals are primarily modeled in two forms: raw waveforms and time-frequency representations. A mono waveform is represented as a 1D temporal sequence $a \in \mathbb{R}^{L}$ sampled at a fixed rate, while multi-channel audio can be expressed as $A \in \mathbb{R}^{C \times L}$. Alternatively, the waveform can be transformed via Short-Time Fourier Transform (STFT) into a spectrogram, whose magnitudes are typically projected onto the Mel scale and logarithmically compressed to obtain a log-Mel spectrogram $S \in \mathbb{R}^{T \times F}$. These two representations dominate the inputs for audio encoders in multimodal foundation systems \citep{ma2024foundation}:

(1) \textit{Global Embeddings}: Encoding an entire audio segment into a single semantic vector can capture high-level acoustic and semantic features~\citep{hsu2021hubert,baevski2022data2vec,mccallum2022supervised}. Formally, an encoder $f_\theta^g$ maps the input signal into a $d$-dimensional embedding: $z = f_\theta^g(x) \in \mathbb{R}^d.$

(2) \textit{Continuous Dense Representations}: Beyond global pooling, many methods preserve temporal (and optionally frequency) structure by producing a latent feature sequence or map~\citep{huang2022masked,chen2023beats}. Let $\tilde{x}$ denote an audio input representation, which may be the waveform itself or a time-frequency transform (e.g., log Mel-spectrogram) $\tilde{x}\in\mathbb{R}^{F\times L}$. An encoder $f_\theta$ produces a (d)-dimensional, downsampled dense representation: $e = f_\theta(\tilde{x}) \in \mathbb{R}^{\frac{F}{s_f}\times \frac{L}{s_t}\times d}$,
where $s_f$ and $s_t$ are downsampling factors along frequency and time (for 1D waveform features, the frequency axis can be treated as $F=1$.

(3) \textit{Discrete Token Representations}: Vector quantization models discretize audio into token sequences, enabling language-model-style sequence modeling and generation~\citep{zeghidour2021soundstream,defossezhigh,jiwavtokenizer}. Using a learned codebook, a tokenizer $f_\theta^{VQ}$ maps $\tilde{x}$ to a sequence of $m$ token IDs: $q = f_\theta^{VQ}(\tilde{x}) \in \mathbb{N}^{m}$.

\subsection{The Visual Modality: Data and Representation}
\label{sec:prel:visual}

Visual data generally refers to images and videos captured as pixel intensities, where an image can be viewed as a projection of the 3D scene onto a 2D image plane (as in a pinhole camera model), and a video is composed of a sequence of image frames. 
Formally, a 3-channel RGB color image is represented as a matrix $I \in \mathbb{R}^{H\times W\times3}$, where $H, W$ refer to height and width, respectively. A video can be represented as an ordered series of $T$ image frames $V = \{I_1, I_2, \dots, I_T\}\in \mathbb{R}^{T\times H\times W\times 3}$, often played at a fixed frame rate (e.g. 30 frames per second) and stored and processed as 4D tensors in computer vision models.

Visual data encompasses a wide variety of forms, including natural images~\citep{ImageNet}, virtual images~\citep{ju2023human}, paintings~\citep{gontier2023delaunay}, etc. In audio-visual applications, human-centric content constitutes a key subset, including human body~\citep{ju2023human}, talking-head~\citep{chen2025talkvid}, and lip-synchronized videos~\citep{jiang2024audio}.
Beyond 2D imagery, many works explore 3D visual representations, such as depth maps~\citep{ming2021deep}, point clouds~\citep{guo2020deep}, surface meshes~\citep{kato2018neural}, NeRFs~\citep{gao2022nerf}, and 3D Gaussian Splatting~\citep{fei20243d}. These formats enable richer modeling of geometry and spatial interaction in multimodal tasks.

Raw image or video data is high-dimensional, and thus various representation methods are employed to extract compact, informative features:

(1) \textit{Global Embeddings}: A common approach encodes an entire image (or video frame) into a single global feature vector capturing semantic content~\citep{radford2021learning}. Models such as convolutional neural networks (CNNs)~\citep{he2016deep} or vision transformers (ViT)~\citep{dosovitskiy2020image} serve as extractor $f_\theta^g$, mapping each image to a $d$-dimensional vector: $z = f_\theta^g(I) \in \mathbb{R}^d$.

(2) \textit{Continuous Dense Representations}: Another strategy is to map images into continuous latent spaces while preserving spatial structures~\citep{he2016deep,dosovitskiy2020image}. Formally, one can denote an encoder function $f_\theta$ that maps an image $I \in \mathbb{R}^{W\times H\times 3}$ to a $d$-dimensional, $s$-downsampled feature map or sequence $e = f_\theta(I) \in \mathbb{R}^{\frac{W}{s}\times \frac{H}{s}\times d}$.

(3) \textit{Discrete Token Representations}: Vector quantization models like VQ-VAE~\citep{van2017neural} can further discretize image features using a learned codebook. Each image is represented by a sequence of $m$ token IDs corresponding to nearest codebook entries: $q = f_\theta^{VQ}(I) \in \mathbb{N}^{m}$.

\section{Task Taxonomy}
\label{sec:task}

\begin{figure*}[!t]
    \centering
    \includegraphics[width=\textwidth]{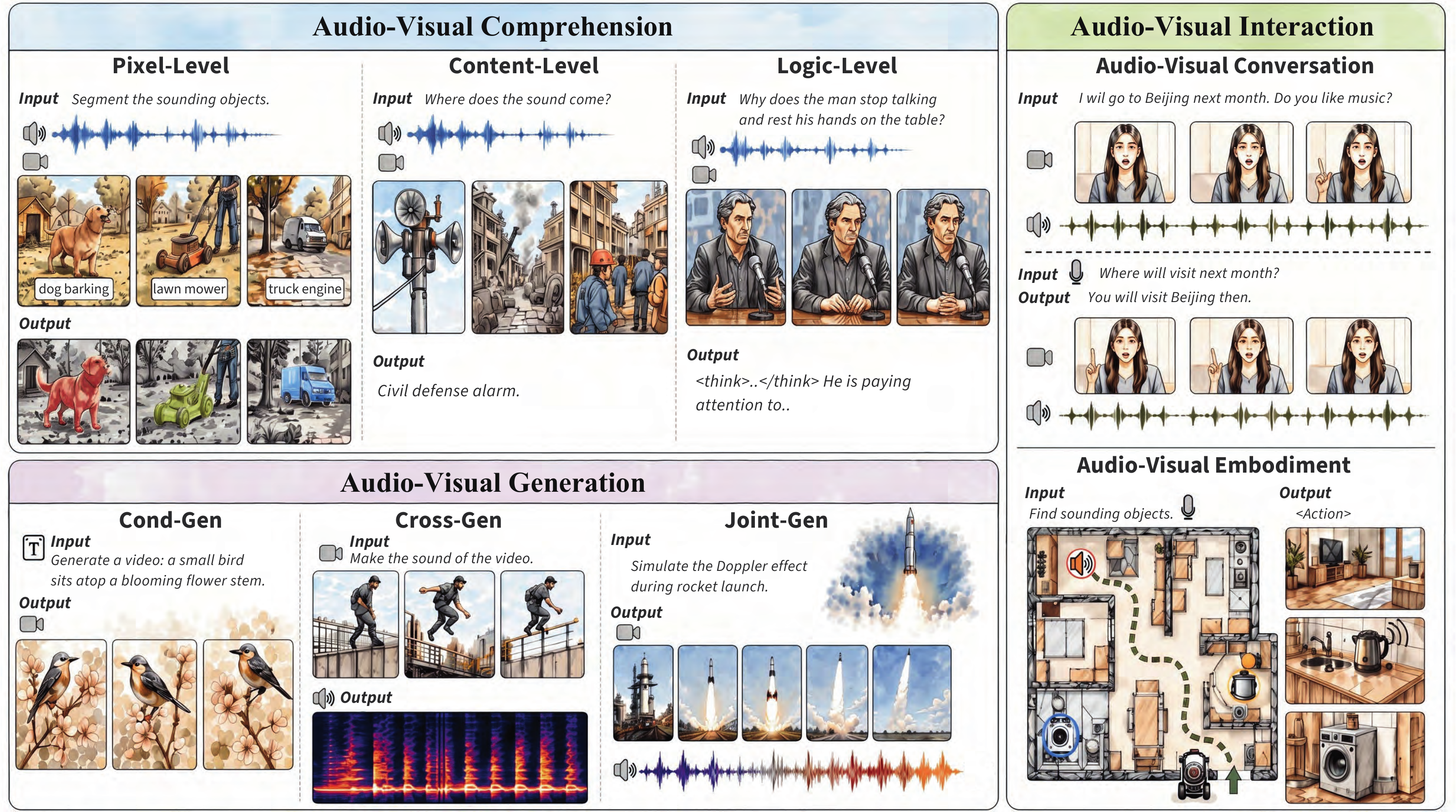}
    \caption{Overview of audio-visual intelligence tasks.}
    \label{fig:task_overview}
\end{figure*}

This section provides a systematic overview of the tasks involved in audio-visual intelligence across three key dimensions: \textbf{understanding}, \textbf{generation}, and \textbf{interaction}. As depicted in \cref{fig:task_overview}, these dimensions outline a roadmap toward the future development of AGI. The specific details of each task will be discussed in \cref{sec:und}, \cref{sec:gen}, and \cref{sec:inter}.

\subsection{Understanding the World: Audio-Visual Perception}
\label{sec:task:und}

Perception is the foundation of audio-visual intelligence, enabling models to sense and interpret the world from raw auditory and visual signals. This section presents a taxonomy of audio-visual perception tasks along three progressively higher levels of abstraction: \textit{pixel-level perception}, \textit{content understanding}, and \textit{logical reasoning}. Together, these task categories outline the perceptual and cognitive capabilities required for building general audio-visual intelligence.

\paratitle{Pixel-Level Perception.} Focusing on learning direct correspondences between raw audio signals and visual pixels, these tasks emphasize low-level sensing and alignment, forming the basis of multimodal perception. 
Representative examples include unimodal tasks such as automatic speech recognition (ASR) in audio~\citep{prabhavalkar2023end} and object detection~\citep{zou2023object} or tracking~\citep{luo2021multiple} in vision, as well as cross-modal tasks such as audio-visual event localization~\citep{tian2018audio} and segmentation~\citep{zhou2022audio}, which associate sounds with their spatial visual origins. 
Overall, pixel-level perception provides fine-grained spatial and temporal grounding of sound sources in visual scenes.

\paratitle{Content Understanding.} This aims to extract high-level semantic information from audio-visual inputs, including objects, events, and their temporal and causal relationships. 
A typical example is audio-visual question answering (AVQA), where models answer questions by jointly understanding audio and video content~\citep{li2022learning}. These tasks require aligning multimodal cues over time and interpreting their semantic interactions.
Compared to pixel-level perception, content understanding operates at a higher level of abstraction and reflects a model’s ability to comprehend what is happening in a multimodal scene.

\paratitle{Logical Reasoning.} It further extends audio-visual understanding by requiring models to perform inference beyond direct observation, incorporating physical commonsense, causality, or mathematical reasoning.
Models are required to infer latent causes or predict outcomes from audio-visual evidence, such as physical audio-visual commonsense reasoning~\citep{yu2022pacs}.
These tasks assess whether models can integrate perception with structured reasoning to explain or anticipate events in the multimodal world.

\subsection{Creating the World: Audio-Visual Generation}
\label{sec:task:gen}
Audio-visual generation aims to synthesize coherent and realistic multimodal content by modeling the joint distribution of visual and auditory signals. Unlike perception tasks that interpret inputs, generation tasks require creating temporally aligned, semantically consistent, and perceptually realistic outputs. This section categorizes generation into three paradigms: conditional, cross-modal, and joint audio-visual generation, which represent progressively deeper integration for building world-creating multimodal systems.

\paratitle{Conditional Audio-Visual Generation.} 
This task produces outputs guided by external control signals such as text, labels, or structured attributes, including (1) conditional audio generation, where models create sounds (e.g., speech, effects) based on text or audio context~\citep{liu2024audioldm,huang2024audiostyletransfer}; and (2) conditional visual generation, where models synthesize images or videos from text, visual cues, or scene representations~\citep{cao2025hunyuanimage,wan2025wan}.
These tasks emphasize controllability and semantic alignment, bridging intent and output through high-level conditioning.

\paratitle{Audio-Visual Cross-Modal Generation.} 
This kind of task synthesizes one modality conditioned on another, requiring fine-grained cross-modal learning capabilities, including (1) audio-to-visual generation, such as talking-head synthesis or audio-driven avatars~\citep{gao2025wan,gaussianheadtalk2025}; and (2) video-to-audio generation, such as generating soundtracks, speech, or spatial audio from visual inputs~\citep{wang2025kling,cheng2025mmaudio}.
These tasks demand strong temporal alignment and causal grounding between audio-visual modalities beyond textual control signals.

\paratitle{Audio-Visual Joint Generation.} 
Unlike the above tasks, it models the coupled evolution of sound and vision during generation, enabling synchronized and causally consistent audio-visual outputs. Key settings include (1) text/image-conditioned joint generation, where both modalities are generated from a shared prompt~\citep{liu2025javisdit,hacohen2026ltx2}; (2) audio-visual editing and extension, modifying existing multimodal content while maintaining coherence~\citep{fu2025objectavedit}; and (3) structured/spatial joint generation, such as text-to-3D audio-visual synthesis involving geometry and spatialized audio~\citep{yang2025streamingtalker}.
This paradigm seeks to capture holistic scene evolution and multimodal consistency.

\subsection{Interacting with the World: Audio-Visual Unified Perception and Generation}
\label{sec:task:inter}

Audio-visual unified perception and generation integrates multimodal understanding with generative decision-making in interactive settings. Unlike isolated perception or offline generation, these systems must interpret ongoing audio-visual inputs, reason over context or intent, and produce timely multimodal outputs or actions. We categorize such systems into two paradigms: \textit{interactive audio-visual conversation} and \textit{interactive audio-visual embodiment} in the digital and physical world, respectively.

\paratitle{Interactive Audio-Visual Conversation.} 
It involves dialogue systems that process multimodal inputs and generate corresponding responses, including conversational editing, instruction-following generation, and multimodal world modeling. Based on output modality, these systems can be divided into: (1) audio-centric, producing speech responses from multimodal context~\citep{huang2024audiogpt}; (2) visual-centric, generating or editing images and videos during conversation~\citep{deng2025emerging,huang2025ming}; and (3) omni-modal, supporting flexible input-output modalities in a unified interface~\citep{liu2025javisgpt,ai2025ming}.

\paratitle{Interactive Audio-Visual Embodiment.} 
It extends into physical environments, where systems convert multimodal understanding into actions. Applications include (1) audio-visual navigation, using sensory cues to guide movement~\citep{liu2024caven,yang2024rila}; (2) embodied QA and manipulation, where agents reason and act based on audio-visual inputs~\citep{zhao2025vlas,wei2025audioVLA}, etc.
These tasks demand tightly integrated perception, reasoning, and action, positioning unified audio-visual models as key enablers of embodied intelligence.

\section{Foundation Techniques}
\label{sec:tech}

This section introduces foundational techniques developed in the era of large foundation models, which provide key technical pathways toward achieving universal audio-visual intelligence, including \textbf{representation-centric}, \textbf{generation-centric}, and \textbf{LLM-centric} methods.

\subsection{Representation-centric Methods}
\label{sec:tech:rep}

\begin{figure}[!t]
    \centering
    \includegraphics[width=\linewidth]{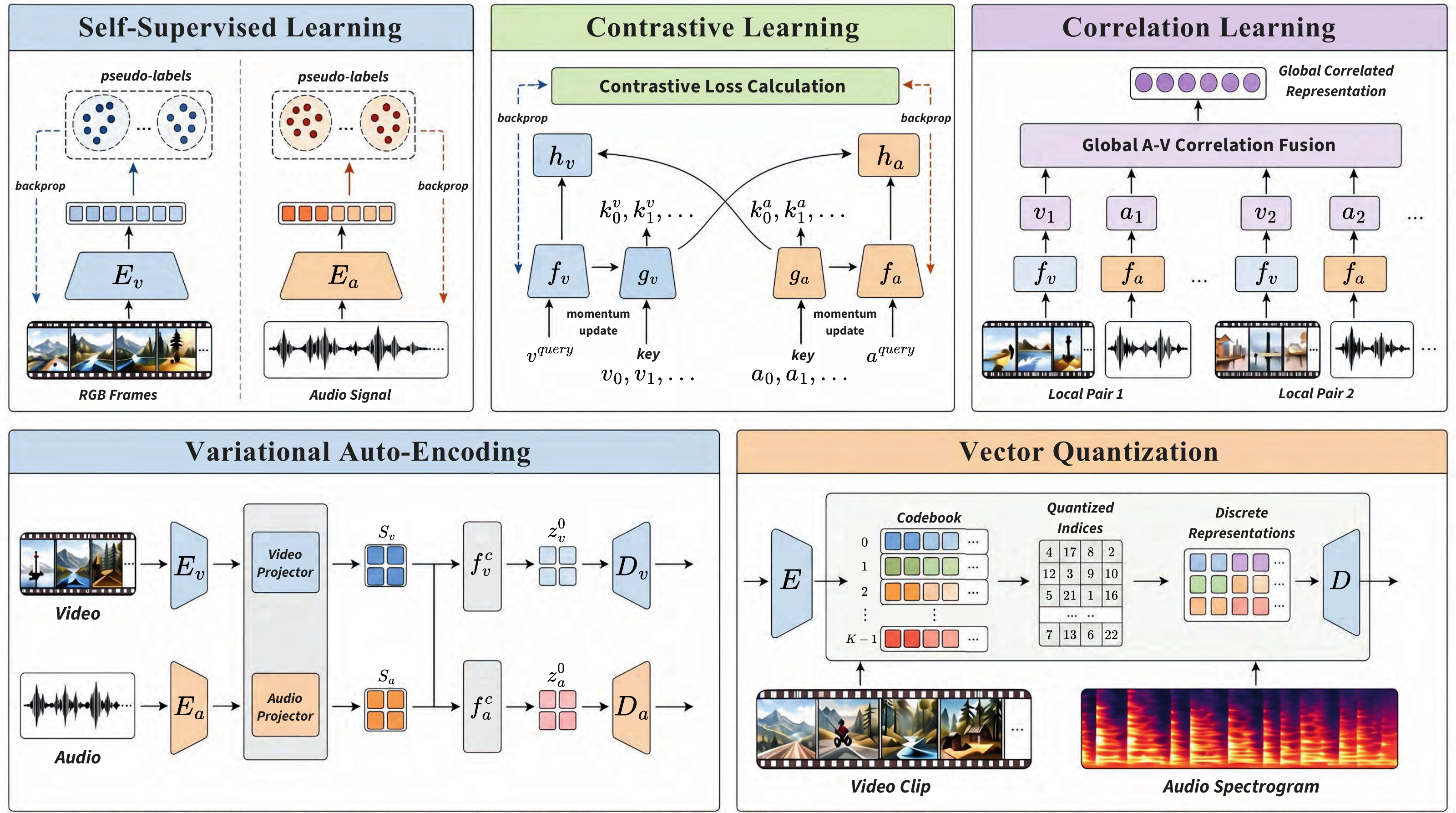}
    \caption{The overview comparison of different representation learning techniques.
    }
    \label{fig:tech_rep}
\end{figure}

As illustrated in \cref{fig:tech_rep}, representation-centric methods aim to transform raw input data (including image pixels, video frames, and audio waveforms) into either continuous embeddings or discrete tokens that can be efficiently processed by neural networks and support relevant artificial intelligence applications.

\subsubsection{Audio-Visual Feature Extraction and Representation}
\label{sec:tech:rep:feat}

Audio-visual representation learning exploits natural synchronization between visual and auditory signals in unlabeled videos to learn cross-modal semantics without manual annotations. Existing approaches mainly rely on self-supervised objectives, contrastive alignment, correlation modeling, or task-oriented encoder designs to learn shared audio-visual embeddings.

\paratitle{Self-Supervised Representation Learning.}
Self-supervised learning leverages cross-modal synchronization as a supervisory signal to learn joint audio-visual representations. Early proxy tasks such as Audio-Visual Correspondence and Temporal Synchronization encourage models to predict cross-modal alignment~\citep{arandjelovic2017look,korbar2018cooperative}, while later methods extend this idea through clustering-based representation learning, contrastive objectives, and scalable transformer architectures~\citep{alwassel2020self,morgado2021audio,akbari2021vatt}. Recent work further explores structured representation decomposition into shared, unique, and synergistic speech information through evolving self-supervised objectives~\citep{zhang2024es3}.

\paratitle{Multi-Modal Contrastive Learning.}
Contrastive learning aligns audio and visual modalities by maximizing similarity between synchronized pairs while separating mismatched samples in a shared embedding space. Representative approaches range from cross-modal semantic transfer and multi-modal alignment frameworks~\citep{wu2022wav2clip,girdhar2023imagebind} to improved contrastive objectives and sequential alignment strategies~\citep{kim2024equiav,tsiamas2025sequential,huang2023mavil}. Additional work relaxes strict synchronization assumptions or introduces task-specific contrastive objectives for downstream tasks such as event localization~\citep{sarkar2023self,bao2023cross}.

\paratitle{Audio-Visual Correlation Modeling.}
Beyond global embedding alignment, correlation modeling incorporates explicit cross-modal interaction mechanisms within the representation architecture to capture temporal and spatial dependencies. Examples include bottleneck-based cross-modal attention~\citep{nagrani2021attention}, bidirectional reconstruction frameworks for audio-visual segmentation~\citep{hao2024improving}, and training strategies that tolerate partial temporal misalignment~\citep{sarkar2023self}.

\paratitle{Task-Oriented Encoders.}
Task-oriented encoders incorporate domain knowledge or task-specific supervision to improve representation learning for particular applications. Audio-visual speech unit learning with masked prediction~\citep{shi2022learning} and modality-aware masked autoencoding for affective understanding~\citep{wu2025avf} are representative examples.

\subsubsection{Audio-Visual Variational Auto-Encoding and Reconstruction}
\label{sec:tech:rep:vae}

In audio-visual intelligence, variational auto-encoding (VAE) provides a unified framework for learning compact representations and enabling reconstruction across modalities~\citep{kingma2013auto,rezende2014stochastic}. As \cref{fig:tech_rep} shows, VAE adopts an encoder-decoder architecture that maps audio or visual inputs into a probabilistic latent space and reconstructs the original signals by sampling from this space~\citep{rezende2014stochastic}. By imposing a structured prior on latent variables, VAEs support both representation learning and generative reconstruction, making them a foundational technique for audio-visual modeling~\citep{qiu2025multimodal}.

In particular, recent multimodal VAEs have explored learning joint latent spaces from synchronized audio and visual inputs~\citep{sadok2024multimodal,qiu2025multimodal}. By structuring latent variables into shared and modality-specific components, these models support cross-modal reconstruction while disentangling modality-invariant semantics from modality-specific information~\citep{sadok2024multimodal,qiu2025multimodal}. Such structured latent designs make VAEs a useful framework for cross-modal reconstruction and generation in audio-visual settings~\citep{sadok2024multimodal}.

\subsubsection{Audio-Visual Discrete Tokenization}
\label{sec:tech:rep:tok}

Audio-visual discrete tokenization converts continuous acoustic and visual signals into compact symbolic units, providing discrete token interfaces commonly used in token-based multimodal models~\citep{zeghidour2021soundstream,van2017neural}. By applying vector quantization to learned latent spaces, structured signals are transformed into token sequences suitable for autoregressive modeling.

In audio, vector-quantization-based codecs such as SoundStream~\citep{zeghidour2021soundstream}, EnCodec~\citep{defossezhigh}, and DAC~\citep{dac2023} progressively improve reconstruction quality, while WavTokenizer~\citep{jiwavtokenizer} reduces token rates. Semantic tokens from self-supervised models such as HuBERT~\citep{hsu2021hubert} capture linguistic content but omit fine acoustic detail, motivating hybrid designs including X-Codec~\citep{ye2025codec}, BiCodec~\citep{bicodec2025}, and Mimi~\citep{defossez2024moshi}. In vision, discrete latent modeling originates from VQ-VAE~\citep{van2017neural} and VQGAN~\citep{esser2020tamingtransformershighresolutionimage}, with subsequent improvements in codebook scalability and efficiency such as MAGVIT-2~\citep{yu2024magvitv2}, VQGAN-LC~\citep{zhu2024vqganlc}, SimVQ~\citep{simvq2024}, and TokenFlow~\citep{tokenflow2025}, and extensions to video tokenization including Divot~\citep{divot2025}.

Recent efforts pursue unified multimodal tokenization, where shared or interleaved vocabularies support cross-modal reasoning and generation. SpeechGPT~\citep{zhang2023speechgpt}, AnyGPT~\citep{anygpt2024}, and Moshi~\citep{defossez2024moshi} exemplify token-level multimodal modeling, highlighting the importance of efficient and semantically grounded tokenizer design for audio-visual foundation models.

\subsection{Generation-centric Methods}
\label{sec:tech:gen}

\begin{figure}[!t]
    \centering
    \includegraphics[width=\linewidth]{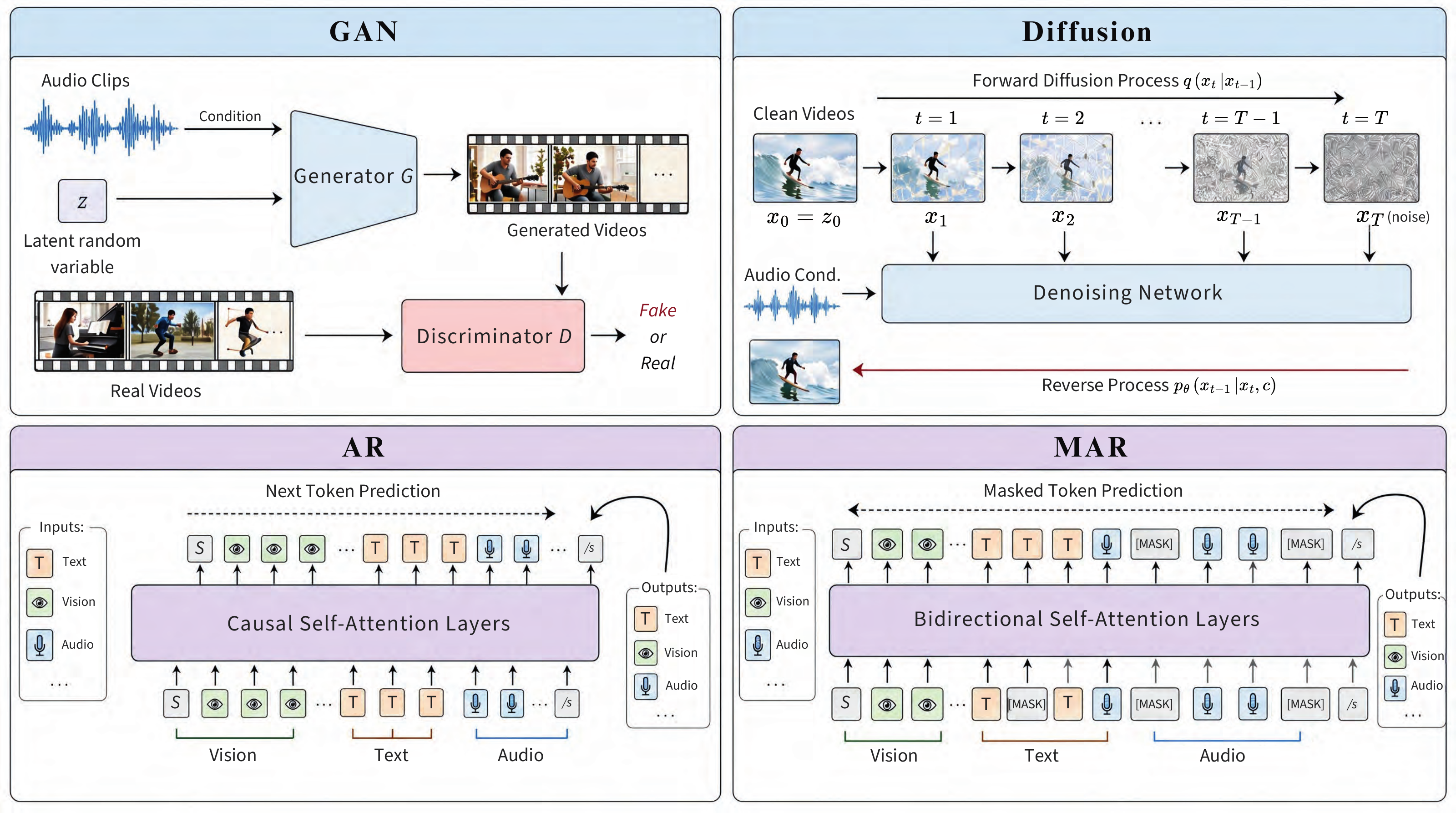}
    \caption{Overview comparison of generation mechanisms, including Generative Adversarial Networks (GANs), diffusion models, autoregressive (AR) models, and masked autoregressive (MAR) models.}
    \label{fig:tech_gen}
\end{figure}

Generation-centric methods support conditional synthesis and editing across unimodal (T2I, T2V, T2A) and cross-modal (A2V, V2A, T2AV) audio-visual tasks, with semantic coherence and temporal synchronization as key requirements. Several dominant generation paradigms are presented in \cref{fig:tech_gen}.

\subsubsection{Generative Adversarial Networks }
\label{sec:tech:gen:gan}

Generative Adversarial Networks (GANs) learn generative models through adversarial training between a generator and a discriminator~\citep{goodfellow2014generative}. Improvements such as DCGAN~\citep{radford2015unsupervised}, WGAN~\citep{arjovsky2017wasserstein}, and StyleGAN~\citep{karras2019style} significantly advanced image synthesis before the rise of diffusion models~\citep{ho2020denoising,rombach2022high}. GANs were also explored for cross-modal audio-visual generation~\citep{pan2019recent}, including text-conditioned synthesis~\citep{reed2016generative,tulyakov2018mocogan}, neural audio generation~\citep{kumar2019melgan,kong2020hifi}, and audio-driven visual animation~\citep{prajwal2020lip,lee2019dancing,owens2016visually,zhou2018visual}.

Despite these early demonstrations~\citep{baltruvsaitis2018multimodal}, progress in diffusion-based generators has made those newer paradigms increasingly common in recent audio-visual generation research~\citep{yang2023diffusion,rombach2022high,blattmann2023align}.

\subsubsection{Diffusion-based Generation }
\label{sec:tech:gen:diff}

Diffusion models generate data by learning a denoising process from noise to the data distribution~\citep{sohl2015deep,ho2020denoising}. Related continuous-time generators such as flow matching and rectified flow learn velocity fields that transport noise to data~\citep{lipman2023flow,liu2023flow}. Architecturally, diffusion and related continuous-time models have evolved from convolutional UNets~\citep{ronneberger2015u} to Diffusion Transformers (DiT)~\citep{peebles2023scalable}, which scale more effectively with data and model size~\citep{esser2024scaling}.

Building on these foundations, diffusion and related continuous-time generators have driven rapid progress in audio-visual and multimodal generation~\citep{yang2023diffusion}. In image and video generation, latent diffusion and rectified flow models underpin recent text-to-image~\citep{cao2025hunyuanimage, wu2025qwenimage, cai2025z} and text-to-video~\citep{kong2024hunyuanvideo,gao2025seedance,wan2025wan,liang2025univauniversalvideoagent} systems, enabling higher resolution, longer duration, and improved temporal consistency. For audio generation, latent-diffusion and related continuous-time models have been applied to speech~\citep{anastassiou2024seed} and sound effects~\citep{liu2024audioldm} synthesis, often operating in learned latent spaces. Recent audio-visual systems also explore aligned latent conditioning across text, audio, and visual inputs for tasks such as audio-conditioned video synthesis~\citep{gao2025wan}, visually grounded sound generation~\citep{cheng2025mmaudio}, and joint audio-video generation~\citep{low2025ovi}. 

Diffusion and related continuous-time models have become a prevalent paradigm in modern audio-visual generation, offering improved stability and scalability over earlier GAN-based approaches~\citep{yang2023diffusion,liu2023flow,yin2024one}.

\subsubsection{Autoregressive Generation
}
\label{sec:tech:gen:ar}

Autoregressive (AR) generators model the data distribution as a sequence of next-token predictions, enabling generation through sequential decoding. With discrete tokenization of images, audio, and video, multimodal generation can be formulated as language modeling over token streams~\citep{oord2016conditionalimagegenerationpixelcnn,oord2016wavenetgenerativemodelraw,van2017neural,zeghidour2021soundstream,defossezhigh}. This paradigm enables token-based generators, including VAR~\citep{tian2024var}, LlamaGen~\citep{sun2024llamagen}, Emu3~\citep{wang2024emu3}/Emu3.5~\citep{wang2025emu35}, and NOVA~\citep{deng2025nova} for images/video generation, as well as codec language models for audio and music generation including AudioLM~\citep{borsos2023audiolm}, MusicLM~\citep{agostinelli2023musiclm}, MusicGen~\citep{copet2023simple}, VALL-E~2~\citep{chen2024valle2}, and CosyVoice~2~\citep{du2024cosyvoice2}.

AR transformers share the next-token prediction interface of large language models~\citep{brown2020language} and can support multimodal prompting by interleaving modality-specific token streams, as demonstrated in systems such as VideoPoet~\citep{kondratyuk2024videopoet} and V-AURA~\citep{viertola2024temporallyalignedaudiovideo} for synchronized audio-visual generation. However, sequential decoding leads to high sampling latency and error accumulation for long sequences, motivating more efficient alternatives such as masked token generation (Section~\ref{sec:tech:gen:mar}) and diffusion-based approaches (Section~\ref{sec:tech:gen:diff}).

\subsubsection{Masked Autoregressive Generation
}
\label{sec:tech:gen:mar}

Masked autoregressive (MAR) generation, a.k.a. mask-and-predict decoding, replaces strictly sequential decoding with iterative masked token prediction, enabling parallel generation over subsets of tokens~\citep{ghazvininejad2019maskpredictparalleldecodingconditional}. Representative systems include MaskGIT~\citep{chang2022maskgit} and Muse~\citep{chang2023muse} for images, MAGVIT~\citep{yu2023magvit,yu2024magvitv2,luo2024openmagvit2}, MaskViT~\citep{gupta2022maskvitmaskedvisualpretraining}, and Phenaki~\citep{villegas2022phenakivariablelengthvideo} for video tokens, and SoundStorm for parallel audio codec generation~\citep{borsos2023soundstorm}. 

Compared to strictly autoregressive decoding, MAR enables faster parallel refinement over discrete visual or audio tokens~\citep{li2024mar}, but its performance often depends on tokenization quality and decoding schedules~\citep{weber2024maskbit,li2024mar}.

\subsection{LLM-Centric Methods}
\label{sec:tech:llm}
As generative large language models (LLMs)~\citep{achiam2023gpt,guo2025deepseek} have achieved strong performance in language understanding and generation, LLM-centric frameworks have become a prominent direction in audio-visual intelligence, where audio and visual inputs are converted into representations that LLMs can process to elicit multimodal intelligence.

\begin{figure}[!t]
    \centering
    \includegraphics[width=\linewidth]{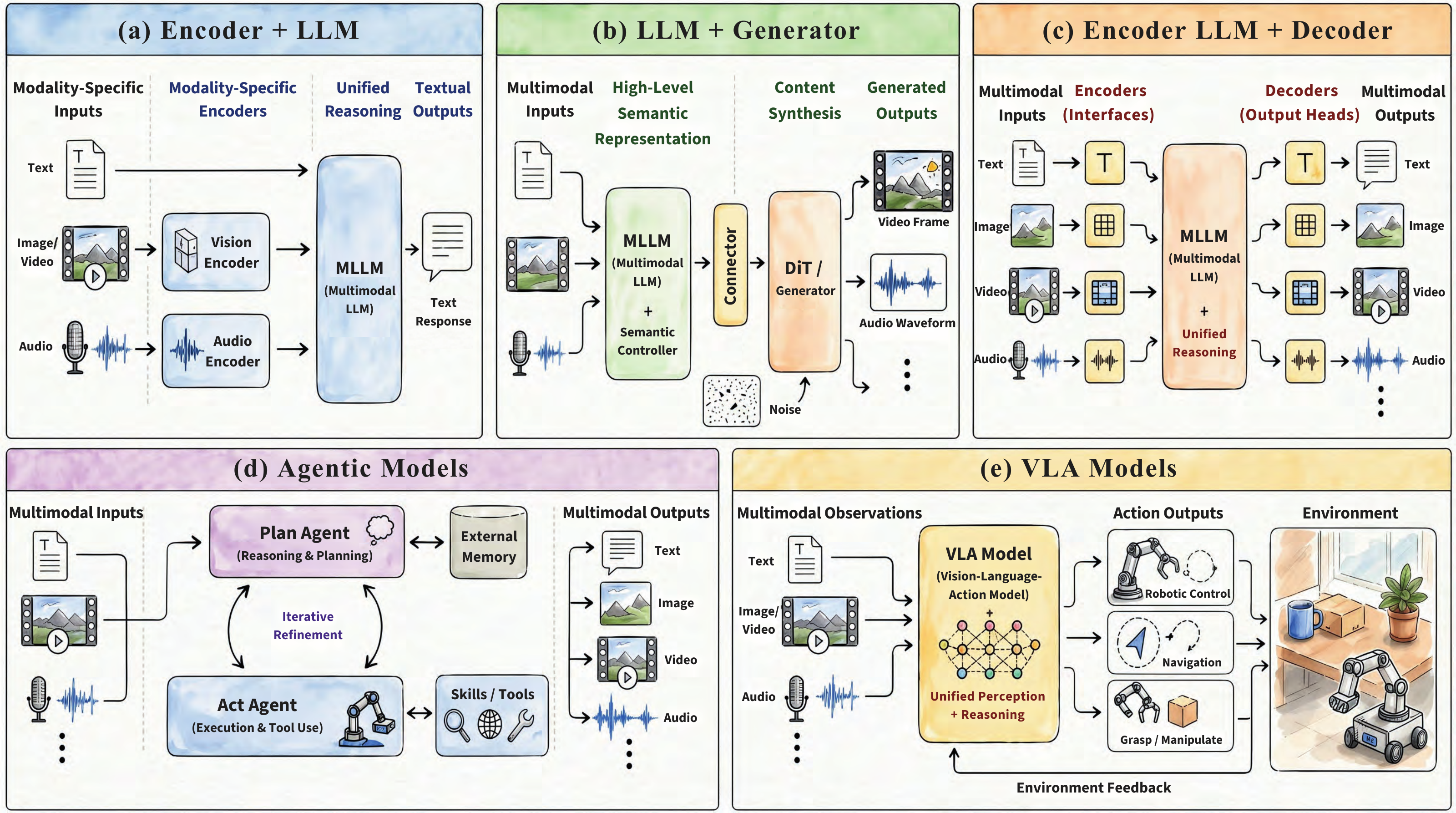}
    \caption{The overview comparison on the mechanism of (a) Encoder + LLM, (b) LLM + Generator, (c) Encoder + LLM + Decoder, (d) Agentic LLM, (e) VLA Model.}
    \label{fig:tech_llm}
\end{figure}

\subsubsection{Encoder+LLM for Multimodal Perception}
\label{sec:tech:llm:und}

\paratitle{Definition.}
Encoder+LLM system connects modality encoders (e.g., vision or audio) to a pretrained LLM backbone. Encoded sensory features are projected into the LLM embedding space through lightweight MLP or attention-based adapters, enabling the language model to interpret and reason over multimodal inputs.

\paratitle{Taxonomy and Representative Systems.}
Encoder+LLM approaches can be grouped by modality awareness.

\textit{(1) Vision-LLM Perception.}
Vision-LLM pipelines project visual features from pretrained encoders such as CLIP~\citep{radford2021learning} and SigLIP~\citep{zhai2023sigmoid,bai2025qwen2} into LLMs through adapters or cross-attention modules, as in Flamingo~\citep{alayrac2022flamingo}, BLIP-2~\citep{li2023blip}, and MiniGPT-style systems~\citep{zhuminigpt}. Extensions to video understanding include Video-ChatGPT~\citep{maaz2024video}, Video-LLaVA~\citep{lin2024video}, Video-LLaMA~\citep{zhang2023video}, Vitron~\cite{fei2024vitron} and LLaMA-VID~\citep{li2024llama}. 
Recent models scale perception with stronger encoders and token compression, such as Qwen2.5/3-VL~\citep{bai2025qwen2,bai2025qwen3}, InternVL~\citep{chen2024expanding,wang2025internvl3}, and long-video frameworks including LongVA~\citep{zhang2024long} and LongVU~\citep{shen2024longvu}. Related inference-acceleration methods such as FastV~\citep{chen2024image} further improve token efficiency at inference time.

\textit{(2) Audio-LLM Perception.}
Audio-LLM systems encode waveforms using pretrained acoustic models such as HuBERT~\citep{hsu2021hubert}, SpeechT5~\citep{ao2022speecht5}, BEATs~\citep{chen2023beats}, and Whisper~\citep{radford2023robust}, whose outputs are projected into the language space. Representative efforts include Pengi~\citep{deshmukh2023pengi}, LLaSM~\citep{shu2023llasm}, GAMA~\citep{ghosh2024gama}, SALMONN~\citep{tang2023salmonn}, and Qwen-Audio series~\citep{chu2023qwenaudio, chu2024qwen2audio}.

\textit{(3) Audio-Visual Encoder-LLM Integration.}
Audio-visual reasoning systems extend this idea by jointly integrating visual and acoustic encoders with an LLM. Representative models include Video-LLaMA 2~\citep{cheng2024videollama}, NExT-GPT~\citep{wu2024next}, OneLLM~\citep{han2024onellm}, AnyGPT~\citep{anygpt2024}, and VITA~\citep{fu2025vita}. Related omni-modal models, such as GPT-4o~\citep{hurst2024gpt}, Qwen2.5-Omni~\citep{xu2025qwen2}, Ming-Omni~\citep{ai2025ming}, and Qwen3-Omni~\citep{xu2025qwen3omni}, further explore tighter joint training and broader multimodal interaction.

\begin{insightbox}[\faLightbulbO~Pros and Challenges: Encoder+LLM as a Modular Perception Framework.]
Encoder+LLM offers a modular framework for multimodal perception by combining strong pretrained encoders with LLM reasoning, enabling flexible integration of vision and audio models while requiring limited training. However, indirect feature projection may limit fine-grained cross-modal alignment, and high-dimensional tokens introduce computational overhead for long videos and continuous audio streams.
\end{insightbox}

\subsubsection{LLM+Generator for Multimodal Generation
}
\label{sec:tech:llm:gen}


\paratitle{Definition.}
LLM+Generator couples an instruction-following LLM with external modality generators (e.g., diffusion models or neural codecs for image/video/audio) to support controllable multimodal synthesis and editing by interpreting and reasoning over user inputs, and enable flexible multimodal generation by reusing off-the-shelf generators and optional perception tools without training a monolithic model.

\paratitle{Taxonomy and Representative Systems.}
Existing LLM+Generator systems fall into two groups.

\textit{(1) Prompt/Tool Orchestration.} The LLM converts user intent into tool calls and prompts with minimal learned interfaces. Visual ChatGPT~\citep{wu2023visualchatgpt} composes multiple visual models for multi-step image editing and generation, while AudioGPT~\citep{huang2024audiogpt} routes requests to specialized audio and talking-head generators, including AudioLDM~\citep{liu2023audioldm}. Although related to agentic tool use, these systems operate over relatively fixed toolsets and remain generator-centered.

\textit{(2) Learned Bridging to Generators.} 
These methods introduce trainable interfaces, such as adapters or learnable queries, to connect LLMs with downstream generators. NExT-GPT~\citep{wu2024next} and CoDi-2~\citep{tang2024codi2} extend this design to multimodal and any-to-any generation, while JavisGPT~\citep{liu2025javisgpt} targets unified audio-visual comprehension and generation and A2-LLM~\citep{chen2026a2llm} focuses on conversational audio-avatar interaction. Standalone models such as Qwen-Image~\citep{wu2025qwenimage} and UniVideo~\citep{wei2025univideo} can also be used as plug-in visual renderers.


\begin{insightbox}[\faLightbulbO~Pros and Challenges: LLM+Generator as a Modular Generation Framework.]
LLM+Generator provides strong modularity by decoupling high-level reasoning from modality-specific generation, enabling the reuse of state-of-the-art audio and visual generators. This design supports flexible multimodal interaction and easy extension to new tools or modalities. However, cascaded pipelines may suffer from interface mismatches, error accumulation, and difficulties maintaining temporal synchronization across audio-visual generators.
\end{insightbox}

\subsubsection{Unified Model for Joint Perception and Generation
}
\label{sec:tech:llm:unify}


\paratitle{Definition.}
Unified models perform multimodal perception and generation within a single end-to-end architecture, integrating modality tokenizers and decoders as native components of a shared backbone. Compared with cascaded LLM+Generator pipelines, they reduce interface mismatches and cascading errors while enabling tighter cross-modal alignment and lower-latency interaction.

\paratitle{Taxonomy and Representative Systems.}
Existing unified systems can be broadly grouped into three threads based on their primary modeling objectives.

\textit{(1) Omni streaming assistants.}
These models target real-time multimodal interaction with tightly coupled perception and generation and support native speech input and output, such as GPT-4o~\citep{hurst2024gpt}, Qwen2.5-Omni~\citep{xu2025qwen2}, Mini-Omni~\citep{xie2024miniomni}, and InteractiveOmni~\citep{tong2025interactiveomni}. Recent additions include Moshi~\citep{defossez2024moshi}, a full-duplex speech-text foundation model achieving 160ms theoretical latency through its Inner Monologue method and multi-stream hierarchical token generation, and Qwen3-Omni~\citep{xu2025qwen3omni}, which introduces a Thinker-Talker MoE architecture with joint multimodal training across 2 trillion tokens.

\textit{(2) Unified token-space / interleaved multimodal models.}
These systems model mixed-modality token sequences within a single transformer, enabling unified understanding and generation of interleaved text-image content. Representative models include Emu~\citep{sun2024emu}, DreamLLM~\citep{dong2024dreamllm}, Janus~\citep{wu2024janus}, Chameleon~\citep{chameleonteam2024chameleon}, and Show-o~\citep{xie2024showo}. Recent advances include Janus-Pro~\citep{chen2025januspro}, which decouples visual encoding into SigLIP for understanding and a VQ tokenizer for generation; BAGEL~\citep{deng2025emerging}, a MoT architecture exhibiting emergent compositional reasoning; Emu3~\citep{wang2024emu3}, which unifies image and video understanding and generation through pure next-token prediction; and Transfusion~\citep{zhou2024transfusion}, which combines autoregressive and diffusion objectives within a single transformer.

\textit{(3) Unified audio-visual generators.}
These models support joint video generation or editing with synchronized audio through closely coupled multimodal generation frameworks. Representative examples include VideoPoet~\citep{kondratyuk2024videopoet} for autoregressive audio-video generation, JavisGPT~\citep{liu2025javisgpt} for unified audio-visual comprehension and generation, and Object-AVEdit~\citep{fu2025objectavedit} for object-level audio-visual editing.
Overall, audio-visual joint modeling remains at an early stage, as both audio and video modeling are still under active exploration.

\begin{insightbox}[\faLightbulbO~Pros and Challenges: Encoder+LLM+Decoder as A Unified Framework.]
Strengths include lower end-to-end latency, more consistent cross-modal alignment, and fewer brittle interfaces between modules. Key challenges include stable and general tokenization for heterogeneous audio (speech/music/sound effects), long-context temporal modeling for video, streaming memory and interruption handling, and evaluation protocols for open-ended multimodal outputs.
\end{insightbox}

\subsubsection{Agentic System for Interactive Perception and Generation
}
\label{sec:tech:llm:agent}

\paratitle{Definition.}
Agentic systems treat a (multimodal) LLM as a planner that decomposes user requests into multi-step actions and interacts with external tools for perception and generation. Unlike fixed cascaded pipelines in \cref{sec:tech:llm:gen}, they dynamically select tools, iterate over intermediate results, and maintain task state, enabling adaptive control for complex multimodal tasks such as audio-visual workflows.

\paratitle{Taxonomy and Representative Systems.}
Existing agentic systems can be grouped by their primary application:

\textit{(1) General tool-routing agents.}
These systems use an LLM to plan tasks and dispatch requests to heterogeneous model hubs or APIs. A representative example is HuggingGPT~\citep{shen2023hugginggpt}, which demonstrates large-scale tool selection and orchestration across diverse models.

\textit{(2) Perception-centric video agents.}
Recent work explores interactive agents for long-horizon video understanding, where the model iteratively queries perception tools and updates reasoning based on intermediate observations, such as DoraemonGPT~\citep{yang2025doraemongpt} and VCA~\citep{yang2025vca}.

\textit{(3) Generation-oriented creative agents.}
Other systems focus on structured media creation through multi-step editing and generation workflows. Examples include video creation agents such as V-Stylist~\citep{yue2025vstylist} and Preacher~\citep{liu2025preacher}, as well as audio-focused frameworks like Audio-Agent~\citep{wang2025audioagent} and AudioToolAgent~\citep{wijngaard2025audiotoolagent}.

\begin{insightbox}[\faLightbulbO~Pros and Challenges: Agentic Multimodal Foundation Systems.]
Agentic systems enable flexible multimodal workflows by decomposing complex tasks into modular tool executions and iterative reasoning steps. This design supports long-horizon interaction, heterogeneous tool integration, and adaptive refinement. However, reliable tool grounding, state management, and error propagation remain major challenges, especially when coordinating multiple perception and generation modules in audio--visual pipelines.
\end{insightbox}

\subsubsection{Visual-Language-Action Models for Embodied Interaction}
\label{sec:tech:llm:vla}

\paratitle{Definition.}
Vision-Language-Action (VLA) models extend LLMs by incorporating action spaces and interacting with physical environments, enabling agents to close the perception-reasoning-action loop. A typical VLA system takes multimodal observations (e.g., images, audios, and language instructions) and outputs actions or policies, which may be discrete navigation commands or continuous robot control signals.

\paratitle{Taxonomy and Representative Systems.}
Recent VLA research focuses on several design directions.

\textit{(1) Language-conditioned policy models.}
RT-2~\citep{zitkovich2023rt} established the modern VLA paradigm by co-training pretrained vision-language models on robot trajectories and internet-scale vision-language data, representing actions as token sequences aligned with language representations. Subsequent extensions such as RT-H~\citep{belkhale2024rt} introduce hierarchical policies to separate high-level planning from low-level control, while RT-Trajectory~\citep{RT-Trajectory} uses hindsight trajectory sketches to improve task generalization.

\textit{(2) Open-source and scalable VLAs.}
OpenVLA~\citep{kimopenvla} demonstrates that large-scale multimodal pretraining combined with robot demonstration datasets can produce strong generalization across tasks and embodiments. Other efficient designs such as TinyVLA~\citep{wen2025tinyvla} and SmolVLA~\citep{shukor2025smolvla} target lightweight deployment, while large industrial systems such as GR00T-N1~\citep{bjorck2025gr00t} and Gemini Robotics~\citep{team2025gemini} extend VLA concepts to whole-body robot control.

\textit{(3) Diffusion and planning-based VLA policies.}
Several works explore generative policy representations and structured planning for complex control. Examples include flow-matching policies such as $\pi_0$~\citep{black2024pi_0}, open-source generalist policies such as Octo~\citep{team2024octo}, unified perception-action models such as GraspVLA~\citep{deng2025graspvla}, and architectures incorporating trajectory reasoning or modular action heads, including TraceVLA~\citep{zheng2024tracevla}, CogACT~\citep{li2024cogact}, and CogVLA~\citep{li2025cogvla}. Recent work also explores robustness and failure recovery, such as FailSafe~\citep{lin2025failsafe}, as well as world-model integration, such as RynnVLA-002~\citep{cen2025rynnvla}.

\begin{insightbox}[\faLightbulbO~Pros and Challenges: Embodied Vision-Language-Action Models.]
VLA models enable embodied agents to integrate perception, reasoning, and action within a unified multimodal framework, benefiting from large-scale vision-language pretraining and language-conditioned planning. However, challenges remain in long-horizon planning, cross-embodiment transfer, and robust execution in dynamic environments. Improving data efficiency, safety, and generalization across robot platforms remains critical for real-world embodied deployment.
\end{insightbox}

\section{Audio-Visual Perception}
\label{sec:und}

\begin{figure*}[!t]
    \centering
    \includegraphics[width=\textwidth]{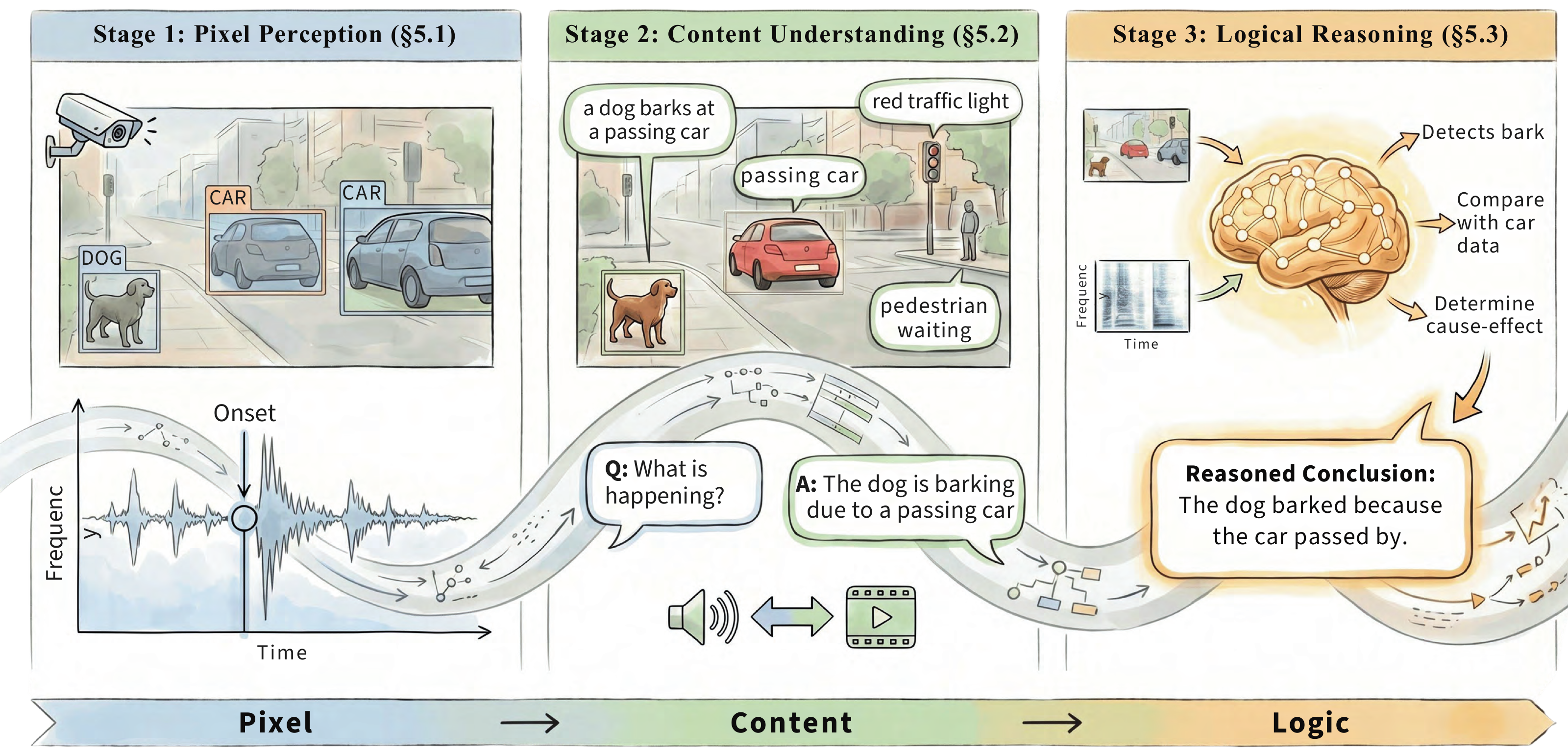}
    \caption{Organization of audio-visual perception and understanding. We organize this section into pixel perception, content understanding, and logical reasoning, moving from low-level signal detection to semantic grounding and causal inference.}
    \label{fig:und_overview}
\end{figure*}

Perception in audio-visual systems begins with \emph{what} can be read off raw signals, then moves to \emph{what} those signals mean in context, and finally to \emph{why} and \emph{how} events are related over time; \cref{fig:und_overview} schematizes that progression.
This section is organized in three parts that mirror that ladder: we start from pixel- and sample-level structure in each modality and in joint settings, then treat semantic content understanding, and last cover reasoning that goes beyond one-step recognition.
Throughout, we distinguish unimodal preview material (brief definitions and pointers) from audio-visual problems where cross-modal alignment, data, and benchmarks are developed in more depth, as the survey scope in \cref{sec:task} requires.

\subsection{Audio-Visual Pixel Perception}
\label{sec:und:pixel}

Pixel-level perception is concerned with time-frequency structure in sound, spatial layout in images and video, and explicit alignment between the two, including when events are jointly visible and audible.
The subsections below treat audio and vision in turn at this granularity, then audio-visual event localization, segmentation, and cross-modal temporal synchronization.


\subsubsection{Audio Perception}
\label{sec:und:pixel:audio}

\paratitle{Definition.}
Audio perception focuses on extracting low- and mid-level structure directly from acoustic signals, operating at the waveform or time-frequency level rather than at the level of semantic interpretation or multi-step reasoning.
Typical outputs include event presence, temporal boundaries, onsets, beats, voice activity, and separated source components across speech, environmental sound, and music.

\paratitle{Representative Tasks and Methods.}
We briefly summarize three representative task families that provide useful acoustic primitives for later audio-visual modeling:

\textit{Audio event detection and clip-level classification}
identify what acoustic events are present in a clip or short time segment.
Self-supervised encoders such as HuBERT~\citep{hsu2021hubert}, BEATs~\citep{chen2023beats}, MERT~\citep{li2024mert}, and MuQ~\citep{zhu2025muq} provide reusable representations for speech classification~\citep{chang2023speechprompt}, music tagging~\citep{ashraf2023hybrid}, cross-modal tagging~\citep{huang2022mulan}, and sound event detection~\citep{martin2023training}.

\textit{Temporal event localization and sequential labeling}
estimate when sounds occur, including voice activity, sound-event boundaries, onsets, and beats.
Representative approaches include spiking networks for low-power VAD~\citep{yang2024svad}, teacher-student transformers for sound event detection~\citep{shao2024fine}, and rhythmic modeling for beat tracking~\citep{zhao2022beat}.
Recent analyses suggest that general-purpose audio-text LLMs still lag behind task-specific models on such fine-grained temporal perception tasks~\citep{ma2025cmiBench}.

\textit{Audio source separation}
recovers individual sources from mixtures by disentangling overlapping acoustic components.
Because separation is both a perceptual primitive and a signal-generation operation, we only note representative systems here, including TF-GridNet~\citep{wang2023tf}, Dual-Path Mamba~\citep{jiang2025dual}, Hybrid Transformer Demucs~\citep{rouard2023hybrid}, and Band-Split RNN~\citep{luo2023music}; its conditional transformation role is revisited in \cref{sec:gen:cond_av:audio}.

\subsubsection{Visual Perception}
\label{sec:und:pixel:visual}


\paratitle{Definition.}
Visual pixel-level perception refers to the fine-grained understanding of visual scenes by predicting spatially resolved outputs for each pixel or region, rather than assigning a single image-level label. These tasks are fundamental to grounding objects, actions, and events in space and time, thereby providing the structural basis for integrating visual content with audio in multimodal systems.

\paratitle{Representative Tasks and Methods.}
In this section, we briefly introduce several core visual perception approaches that are especially amenable to audio-visual integration.
Other purely visual perception methods, such as depth estimation~\citep{ming2021deep}, pose estimation~\citep{zheng2023deep}, optical flow~\citep{zhai2021optical}, or salient object detection/segmentation~\citep{borji2019salient} are not included due to space limitations.

\textit{Object Detection and Grounding}, which aims to localize and classify objects using bounding boxes~\citep{zou2023object}. Classic approaches include two-stage detectors (e.g., Faster R-CNN~\citep{ren2016faster}) and one-stage detectors (e.g., YOLO~\citep{terven2023comprehensive}, DETR~\citep{zhu2020deformable}), and recent advances move toward open-vocabulary and grounded detection, allowing models to detect arbitrary objects via natural language prompts~\citep{minderer2023scaling}. Representative works include GLIP~\citep{zhang2022glipv2}, Grounding DINO~\citep{liu2024grounding}, and MLLM-based systems~\citep{zhang2023next} capable of visual question-answering.

\textit{Object Segmentation}, which assigns a class label or instance ID to each pixel, delineating object boundaries in the scene~\citep{yu2023techniques}. This includes: (1) semantic segmentation~\citep{hao2020brief}: labeling each pixel with a category (e.g., road, car, person); (2) instance segmentation~\citep{hafiz2020survey}: distinguishing individual object instances within each class; and (3) panoptic segmentation~\citep{kirillov2019panoptic}: unifying both, by assigning every pixel to either a specific object instance or a background category. Notable models include Mask2Former~\citep{cheng2022masked} for universal image segmentation and Segment Anything Model (SAM)~\citep{kirillov2023segment,ravi2024sam,carion2025sam} for promptable, general-purpose segmentation.

\textit{Object Tracking,} which aims to follow objects across video frames, producing their spatio-temporal trajectories (bounding box and identity sequences)~\citep{luo2021multiple}. Single-object trackers (e.g., SUTrack~\citep{chen2025sutrack}) and multi-object trackers (e.g., ByteTrack~\citep{zhang2022bytetrack}) maintain object identity over time. Recent models like the Track Anything Model (TAM)~\citep{yang2023track} support prompt-based or language-guided tracking~\citep{maalouf2024follow}, offering potential interactive capabilities useful for audio-visual synchronization and temporal reasoning.

\textit{Temporal Action Detection,} which involves identifying and localizing action instances within untrimmed video sequences, typically by predicting their start and end times~\citep{liu2022empirical}. Unlike image-based action classification, this task requires understanding temporal structure and motion cues, often at frame-level resolution. Representative works include TriDet~\citep{shi2023tridet} and transformer-based methods like TadTR~\citep{liu2022end} and ActionFormer~\citep{zhang2022actionformer}. This task lays the groundwork for aligning visual events with audio streams, such as matching gestures to speech or locating sound-emitting activities.


\subsubsection{Audio-Visual Event Localization/Grounding}
\label{sec:und:pixel:av_ground}


\paratitle{Definition.}
Audio-visual event localization (AVEL) aims to temporally localize events in a video that are both visible and audible (\ie events that are simultaneously present in visual and audio modalities). In other words, given an unconstrained video containing various sounds and scenes, the goal is to identify when and where an event occurs that can be perceived in both vision and sound.

\paratitle{Representative Tasks and Methods.}
For AVEL, \citet{tian2018audio} first introduced audio-guided visual attention and a dual multimodal residual network. Subsequent work has improved robustness and accuracy through several main directions: (1) cross-modal attention strengthens discriminative feature extraction and modality correspondence, including audio-guided spatial-channel attention~\citep{zhao2018sound, xu2020cross}, cross-modal co-attention~\citep{wu2019dual, xuan2020cross}, and gated attention~\citep{lin2019dual}; (2) adaptive fusion methods exploit complementary information across modalities, such as multimodal attention fusion~\citep{zhou2021positive} and spatiotemporal fusion encoders~\citep{rao2022dual}; and (3) semantic modeling further improves localization via semantic relation modulation~\citep{xu2020cross}, video-level semantic consistency~\citep{tian2020unified}, and dense modality interaction~\citep{rao2022dual}.

Pre-trained models have further advanced AVEL and can be roughly divided into two groups. One line adopts pre-trained audio and video encoders to provide semantically rich modality-specific features before fusion, improving robustness and generalization~\citep{mahmud2023ave}. The other re-purposes large pre-trained image models as visual backbones and equips them with lightweight cross-modal adapters for parameter-efficient audio-visual learning~\citep{lin2023vision, wang2024towards, wang2025prompt}.

More recently, open-set or open-vocabulary AVEL introduces semantic embeddings to relax the closed-set assumption and localize unseen event categories~\citep{yu2024openave, zhou2025towards}. Dense AVEL further extends the task from trimmed to untrimmed videos with multiple overlapping events, requiring joint reasoning over temporal boundaries and audio-visual correspondence~\citep{geng2023dense,zhou2025dense}. Overall, these advances push AVEL toward more scalable, robust, and semantically flexible audio-visual understanding.

\paratitle{Benchmarks.}
For audio-visual event localization (AVEL), the standard benchmark is the AVE dataset~\citep{tian2018audio}, which contains 4,143 trimmed 10-second videos with segment-level labels over 28 event categories. For weakly supervised settings, the LLP dataset~\citep{tian2020unified} provides only video-level labels for training while using segment-level annotations for evaluation, covering audio-only, visual-only, and audio-visual events. For dense AVEL in untrimmed videos, datasets such as UnAV-100~\citep{geng2023dense} include longer videos with multiple overlapping events, supporting evaluation of fine-grained temporal boundary localization.

\begin{insightbox}[\faLightbulbO~Trend: Towards Open-Vocabulary Audio-Visual Event Localization/Grounding.]
Recent work shows a clear shift from closed-set to open-vocabulary AVEL, aiming at more scalable and flexible multimodal understanding. By leveraging semantic embeddings, large-scale pre-training, and language-aligned representations, these methods can localize unseen events at inference time. This trend reflects the broader move in multimodal learning toward semantic generalization and deployment in open-world environments.
\end{insightbox}

\begin{table}[!t]
    \centering
    \caption{Representative benchmarks for audio-visual event localization (AVEL).}
    \label{tab:bench_avel}
    \benchmarktablestyle
    \begin{fullwidthtabular}{8}{l l l l}
    \toprule
    \benchhead \tblhead{Dataset} & \tblhead{Scale} & \tblhead{Annotation} & \tblhead{Primary Use} \\
    \midrule

    AVE~\citep{tian2018audio}
    & 4,143 videos (10s)
    & Segment-level AV labels
    & Supervised AVEL \\

    LLP~\citep{tian2020unified}
    & 11K videos
    & Video-level weak labels
    & Weakly supervised AV parsing \\

    UnAV-100~\citep{geng2023dense}
    & 100 untrimmed videos
    & Dense temporal boundaries
    & Dense long-video localization \\

    AVVP~\citep{tian2020unified}
    & 11K videos
    & Multi-label temporal events
    & Unified AV parsing \\

    OpenAVE~\citep{yu2024openave}
    & Large-scale web videos
    & Open-set semantic labels
    & Open-vocabulary localization \\

    \bottomrule
    \end{fullwidthtabular}
\end{table}

\begin{table}[!t]
\centering
\caption{Performance comparison of representative AVEL methods on the AVE dataset.}
\label{tab:avel_sota}
\performancetablestyle
\begin{fullwidthtabular}{8}{l l l c}
\toprule
\perfhead \tblhead{Method} & \tblhead{Pre-training Data} & \tblhead{Backbone} & \tblhead{Acc. (\%)} \\
\midrule

PSP~\citep{zhou2021positive} & ImageNet + AudioSet & VGG-19 + VGGish & 77.8 \\
DAM~\citep{wu2019dual} & ImageNet + AudioSet & VGG-19 + VGGish & 74.5 \\
DPNet~\citep{rao2022dual} & ImageNet & VGG-19 + VGGish & 78.9 \\
AVEL~\citep{tian2018audio} & ImageNet + AudioSet & ResNet-152 + VGGish & 74.0 \\
CMRAN~\citep{xu2020cross} & ImageNet + AudioSet & ResNet-152 + VGGish & 78.3 \\
AVSDN~\citep{lin2019dual} & ImageNet + AudioSet & ResNet-152 + VGGish & 75.4 \\
LAViSH~\citep{lin2023vision} & ImageNet & Swin-L (shared AV encoder) & 81.1 \\
STG-CMA~\citep{wang2024towards} & CLIP & ViT-L/14 (shared AV encoder) & \textbf{83.3} \\
AV-STFP~\citep{wang2025prompt}  & CLIP & ViT-L/14 (shared AV encoder) & 83.0 \\

\bottomrule
\end{fullwidthtabular}
\end{table}

\subsubsection{Audio-Visual Segmentation}
\label{sec:und:pixel:av_seg}

\paratitle{Definition.}
Audio-visual segmentation (AVS) aims to identify and delineate the visual regions (pixel-level maps) of objects that are actively producing sound at the time of the image frame~\citep{zhou2022audio,zhou2025audio}, given the audio track and optionally the object category. The task requires learning fine-grained correspondences between audio cues and visual regions, while ignoring silent or off-screen sources.

\paratitle{Methods.}
Recent progress in audio-visual segmentation (AVS) has evolved from early modality fusion and weakly supervised learning to transformer-based methods and large foundation models.

A major line of work learns \textit{fused audio-visual representations} for pixel-wise segmentation. TPAVI~\citep{zhou2022audio} injects temporal audio semantics into a visual feature pyramid, RAVS~\citep{liu2025robust} mitigates audio ambiguity via semantic-density-aware visual grouping, and DDESeg~\citep{liu2025dynamic} disentangles mixed audio cues and filters noise with visual context.

To reduce annotation cost, \textit{weakly- and unsupervised} AVS methods have also been explored. WS-AVS~\citep{mo2023weakly} uses instance-level labels with multi-scale alignment and contrastive learning, while SAMA-AVS~\citep{liu2024annotation} and MoCA~\citep{bhosale2025unsupervised} leverage pretrained models~\citep{kirillov2023segment,oquab2024dinov2} to generate pseudo labels.

Recent \textit{transformer-based} methods adopt query-driven designs for object-level audio-visual alignment~\citep{li2024qdformer}. AQFormer~\citep{huang2023discovering} uses audio-conditioned object queries for mask prediction, CPM~\citep{chen2024cpm} introduces class-aware prompts, AVSegFormer~\citep{gao2024avsegformer} implicitly separates audio sources to match visual features, and DeepAVFusion~\citep{mo2024unveiling} integrates audio-visual patches with learnable tokens.

More recently, AVS has increasingly benefited from \textit{large foundation models}, where audio serves as a high-level prompt rather than fused feature input. GAVS~\citep{wang2024prompting} and COMBO~\citep{yang2024cooperation} build on SAM-like segmentation priors~\citep{kirillov2023segment,ravi2024sam}; DiffusionAVS~\citep{mao2025contrastive} and BAVS~\citep{liu2024bavs} exploit diffusion-based generative priors~\citep{bao2023one}; and TeSO~\citep{wang2024can} and SSP~\citep{lee2025optical} incorporate MLLMs for higher-level reasoning.

Overall, AVS is shifting from pixel-level fusion toward audio-guided prompting, reasoning, and generation on top of foundation models, improving generalization and reducing reliance on dense annotations.

\begin{table}[!t]
    \centering
    \caption{Performance comparison on AVSBench~\citep{zhou2022audio} and AVSBench-Semantic~\citep{zhou2025audio} with $\mathcal{J}$ and $\mathcal{F}$ score metrics. Best results are marked in \textbf{bold}.}
    \label{tab:perf_avs}
    \performancetablestyle
    \begin{fullwidthtabular}{14}{l c c c c c c}
    \toprule
    \perfhead
    & \multicolumn{2}{c}{\tblhead{S4}} & \multicolumn{2}{c}{\tblhead{MS3}} & \multicolumn{2}{c}{\tblhead{AVSS}} \\
    \perfhead
    \multirow{-2}{*}{\tblhead{Model}} & \tblhead{$\mathcal{J}\uparrow$} & \tblhead{$\mathcal{F}\uparrow$} & \tblhead{$\mathcal{J}\uparrow$} & \tblhead{$\mathcal{F}\uparrow$} & \tblhead{$\mathcal{J}\uparrow$} & \tblhead{$\mathcal{F}\uparrow$} \\
    \midrule
    AVSBench~\citep{zhou2022audio}    & 78.74 & 87.90 & 54.00 & 64.50 & 29.77 & 35.20 \\
    BAVS~\citep{liu2024bavs}             & 82.68 & 89.75 & 59.63 & 65.89 & 33.59 & 37.52 \\
    CPM~\citep{chen2024cpm}              & 81.37 & 90.47 & 59.80 & 71.00 & 34.53 & 39.57 \\
    TESO~\citep{wang2024can}             & 83.27 & 93.30 & 66.02 & 80.10 & 38.96 & 45.10 \\
    WS-AVS~\citep{mo2023weakly}          & 34.13 & 51.76 & 30.85 & 46.87 & - & - \\
    QDFormer~\citep{li2024qdformer}      & 79.50 & 88.20 & 61.90 & 66.10 & 53.40 & - \\
    COMBO~\citep{yang2024cooperation}    & 84.70 & 91.90 & 59.20 & 71.20 & 42.10 & 46.10 \\
    VPO~\citep{chen2024unraveling}       & 85.77 & 92.86 & 62.39 & 73.62 & 44.70 & 57.76 \\
    DeepAVFusion~\citep{mo2024unveiling} & 89.94 & 92.34 & 52.05 & 58.29 & - & - \\
    RAVS~\citep{liu2025robust}           & \textbf{93.10} & 93.80 & 70.60 & 82.10 & 60.80 & 70.60 \\
    DDESeg~\citep{liu2025dynamic}        & 92.40 & \textbf{95.90} & \textbf{72.30} & \textbf{83.40} & \textbf{63.40} & \textbf{72.30} \\
    \bottomrule
    \end{fullwidthtabular}
\end{table}

\paratitle{Extension.}
Beyond conventional AVS, recent work extends the task to referring audio-visual segmentation (\textbf{Ref-AVS}) and omnimodal audio-visual segmentation (\textbf{OmniAVS}).

\textit{Ref-AVS}~\citep{wang2024ref} segments sounding objects conditioned on textual referring expressions, enabling finer disambiguation than standard AVS. EEMC~\citep{wang2024ref} provides a baseline with modality-specific encoders and a query-based decoder; later methods improve this setting via reinforcement learning (Omni-R1~\citep{zhong2025omni}), SAM-based prompting (SAM2-LOVE~\citep{wang2025sam2}; TSAM~\citep{radman2025tsam}), and agentic reasoning (TGS-Agent~\citep{zhou2025think}).

\textit{OmniAVS}~\citep{OmniAVS} further generalizes Ref-AVS to multimodal referring expressions composed of text, speech, sound, and visual cues, with OISA~\citep{OmniAVS} as a baseline combining an MLLM with a flexible mask head.

These trends indicate a shift from task-specific audio-visual fusion to reasoning-centric segmentation, where MLLMs leverage world knowledge and multi-step reasoning to handle open-vocabulary sounds, abstract references, and complex multimodal scenes.

\paratitle{Benchmarks.}
As summarized in \cref{tab:bench_av_seg}, AVS benchmarks have evolved toward more challenging settings. AVSBench-Object~\citep{zhou2022audio} introduces frame-level masks for sounding objects, and AVSBench-Semantic~\citep{zhou2025audio} further adds class labels. VPO~\citep{chen2024unraveling} studies visually diverse multi-source scenes, AVSBench-Robust~\citep{li2025audio} evaluates robustness to distractors and off-screen sounds, and AVISeg~\citep{guo2025audio} targets audio-visual instance segmentation in long videos. Beyond audio-only queries, Ref-AVS~\citep{wang2024ref} and OmniAVS~\citep{OmniAVS} extend benchmarking to text-guided and omnimodal referring settings.

\begin{insightbox}[\faLightbulbO~Trend: Segmentation Becomes Compositional]
Audio-visual segmentation is shifting from low-level fusion to \emph{audio-driven multimodal reasoning} built on large foundation models. Future advances will focus on robustness in complex scenarios, \emph{open-vocabulary and compositional segmentation} with flexible queries, and reduced reliance on dense annotations via self-supervised or generative learning. Integrating AVS with interactive or embodied systems further opens new opportunities for real-time grounding in practical applications.
\end{insightbox}

\begin{table*}[!t]
\centering
\caption{Common benchmarks for audio-visual segmentation.
}
\label{tab:bench_av_seg}
\benchmarktablestyle
\begin{fullwidthtabular}{10}{l c c c c}
\toprule
\benchhead \tblhead{Dataset} & \tblhead{Task}& \tblhead{Videos} & \tblhead{Duration} & \tblhead{Classes} \\
\midrule
AVSBench-Object~\citep{zhou2022audio}
& Obj-AVS
& 5,356
& 5.0s
& 23
\\

AVSBench-Semantic~\citep{zhou2025audio}
& Sem-AVS
& 12,356
& 7.8s
& 70
\\

VPO~\citep{chen2024unraveling}
& Obj\&Sem-AVS
& 25,057
& 1 frame
& 21
\\

AVSBench-Robust~\citep{li2025audio}
& Obj-AVS
& 5,356
& 5.0s
& 23
\\

AVISeg~\citep{guo2025audio}
& Ins-AVS
& 926
& 61.4s
& 26
\\

Ref-AVS~\citep{wang2024ref}
& Ref-AVS
& 4,002
& 10.0s
& 51
\\

OmniAVS~\citep{OmniAVS}
& Omni-AVS
& 2,104
& 10.6s
& -
\\

\bottomrule
\end{fullwidthtabular}
\end{table*}

\subsubsection{Audio-Visual Synchronization}
\label{sec:und:pixel:avsync_det}

\paratitle{Definition.}
Audio-visual synchronization (AV-Sync) studies whether audio and visual streams in a video are temporally aligned and, if not, estimates their relative offset. Existing formulations mainly include three settings: \textit{binary synchronization detection}, predicting whether the two modalities are aligned~\citep{ebenezer2021detectionaudiovideosynchronization,chen2021audiovisualsynchronisation}; \textit{temporal offset estimation}, regressing the time shift directly~\citep{raina2022syncnet,park2024interpretableSyncNet}; and \textit{offset classification}, predicting a predefined offset interval~\citep{voas2025temporallyRealSync}.

\paratitle{Representative Tasks and Methods.}
Early AV synchronization work mainly focused on lip-speech alignment, exploiting the strong correlation between speech and mouth motion~\citep{halperin2019dynamictemporallips,kadandale2022vocalist,kim2021endtoendlipsynchronisation}. This lip-centric paradigm also supported related tasks such as lip-reading~\citep{son2017lip}, lip-shape generation~\citep{prajwal2020lip}, and active speaker detection~\citep{hershey1999audiovision,gao2021visualvoice}.

Later work extends synchronization to general audio-visual alignment. Most methods adopt dual-encoder architectures with CNNs~\citep{chung2016outoftime,park2024interpretableSyncNet,chen2021audiovisualsynchronisation,ebenezer2021detectionaudiovideosynchronization} or Transformers~\citep{fernandez2024divas}, and fuse modality-specific features via attention, encoder-decoder structures, or fully cross-modal Transformers~\citep{khosravan2019attentionAudioVisual,kadandale2022vocalist,chen2021audiovisualsynchronisation}.

For training objectives, contrastive learning has become dominant~\citep{chung2016outoftime,fernandez2024divas}, using synchronized pairs as positives and shifted or cross-video pairs as negatives. Representative formulations include multi-way offset classification in Perfect Match~\citep{chung2019perfectmatch}, InfoNCE in AVST~\citep{chen2021audiovisualsynchronisation}, BCE in VocaLiST~\citep{kadandale2022vocalist}, and improved negative sampling in ModEFormer~\citep{gupta2023modeformer} and later balanced-BCE designs~\citep{park2024interpretableSyncNet}. 

Recent work increasingly formulates AV synchronization as offset classification, which often yields more stable training and better matches perceptual tolerance to small misalignment~\citep{voas2025temporallyRealSync}.

\begin{table}[!ht]
\centering
\caption{Statistics for common datasets in AV Sync. $N_{\text{clips}}$ is the total number of clips within the dataset; $T_{\text{clips}}$ is the average duration of each clip; $T_{\text{total}}$ is the aggregate duration of the dataset; AVC indicates whether the audio and video components correspond. $\Delta$ indicates that the sound source is visually discernible in the video, though synchronization is not assured.}
\label{tab:av_sync_datasets}
\benchmarktablestyle
\begin{fullwidthtabular}{12}{l c c c c c}
\toprule
\benchhead \tblhead{Dataset} & \tblhead{$\text{N}_{\text{clips}}$} & \tblhead{$\text{T}_{\text{clips}}$} & \tblhead{$\text{T}_{\text{total}}$} & \tblhead{Domain} & \tblhead{AVC} \\
\midrule
{AudioSet}~\citep{gemmeke2017audioset} & 2.1M & 10s & 243d & General & \crossmark \\
{AVE}~\citep{tian2018audio} & 4.1K & 10s & 11.5h & General & \checkmark \\
{TennisED}~\citep{ebenezer2021detectionaudiovideosynchronization} & 4 & 1.5h & 6h & Sports & $\Delta$ \\
{VGGSound}~\citep{chen2020vggsound} & 200K & 10s & 550h & General & $\Delta$ \\
{VGGSync}~\citep{chen2021audiovisualsynchronisation} & 100K & 10s & 275h & General & \checkmark \\
{LRS2}~\citep{son2017lip} & 118K & 6.8s & 224h & News & \checkmark \\
{LRS3}~\citep{afouras2018lrs3} & 74.5k & 22.9s & 474h & TED Talks & \checkmark \\
{RealSync}~\citep{voas2025temporallyRealSync} & 11.2K & 5m & 927h & Sports/News & $\Delta$ \\
\bottomrule
\end{fullwidthtabular}
\end{table}

\paratitle{Benchmarks.}
As summarized in \cref{tab:av_sync_datasets}, large-scale datasets such as AudioSet~\citep{gemmeke2017audioset} and VGGSound~\citep{chen2020vggsound} provide diverse audio-visual events and are widely used for representation learning. In contrast, synchronization-focused datasets, including AVE~\citep{tian2018audio}, VGGSync~\citep{chen2021audiovisualsynchronisation}, and the LRS series~\citep{son2017lip,afouras2018lrs3}, enforce stronger correspondence, often in speech-driven settings with clear lip-motion cues. More recent datasets such as RealSync~\citep{voas2025temporallyRealSync} target realistic broadcast and streaming scenarios for practical synchronization evaluation.

\begin{insightbox}[\faLightbulbO~Trend: Sync Moves Beyond Lips]
Audio-visual synchronization has progressed from lip-centric alignment to general cross-modal temporal correspondence learning. Recent work favors more scalable and perceptually grounded formulations, such as offset classification and contrastive pretraining. Future directions include improving robustness in real-world settings, modeling finer temporal dynamics, leveraging foundation models, and integrating AV-Sync into broader multimodal systems for more reliable temporal reasoning.
\end{insightbox}

\subsection{Audio-Visual Content Understanding}
\label{sec:und:cont}

Once low-level structure is in place, models must map signals to \emph{meaning}: objects, events, questions, and retrieval targets stated in language or other structured form.
We cover unimodal audio and visual content understanding, then the central audio-visual problems of question answering and cross-modal retrieval, where the emphasis shifts from per-modality recognition to shared semantics and spatio-temporal evidence.

\subsubsection{Audio Understanding }
\label{sec:und:cont:audio}

\paratitle{Definition.}
Audio content understanding aims to extract semantic information from audio signals, including speaker traits, emotional states, sound-event descriptions, and question-relevant acoustic evidence.
Unlike audio perception, which emphasizes signal-level detection and localization, audio understanding maps acoustic patterns to linguistic or semantic interpretations; explicit multi-step inference is discussed separately in \cref{sec:und:reason:audio}.

\paratitle{Representative Tasks and Methods.}
Representative tasks include speaker recognition, emotion analysis, audio captioning, and audio question answering, which capture different levels of semantic interpretation:

\textit{Speaker recognition}
identifies or verifies speaker identity from speech signals.
Early systems relied on i-vectors with probabilistic linear discriminant analysis, while modern methods learn neural speaker embeddings.
Representative approaches include d-vector~\citep{variani2014deep}, x-vector~\citep{snyder2018xvector}, ECAPA-TDNN~\citep{desplanques2020ecapa}, and self-supervised speech representations such as WavLM~\citep{chen2022wavlm}.

\textit{Emotion and paralinguistic analysis}
infers affective states and non-verbal cues from speech or general audio.
Traditional systems relied on handcrafted acoustic features with classical classifiers, while modern approaches learn representations directly from spectrograms or raw waveforms using deep neural networks~\citep{pepino2021emotion}.
Self-supervised speech models, including Wav2Vec~2.0~\citep{baevski2020wav2vec} and Emotion2Vec~\citep{ma2024emotion2vec}, have significantly improved robustness and transferability.
Multimodal fusion with text or visual signals further enhances performance in affective computing settings~\citep{tsai2019multimodal}.

\textit{Audio captioning}
generates natural language descriptions of general audio content.
Most approaches adopt encoder--decoder architectures that combine pretrained audio encoders such as PANNs~\citep{kong2020panns} and HTSAT~\citep{chen2022htsat} with sequence decoders~\citep{mei2021audiocaptioning}.
Recent work explores retrieval-augmented generation~\citep{kim2024retrieval} and large audio-language models such as SALMONN~\citep{tang2023salmonn}, Audio Flamingo~\citep{kong2024audioflamingo}, and Qwen-Audio~\citep{chu2023qwenaudio}.

\textit{Audio question answering (AQA)}
requires models to answer natural language questions grounded in audio content, often combining event recognition with temporal comparison or counting.
Early methods used joint audio--text encoders with fusion modules~\citep{fayek2020temporal}, while recent work leverages Audio LLMs~\citep{gong2024audiollm,chu2024qwen2audio}.
Training strategies such as curriculum learning~\citep{wang2024curriculum} have been explored to improve question-conditioned acoustic grounding.

\subsubsection{Visual Understanding }
\label{sec:und:cont:visual}

\paratitle{Definition.}
Visual content understanding extracts semantic information from images and videos, including objects, attributes, actions, and spatial relations, to form coherent interpretations of scenes and events in structured~\citep{ms-coco-2014} or textual~\citep{antol2015vqa} formats. 

\paratitle{Representative Tasks and Methods.}
Researchers addressed this goal through several typical tasks:

\textit{Visual Recognition,}
which assigns semantic labels to visual inputs, including image classification, object detection, and segmentation.
Classical supervised systems relied on convolutional or transformer architectures such as ResNet \citep{he2016deep} and Vision Transformer (ViT) \citep{dosovitskiy2020image}.
Large-scale representation learning further enabled open-vocabulary recognition through vision-language alignment, exemplified by CLIP \citep{radford2021learning}.
Subsequent multimodal pretraining models, including BLIP \citep{li2023blip}, CoCa \citep{yu2022coca}, and PaLI \citep{chen2022pali}, as well as instruction-tuned MLLMs such as LLaVA \citep{liu2023visual}, Qwen2.5/3-VL \citep{bai2025qwen2,bai2025qwen3}, and InternVL~3.5 \citep{wang2025internvl3}, integrate recognition with multimodal reasoning, and recognition is often done in the form of QA question answering.

\textit{Visual Grounding,}
which links language expressions to image regions or video segments.
Early referring expression models such as MAttNet \citep{yu2018mattnet} relied on region proposals and modular reasoning \citep{mao2016generation}.
Transformer-based models later unified detection and grounding, including MDETR \citep{kamath2021mdetr}, GLIP \citep{li2022grounded}, and GroundingDINO \citep{liu2024grounding}.
Recent multimodal LLMs incorporate grounding capabilities into instruction-following frameworks, including MiniGPT-v2 \citep{chen2023minigpt}, CogVLM \citep{wang2024cogvlm}, and Qwen2.5-VL \citep{bai2025qwen2}.
Video grounding introduces temporal localization challenges, studied by Moment-DETR \citep{lei2021detecting}, Vid2Seq \citep{yang2023vid2seq}, and time-aware VideoLLMs such as TimeChat \citep{ren2024timechat}, VTimeLLM \citep{huang2024vtimellm}, TimeSuite \citep{zeng2024timesuite}, Chrono \cite{meinardus2024chrono}, Time-R1 \citep{liu2025time}, and VideoChat-R1/1.5 \citep{li2025videochat,yan2025videochat}, though robustness issues remain under temporal perturbations or view variations \citep{Jung_2025_CVPR,jung2025egoexoconexploringviewinvariantvideo}.

\textit{Visual Captioning,}
which generates natural language descriptions for images or videos.
Early CNN-RNN architectures trained on MS-COCO \citep{ms-coco-2014} used attention and region features to improve grounding \citep{xu2015show,anderson2018bottom}.
Large-scale vision-language pretraining later enabled stronger caption generation, including SimVLM \citep{wang2021simvlm}, CoCa \citep{yu2022coca}, BLIP \citep{li2022blip}, and BLIP-2 \citep{li2023blip}.
Recent multimodal large language models such as LLaVA series \citep{li2024llava,an2025llava,zhang2024llava}, Qwen-VL series \citep{bai2025qwen2,bai2025qwen3} and InternVL series \citep{chen2024internvl,wang2025internvl3} treat captioning as instruction-following multimodal generation, though hallucination and evaluation limitations remain \citep{bai2024hallucination,vedantam2015cider,zhao2025sycophancy}.

\textit{Visual Question Answering (VQA),}
which requires answering natural language questions grounded in visual content and is widely used to evaluate vision-language reasoning.
Early systems relied on convolutional encoders and attention-based reasoning modules \citep{zhong2022video}.
Modern approaches adopt the multimodal LLM paradigm, coupling visual encoders \citep{radford2021learning} with language models \citep{touvron2023llama}, such as Flamingo \citep{alayrac2022flamingo}, BLIP-2 \citep{li2023blip}, and LLaVA \citep{liu2023visual,an2025llava}.
Video QA introduces additional temporal reasoning challenges, addressed by systems such as Video-ChatGPT \citep{maaz2024video}, VideoChat \citep{li2024mvbench}, LLaVA-Video \citep{zhang2024llava}, and Qwen \citep{qwen35blog}. However, 
However, current systems still struggle with temporal reasoning, long-context memory, egocentric perpection, and answer consistency \citep{di2025streaming,yang2025streammem,xiao2025egoblind,zhou2025egotextvqa,xiao2024can,xiao2025videoqa,Jung_2025_CVPR}.

\begin{insightbox}[\faLightbulbO~Trend: Vision Needs Longer Memory]
Advanced MLLMs have demonstrated surprisingly good performace for general-purpose object and event understanding in image and videos. However, their capacities for spatial intelligence, temporal dynamics, long-term context modeling, and reliability remain weak. 
\end{insightbox}

\subsubsection{Audio-Visual Question Answering}
\label{sec:und:cont:avqa}

\paratitle{Definition.}
Audio-Visual Question Answering (AVQA) \citep{li2022learning,yang2022avqa} aims to answer natural-language questions by jointly understanding over visual content, audio signals, and language, requiring models to associate sounds with visual events and temporal dynamics. Compared with visual-only VQA \citep{antol2015vqa,xiao2021next}, AVQA introduces modality-specific challenges such as sound source localization, cross-modal temporal alignment \citep{alamri2018avsd}, and audio-visual causal comprehension between what is seen and heard.

\paratitle{Methods.}
Recent AVQA methods can be roughly organized into three lines:

The first line improves question-conditioned clue extraction and bias control before final answering: TSPM~\citep{li2024boosting} converts questions into declarative prompts to guide temporal-spatial perception, MCD~\citep{ye2024mcd} attaches key audio-visual clues to textual queries through mutual correlation distillation, QA-TIGER~\citep{kim2025question} models continuous question-aware temporal dynamics with Gaussian experts, RAVEN~\citep{biswas2025raven} performs query-guided cross-modal representation alignment to suppress distractors, and AV-Master~\citep{zhang2025av} combines adaptive sampling with modality-preference modeling for complex audio-visual scenes. 

The second line moves AVQA from task-specific classifiers to generative AV-LLM backbones. Audio-Visual LLM~\citep{Shu_2025_ICCV} introduces a unified audio-visual representation for language-centric video understanding, CAT~\citep{ye2024cat} strengthens fine-grained grounding via cross-modal alignment transformers, VideoLLaMA 2~\citep{cheng2024videollama} enhances joint spatial-temporal and audio modeling in video LLMs, video-SALMONN~\citep{shu2024videosalmonn} injects speech-aware audio perception into video-language modeling, and Meerkat~\citep{chowdhury2024meerkat} emphasizes space-time grounding for audio-visual instruction following and QA. 

The third line is formed by industrial omni models that explicitly support joint audio-video input and therefore serve as increasingly relevant general-purpose AVQA baselines. GPT-4o~\citep{hurst2024gpt} and Gemini 2.5 Pro~\citep{comanici2025gemini} represent proprietary large-scale omni systems, while Qwen3-Omni~\citep{xu2025qwen3omni}, Baichuan-Omni-1.5~\citep{li2025baichuanomni15}, Ming-Omni~\citep{ai2025ming}, Ming-Flash-Omni~\citep{ma2025ming}, and LongCat-Flash-Omni~\citep{wang2025longcat} extend this trend to open or domestic industrial ecosystems.

\paratitle{Benchmarks.}
As shown in \cref{tab:avqa_datasets}, benchmark development has similarly evolved from closed-set short-video QA toward robustness, openness, and audio indispensability: representative datasets now include AVQA~\citep{yang2022avqa} and MUSIC-AVQA~\citep{li2022learning} for foundational evaluation, FortisAVQA~\citep{ma2025fortisavqa} for debiasing and robustness, AVQACL~\citep{wu2025avqacl} for continual learning, Valor32k-AVQA v2.0~\citep{riahi2025valor32k} for large-scale open-ended assessment, and DAVE~\citep{radevski2025dave}, AVUT~\citep{yang2025avut}, and AVHBench~\citep{kim2024avhbench} for diagnosing genuine audio necessity, shortcut resistance, and hallucination in modern AV-LLMs.

\begin{table}[t]
\centering
\caption{Comparison of representative AVQA benchmarks. Ans. uses three categories: multiple-choice (MC), open-ended (OE), and mixed formats.
}
\label{tab:avqa_datasets}
\benchmarktablestyle
\begin{fullwidthtabular}{12}{l c c c c c}
\toprule
\benchhead \tblhead{Dataset} & \tblhead{\# Vid} & \tblhead{\# QAs} & \tblhead{Ans.} & \tblhead{Len.} & \tblhead{Domain} \\
\midrule
AVQA~\citep{yang2022avqa} & 17K & 17K & MC & 10s & Real-life \\
MUSIC-AVQA~\citep{li2022learning} & 9.2K & 9.2K & MC & 60s & Music \\
AVQACL~\citep{wu2025avqacl} & 11K & 11K & MC & 10s/60s & Continual learning \\
FortisAVQA~\citep{ma2025fortisavqa} & 211.6K & 211.6K & MC & 60s & Robustness \\
Valor32k-AVQA v2.0~\citep{riahi2025valor32k} & 3.5K & 26.1K & Mixed & 10s & Open domain \\
DAVE~\citep{radevski2025dave} & 2.4K & 2.4K & MC & $\leq$60s & Egocentric \\
AVUT~\citep{yang2025avut} & 2.7K & 11.6K & Mixed & 67.8s & Audio-centric \\
AVHBench~\citep{kim2024avhbench} & 2.1K & 5.3K & Mixed & - & Hallucination \\
\bottomrule
\end{fullwidthtabular}
\end{table}

\begin{table}[t]
\centering
\caption{Open-ended audio-visual QA on MUSIC-AVQA, AVSD, and VGGSound, using the OE-AVQA setting and GPT-based scoring for AVSD and VGGSound as in \citet{cheng2024videollama} (Table~8). MUSIC-AVQA and AVSD baselines are zero-shot; VGGSound follow their protocol (first three models zero-shot, remaining in-domain). N/A indicates results not reported in that table. FastAV \citep{jung2026fastav} reports MUSIC-AVQA for pruned VideoLLaMA2.1-AV; AVSD and VGGSound are not reported there.}
\label{tab:avqa_comparison}
\performancetablestyle
\begin{fullwidthtabular}{10}{l c c c c}
\toprule
\perfhead \tblhead{Method} & \tblhead{Size} & \tblhead{MUSIC-AVQA} & \tblhead{AVSD} & \tblhead{VGGSound} \\
\midrule
PandaGPT \citep{su2023pandagpt}      & 13B & 33.7 & 26.1 & 32.7 \\
Macaw-LLM \citep{lyu2023macaw}     & 7B & 31.8 & 34.3 & 36.1 \\
VideoLLaMA \citep{zhang2023video}     & 7B & 36.6 & 36.7 & 40.8 \\
X-InstructBLIP \citep{panagopoulou2023x} & 13B & 44.5 & N/A & N/A \\
AV-LLM \citep{Shu_2025_ICCV}        & 13B & 45.2 & 52.6 & 47.6 \\
OneLLM \citep{han2024onellm}        & 7B & 47.6 & N/A & N/A \\
AVicuna \citep{tang2025empowering}        & 7B & 49.6 & 53.1 & N/A \\
CREMA \citep{yu2024crema}        & 4B & 52.6 & N/A & N/A \\
VideoLLaMA2-AV \citep{cheng2024videollama}   & 7B & 79.2 & \bf{57.2} & 70.9 \\
VideoLLaMA2.1-AV \citep{cheng2024videollama}   & 7B & 80.9 & \bf{57.2} & \bf{71.4} \\
FastAV \citep{jung2026fastav}   & 7B & \bf{81.2} & N/A & N/A \\
\bottomrule
\end{fullwidthtabular}
\end{table}

\begin{insightbox}[\faLightbulbO~Trend: Answers Become Open-Ended]
AVQA is shifting from task-specific fusion models to MLLM-based, language-centric reasoning, where fine-grained cross-modal alignment and evidence grounding have become central. Recent work also moves beyond raw accuracy toward explicit comprehension, modality-aware routing, bias mitigation, and efficient inference, suggesting that the field is increasingly concerned with whether models truly reason over audio-visual evidence.
\end{insightbox}

\subsubsection{Audio-Visual Cross-Modal Retrieval}
\label{sec:und:cont:av_retrieve}

\paratitle{Definition.}
Audio-visual cross-modal retrieval retrieves semantically corresponding samples across audio and visual modalities, such as videos from audio queries or audio from visual scenes. Its core challenge is learning shared representations over heterogeneous signals while handling temporal and spatial misalignment. Recent work has progressed from coarse clip-level matching to finer-grained retrieval.

\paratitle{Methods.}
Unlike unimodal retrieval, audio-visual retrieval requires learning a shared embedding space where heterogeneous modalities can be meaningfully compared. Early representation learning approaches relied on \emph{knowledge distillation} from visual networks to audio models \citep{aytar2016soundnet,owens2016ambient}. Later work adopted \emph{paired sample discrimination} to distinguish matched and mismatched audio-video pairs \citep{arandjelovic2017look,arandjelovic2018objects,korbar2018cooperative,owens2018audio}. Modern methods formulate it as \emph{contrastive learning} \citep{rouditchenko2020avlnet,sun2023learning}, improving robustness through hard-negative mining, data augmentation, and multi-view objectives \citep{morgado2021audio,morgado2021robust,recasens2021broaden,wang2021multimodalaudio,zeng2021contrastivegloballocal}, while newer studies address asynchronous audio-visual relations and equivariant constraints \citep{sarkar2023self,kim2024equiav}.

In parallel, \emph{masked modeling} approaches learn representations by reconstructing masked inputs or latent features \citep{IshikawaNMSKA25}. Models such as CAV-MAE \citep{gong2022CAVMAE}, AV-MAE \citep{georgescu2023AVMAE}, MAViL \citep{huang2023mavil}, and CrossMAE \citep{guo2024crossmae} combine reconstruction with contrastive objectives to capture cross-modal semantics. Extensions including PCAV-MAE \citep{yi2023pcavmae}, AVSiam \citep{lin2024avsiam}, VAB \citep{SuLS24VAB}, and CAV-MAE Sync \citep{araujo2025cavmaesync} further improve alignment, efficiency, and spatial localization.

Beyond pairwise alignment, recent work explores unified multimodal embeddings. ImageBind \citep{girdhar2023imagebind} aligns multiple modalities using images as a semantic anchor, while LanguageBind \citep{zhu2023languagebind} maps modalities into a shared language space. DenseAV \citep{mark2024denseav} further demonstrates that aligning dense audio-visual features with DINO \citep{CaronTMJMBJ21dino} and HuBERT \citep{hsu2021hubert} significantly improves fine-grained localization and retrieval.

\paratitle{Benchmarks.}
Audio-visual cross-modal retrieval is commonly evaluated on large-scale datasets with paired or weakly aligned audio-video signals. AudioSet \citep{gemmeke2017audioset} is widely used for its scale and diversity, while VGGSound \citep{chen2020vggsound} offers cleaner correspondence and balanced categories. Video datasets such as MSR-VTT \citep{xu2016msrvtt} and YouCook2 \citep{zhou2018YouCook2} are also adapted for retrieval with audio tracks, and HowTo100M \citep{miech2019howto100m} supports large-scale weakly aligned pretraining. More recent datasets, including MUGEN \citep{hayes2022mugen} and AVSET-10M \citep{cheng2025vset10m}, further reflect the trend toward larger and more structured multimodal benchmarks.

\begin{table*}[!t]
\centering
\caption{Common benchmarks for audio-visual cross-modal retrieval.
\textbf{Scale} reports the approximate number of video clips.
\textbf{Alignment} indicates whether audio and visual signals are explicitly synchronized.}
\label{tab:av_retrieval_datasets}
\benchmarktablestyle
\begin{fullwidthtabular}{10}{l c c c c}
\toprule
\benchhead \tblhead{Dataset} & \tblhead{Scale} & \tblhead{Duration} & \tblhead{Alignment} & \tblhead{Modalities} \\
\midrule
AudioSet~\citep{gemmeke2017audioset}
& $\sim$2M clips
& 10s
& Weak
& \audtag{Audio} \audtag{Video} \\

VGGSound~\citep{chen2020vggsound}
& $\sim$200K clips
& 10s
& Strong
& \audtag{Audio} \audtag{Video} \\

MSR-VTT~\citep{xu2016msrvtt}
& 10K videos
& $\sim$20s
& Weak
& \audtag{Video} \audtag{Audio} \audtag{Text} \\

YouCook2~\citep{zhou2018YouCook2}
& $\sim$2K videos
& Long-form
& Weak
& \audtag{Video} \audtag{Audio} \audtag{Text} \\

HowTo100M~\citep{miech2019howto100m}
& 136M clips
& Variable
& Weak
& \audtag{Video} \audtag{Audio} \audtag{Text} \\

MUGEN~\citep{hayes2022mugen}
& $\sim$2M clips
& Short
& Strong
& \audtag{Video} \audtag{Audio} \audtag{Text} \\

AVSET-10M~\citep{cheng2025vset10m}
& 10M clips
& Short
& Strong
& \audtag{Audio} \audtag{Video} \\
\bottomrule
\end{fullwidthtabular}
\end{table*}



\begin{insightbox}[\faLightbulbO~Trend: Retrieval Gets Finer]
Audio-visual cross-modal retrieval is moving from pairwise alignment to scalable multimodal representation learning. Recent work shows that contrastive and masked modeling, combined with large-scale pretraining, effectively improve alignment and generalization. The field is also shifting toward finer and temporally grounded retrieval. Future directions include handling weakly aligned signals, denser correspondence, and more open-ended retrieval with foundation models.
\end{insightbox}

\subsection{Audio-Visual Logical Reasoning}
\label{sec:und:reason}

Reasoning subsections require models to combine evidence, follow constraints, and justify conclusions rather than only label inputs.
We address audio-only and vision-only settings first, then audio-visual reasoning, where sound and image streams jointly support multi-step and causal inferences.

\subsubsection{Audio Reasoning }
\label{sec:und:reason:audio}

\paratitle{Definition.}
Audio reasoning goes beyond basic audio understanding by requiring multi-step inference, world knowledge, and causal deduction from acoustic signals. Unlike recognition tasks that identify “what” is present, audio reasoning addresses “why”, “how”, and “what if” questions about events and their relationships. This capability is important for applications requiring situational awareness, including autonomous driving, medical diagnosis from acoustic signals, and intelligent assistants~\citep{ma2025mmar}.

\paratitle{Methods.}
Recent audio reasoning methods adapt LLM-style inference to acoustic evidence, but the area remains less mature than text or vision reasoning:

\textit{Prompt Engineering.}
Borrowing from text-LLM chain-of-thought prompting~\citep{wei2022chain}, recent work instructs Audio LLMs to reason over acoustic evidence before answering.
Audio-specific prompting strategies often decompose questions into perception (``What sounds are present?'') and inference (``Given these sounds, what is happening?'') stages, which helps reduce shortcut answers on complex questions.

\textit{Instruction tuning and supervised fine-tuning.}
Training Audio LLMs on instruction or reasoning-annotated data teaches models to connect acoustic events with textual explanations.
GAMA~\citep{ghosh2024gama} demonstrates that fine-tuning on audio instruction data with diverse reasoning requirements improves generalization to unseen reasoning tasks.
Multi-task training in systems such as SALMONN~\citep{tang2023salmonn} and Audio Flamingo~2~\citep{ghosh2025audioflamingo2} further suggests that broad audio-language supervision can elicit stronger temporal and commonsense reasoning.

\textit{Reasoning-oriented test-time behavior.}
Compared with audio-visual reasoning, audio-only reinforcement learning and test-time scaling are still emerging.
Current systems more commonly rely on prompting, instruction tuning, and benchmark-driven diagnosis, while RL-style post-training is most clearly developed in audio-visual reasoning settings discussed in \cref{sec:und:reason:av}.

\paratitle{Benchmarks.}
Several benchmarks evaluate audio reasoning across perception and inference. CLEAR~\citep{abdelnour2018clear} targets compositional acoustic reasoning, MMAR~\citep{ma2025mmar} spans speech, music, environmental sounds, and mixtures, and CMDAR~\citep{wang2024cmdar} focuses on multi-step reasoning such as temporal ordering, causal inference, and domain knowledge.
Results show that current Audio LLMs perform better on perception than on complex temporal and causal reasoning~\citep{ghosh2025audioflamingo2}.

\begin{insightbox}[\faLightbulbO~Trend: Audio Learns to Reason]
Audio reasoning is emerging from audio perception through the adaptation of LLM paradigms to acoustic signals. Recent progress shows that prompting, instruction tuning, and reasoning-oriented training can elicit multi-step inference, but current models still struggle with complex temporal and causal relationships. Future work will likely focus on improving reasoning faithfulness and consistency, strengthening temporal-causal modeling, and developing better training signals and benchmarks.
\end{insightbox}

\subsubsection{Visual Reasoning}
\label{sec:und:reason:visual}

\paratitle{Definition.}
Visual reasoning refers to a model’s ability to perform multi-step, compositional, and often abstract inference grounded in visual inputs \citep{johnson2017clevr,girdharcater,yi2019clevrer,0001W0ZZLH24}. Unlike recognition-centric tasks, visual reasoning emphasizes how conclusions are derived from visual evidence, involving capabilities such as relational understanding \citep{hudson2019gqa, ji2020action, xiao2021next}, counting \citep{jang2017tgif}, causal inference \citep{xiao2021next,li2022invariant,niu2021counterfactual}, and commonsense reasoning \citep{zellers2019recognition,wang2025multimodal}. 

\paratitle{Methods.}
Early visual reasoning models were largely task-specific and symbolic or neuro-symbolic \citep{yi2018neural}. Benchmarks such as CLEVR \citep{johnson2017clevr} and GQA \citep{hudson2019gqa} inspired methods that modeled reasoning via functional programs \citep{subramanian2023modular,suris2023vipergpt}, scene graphs \citep{shi2019explainable,li2019relation,hildebrandt2020scene}, and modular networks \citep{andreas2016neural,chen2021meta}. Although effective in controlled settings, their reliance on synthetic data and predefined reasoning primitives limited real-world generalization.

Vision-language foundation models shifted this paradigm by leveraging LLM reasoning. Systems such as Flamingo and BLIP-2 showed that pairing visual encoders with frozen LLMs can yield emergent visual reasoning, while instruction-tuned MLLMs (including the LLaVA series \citep{liu2023visual,liu2024improved,li2024llava,an2025llava}, Qwen-VL series \citep{bai2023qwenvlversatilevisionlanguagemodel,bai2025qwen2}, and InternVL series \citep{chen2024internvl,wang2025internvl3}) demonstrate strong reasoning via prompting and chain-of-thought supervision \citep{han2025videoespresso,qin2025chain,sun2025visual}. Architectures such as QvQ \citep{qvq-72b-preview} further highlight how model design and reasoning-oriented training shape visual reasoning.

To improve robustness beyond supervised reasoning signals, recent work explores reinforcement learning (RL). Methods such as Reason-RFT \citep{tan2025reason}, Point-RFT \citep{ni2025point}, Visionary-R1 \citep{xia2025visionary}, and VTool-R1 \citep{wu2025vtool} apply RL-based fine-tuning to strengthen grounding, reasoning consistency, and tool-assisted multimodal reasoning, achieving improved performance on complex tasks including ChartXiv \citep{wang2024charxiv}.
Visionary-R1 \citep{xia2025visionary} trains VLMs with RL to encourage models to first interpret images and then reason, mitigating shortcut paths and improving performance against strong multimodal baselines such as GPT-4o \citep{hurst2024gpt} and Claude3.5-Sonnet \citep{claude35}.


Video reasoning introduces additional challenges due to temporal structure and event dependencies. Recent RL-based methods encourage temporally grounded reasoning, including Video-R1 \citep{feng2025video} with T-GRPO \citep{guo2025deepseek}, VideoRFT \citep{wang2025videorft}, VideoChat-R1/R1.5 \citep{li2025videochat,yan2025videochat}, and Time-R1 \citep{wang2025time}. These approaches optimize temporal-aware rewards to align reasoning with relevant video moments and improve long-video spatio-temporal inference.

\begin{frontierbox}[\faLightbulbO~Open Problem: Reasoning Must Stay Grounded]
These trends suggest a shift from step-wise chain-of-thought to more grounded multimodal reasoning with reinforcement-based training. Yet models still suffer from hallucinations, spurious correlations, and unfaithful reasoning. Recent work also frames reasoning as inference over latent chains, emphasizing diversity and uncertainty. Future progress will require tighter perception-reasoning integration, better grounding rewards, and benchmarks for both correctness and faithfulness.
\end{frontierbox}

\paratitle{Benchmarks.}
Several benchmarks evaluate visual reasoning across perception and inference. MMMU \citep{Yue_2024_CVPR} and MathVista \citep{lu2024mathvista} target reasoning over diverse visual formats and mathematical visual contexts, while MATH-V \citep{wang2024measuring}, Visual CoT \citep{shao2024visualcot}, VisuLogic \citep{xu2025visulogic}, MuSLR \citep{xu2025MuSLR} and MMReason \citep{Yao_2025_ICCV} focus on multimodal mathematical reasoning, grounded chain-of-thought reasoning, vision-centric reasoning, and open-ended multi-step reasoning.

\subsubsection{Audio-Visual Reasoning} 
\label{sec:und:reason:av}

\paratitle{Definition.}
Audio-visual reasoning requires models to infer why, when, and how events unfold by jointly interpreting video and audio. Beyond recognition or simple question answering, it emphasizes temporal alignment, causal inference, cross-modal grounding, and faithful rationale generation.

\paratitle{Methods.}
Inspired by OpenAI-o1~\citep{openai2024o1} and DeepSeek-R1~\citep{guo2025deepseek}, recent work starts to optimize reasoning explicitly for audio-video inputs. video-SALMONN-o1~\citep{sun2025video} introduces process-level preference optimization for general audio-visual reasoning, while EchoInk-R1~\citep{xing2025echoink}, Omni-R1~\citep{zhong2025omni}, and AVATAR~\citep{kulkarni2025avatar} apply RL-style post-training to improve grounded reasoning, self-correction, and long-horizon credit assignment. In parallel, ThinkOmni~\citep{guan2026thinkomni} transfers textual slow-thinking ability to omni-modal inputs without extra training.

Another line targets reasoning reliability at inference time. AVCD~\citep{jung2025avcd} suppresses trimodal hallucination through contrastive decoding, while Fork-Merge Decoding~\citep{jung2025fork} explicitly separates early audio-only and video-only reasoning before late fusion, reducing visually dominated shortcuts.

At the current frontier, industrial systems are also becoming explicitly reasoning-centric. As OpenAI-o3~\citep{openai2025o3} extends strong reasoning to visual analysis, and Gemini 2.5/3.0 Pro~\citep{comanici2025gemini} is positioned as a thinking model with native multimodal understanding over audio and video. Qwen3-Omni-Thinking~\citep{xu2025qwen3omni} also explicitly strengthens audio-visual reasoning in an open omni-modal setting. Together, these models indicate a broader transition from multimodal perception to reasoning-centric omni systems.

\paratitle{Benchmarks.}
Recent benchmarks increasingly target audiovisual reasoning rather than general multimodal understanding. OmniBench~\citep{li2024omnibench} and AVQA-R1-6K~\citep{xing2025echoink} capture audio-image reasoning, Daily-Omni~\citep{zhou2025daily} emphasizes temporal alignment, AURA~\citep{galougah2025aura} measures reasoning fidelity, 
JointAVBench~\citep{chao2025jointavbench} emphasizes cross-modal dependency, OmniVideoBench~\citep{li2025omnivideobench} extends evaluation to long-form audio-video understanding, and AV-ConfuseBench~\citep{ye2026eyes} tests resistance to visually plausible but audio-absent distractors.

\begin{table}[t]
\centering
\caption{Comparison of representative audio-visual reasoning benchmarks.
Recent benchmarks move beyond short closed-set QA, and increasingly emphasize temporal alignment, modality complementarity, hallucination diagnosis, and long-form reasoning.}
\label{tab:bench_av_reason}
\benchmarktablestyle
\begin{fullwidthtabular}{12}{l c c c c c}
\toprule
\benchhead \tblhead{Benchmark} & \tblhead{\# Source} & \tblhead{\# QA} & \tblhead{Format} & \tblhead{Duration} & \tblhead{Domain} \\
\midrule
OmniBench~\citep{li2024omnibench}
& 1.1K
& 1.1K
& MCQA
& $\sim$ 9s
& General \\

AVQA-R1-6K~\citep{xing2025echoink}
& 1.9K
& 1.9K
& MCQA
& $\sim$ 9s
& Synthetic \\

Daily-Omni~\citep{zhou2025daily}
& 684
& 1.2K
& MCQA
& 30s / 60s
& Daily \\

JointAVBench~\citep{chao2025jointavbench}
& 1.0K
& 2.9K
& MCQA
& $\sim$ 97s
& Movie \\

OmniVideoBench~\citep{li2025omnivideobench}
& 628
& 1.0K
& QA+rationale
& up to 30 min
& Long-form \\

AV-ConfuseBench~\citep{ye2026eyes}
& 59
& 173
& Mixed
& --
& Confusion \\

\bottomrule
\end{fullwidthtabular}
\end{table}

\begin{table}[t]
\centering
\caption{Performance comparison of omni-modal models on Daily-Omni~\citep{zhou2025daily}.
We report AV Align, Reasoning, and overall average accuracy (Avg, \%).}
\label{tab:avr_dailyomni_omni}
\performancetablestyle
\begin{fullwidthtabular}{10}{l c c c c}
\toprule
\perfhead \tblhead{Method} & \tblhead{Size} & \tblhead{AV Align} & \tblhead{Reasoning} & \tblhead{Avg} \\
\midrule
Unified-IO-2 XL~\citep{lu2024unified} & 3B & 30.25 & 21.71 & 28.32 \\
VideoLLaMA2~\citep{cheng2024videollama} & 7B & 35.71 & 34.29 & 35.17 \\
Ola~\citep{liu2025ola} & 7B & 40.34 & 69.71 & 50.71 \\
Qwen2.5-Omni-3B-Instruct~\citep{xu2025qwen2} & 3B & 50.84 & 70.29 & 60.23 \\
Qwen2.5-Omni-7B-Instruct~\citep{xu2025qwen2} & 7B & 48.32 & 73.14 & 62.07 \\
Qwen3-Omni-30B-A3B-Instruct~\citep{xu2025qwen3omni} & 30B & 66.81 & 81.14 & 71.85 \\
Qwen3-Omni-30B-A3B-Thinking~\citep{xu2025qwen3omni} & 30B & 65.97 & 80.57 & \textbf{73.60} \\
Gemini 2.5 Flash~\citep{comanici2025gemini} & -- & \textbf{73.82} & \textbf{81.87} & 73.06 \\
\bottomrule
\end{fullwidthtabular}
\end{table}

\begin{insightbox}[\faLightbulbO~Trend: Towards Grounded and Faithful Audio-Visual Reasoning.]
Audio-visual reasoning is beginning to move beyond perception-oriented fusion toward MLLM-based reasoning over temporal, causal, and grounded evidence. Recent methods and benchmarks have expanded the scope from short synchronized QA to grounding, hallucination control, and long-video reasoning. However, the field is still at an early stage, as current systems remain limited in faithful cross-modal reasoning, robust temporal understanding, and reliable evaluation.
\end{insightbox}

\section{Audio-Visual Generation}
\label{sec:gen}

\begin{figure*}[!t]
    \centering
    \includegraphics[width=\textwidth]{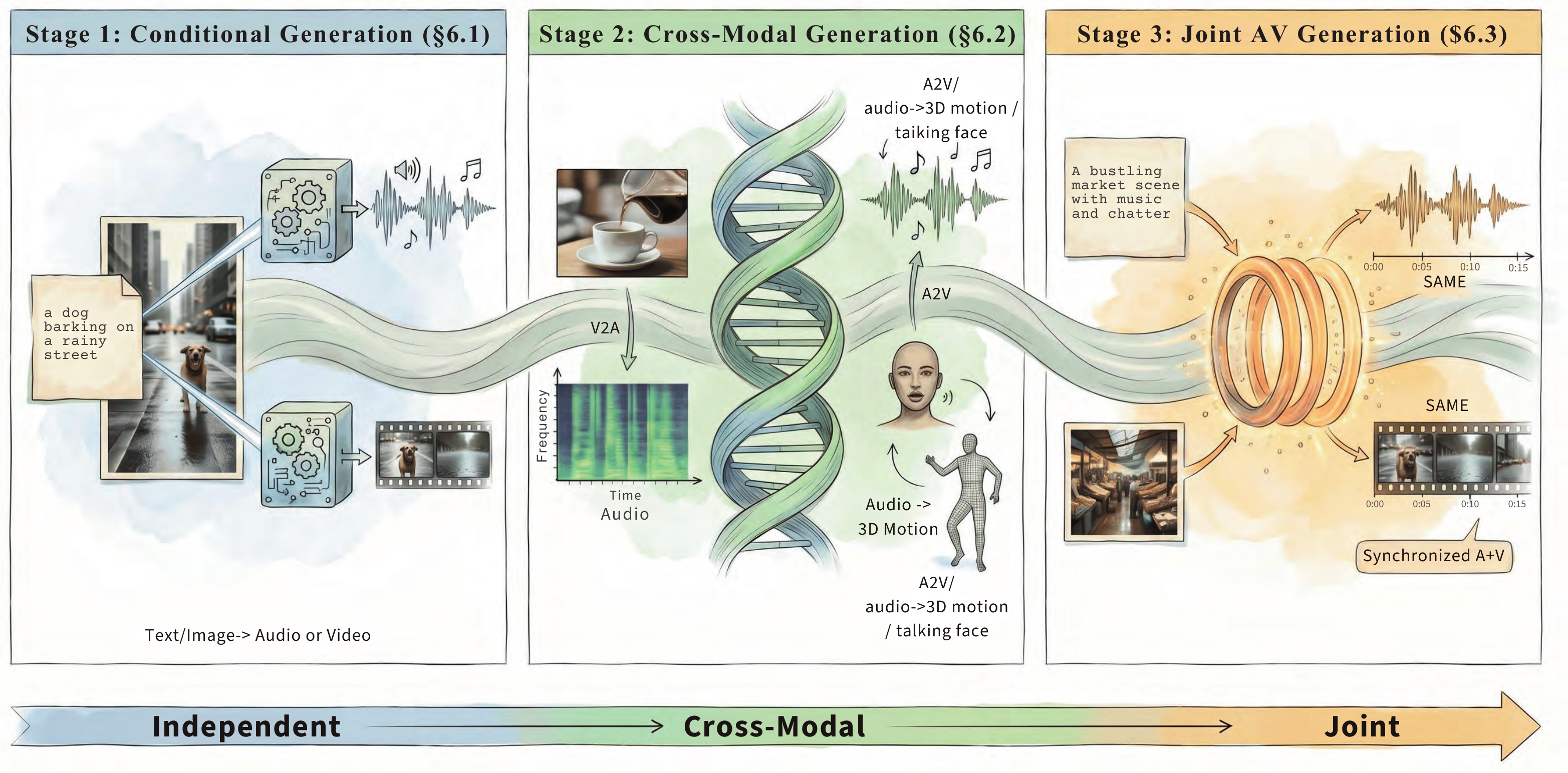}
    \caption{Organization of audio-visual generation tasks. We organize this section into conditional generation, cross-modal generation, and joint audio-visual generation, reflecting the transition from independent modality synthesis to tightly synchronized audio-video creation.}
    \label{fig:gen_overview}
    \vspace{-3mm}
\end{figure*}

Audio-visual generation studies how to synthesize temporally aligned sound and imagery from text, images, video, or audio. We organize this section around three increasingly coupled regimes: conditional unimodal generation, cross-modal translation, and joint audio-visual generation, as summarized in~\cref{fig:gen_overview}. The field has moved from composing strong single-modality generators to explicitly coupled and, more recently, unified architectures, with diffusion, flow matching, and multimodal transformers as the dominant design choices.

\subsection{Conditional Audio/Visual Generation}
\label{sec:gen:cond_av}

This part concerns generating or transforming one modality at a time under external control (text, reference media, or instructions), without yet requiring a full joint audio-video model.
We treat conditional audio and conditional visual generation in separate subsections, both as building blocks for the cross-modal and joint settings that follow.

\subsubsection{Conditional Audio Generation }
\label{sec:gen:cond_av:audio}

\paratitle{Definition.}
Conditional audio generation aims to synthesize or transform audio signals under external conditioning information, such as text descriptions, reference audio, mixture recordings, or editing instructions.

\paratitle{Representative Tasks and Methods.}
We keep this single-modality subsection concise and focus on three common conditioning regimes:

\textit{Text-conditioned audio, music, and speech generation}
synthesizes sound effects, ambient scenes, music, or speech from natural-language prompts.
Representative approaches include latent diffusion models such as AudioLDM and Stable Audio Open~\citep{liu2024audioldm,stableaudioopen2025}, codec-token language models such as AudioLM~\citep{borsos2023audiolm}, and broader conditional frameworks such as AudioX~\citep{audiox2025}.
Across these systems, neural audio tokenizers and latent representations remain central to fidelity, controllability, and efficiency.

\textit{Conditional audio transformation and separation}
modify or decompose an existing recording, covering speech enhancement, source separation, remixing, and restoration.
TF-GridNet~\citep{wang2023tf} and Dual-Path Mamba~\citep{jiang2025dual} improve temporal modeling for separation, while Audio Prompt Tuning~\citep{liu2024audio} moves toward promptable and open-set separation.
In music, Band-Split RNN and Hybrid Transformers~\citep{luo2023music,rouard2023hybrid} show how frequency-structured modeling supports instrument-level control.

\textit{Instruction-guided audio editing}
uses natural language or reference cues to change selected attributes while preserving non-edited content.
Representative systems include InstructAudioEdit~\citep{wang2023instructaudioedit}, SAO-Instruct~\citep{ungersbock2025saoinstruct}, Guiding Audio Editing with Audio Language Model~\citep{lan2025guidingaudioediting}, SteerMusic~\citep{niu2025steermusic}, Step-Audio-EditX~\citep{yan2025stepaudioeditx}, and EmoCorrector~\citep{liu2025emocorrector}.
Related work on voice conversion, audio inpainting, and style transfer~\citep{popov2022diffusion,moliner2023inpainting,huang2024audiostyletransfer} points toward controllable audio editors that behave more like instruction-following generation systems.

\subsubsection{Conditional Visual Generation}
\label{sec:gen:cond_av:visual}

\paratitle{Definition.}
Conditional visual generation synthesizes or transforms images and videos under external conditions, such as text prompts, reference images, structured controls, or editing instructions.

\paratitle{Representative Tasks and Methods.}
Depending on the conditioning signal, this paradigm includes several related subtasks.

\textit{Text to Image/Video Generation}.
Text-to-image generation progressed from DALL-E~\citep{ramesh2022dalle} and Stable Diffusion~\citep{rombach2022high} to stronger native multimodal and industrial systems, such as GPT-4o image generation~\citep{hurst2024gpt}, Seedream 5.0 Lite~\citep{bytedance2026seedream5lite}, HunyuanImage-3.0~\citep{cao2025hunyuanimage}, Qwen-Image~\citep{wu2025qwenimage}, and JoyAI-Image~\citep{jd2026joyaiimage}. 
Text-to-video generation likewise advanced from Make-A-Video~\citep{singer2022makeavideo} and Imagen Video~\citep{ho2022imagenvideo} to newer foundation and product models, including Sora~\citep{openai2024sora}, Veo~\citep{google2025veo3}, Step-Video-T2V~\citep{ma2025stepvideot2v}, HunyuanVideo-1.5~\citep{kong2024hunyuanvideo}, Wan-2.1/2.2~\citep{wan2025wan}, LongCat-Video~\citep{longcatteam2025longcatvideo}, Any2caption~\citep{wu2025Any2Caption}, Hailuo 2.3~\citep{minimax2025hailuo23}, Kling 3.0~\citep{kuaishou2026kling30}, and Seedance 2.0~\citep{bytedance2026seedance}. 
Recent work also extends short-clip synthesis toward long and online generation, including StreamingT2V~\citep{henschel2024streamingt2v}, Stable Video Infinity~\citep{li2025stablevideoinfinity}, StreamDiT~\citep{kodaira2025streamdit}, Self-Forcing~\citep{huang2025self}, and LoL~\citep{cui2026lol}.

\textit{Controllable Video Generation}.
Controllable generation increasingly disentangles appearance, structure, motion, and temporal continuation. ControlNet~\citep{zhang2023controlnet} and ControlVideo~\citep{zhang2023controlvideo} showed that pretrained diffusion models can absorb structured controls without losing their generative prior. 
For image-to-video generation, Motion-I2V~\citep{shi2024motioni2v}, HunyuanVideo-I2V~\citep{kong2024hunyuanvideo}, and Step-Video-TI2V~\citep{huang2025stepvideoti2v} improve first-frame fidelity and motion control, while product systems such as Kling 3.0~\citep{kuaishou2026kling30}, Veo~\citep{google2025veo3}, and Hailuo 2.3~\citep{minimax2025hailuo23} further strengthen practical shot-level control. 
For video interpolation, Generative Inbetweening~\citep{wang2024generativeinbetweening}, large-motion interpolation via I2V diffusion adaptation~\citep{jin2024largemotioninterpolation}, and ViBiDSampler~\citep{yang2024vibidsampler} adapt generative backbones for keyframe-conditioned temporal completion.

\textit{Image/Video Editing}.
Image and video editing modify semantic content while preserving scene identity and temporal coherence. InstructPix2Pix~\citep{brooks2023instructpix2pix} established instruction-based image editing, and Video-P2P~\citep{liu2024videop2p} extended this idea to video. 
Recent systems, including Qwen-Image-Edit~\citep{wu2025qwenimage}, JoyAI-Image-Edit~\citep{jd2026joyaiimage}, DreamVE~\citep{dreamve2025}, InsViE-1M~\citep{wu2025insvie1m}, Nano-Banana~\citep{nanobanana2026}, and Wan-2.7~\citep{wan2025wan} further combine understanding and editing in shared training pipelines.

\subsection{Audio-Visual Cross-Modal Generation}
\label{sec:gen:cross_av}

Cross-modal generation uses one realized sensory stream to drive another: adding sound to video, synthesizing video or motion from audio, animating a still image with audio, or lifting audio into explicit 3D structure.
The subsections below follow that scope from video-to-audio through audio-to-video, audio-synchronized image animation, and audio-driven 3D visual generation.

\subsubsection{Video-to-Audio Generation }
\label{sec:gen:cross_av:v2a}


\paratitle{Definition.}
Video-to-audio (V2A) generation adds a plausible soundtrack to silent video, covering Foley, ambience, speech, or music depending on the scene~\citep{comunita2024diff}. The task is difficult because models must jointly satisfy semantic relevance, event-level synchronization, and acoustic plausibility under physical constraints. Compared with early Foley synthesis, current V2A models are increasingly expected to handle open-domain events, user control, and longer clips rather than only short impact sounds.

\paratitle{Methods.}
Recent V2A methods are best grouped by how they enforce alignment and how much explicit reasoning they place before waveform generation.

\textit{Alignment-first diffusion and flow models.}
Diff-Foley~\citep{comunita2024diff} established the modern recipe of contrastive AV pretraining plus latent diffusion conditioned on video features. FoleyCrafter~\citep{zhang2024foleycrafter} adds explicit onset conditioning for cleaner event timing, while FRIEREN and Tri-Ergon~\citep{wang2024frieren,liu2025triergon} replace diffusion-style denoising with rectified or flow-matching objectives to improve efficiency without giving up fine-grained correspondence.

\textit{Jointly trained and controllable V2A systems.}
The recent wave shifts the problem from narrow Foley generation to more controllable multimodal synthesis. MMAudio~\citep{cheng2025mmaudio} shows the value of large-scale multimodal joint training, Smooth-Foley~\citep{yang2025smoothfoley} adds text-guided style control, Kling-Foley~\citep{wang2025kling} pushes quality with a stronger multimodal diffusion transformer, and HunyuanVideo-Foley~\citep{lin2025hunyuanfoley} scales the recipe to stronger video backbones and longer temporal windows.

\textit{Reasoning-guided, open-world, and long-form generation.}
Another recent trend is to insert higher-level reasoning before audio decoding. ThinkSound~\citep{thinksound2025} uses multimodal chain-of-thought to infer what should sound when, PrismAudio~\citep{prismaudio2025} uses multimodal LLMs to parse richer scene semantics, and DreamFoley~\citep{dreamfoley2025} pushes toward higher-fidelity generation by leveraging scalable vision-language models. The frontier is now moving beyond short clip synthesis: Omni2Sound~\citep{omni2sound2026} unifies video-text-to-audio generation under a single any-to-audio framework, while ALIVE and Echoes Over Time~\citep{alive2026,echoesovertime2026} target joint AV generation and length generalization over much longer horizons. AV-Link~\citep{avlink2024} shows that temporally aligned diffusion features can support unified V2A/A2V generation.

\begin{insightbox}[\faLightbulbO~Trend: Sound Needs Planning]
The main difference between recent V2A systems is no longer only diffusion versus flow matching. The more important split is whether the model directly maps video features to sound, or first constructs a higher-level hypothesis about events, sources, and timing. This is why multimodal reasoning modules now matter: open-domain V2A increasingly looks like structured audiovisual planning followed by audio rendering, not just conditional denoising.
\end{insightbox}

\paratitle{Benchmarks.}
V2A benchmarks remain fragmented by sound type and synchronization granularity. VGGSound~\citep{chen2020vggsound} and AudioSet~\citep{gemmeke2017audioset} remain the default open-domain resources; Greatest Hits~\citep{owens2016visually} stresses impact timing; AVSync15~\citep{zhang2024audio} explicitly targets fine-grained synchronization; and newer resources such as FoleyBench, AudioCanvas, and VGGSound-Omni~\citep{foleybench2025,prismaudio2025,omni2sound2026} move evaluation toward harder, more diverse, and longer-horizon settings. Common metrics now mix perceptual audio quality, semantic alignment, and synchronization quality rather than relying on a single score.

\begin{table}[!t]
    \centering
    \caption{Representative benchmarks for Video-to-Audio generation.}
    \label{tab:bench_v2a}
    \benchmarktablestyle
    \begin{fullwidthtabular}{8}{l l l l}
        \toprule
        \benchhead \tblhead{Benchmark} & \tblhead{Signal} & \tblhead{Focus} & \tblhead{Metrics} \\
        \midrule
        VGGSound~\citep{chen2020vggsound} & Open-domain clips & Training/testing & FD, KL, IS, CLAP \\
        Greatest Hits~\citep{owens2016visually} & Contact clips & Impact/onset timing & Onset, sync \\
        AVSync15~\citep{zhang2024audio} & Sync stress test & Temporal alignment & AVSync, VGGSync \\
        FoleyBench~\citep{foleybench2025} & Foley clips & Foley fidelity & Human pref., sync, quality \\
        AudioCanvas~\citep{prismaudio2025} & Open-world scenes & Multi-event soundscapes & Semantic/spatial consistency \\
        VGGSound-Omni~\citep{omni2sound2026} & V/T/A conditions & Omni-condition V2A & Quality, align., instruction \\
        \bottomrule
    \end{fullwidthtabular}
\end{table}

\begin{table}[!t]
    \centering
    \caption{Performance comparison of representative Video-to-Audio generation methods on the VGGSound-Test leaderboard. Metrics follow the official HunyuanVideo-Foley~\citep{lin2025hunyuanfoley} evaluation protocol with PANNs-based FD/KL and Inception Score (IS). Best result in \textbf{bold}.}
    \label{tab:perf_v2a}
    \performancetablestyle
    \begin{fullwidthtabular}{10}{l l c c c}
        \toprule
        \perfhead \tblhead{Method} & \tblhead{Family} & \tblhead{FD$_\text{P}$ $\downarrow$} & \tblhead{KL$_\text{P}$ $\downarrow$} & \tblhead{IS$\uparrow$} \\
        \midrule
        Diff-Foley~\citep{comunita2024diff} & Latent Diffusion & 6.34 & 6.15 & 8.52 \\
        FoleyCrafter~\citep{zhang2024foleycrafter} & Diffusion + Onset Control & 4.12 & 4.01 & 14.59 \\
        FRIEREN~\citep{wang2024frieren} & Rectified Flow & 4.18 & 4.16 & 12.80 \\
        MMAudio~\citep{cheng2025mmaudio} & Joint Multimodal Training & 2.89 & 2.82 & 15.70 \\
        ThinkSound~\citep{thinksound2025} & MLLM-guided V2A & 2.70 & 2.38 & 16.48 \\
        HunyuanVideo-Foley~\citep{lin2025hunyuanfoley} & Scaled Controllable V2A & \textbf{2.51} & \textbf{2.17} & \textbf{16.87} \\
        \bottomrule
    \end{fullwidthtabular}
\end{table}

\subsubsection{Audio-to-Video Generation}
\label{sec:gen:cross_av:a2v}

\paratitle{Definition.}
Audio-to-video (A2V) generation synthesizes video conditioned on sound, speech, or music. Compared with V2A, the ambiguity is larger: the same audio can support many visually plausible motions, so models must balance semantic faithfulness, rhythmic alignment, and controllability.

\paratitle{Representative Tasks and Methods.}
A2V methods fall into three broad families.

\textit{Backbone adaptation.}
These methods retrofit pretrained video generators with audio-conditioning modules. MusicInfuser~\citep{hong2025musicinfuser} injects music features through cross-attention and low-rank adaptation, while Diverse and Aligned A2V~\citep{yariv2024diverse} maps audio into the token space of pretrained text-to-video models. AV-Link~\citep{avlink2024} further shows that a single temporally aligned framework can support both A2V and V2A generation.

\textit{Motion-mediated pipelines.}
Rather than generating pixels directly, these systems first predict intermediate motion such as keypoints or SMPL trajectories and only then render video. ChoreoMuse~\citep{wang2025choreomuse} uses this decomposition for music-to-dance generation, and earlier works~\citep{zhu2021let} follow a similar two-stage design. The advantage is stronger beat alignment; the cost is higher pipeline complexity and weaker end-to-end flexibility.

\textit{Identity-conditioned generation.}
Reference-image-based models reduce ambiguity by anchoring appearance. DiffTalk~\citep{shen2023difftalk} produces speech-driven talking heads with landmark-guided diffusion, and AudCast~\citep{guan2025audcast} extends the setting to full-body avatars with cascaded diffusion transformers for facial motion, gestures, and body dynamics.

\begin{frontierbox}[\faLightbulbO~Open Problem: Audio Underspecifies Video]
The central difficulty in A2V is not just quality but underdetermination: many videos can be faithful to the same soundtrack. Recent systems reduce this ambiguity with reference images, motion priors, or explicit control signals, but none fully resolves the trade-off between diversity and user steerability. In practice, controllability is becoming as important as realism because downstream use cases rarely want just any plausible video.
\end{frontierbox}

\paratitle{Benchmarks.}
Benchmarks are task-specific. AIST++~\citep{li2021aist} and AIOZ-GDance~\citep{le2023music} dominate music-to-dance evaluation, VoxCeleb2~\citep{chung2018voxceleb2} and HDTF~\citep{zhang2021hdtf} are standard for speech-driven avatars, and VGGSound~\citep{chen2020vggsound} remains the broadest open-domain benchmark for general A2V generation. Many recent systems additionally report zero-shot or curated in-the-wild evaluations~\citep{guan2025audcast}.

\begin{table}[!t]
    \centering
    \caption{Representative benchmarks for Audio-to-Video generation.}
    \label{tab:bench_a2v_gen}
    \benchmarktablestyle
    \setlength{\tabcolsep}{2.5pt}
    \begin{fullwidthtabular}{10}{l c c c l}
    \toprule
    \benchhead \tblhead{Dataset} & \tblhead{Scale} & \tblhead{Input Audio} & \tblhead{Visual Target} & \tblhead{Primary Use} \\
    \midrule

    AIST++~\citep{li2021aist} & 1.4K clips & Music & Human motion/video & Music-to-dance generation \\
    AIOZ-GDance~\citep{le2023music} & 19.8K clips & Music & Group dance & Choreography and formation control \\
    VoxCeleb2~\citep{chung2018voxceleb2} & 1M+ utterances & Speech & Talking face & Identity-generalized speech animation \\
    HDTF~\citep{zhang2021hdtf} & 362 videos & Speech & Talking face & High-resolution talking-head quality \\
    VGGSound~\citep{chen2020vggsound} & 200K clips & General sound & Open-domain video & Broad audio-conditioned video \\
    \bottomrule
    \end{fullwidthtabular}
\end{table}

\subsubsection{Audio-Synchronized Image Animation}
\label{sec:gen:cross_av:ai2v}

\paratitle{Definition.}
Audio-synchronized image animation, or audio-driven image-to-video (AI2V) generation, synthesizes video from a reference image and driving audio while preserving identity and temporal coherence. Beyond the classical talking-head setting~\citep{prajwal2020lip,cui2025hallo3}, recent models also support general audio+image+text generation~\citep{zhang2024audio,gao2025wan}, where text serves as auxiliary control over motion, emotion, or scene behavior.

\paratitle{Representative Tasks and Methods.} Recent audio-synchronized image animation methods can be divided into the following aspects:

\textit{Human-centric AI2V.}
Most existing work remains human-centric. In talking-head and portrait animation, Wav2Lip~\citep{prajwal2020lip} emphasizes lip accuracy; DiffTalk~\citep{shen2023difftalk} and SadTalker~\citep{zhang2023sadtalker} add head pose and expression; EMO~\citep{tian2024emo}, VASA-1~\citep{xu2024vasa}, Hallo2~\citep{cui2024hallo2}, Hallo3~\citep{cui2025hallo3}, FantasyTalking~\citep{wang2025fantasytalking}, and StableAvatar~\citep{tu2025stableavatar} further improve expressiveness, duration, and fidelity. Beyond talking heads, EchoMimicV2/V3~\citep{meng2025echomimicv2,meng2026echomimicv3} extends to semi-body animation, EMO2~\citep{tian2025emo2} adds gesture-aware control, and OmniAvatar~\citep{gan2025omniavatar}, OmniHuman-1~\citep{lin2025omnihuman}, HuMo~\citep{chen2025humo}, AudCast~\citep{guan2025audcast}, and Playmate2~\citep{ma2025playmate2} expand toward full-body, multi-character, and richer human motion. Recent industrial systems such as HunyuanVideo-Avatar~\citep{chen2025hunyuanvideo}, Kling-Avatar~\citep{ding2025kling,team2025klingavatar}, and JoyAvatar~\citep{wang2026joyavatar} further combine audio, image, and optional text to improve controllability and long-form avatar generation.

\textit{General AI2V.}
General AI2V is still much smaller than the human-centric branch, but it is not empty. ASVA~\citep{zhang2024audio} explicitly formulates static-image animation from audio across multiple object and scene classes, and KeyVID~\citep{wang2025keyvid} improves this setting with keyframe-aware generation for highly dynamic motions. Scaling Up ASVA~\citep{zhang2025scaling} extends the task toward open-domain audio-synchronized visual animation, while TIA2V~\citep{zhao2025tia2v} and Syncphony~\citep{song2025syncphony} achieve more controllable general video generation. In industry, Wan-S2V~\citep{gao2025wan} sits near the boundary between human-centric and general AI2V, while broader multimodal foundation models such as SkyReels-V4~\citep{chen2026skyreels}, Seedance 2.0~\citep{bytedance2026seedance}, and Veo 3.1~\citep{google2025veo3} already support image-conditioned video generation with audio or sound.

\paratitle{Benchmarks.}
As summarized in \cref{tab:bench_ai2v_gen}, VoxCeleb/VoxCeleb2~\citep{nagrani2017voxceleb,chung2018voxceleb2}, LRS2/LRS3~\citep{son2017lip,afouras2018lrs3}, HDTF~\citep{zhang2021hdtf}, and CelebV-HQ~\citep{zhu2022celebv} remain standard for identity preservation, speech synchronization, and portrait quality. Recent resources broaden the scope: TalkVid-Bench~\citep{chen2025talkvid} target more diverse talking-head training and evaluation, TalkVerse~\citep{wang2025talkverse} emphasizes minute-long generation, TalkCuts~\citep{chen2025talkcuts} targets multi-shot speech videos, MIT~\citep{zhu2025mit} benchmarks multi-human interaction, and THEval~\citep{quignon2025theval} and EvalTalker~\citep{zhou2025evaltalker} provide finer-grained perceptual evaluation.

\begin{table}[!t]
    \centering
    \caption{Representative datasets and evaluation benchmarks for audio-driven image animation.}
    \label{tab:bench_ai2v_gen}
    \benchmarktablestyle
    \begin{fullwidthtabular}{8}{l c c l}
    \toprule
    \benchhead \tblhead{Dataset / Benchmark} & \tblhead{Scale} & \tblhead{Target} & \tblhead{Primary Use} \\
    \midrule

    VoxCeleb2~\citep{chung2018voxceleb2} & 1.13M utter., 2.44k h & Face & Identity generalization \\
    LRS2~\citep{son2017lip} & 144k utter. & Face & Lip-sync alignment \\
    LRS3~\citep{afouras2018lrs3} & 400h+ & Face & AV speech alignment \\
    HDTF~\citep{zhang2021hdtf} & 15.8h, 300+ subj. & Portrait & Talking-head synthesis \\
    CelebV-HQ~\citep{zhu2022celebv} & 35.7k clips & Portrait & Expressive portrait animation \\
    TalkVid-Bench~\citep{chen2025talkvid} & 1,244h + 500 eval clips & Talking head & Fairness-aware evaluation \\
    TalkVerse~\citep{wang2025talkverse} & 2.3M clips, 6.3k h & Single-person & Minute-long generation \\
    TalkCuts~\citep{chen2025talkcuts} & 164k clips, 500h+ & Multi-human & Multi-shot speech videos \\
    MIT~\citep{zhu2025mit} & 12h, 2--4 speakers & Multi-human & Multi-speaker interaction \\
    THEval~\citep{quignon2025theval} & 85k generated videos & Talking head & Fine-grained evaluation \\
    EvalTalker~\citep{zhou2025evaltalker} & 5,492 samples & Multi-human & Perceptual quality assessment \\
    \bottomrule
    \end{fullwidthtabular}
\end{table}

\subsubsection{Audio-Driven 3D Visual Generation}
\label{sec:gen:cross_av:a2v3d}

\paratitle{Definition.}
Audio-driven 3D visual generation maps sound into dynamic 3D facial, head, or body representations $\mathcal{V}_{3D}(t)$. Relative to 2D animation, the 3D formulation offers viewpoint consistency, explicit geometry, and better compatibility with downstream rendering, but it also makes subtle motion realism harder to learn.

\paratitle{Representative Tasks and Methods.}
Audio-driven 3D synthesis spans five recurring directions.

\textit{3D Face Animation \& Talking Heads} deform explicit mesh topologies such as FLAME or SMPL. Early regression systems (VOCA~\citep{cudeiro2019voca}, MeshTalk~\citep{richard2021meshtalk}) often over-smoothed motion, whereas FaceFormer~\citep{fan2022faceformer}, CodeTalker~\citep{xing2023codetalker}, and StreamingTalker~\citep{yang2025streamingtalker} improve co-articulation and streaming behavior through stronger sequence modeling.

\textit{Neural Rendering \& Dynamic Avatars} focus on photorealism. AD-NeRF~\citep{guo2021adnerf} pioneered audio-conditioned NeRFs, GeneFace~\citep{ye2023geneface} improved cross-identity transfer, and recent 3D Gaussian Splatting systems such as GaussianTalker~\citep{zhang2024gaussiantalker} and SynGauss~\citep{zhu2025syngauss} make interactive rendering increasingly practical.

\textit{Emotion \& Expression Enhancement} explicitly combats ``dead-face'' artifacts. EVP~\citep{ji2021evp}, DreamTalk~\citep{ma2023dreamtalk}, and EmoGene~\citep{wang2025emogene} disentangle content from style or emotion so that non-verbal dynamics are not washed out by phoneme tracking alone.

\textit{Co-speech Gestures \& Full-Body Motion} map speech to skeletal kinematics. TalkSHOW~\citep{yi2023talkshow} separates facial and bodily channels, while EMAGE~\citep{liu2024emage} improves gesture diversity and beat consistency with masked gesture transformers and compositional VQ-VAEs.

\textit{Holistic Human Synthesis} targets full 3D humans. Stereo-Talker~\citep{deng2024stereotalker} and Real3D-Portrait~\citep{ye2024real3dportrait} move beyond the head-only regime by coupling portrait reconstruction with coordinated body and background dynamics.

\paratitle{Benchmarks.}
Benchmarks depend on the output representation: VoxCeleb2~\citep{chung2018voxceleb2} and HDTF~\citep{zhang2021hdtf} support portrait synthesis, MEAD~\citep{wang2020mead} adds controlled emotion, VOCASET~\citep{cudeiro2019voca} and BIWI serve mesh-level evaluation, and BEAT2~\citep{liu2024emage} targets holistic gesture generation. Metrics span geometric accuracy (LVE, vertex error), lip-sync quality (LSE-C/LSE-D), perceptual realism (FID), and body-motion alignment such as Beat Constancy or FGD~\citep{liu2024emage}.

\begin{table}[!t]
    \centering
    \caption{Representative datasets for audio-driven 3D visual generation.}
    \label{tab:datasets:a2v3d}
    \benchmarktablestyle
        \begin{fullwidthtabular}{8}{l c c l}
        \toprule
        \benchhead \tblhead{Dataset} & \tblhead{Modality} & \tblhead{Hours} & \tblhead{Focus} \\
        \midrule
        VoxCeleb2~\citep{chung2018voxceleb2} & Face (2D video) & 2,442 & In-the-wild pretraining and identity robustness \\
        MEAD~\citep{wang2020mead} & Face (2D video) & $\sim$40 & Multi-view emotion control \\
        HDTF~\citep{zhang2021hdtf} & Face (2D video) & 15.8 & High-resolution talking-head rendering \\
        VOCASET~\citep{cudeiro2019voca} & Face (3D mesh) & 0.48 & Registered 4D scans and mesh-level evaluation \\
        BEAT2~\citep{liu2024emage} & Full body (3D) & 60 & Holistic gesture/body generation with semantics \\
        \bottomrule
        \end{fullwidthtabular}
\end{table}

\subsection{Joint Audio-Visual Generation}
\label{sec:gen:joint_av}

Joint generation optimizes for coupled audio and video (or related edits) in a single objective or training recipe, rather than stapling a unimodal output to another model as an afterthought.
The following subsections cover text- and image-conditioned joint synthesis as well as joint audio-video editing workflows.


\subsubsection{Text-to-Audio-Video Generation}
\label{sec:gen:joint_av:t2av}

\paratitle{Definition.}
Text-to-audio-video (T2AV) generation synthesizes both video and synchronized audio from a textual prompt~\citep{liu2025javisdit}. A strong model must satisfy three conditions at once: within-modality quality, faithfulness to the prompt, and cross-modal synchronization.

\paratitle{Methods.}
MM-Diffusion~\citep{ruan2023mm} first showed joint audio-video diffusion, but only in narrow domains such as landscapes and dance~\citep{lee2022sound,li2021aist}. The text-conditional era begins with JavisDiT~\citep{liu2025javisdit}, after which several design lines emerge.

\textit{Architecture Evolution.}
Early attempts such as JavisDiT~\citep{liu2025javisdit}, UniVerse-1~\citep{wang2025universe}, BridgeDiT~\citep{guan2025taming}, and Ovi~\citep{low2025ovi} keep audio and video branches partially separate while exchanging information through specific synchronization modules (e.g., cross-attention). Recent approaches including JoVA~\citep{huang2025jova} and JavisDiT++~\citep{liu2026javisdit++} move toward more unified modeling by shared or tightly coupled attention blocks to improve cross-modal coherence.

\textit{Audio-Visual Synchronization.}
Commonly-used strategies includes text-derived temporal priors~\citep{liu2025javisdit}, frame-level cross-attention~\citep{zhao2025uniform,wang2025universe,hu2025harmony,low2025ovi}, or aligned RoPE position IDs~\citep{huang2025jova,liu2026javisdit++} keep both modalities in step. On the other hand, data scaling~\citep{team2026mova} also plays a vital role in achieving high-quality synchronization.

\textit{Multi-Task Unification.}
Beyond the standalone text-to-audio-video (T2AV) generation, recent works (e.g., Harmony~\citep{hu2025harmony}, UniAVGen~\citep{zhang2025uniavgen}, and Apollo~\citep{wang2026klear}) widely explore multi-task integration (e.g., A2V, V2A, TI2AV, etc.) to achieve better modality learning and cross-task synergy, and OmniForcing~\citep{su2026omniforcing} further extends to real-time joint audio-video generation.

\textit{Post-Training Application.}
Preference optimization via DPO~\citep{liu2025improving} or GRPO~\citep{liu2025flow} is also becoming standard for improving prompt following and subjective quality~\citep{chen2025seedance}. For instance, JavisDiT++~\citep{liu2026javisdit++} derives the AV-DPO strategy to investigate DPO adaptation to joint audio-video generation.

\textit{Agentic Workflow.}
In contrast to end-to-end T2AV models, agentic approaches decompose audio-visual generation into cascaded stages, typically \textit{text-to-video} plus \textit{video-to-audio}. MV-Crafter~\citep{chen2025mvcrafter} is an early workflow-style example, while ReelWave~\citep{wang2025reelwave}, LVAS-Agent~\citep{zhang2025lvasagent}, and AutoMV~\citep{tang2025automv} more explicitly formulate soundtrack or music-video generation as multi-agent pipelines. Related systems such as AesopAgent~\citep{wang2024aesopagent}, MM-StoryAgent~\citep{xu2025mmstoryagent}, MovieAgent~\citep{wu2025movieagent}, AniME~\citep{zhang2025anime}, and MAViS~\citep{wang2026mavis} further support the value of explicit planning for long-form generation.

Nowadays, commercial systems (including Sora2, Veo3.1~\citep{google2025veo3}, Kling3.0, Wan2.6, and Seedance 2.0 Pro) are still ahead because they combine larger audio-video corpora, stronger base generators, and heavier post-training. On the other hand, open-source models are catching up with proprietary systems in performance, where LTX-2~\citep{hacohen2026ltx2} and MOVA~\citep{team2026mova} demonstrate promising potential and scalability.

\begin{frontierbox}[\faLightbulbO~Open Problem: Scale Still Decides]
T2AV currently has the widest gap between public research prototypes and frontier product systems. The difference is not a single architectural trick; it is data scale, base-model maturity, and post-training depth combined. Open models have largely solved the question of \emph{how} to build joint AV generators, but not yet the question of how to train them at product-level scale.
\end{frontierbox}


\begin{table}[!t]
\centering
\caption{Representative datasets and curated evaluation suites for text-to-audio-video generation.
}
\label{tab:bench_t2av_gen}
\benchmarktablestyle
\begin{fullwidthtabular}{10}{l c c c l}
\toprule
\benchhead \tblhead{Name} & \tblhead{Type} & \tblhead{Scale} & \tblhead{Audio} & \tblhead{Primary Focus} \\
\midrule

Greatest Hits~\citep{owens2016visually}
& Dataset
& 977 videos
& \audtag{Sound}
& Impact/onset sync \\

Landscape~\citep{lee2022sound}
& Dataset
& 928 videos
& \audtag{Sound}
& Ambient scene realism \\

VGGSound~\citep{chen2020vggsound}
& Dataset
& 200K clips
& \audtag{Sound}
& Open-domain coverage \\

AVSync15~\citep{zhang2024audio}
& Suite
& 150 clips
& \audtag{Mixed}
& Fine-grained AV sync \\

JavisBench~\citep{liu2025javisdit}
& Suite
& 10,140 prompts
& \audtag{Sound}
& Quality, semantics, sync \\

Verse-Bench~\citep{wang2025universe}
& Suite
& 600 prompts
& \audtag{Sound} \audtag{Speech}
& Speech-inclusive T2AV \\

Harmony-Bench~\citep{hu2025harmony}
& Suite
& 150 prompts
& \audtag{Sound} \audtag{Speech}
& Prompt following; sync \\

VABench~\citep{hua2025vabench}
& Suite
& 778 prompts
& \audtag{Sound} \audtag{Music} \audtag{Speech}
& Broad audio coverage \\

T2AV-Compass~\citep{cao2025t2av}
& Suite
& 500 prompts
& \audtag{Sound} \audtag{Music} \audtag{Speech}
& Unified multimodal scoring \\

PhyAVBench~\citep{xie2025phyavbench}
& Suite
& 1,000 prompts
& \audtag{Sound} \audtag{Music} \audtag{Speech}
& Physical/causal grounding \\

\bottomrule
\end{fullwidthtabular}
\end{table}

\paratitle{Benchmarks.}
As shown in \cref{tab:bench_t2av_gen}, early benchmarks (Greatest Hits~\citep{owens2016visually}, Landscape~\citep{lee2022sound}, VGGSound~\citep{chen2020vggsound}, AVSync15~\citep{zhang2024audio}) cover limited scenarios without standardized metrics.
Curated suites such as JavisBench~\citep{liu2025javisdit}, Verse-Bench~\citep{wang2025universe}, Harmony-Bench~\citep{hu2025harmony}, VABench~\citep{hua2025vabench}, and T2AV-Compass~\citep{cao2025t2av} provide broader prompt coverage and more systematic scoring, while PhyAVBench~\citep{xie2025phyavbench} explicitly measures physical plausibility. The evaluation trend is moving from modality-specific metrics toward composite scores that jointly assess prompt following, perceptual quality, and temporal coherence.


\begin{table}[!t]
    \centering
    \caption{Performance comparison of T2AV models on T2AV-Compass~\citep{cao2025t2av}. Best results in \textbf{bold}.}
    \label{tab:perf_t2av}
    \performancetablestyle
    \setlength{\tabcolsep}{3pt}
    \begin{fullwidthtabular}{22}{l c c c c c c c c c c}
    \toprule
    \perfhead
    & & \multicolumn{2}{c}{\tblhead{Video Quality}} & \multicolumn{2}{c}{\tblhead{Audio Quality}} & \multicolumn{5}{c}{\tblhead{Cross-modal Alignment}} \\
    \perfheadersep{3-11}
    \perfhead
    \multirow{-2}{*}{\tblhead{Model}} & \multirow{-2}{*}{\tblhead{Open-Source}} & \tblhead{VT$\uparrow$} & \tblhead{VA$\uparrow$} & \tblhead{PQ$\uparrow$} & \tblhead{CU$\uparrow$} & \tblhead{A-V$\uparrow$} & \tblhead{T-A$\uparrow$} & \tblhead{T-V$\uparrow$} & \tblhead{DS$\downarrow$} & \tblhead{LS$\uparrow$} \\
    \midrule
    Veo-3.1~\citep{google2025veo31} & \crossmark & \textbf{13.39} & \textbf{5.425} & 7.015 & 6.621 & 0.2856 & 0.2335 & 0.2438 & 0.6776 & 1.509 \\
    Sora-2~\citep{openai2025sora2} & \crossmark & 7.568 & 4.112 & 5.827 & 5.340 & 0.2419 & 0.2484 & 0.2432 & 0.8100 & 1.331 \\
    Kling-2.6~\citep{kuaishou2025kling26} & \crossmark & 11.41 & 5.417 & 6.882 & 6.449 & 0.2495 & 0.2495 & 0.2449 & 0.7852 & 1.502 \\
    Wan-2.6~\citep{alibaba2025wan26} & \crossmark & 11.87 & 4.605 & 6.658 & 6.222 & 0.2149 & \textbf{0.2572} & \textbf{0.2451} & 0.8818 & 1.081 \\
    Seedance-1.5~\citep{chen2025seedance} & \crossmark & 12.74 & 5.007 & \textbf{7.555} & \textbf{7.250} & \textbf{0.2875} & 0.2320 & 0.2370 & 0.8650 & \textbf{1.560} \\
    PixVerse-V5.5~\citep{pixverse2025v55} & \crossmark & 11.54 & 4.558 & 6.108 & 5.855 & 0.1816 & 0.2305 & 0.2431 & \textbf{0.6627} & 1.306 \\
    Ovi-1.1~\citep{low2025ovi}       & \checkmark & 9.336 & 4.368 & 6.569 & 6.492 & 0.1620 & 0.1756 & 0.2391 & 0.9624 & 1.191 \\
    JavisDiT~\citep{liu2025javisdit} & \checkmark & 6.850 & 3.575 & 4.299 & 5.204 & 0.1284 & 0.1257 & 0.2320 & 1.3220 & - \\
    \bottomrule
    \end{fullwidthtabular}
\end{table}

\subsubsection{Image-to-Audio-Video Generation}
\label{sec:gen:joint_av:i2av}


\paratitle{Definition.}
Image-to-audio-video (I2AV) generation turns a single image into a sounding video clip. Unlike T2AV, appearance is anchored by the reference image, while motion and sound must be inferred jointly under a much tighter spatial constraint.

\paratitle{Methods.}
Early I2AV solutions are mostly cascaded. \textit{Independent generation} combines an image-to-video model such as Stable Video Diffusion~\citep{blattmann2023stable} with an image-to-audio model such as IM2Wav~\citep{sheffer2023hear}, but provides weak audio-video coupling. \textit{Video-first} pipelines animate the image and then add sound with video-to-audio models such as Diff-Foley~\citep{comunita2024diff} and FoleyCrafter~\citep{zhang2024foleycrafter}, while \textit{caption-mediated} variants rely on image-derived prompts for separate video and audio generators~\citep{hong2022cogvideo,kong2024hunyuanvideo,liu2024audioldm}. 

A dedicated line emerged with Animate and Sound an Image~\citep{wang2025animate}, which explicitly formulates image-to-sounding-video generation with joint transformer conditioning. Recent unified AV models further absorb image-conditioned generation into a single framework, including UniVerse-1~\citep{wang2025universe}, Ovi~\citep{low2025ovi}, Klear~\citep{wang2026klear}, ALIVE~\citep{alive2026}, and SkyReels-V4~\citep{chen2026skyreels}. The same trend appears in production systems such as Kling Video 2.6/3.0/3.0 Omni~\citep{kuaishou2026kling30}, Wan2.6~\citep{wan2025wan}, Seedance 1.5 pro/2.0~\citep{chen2025seedance,bytedance2026seedance}, Vidu Q3~\citep{vidu2026q3}, Sora 2~\citep{openai2024sora}, Veo 3.1~\citep{google2025veo3}, and Runway Gen-4.5~\citep{runway2025nativeaudio}, which all expose image-conditioned or reference-guided audio-video generation in an integrated workflow. 

\paratitle{Benchmarks.}
Existing benchmarks for image-conditional I2AV generation remain scarce. AVSync15~\citep{zhang2024audio} is the most used benchmark for synchronized sounding-video generation, and Verse-Bench~\citep{wang2025universe} further supports image-conditioned joint audio-video evaluation. More standardized benchmarks are still needed for this setting.




\subsubsection{Joint Audio-Video Editing}
\label{sec:gen:joint_av:tav2av}

\paratitle{Definition.}
Joint audio-video editing edits video and sound \emph{together} so that the result remains synchronized and semantically consistent across modalities. Given $(V, A)$ and an edit condition $c$, the model produces $(V', A')$ that applies the intended change to both streams while preserving alignment and minimizing changes to unrelated content.

\paratitle{Representative Tasks and Methods.} Early work, such as one-shot adaptation~\citep{liang2024avedit} and AvED~\citep{lin2025aved}, establishes instruction-based and zero-shot joint editing, mainly targeting prompt-consistent changes without strong paired supervision.
Meanwhile, more recent methods move to finer granularity. Object-AVEdit~\citep{fu2025objectavedit} performs object-level addition, replacement, and removal. VAInpaint~\citep{wu2025vainpaint} addresses zero-shot joint inpainting and removal with visual masks and LLM-assisted audio querying. SAVE~\citep{xu2025save} formulates paired source-to-target audiovisual removal, while AVI-Edit~\citep{zheng2025aviedit} further supports instance-level masked editing with refined spatial control. For specialized settings, Coherent Audio-Visual Editing~\citep{ishii2025coherent} edits audio conditionally after video edits, and JUST-DUB-IT~\citep{chen2026just} adapts joint AV diffusion for dubbing-style speech and facial-motion editing.

A broader recent trend is to absorb editing into unified joint AV backbones rather than treating it as a standalone task. UniAVGen~\citep{zhang2025uniavgen} and SkyReels-V4~\citep{chen2026skyreels} suggest this direction by supporting continuation, inpainting, and editing-like workflows within a shared audio-video generation framework.

\paratitle{Benchmarks.}
Dedicated joint AV editing benchmarks are still limited, but they have evolved from small prompt-based sets to large-scale mask-aware and paired-data benchmarks. As organized in \cref{tab:bench_joint_av_edit}, OAVE~\citep{liang2024avedit} is an early benchmark for one-shot joint editing, and AvED-Bench~\citep{lin2025aved} evaluates full-video zero-shot editing. More recently, AVISet~\citep{zheng2025aviedit} and SAVEBench~\citep{xu2025save} provide instance masks or paired source-target supervision, enabling evaluation of edit faithfulness, preservation, and synchronization under localized edits.

\begin{table}[!t]
    \centering
    \caption{Representative benchmarks for joint audio-video editing.}
    \label{tab:bench_joint_av_edit}
    \benchmarktablestyle
    \begin{fullwidthtabular}{8}{l c c l}
    \toprule
    \benchhead \tblhead{Dataset} & \tblhead{Scale} & \tblhead{Edit Scope} & \tblhead{Key Feature} \\
    \midrule
    OAVE~\citep{liang2024avedit} & 44 events & One-shot event editing & Early dedicated benchmark \\
    AvED-Bench~\citep{lin2025aved} & 110 videos & Full-video editing & Zero-shot prompt-based evaluation \\
    AVISet~\citep{zheng2025aviedit} & 73k clips & Instance-level editing & Masks and paired edit instructions \\
    SAVEBench~\citep{xu2025save} & 17k pairs & Object-level removal & Paired source-target supervision \\
    \bottomrule
    \end{fullwidthtabular}
\end{table}

\begin{insightbox}[\faLightbulbO~Trend: From Joint Editing to Precise Intervention]
The field is shifting from coarse joint modification to precise cross-modal intervention. The real challenge is not just editing audio and video together, but doing so locally, synchronously, and with minimal side effects. This trend is pushing joint AV editing toward finer-grained supervision and unified AV foundation models.
\end{insightbox}

\section{Audio-Visual Interaction}
\label{sec:inter}

Audio-visual interaction concerns systems that do not merely perceive multimodal content, but must respond to it in real time. We distinguish two settings: \emph{conversational interaction}, where the response is language or speech, and \emph{embodied interaction}, where the response is an action in the physical world. Both require tightly coupled perception, memory, reasoning, and generation under latency constraints, making interaction a stricter test of AV intelligence than static understanding alone.

\subsection{Interactive Audio-Visual Conversation}
\label{sec:inter:av_conver}

\begin{figure*}[!t]
    \centering
    \includegraphics[width=\textwidth]{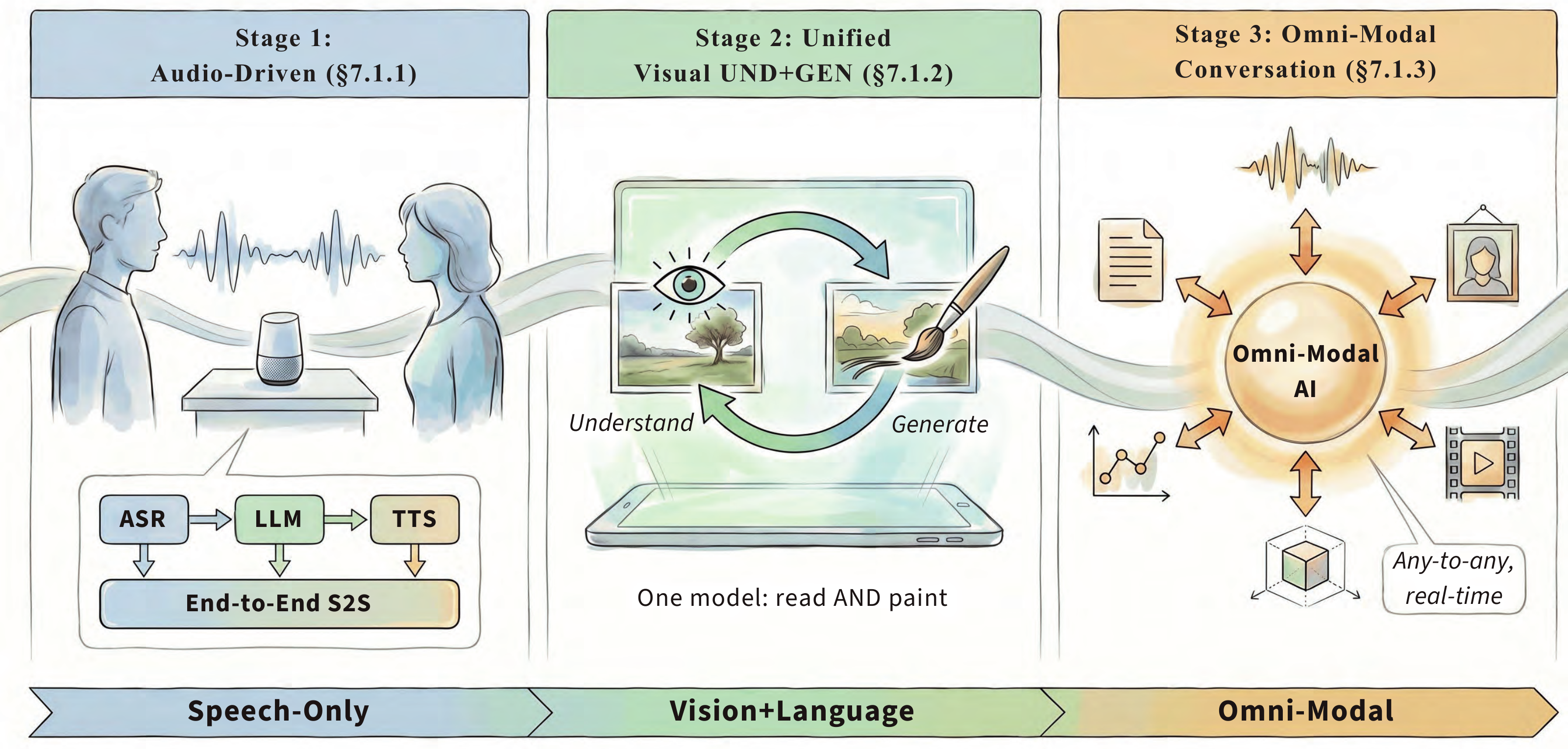}
    \caption{Organization of interactive audio-visual conversation. We organize this subsection into audio-driven speech interaction, unified visual understanding and generation, and omni-modal conversation for real-time any-to-any multimodal exchange.}
    \label{fig:av_conver_overview}
\end{figure*}

Conversational AV systems aim for natural multi-turn interaction over speech, images, video, and text. We organize this subsection around three stages of conversational capability: speech-first interaction, unified visual understanding and generation, and omni-modal conversation that handles arbitrary input/output combinations, as summarized in~\cref{fig:av_conver_overview}. The persistent constraints are latency, context retention, and the preservation of non-verbal cues such as prosody, emotion, and temporal grounding.

\subsubsection{Audio-Driven Conversation }
\label{sec:inter:av_conver:audio}

\paratitle{Definition.}
Audio-driven conversation refers to dialogue systems that take speech, environmental sound, or music as conversational input and respond in text or speech. Unlike text-only chat, these systems must preserve acoustic content, linguistic semantics, and paralinguistic cues such as prosody, emotion, and speaker identity. Recent AudioLLMs~\citep{ji2024wavchat} make this formulation practical for assistants, agents, and streaming multimodal interaction.

\paratitle{Methods.}
Current systems under this category can be organized by how they pass acoustic information through the dialogue loop:

\textit{Cascaded ASR--LLM--TTS pipelines}
explicitly transcribe speech, reason in text, and synthesize speech again, as in AudioGPT~\citep{huang2024audiogpt}. Cascades are modular and easy to debug, but they accumulate latency and lose prosodic information at the text bottleneck.

\textit{Continuous-feature audio-to-text models}
feed speech or audio representations into an LLM and usually return text. They use pretrained encoders such as Wav2Vec~\citep{schneider2019wav2vec}, HuBERT~\citep{hsu2021hubert}, Whisper~\citep{radford2023robust}, or WavLM~\citep{chen2022wavlm}, with representative systems including Qwen-Audio/Qwen2-Audio~\citep{chu2023qwenaudio,chu2024qwen2audio}, SpeechVerse~\citep{das2024speechverse}, and E-Chat~\citep{xue2024echat}.

\textit{Discrete-token and end-to-end speech-to-speech models}
represent speech with semantic or codec tokens and generate spoken responses directly or through a vocoder~\citep{du2024cosyvoice}. SpeechGPT~\citep{zhang2023speechgpt}, Mini-Omni~\citep{xie2024miniomni}, LLaMA-Omni~\citep{fang2024llamaomni}, Moshi~\citep{defossez2024moshi}, IntrinsicVoice~\citep{zhang2024intrinsicvoice}, OmniFlatten~\citep{zhang2025omniflatten}, Freeze-Omni~\citep{wang2024freezeomni}, and PSLM~\citep{mitsui2024pslm} are representative of this direction.

\textit{Acoustic-oriented and cross-domain audio models}
try to preserve non-linguistic information such as emotion, timbre, speaker identity, and environmental sound. Some models use unified encoders for speech, sound, and music, such as Audio Flamingo~\citep{kong2024audioflamingo}, Audio Flamingo~2~\citep{ghosh2025audioflamingo2}, and Qwen-Audio~\citep{chu2023qwenaudio}; others use multiple domain-specific encoders or projectors, such as SALMONN~\citep{tang2023salmonn} and GAMA~\citep{ghosh2024gama}. LauraGPT~\citep{du2023lauragpt} and SpeechGPT-Gen~\citep{zhang2024speechgptgen} further show how audio regeneration can preserve richer acoustic detail.

\begin{insightbox}[\faLightbulbO~Trend: Speech Drops the Cascade]
Speech interaction is moving away from hard ASR$\rightarrow$LLM$\rightarrow$TTS decomposition, but cascades are not obsolete. End-to-end systems are better at latency and prosody preservation, whereas cascades still win on modularity and controllable intermediate text. The most practical near-term compromise is token-based hybrids that keep more acoustic detail without giving up LLM-compatible interfaces.
\end{insightbox}

\paratitle{Benchmarks.}
Evaluation is still fragmented across sound, music, and speech. Domain-specific suites include Clotho-AQA~\citep{lipping2022clothoaqa} and AudioCaps~\citep{kim2019audiocaps} for environmental sounds, MuChoMusic~\citep{weck2024muchomusic}, MusicBench~\citep{melechovsky2024mustango}, and CMI-Bench~\citep{ma2025cmiBench} for music, and LibriSQA~\citep{zhao2024librisqa} plus Dynamic-SUPERB~\citep{huang2024dynamicsuperb} for speech. More recent cross-domain suites such as AudioBench~\citep{wang2025audiobench}, AIR-Bench~\citep{yang2024airbench}, MMAU~\citep{sakshi2024mmau}, and MMAR~\citep{ma2025mmar} better reflect the breadth expected from conversational audio models.

\begin{table}[!t]
    \centering
    \caption{Representative evaluation suites relevant to audio-driven conversation and audio-input dialogue models. Cross-domain denotes benchmarks that intentionally mix multiple audio domains or reasoning settings.}
    \label{tab:bench_audio_conv}
    \benchmarktablestyle
    \begin{fullwidthtabular}{12}{l c c c c l}
    \toprule
    \benchhead \tblhead{Dataset} & \tblhead{Sound} & \tblhead{Music} & \tblhead{Speech} & \tblhead{Cross-domain} & \tblhead{Primary Focus} \\
    \midrule
    Clotho-AQA~\citep{lipping2022clothoaqa} & \checkmark & \crossmark & \crossmark & \crossmark & Audio question answering \\
    AudioCaps~\citep{kim2019audiocaps} & \checkmark & \crossmark & \crossmark & \crossmark & Grounded audio captioning \\
    MuChoMusic~\citep{weck2024muchomusic} & \crossmark & \checkmark & \crossmark & \crossmark & Music understanding \\
    MusicBench~\citep{melechovsky2024mustango} & \crossmark & \checkmark & \crossmark & \crossmark & Music instruction following \\
    CMI-Bench~\citep{ma2025cmiBench} & \crossmark & \checkmark & \crossmark & \crossmark & Comprehensive music understanding \\
    LibriSQA~\citep{zhao2024librisqa} & \crossmark & \crossmark & \checkmark & \crossmark & Spoken question answering \\
    Dynamic-SUPERB~\citep{huang2024dynamicsuperb} & \crossmark & \crossmark & \checkmark & \crossmark & Robust speech interaction \\
    AudioBench~\citep{wang2025audiobench} & \checkmark & \crossmark & \checkmark & \checkmark & Broad audio understanding \\
    AIR-Bench~\citep{yang2024airbench} & \checkmark & \checkmark & \checkmark & \checkmark & Instruction following and reasoning \\
    MMAU~\citep{sakshi2024mmau} & \checkmark & \checkmark & \checkmark & \checkmark & Multiple-choice audio understanding \\
    MMAR~\citep{ma2025mmar} & \checkmark & \checkmark & \checkmark & \checkmark & Multi-step audio reasoning \\
    \bottomrule
    \end{fullwidthtabular}
\end{table}


\subsubsection{Unified Visual Understanding and Generation}
\label{sec:inter:av_conver:visual}

\paratitle{Definition.}
Unified visual understanding and generation seeks one model that can both interpret and synthesize images or videos. The appeal is architectural consistency: a shared representation can support perception, editing, and generation within a single conversational loop.

\paratitle{Representative Tasks and Methods.}
We discuss unified image and video modeling separately.

\textit{Unified image understanding and generation.}
Recent image models can be grouped by where understanding and generation are coupled:

\qquad\textit{(1) MLLM-to-diffusion bridging.}
This line keeps an MLLM as the semantic core and links it to a diffusion or DiT-style generator through an explicit interface. Representative examples include MetaQuery~\citep{pan2025metaquery}, BLIP3o-NEXT~\citep{chen2025blip3o}, and Skywork UniPic 2.0~\citep{wei2025skywork}. These models retain strong reasoning and generation modules, but the interface can limit deeper transfer.

\qquad\textit{(2) Block-level fusion of MLLM and diffusion-style generation modules.}
A second family couples understanding and generation more tightly by sharing transformer computation rather than using only a lightweight connector. BAGEL~\citep{deng2025emerging}, JanusFlow~\citep{ma2025janusflow}, and Show-o2~\citep{xie2025show} are representative examples.

\qquad\textit{(3) Native all-modal next-token autoregression.}
A third line maps all modalities into a shared token space and trains them with next-token prediction. Earlier representatives such as Chameleon~\citep{chameleonteam2024chameleon} and Janus-Pro~\citep{chen2025januspro} established this route. Recent work, including SelfTok~\citep{wang2025selftok}, Ming-UniVision~\citep{huang2025ming}, and LongCat-Next~\citep{team2026longcat}, further strengthens it with improved tokenization and interleaved multimodal generation.

\qquad\textit{(4) Text autoregression with visual masked autoregression.}
Another line keeps causal autoregression for text but adopts masked autoregressive modeling for images. Harmon~\citep{wu2025harmonizing} is a representative example, and Skywork UniPic 1.0~\citep{wang2025skywork} is closely related. This design keeps a unified semantic framework without forcing all modalities to share the same decoding rule.

\qquad\textit{(5) Masked discrete diffusion.}
A newer line replaces autoregression with mask-based discrete diffusion. Omni-Diffusion~\citep{li2026omni} is the clearest representative, while LLaDA-o~\citep{you2026llada} further develops this direction by combining discrete masked diffusion for understanding with continuous diffusion for visual generation.

\textit{Unified video understanding and generation.}
Compared with the image side, unified video modeling remains less diverse and is still dominated by hybrid designs. Most current systems pair an MLLM with a dedicated video generator, as in Omni-Video~\citep{tan2025omni}, UniVid~\citep{luo2025univid}, and UniVideo~\citep{wei2025univideo}. Tokenizer-centric work such as Divot~\citep{divot2025} explores shared video representations for both comprehension and generation, while Uni-ViGU~\citep{qin2026uni} moves toward a generation-centric design. Overall, unified video modeling is still at an early stage.

\paratitle{Benchmarks.}
Image understanding is commonly tested on COCO Captioning~\citep{chen2015microsoft}, NoCaps~\citep{agrawal2019nocaps}, VQAv2~\citep{goyal2017making}, GQA~\citep{hudson2019gqa}, ScienceQA~\citep{saikh2022scienceqa}, MMBench~\citep{liu2024mmbench}, and SEED-Bench~\citep{li2023seed}, while image generation uses MS-COCO~\citep{ms-coco-2014}, DrawBench~\citep{saharia2022photorealistic}, and GenEval~\citep{ghosh2023geneval}. Video understanding uses MSVD-QA~\citep{xu2017video}, ActivityNet-QA~\citep{yu2019activitynet}, and TVQA~\citep{lei2018tvqa}; video generation is typically measured on WebVid~\citep{Bain21}, MSR-VTT~\citep{xu2016msrvtt}, and VBench~\citep{huang2024vbench}. What is still lacking is a benchmark that jointly punishes failures in both understanding and generation over long visual contexts.

\begin{differencebox}[\faLightbulbO~Key Difference: Quality Still Favors Hybrids]
The trade-off is straightforward: autoregressive unification offers conceptual simplicity, whereas hybrid systems offer better visual quality. In practice, most high-performing systems now accept extra complexity because the quality gap remains large, especially for video. A promising middle ground is emerging around continuous or latent-token autoregression, which keeps the single-model view while relaxing the worst bottlenecks of discrete visual tokenization.
\end{differencebox}


\subsubsection{Omni-Modal Audio-Visual Conversation}
\label{sec:inter:av_conver:av}

\paratitle{Definition.}
Omni-modal audio-visual conversation concerns dialogue systems that jointly ground responses in both what the model sees and what it hears, while returning text or, increasingly, streaming speech. Compared with video chat models that answer in text only, the target here is end-to-end or tightly integrated audiovisual interaction, often under low-latency, multi-turn, and real-time constraints.

\paratitle{Methods.}
Recent work in this area follows two converging lines.

The first line focuses on omni assistants for real-time audio-visual dialogue with speech output. Early open efforts moved from the speech-centric Mini-Omni~\citep{xie2024miniomni} to the visual-audio Mini-Omni2~\citep{xie2024miniomni2}, and then to stronger vision-speech systems such as VITA-1.5~\citep{fu2025vita}, InteractiveOmni~\citep{tong2025interactiveomni}, and Baichuan-Omni-1.5~\citep{li2025baichuanomni15}. More recent omni models, including GPT-4o~\citep{hurst2024gpt}, Gemini 3.1 Flash~\citep{gemini31_flash_live}, Qwen2.5-Omni~\citep{xu2025qwen2}, Qwen3-Omni~\citep{xu2025qwen3omni}, Ming-Omni~\citep{ai2025ming}, and LongCat-Flash-Omni~\citep{wang2025longcat}, further unify video, audio, and speech generation within a single conversational stack. Architecturally, these systems usually combine visual and audio encoders with an LLM backbone and a streaming speech decoder, with representative designs such as block-wise streaming encoders, interleaved audio-video tokenization, and Thinker-Talker style decoupling between reasoning and speech synthesis.

The second line studies unified understanding-generation models, where conversation is treated as one instance of broader any-to-any multimodal interaction. MultiDialog~\citep{park2024let} introduced an early direct face-to-face dialogue model that maps audio-visual speech to audio-visual speech without relying on intermediate text. VideoPoet~\citep{kondratyuk2024videopoet}, NExT-GPT~\citep{wu2024next}, and Unified-IO 2~\citep{lu2024unified} broadened this agenda by leveraging pretrained components or autoregressively modeling multimodal inputs and outputs across video, audio, image, and text, making generation and comprehension part of one framework. More recent systems, such as JavisGPT~\citep{liu2025javisgpt} and X-Streamer~\citep{xie2025x}, move closer to genuine omni-modal audio-visual conversation by coupling sounding-video comprehension with synchronized audiovisual generation and real-time interaction. AR-Omni~\citep{cheng2026ar} is also relevant as a fully autoregressive bridge toward unified multimodal generation, although its current formulation mainly centers on text, image, and speech rather than strict video-grounded dialogue.

\paratitle{Benchmarks.}
Naturalistic AV dialogue data is still scarce. MultiDialog~\citep{park2024let} provides 340 hours of face-to-face conversation with aligned video, audio, and affective annotations, while AVInstruct~\citep{ye2024cat} and Video-ChatGPT~\citep{maaz2024video} expand instruction-style supervision. 
For evaluation, Video-MME~\citep{fu2025video} probes long-video reasoning with audio transcripts, and OmniVideoBench~\citep{li2025omnivideobench} targets omni-modal AV understanding more explicitly. Even with these additions, open-ended multi-turn AV dialogue remains under-benchmarked.
Very recently, AVI-Bench~\citep{wang2026avibench} categorizes the evaluation of audio-visual tasks into three stages: perception, understanding, and reasoning, inspired by human perceptual processes.

\begin{frontierbox}[\faLightbulbO~Open Problem: Real Dialogue Is Missing]
The biggest bottleneck is still data, not architecture. Synthetic instruction data scales, but it rarely captures authentic conversational timing, grounded audio events, or the spontaneity of human-human interaction. Until naturalistic AV dialogue corpora become much larger, omni-modal conversation will likely continue to underperform its architectural promise.
\end{frontierbox}

\subsection{Interactive Audio-Visual Embodiment}
\label{sec:inter:av_embodi}

\begin{figure*}[!t]
    \centering
    \includegraphics[width=\textwidth]{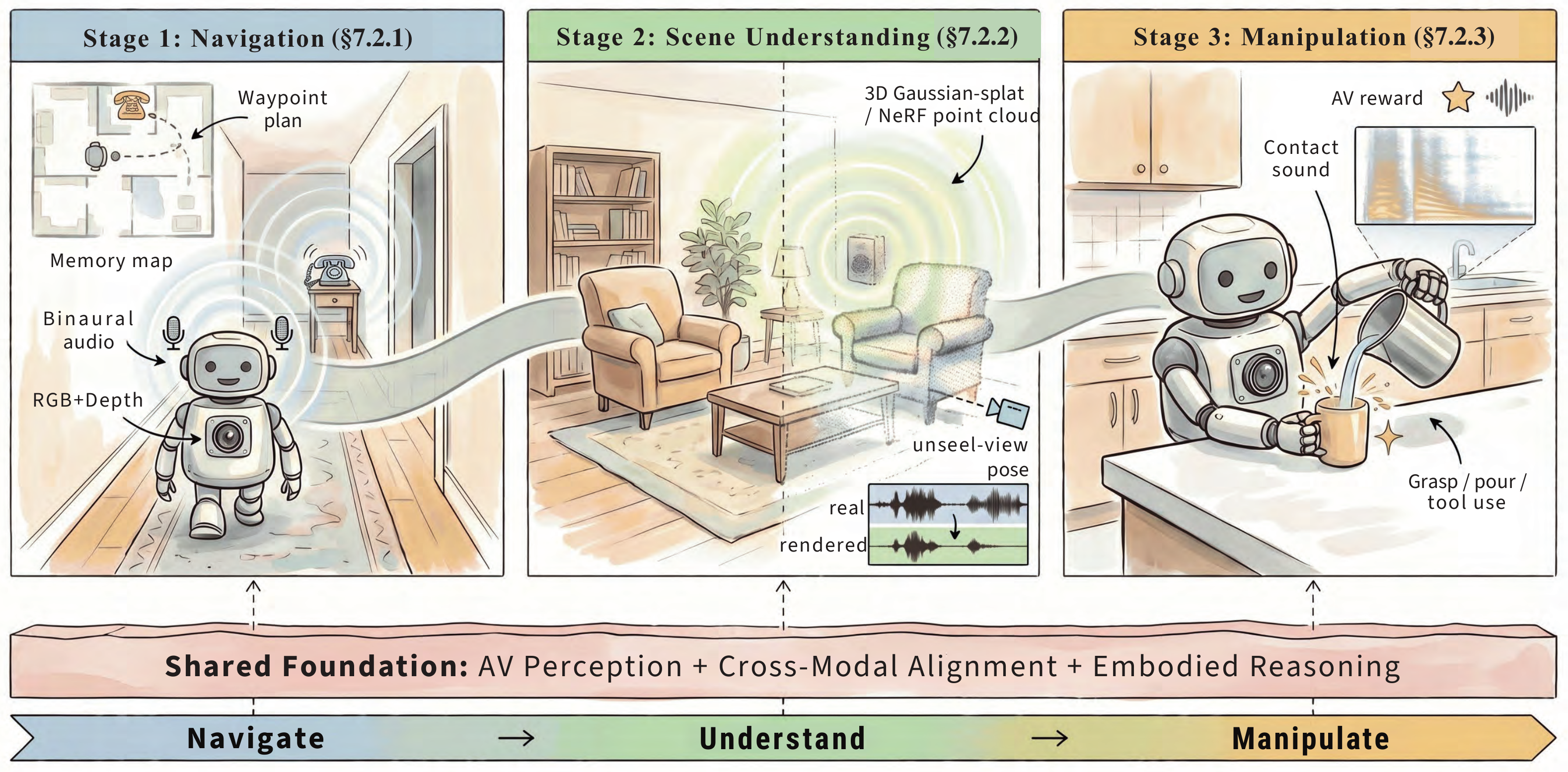}
    \caption{Organization of audio-visual embodied interaction. We organize this subsection into navigation, scene understanding, and manipulation, grounded in the shared foundations of audio-visual perception, cross-modal alignment, and embodied reasoning.}
    \label{fig:av_embodi_overview}
\end{figure*}

Embodied AV intelligence grounds multimodal perception in action. We organize this subsection around three abilities: navigating toward sound sources, reconstructing environments from visual and acoustic cues, and manipulating objects using contact sounds as feedback, as summarized in~\cref{fig:av_embodi_overview}. Across all three, the key challenge is turning multimodal understanding into control policies that remain robust outside clean simulation.

\subsubsection{Audio-Visual Navigation}
\label{sec:inter:av_embodi:av_navig}

\paratitle{Definition.}
Audio-visual navigation (AVN) asks an embodied agent to reach a sound-emitting target by jointly using what it sees and hears. Sound complements vision in two ways: it exposes out-of-view goals and carries structural cues through reverberation and propagation. The standard task variants are \textit{AudioGoal}, where the sound itself defines the target, and \textit{AudioPointGoal}, where a coarse directional hint is also provided~\citep{ChenJSGAIRG20SoundSpaces}.

\paratitle{Methods.}
AVN has evolved through several complementary lines. \emph{1) End-to-end RL} methods~\citep{li2025audioDynamicModality,ChenWLLY23,ChenMAGRG21AVWaN,ChenJSGAIRG20SoundSpaces} fuse visual and binaural features and learn recurrent navigation policies directly from reward. \emph{Memory-augmented and hierarchical} methods~\citep{ChenMAGRG21AVWaN,GanZ0GT20,li2023transformermemory} introduce waypoint planning, spatial memory, or transformer recurrence to reduce myopic behavior.
\emph{2) Semantic navigation}~\citep{ChenAG21SAVi} shifts the task from tracking an active sound source to reasoning about semantic sound events whose source may become silent. \emph{Robust navigation} further addresses noise, moving targets, and adversarial conditions through dynamic modality weighting, stereo-aware fusion, and explicit robustness objectives~\citep{zhao2025audioAntiBacktracking,shi2025towardsNoisyEnvironments,wang2025boostingChannelAttention,li2025audioDynamicModality,younes2022dynamical,yu2022soundadversarial}.
\emph{3) Language-augmented} systems decouple perception from planning. AVLEN~\citep{paul2022avlen} introduces language queries, and RILA-style agents~\citep{yang2024rila,liu2024caven} use higher-level reasoning and self-correction to improve zero-shot generalization.
\emph{4) Sim-to-real and acoustic-field modeling} address the gap between clean simulation and physical deployment.
SoundSpaces~2.0~\citep{chen2022soundspaces20} improves continuous acoustic rendering, RAF~\citep{ChenGR0LOR24RAF} provides real dense room impulse responses, Sonicverse~\citep{gao2023sonicverse} supports multisensory embodied simulation, and noisy navigation settings such as BeDAViN~\citep{shi2025towardsNoisyEnvironments} stress robustness under more realistic sound conditions.

\paratitle{Benchmarks.}
SoundSpaces~\citep{ChenJSGAIRG20SoundSpaces} provides geometric acoustic simulation over Replica and Matterport3D scenes; its PanoIR extension adds panoramic impulse responses across 100+ scenes; and SoundSpaces~2.0~\citep{chen2022soundspaces20} supports continuous acoustic rendering on arbitrary 3D meshes.
Real Acoustic Fields (RAF)~\citep{ChenGR0LOR24RAF} offers real-world dense RIRs with multi-view images and 6DoF pose tracking. Sonicverse~\citep{gao2023sonicverse} is a multimodal simulation platform for embodied AV tasks, and BeDAViN~\citep{shi2025towardsNoisyEnvironments} extends standard benchmarks with realistic noise conditions.

\begin{table}[!ht]
    \centering
    \caption{Summary of AVN benchmarks.}
    \label{tab:bench_av_embodi_nav}
    \benchmarktablestyle
    \setlength{\tabcolsep}{2pt}%
    \begin{fullwidthtabular}{16}{l c c c c c c c}
        \toprule
        \benchhead \tblhead{Dataset} & \tblhead{\#Scenes} & \tblhead{Resolution} & \tblhead{\shortstack{Sampling\\Rate}} & \tblhead{\shortstack{Real-\\world}} & \tblhead{\shortstack{Avg.\\Area}} & \tblhead{\shortstack{\#Training\\Episodes}} & \tblhead{\shortstack{\#Test\\Episodes}} \\
        \midrule
        Replica~\citep{ChenJSGAIRG20SoundSpaces} & 18 & $0.5\text{m}$ & $44100\text{Hz}$ & \crossmark & $47.24\text{m}^2$ & $0.1\text{M}$ & 1000 \\
        Matterport3D~\citep{ChenJSGAIRG20SoundSpaces} & 85 & $1\text{m}$ & $16000\text{Hz}$ & \crossmark & $517.34\text{m}^2$ & $2\text{M}$ & 1000 \\
        RAF~\citep{ChenGR0LOR24RAF} & 2 & Continuous & $48000\text{Hz}$ & \checkmark & $45.1\text{m}^2$ & 3000 & 500 \\
        Sonicverse~\citep{gao2023sonicverse} & 103 & Continuous & $44100\text{Hz}$ & \crossmark & $215.3\text{m}^2$ & $5\text{M}$ & 1000 \\
        SoundSpaces 2.0~\citep{chen2022soundspaces20} & 103 & Continuous & $16000\text{Hz}$ & \crossmark & $435.4\text{m}^2$ & $10\text{M}$ & 1000 \\
        BeDAViN~\citep{shi2025towardsNoisyEnvironments} & 85 & $1\text{m}$ & $96000\text{Hz}$ & \crossmark & $517.3\text{m}^2$ & $1.5\text{M}$ & 3000 \\
        \bottomrule
    \end{fullwidthtabular}
\end{table}

\begin{table}[!t]
    \centering
    \caption{Audio-visual navigation performance on SoundSpaces~\citep{ChenJSGAIRG20SoundSpaces}. Top: \textit{AudioGoal} (continuous sound) on Replica (Heard). Bottom: \textit{Semantic AudioGoal} (intermittent sound) on MP3D (Unheard). The two task settings differ substantially in difficulty. SPL and SR are in \%. $\ddagger$: zero-shot (no training trajectories). Best per-task results in \textbf{bold}.}
    \label{tab:perf_av_nav}
    \performancetablestyle
    \setlength{\tabcolsep}{5pt}
    \begin{fullwidthtabular}{8}{l l c c}
    \toprule
    \perfhead \tblhead{Method} & \tblhead{Paradigm} & \tblhead{SPL$\uparrow$} & \tblhead{SR$\uparrow$} \\
    \midrule
    \perfgroup{4}{$\blacktriangleright$ AudioGoal: Replica (Heard)}
    \quad SoundSpaces~\citep{ChenJSGAIRG20SoundSpaces} & End-to-End RL & 74.4 & 91.4 \\
    \quad AV-WaN~\citep{ChenMAGRG21AVWaN} & Waypoint Nav & \textbf{86.6} & \textbf{98.7} \\
    \perfgroup{4}{$\blacktriangleright$ Semantic AudioGoal: MP3D (Unheard)}
    \quad SoundSpaces~\citep{ChenJSGAIRG20SoundSpaces} & End-to-End RL & 15.5 & 16.5 \\
    \quad AV-WaN~\citep{ChenMAGRG21AVWaN} & Waypoint Nav & 13.2 & 17.2 \\
    \quad SAVi~\citep{ChenAG21SAVi} & Semantic Nav & 17.2 & 24.8 \\
    \quad AVLEN~\citep{paul2022avlen} & Language-Aug & 17.6 & 26.2 \\
    \quad RILA$^\ddagger$~\citep{yang2024rila} & LLM Agent & 11.8 & \textbf{35.4} \\
    \bottomrule
    \end{fullwidthtabular}
\end{table}

\begin{frontierbox}[\faLightbulbO~Open Problem: Agents Need Grounding]
AVN is shifting from reactive control toward deliberative control. LLM-based agents bring stronger priors about rooms, objects, and human goals, which improves semantic generalization, but they are only useful when tightly grounded in low-latency sensory feedback. The next gains will likely come from better interfaces between symbolic planning and continuous audio-visual control, not from larger language models alone.
\end{frontierbox}

\subsubsection{Audio-Visual Scene Understanding and Reconstruction} 
\label{sec:inter:av_embodi:av_scene}

\paratitle{Definition.}
In the embodied setting, audio-visual scene understanding and reconstruction aim to build an \emph{actionable} spatial representation for an agent rather than a purely descriptive scene label. The goal is to estimate geometry, acoustic response, and language-grounded semantics tightly enough that a robot can localize sound sources, reason about out-of-view structure, and update its spatial memory while acting~\citep{chen2022soundspaces20,luo2022learning}. This is therefore narrower than generic AV segmentation or event classification: the output must be useful for navigation, manipulation, or world modeling.

\paratitle{Representative Tasks and Methods.}
Related approaches can be categorized as follows:

\textit{Echo-based geometry and acoustic reconstruction.}
Early work such as BatVision, Beyond2D, and few-shot acoustic reconstruction~\citep{christensen2020batvision,parida2021beyond,majumder2022few} showed that echoes can recover layout cues when vision is degraded.
In embodied AV systems, this line has evolved into queryable acoustic fields: SoundSpaces~2.0~\citep{chen2022soundspaces20} provides the simulation substrate, and Neural Acoustic Fields (NAF)~\citep{luo2022learning} represent room acoustics as continuous functions that agents can query for localization, reverberation, or planning.

\textit{Joint visual-acoustic scene fields.}
Recent work explicitly couples geometry, appearance, and acoustics in one scene representation. AV-NeRF~\citep{liang2024avnerf}, AV-GS~\citep{bhosale2024avgs}, NeRAF~\citep{cai2024neraf}, and related audio-aware neural field models move toward physically grounded visual-acoustic rendering. The important embodied point is not photorealism by itself, but whether the learned field stays spatially consistent enough to support downstream decisions under viewpoint change.

\textit{Language-grounded spatial memory.}
Audio-Visual Language Maps~\citep{huang2024avlmaps} make this actionability explicit by grounding language, vision, and audio in a shared 3D frame that can be queried during navigation. The extension to Multimodal Spatial Language Maps~\citep{huang2025mslmaps} broadens the same map abstraction from navigation toward manipulation, suggesting that embodied AV scene understanding is converging on reusable spatial memory rather than task-specific perception heads.

\textit{Toward interactive AV world models.}
The frontier is shifting from static maps to updateable world models. X-Streamer~\citep{xie2025x} is an example of unified audiovisual world modeling, and even though it is not limited to robot mapping, it points to the next step for embodiment: scene representations that evolve with interaction, preserve long-horizon audiovisual memory, and expose state variables useful for control.

\paratitle{Benchmarks.}
Embodied evaluation should reflect spatial usefulness rather than generic AV recognition. 
The most relevant resources are therefore simulators, acoustic-field datasets, and map-based evaluation settings: SoundSpaces~2.0~\citep{chen2022soundspaces20} and SonicVerse~\citep{gao2023sonicverse} support embodied simulation, RAF~\citep{ChenGR0LOR24RAF} measures real acoustic reconstruction, and language-grounded map settings such as AVLMaps/MSLMaps~\citep{huang2024avlmaps,huang2025mslmaps} test whether the representation can support embodied queries and downstream success. 
Typical metrics include localization error, RT60 or field-reconstruction error, success rate, and task-completion metrics on navigation/manipulation splits.

\begin{table}[!t]
    \centering
    \caption{Representative resources and evaluation settings for audio-visual scene understanding and reconstruction. The emphasis is on spatial memory and field estimation rather than generic AV perception.}
    \label{tab:bench_av_scene}
    \benchmarktablestyle
    \begin{fullwidthtabular}{8}{l l l l}
        \toprule
        \benchhead \tblhead{Resource} & \tblhead{Signal} & \tblhead{Use} & \tblhead{Metrics} \\
        \midrule
        SoundSpaces~2.0~\citep{chen2022soundspaces20} & 3D scans + sim audio & AV sim & SPL/SR + loc err \\
        RAF~\citep{ChenGR0LOR24RAF} & Real RIRs + RGB & Acoustic recon & RT60 + loc/recon err \\
        SonicVerse~\citep{gao2023sonicverse} & Sim AV rooms & Sim2real & Success + acoustic err \\
        AVLMaps / MSLMaps~\citep{huang2024avlmaps,huang2025mslmaps} & AV-lang maps & Grounded QA & Success + grounding \\
        \bottomrule
    \end{fullwidthtabular}
\end{table}

\begin{differencebox}[\faLightbulbO~Key Difference: State Must Guide Action]
This subsection should not collapse into sounding-object segmentation or generic AV scene parsing. For embodiment, the useful representation is the one that helps an agent decide where to move, what is out of view, and how the environment will sound after an action. That is why map-like spatial memory, queryable acoustic fields, and world-model interfaces matter more here than standalone AV classification accuracy.
\end{differencebox}

\subsubsection{Audio-Visual Embodiment Interaction and Manipulation}
\label{sec:inter:av_embodi:av_action}
\vspace{-3mm}

\paratitle{Definition.}
Audio-visual embodiment interaction and manipulation study how agents use sound together with vision for contact-rich control. Contact sounds reveal events such as slip, collision, pouring progress, or material change that are often ambiguous in RGB alone, making audio a useful proxy for tactile feedback.
Acoustic sensing can be passive, where the agent listens to naturally occurring contact sounds, or active, where the agent emits vibrations or probing signals and interprets the returned acoustic response.

\vspace{-2mm}
\paratitle{Representative Tasks and Methods.}
Research has evolved through four related directions. \emph{Contact representation and feature enhancement}~\citep{mejia2024hearingtouch,li2022seehearfeel,wang2022audiovisualgrounding} use audio to enrich visual perception, especially for contact events that are visually subtle.
\emph{Active and passive acoustic sensing} studies how contact sounds or controlled acoustic probing reveal object state, material, and interaction dynamics; SonicSense~\citep{yu2024sonicsense} is representative of in-hand acoustic object perception.
\emph{World-model and generative approaches}~\citep{zhang2025learningAudioWorld, huang2025unifiedForceful, wang2025soundSimulation, yi2024visualauditory} integrate audio into diffusion policies or audio-centric predictive models so that agents can anticipate physical interactions.
\emph{VLA integration}~\citep{zhao2025vlas,wei2025audioVLA,wang2024allinone} injects auditory tokens directly into vision--language--action backbones for closed-loop multimodal control.

\vspace{-2mm}
\paratitle{Benchmarks.}
Benchmarks are now appearing across both real and simulated settings. ManiWAV~\citep{liu2024maniwav} collects in-the-wild AV demonstrations through an ear-in-hand setup, ARIO~\citep{wang2024allinone} aggregates multi-robot trajectories into a unified format, Kaiwu~\citep{jiang2025kaiwu} targets real-world assembly with rich synchronized sensing, and AV grounding-and-act~\citep{wang2022audiovisualgrounding} stresses perception-action coupling. Audio-VLA~\citep{wei2025audioVLA} augments RLBench and LIBERO with collision audio and introduces Task Completion Rate (TCR) for dynamic process tracking, while sim-to-real evaluation increasingly uses intentionally hard-to-simulate audio~\citep{wang2025soundSimulation}.

\begin{table}[!t]
    \centering
    \caption{Representative benchmarks for audio-visual embodied manipulation.}
    \label{tab:bench_av_embodi_manip}
    \benchmarktablestyle
    \begin{fullwidthtabular}{10}{l c l l l}
        \toprule
        \benchhead \tblhead{Benchmark} & \tblhead{Setting} & \tblhead{Modalities} & \tblhead{Scale} & \tblhead{Primary Use} \\
        \midrule
        ManiWAV~\citep{liu2024maniwav} & \realcell & RGB, audio, state & In-wild demos & Contact manipulation \\
        ARIO~\citep{wang2024allinone} & \realsimcell & Multimodal robot traj. & Millions episodes & Embodied pretraining \\
        Kaiwu~\citep{jiang2025kaiwu} & \realcell & Audio, video, mocap & Assembly data & Industrial manipulation \\
        Audio-VLA~\citep{wei2025audioVLA} & \simcell & RGB, audio, action & RLBench/LIBERO & TCR process eval \\
        \bottomrule
    \end{fullwidthtabular}
\end{table}

\begin{frontierbox}[\faLightbulbO~Open Problem: Simulation Still Misleads]
Sim-to-real transfer remains the defining obstacle for embodied AV systems. Visual simulation is already imperfect; acoustic simulation is usually worse, especially for contact-rich manipulation where tiny sound differences signal large changes in task state. Progress will depend on better acoustic rendering, more domain-invariant representations, and training recipes that use simulation for scale without overfitting to its unrealistic sound statistics.
\end{frontierbox}

\begin{figure}[!t]
    \centering
    \includegraphics[width=\linewidth]{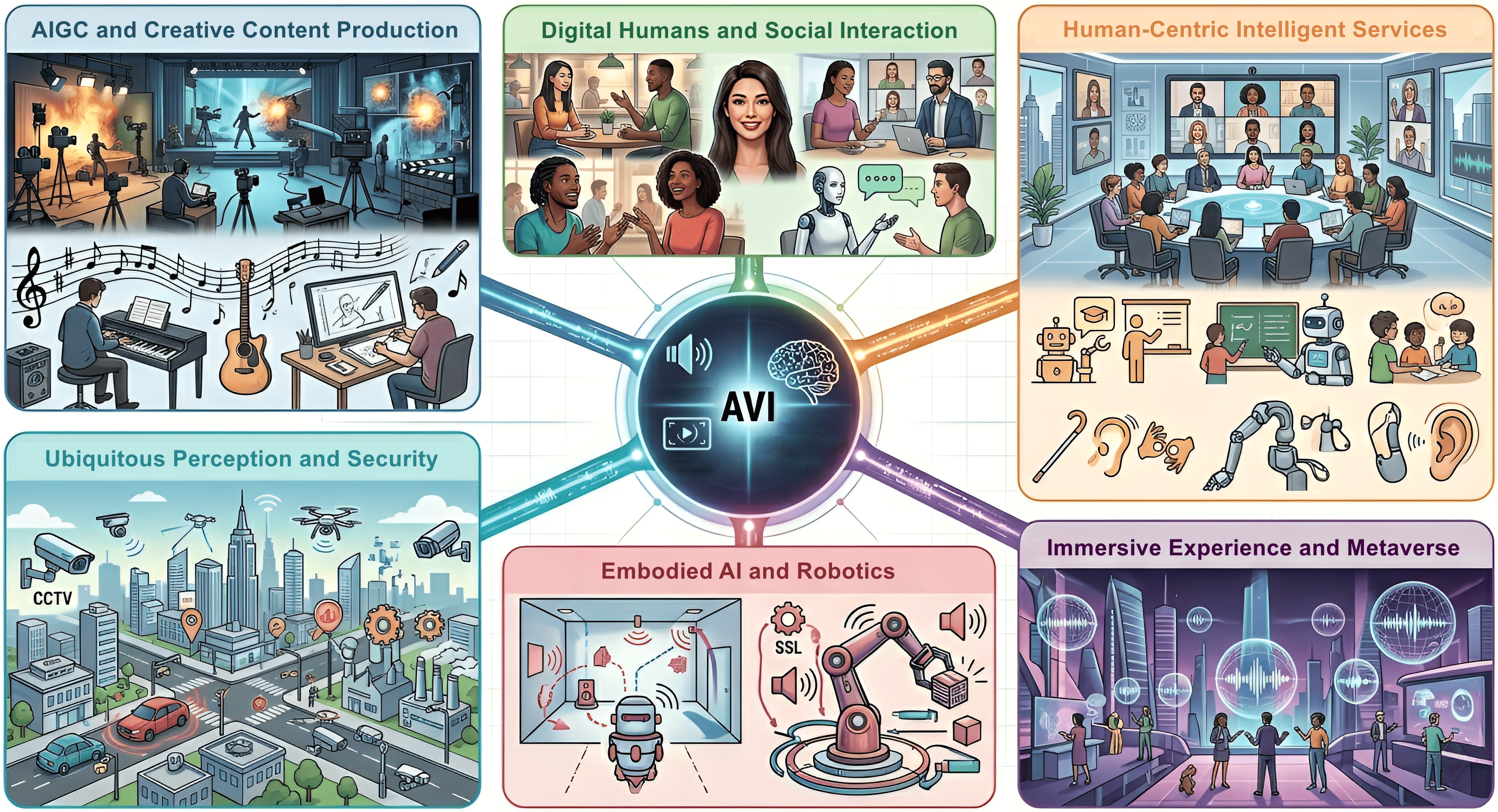}
    \caption{The landscape of AVI applications. We here highlight how audio-visual foundation models act as a core intelligence layer for AIGC, human-centric services, immersive environments, and ubiquitous perception, etc.}
    \label{fig:application}
\end{figure}

\section{Applications}
\label{sec:app}

Audio-visual intelligence powered by foundation models has enabled a wide range of real-world applications spanning creative industries, human-computer interaction, immersive technologies, and embodied systems. This section surveys the major domains where these advances have shown substantial impact.

\subsection{AIGC and Creative Content Production}

Generative tools are already changing how video and sound are produced, from automated Foley and short-form co-creation to long-standing workflows such as post-production, music scoring, and asset iteration.
The discussion below links foundation-model capabilities in generation and editing to these creative pipelines, using film and game-oriented examples before broadening to music and sound design.

\paratitle{Film and Video Post-Production.}
The film and video production industry has been transformed by foundation models capable of generating and manipulating audio-visual content~\citep{ruan2023mm,liu2025javisdit}. Foley synthesis, traditionally reliant on manual labor and specialized studios, can now be automated by video-to-audio models such as Diff-Foley~\citep{comunita2024diff}, MMAudio~\citep{cheng2025mmaudio}, and FoleyCrafter~\citep{zhang2024foleycrafter}, which generate temporally aligned sound effects and ambient audio directly from silent video, reducing post-production time and cost. These models also support dubbing and voice-over workflows: speech-driven talking head generation~\citep{prajwal2020lip,ye2023geneface} enables cross-lingual lip-synchronized dubbing, while voice conversion systems~\citep{popov2022diffusion} preserve emotional nuance during voice transformation. More recently, joint text-to-audio-video generation models such as JavisDiT~\citep{liu2025javisdit} and commercial systems including Veo-3 and Seedance~\citep{chen2025seedance} have opened the possibility of generating complete scenes with synchronized audio and visuals from text, though current performance is still strongest for short-form content.

\paratitle{Music and Sound Design.}
Foundation models have also reshaped music production and sound design workflows~\citep{copet2023simple,borsos2023audiolm,agostinelli2023musiclm}. Text-to-audio systems such as AudioLDM~\citep{liu2023audioldm}, MusicGen~\citep{copet2023simple}, and Stable Audio allow creators to rapidly prototype audio from natural language prompts, while source separation models including TF-GridNet~\citep{wang2023tf} and Band-Split RNN~\citep{luo2023music} can isolate instruments from mixed recordings for karaoke generation, remixing, and restoration. These capabilities are increasingly integrated into digital audio workstations and video editing tools~\citep{kreuk2022audiogen}. In addition, audio manipulation models support style transfer, inpainting of corrupted segments~\citep{moliner2023inpainting}, and instruction-based editing~\citep{wang2023instructaudioedit}. In gaming, procedural sound generation conditioned on game state and visual context reduces dependence on large prerecorded libraries while enabling more dynamic and context-aware audio experiences.

\subsection{Digital Humans and Social Interaction}

Synthesized faces, bodies, and consistent personas sit at the boundary of entertainment, communication, and service interfaces; audio-driven motion and identity-locked output are the usual technical levers.
We first sketch avatar and talking-head technology, then comment on applications that require the same person (or character) to persist across sessions and media.

\paratitle{Talking Head and Avatars.}
Digital human synthesis has progressed from simple 2D lip-syncing to photorealistic 3D avatar generation~\citep{cudeiro2019voca,richard2021meshtalk}. Early methods such as Wav2Lip~\citep{prajwal2020lip} focused on robust audio-lip synchronization at the pixel level, whereas newer approaches use explicit 3D representations. Mesh-based methods such as FaceFormer~\citep{fan2022faceformer} and CodeTalker~\citep{xing2023codetalker} animate parametric face models, while neural rendering approaches, including AD-NeRF~\citep{guo2021adnerf} and GaussianTalker~\citep{zhang2024gaussiantalker}, achieve photorealistic synthesis through implicit representations. Recent work further addresses the ``dead-face'' problem~\citep{ji2021evp} by separating emotional state from speech content. Models such as DreamTalk~\citep{ma2023dreamtalk} and EmoGene~\citep{wang2025emogene} improve expressive control, and full-body systems such as EMAGE~\citep{liu2024emage} and Stereo-Talker~\citep{deng2024stereotalker} generate synchronized speech, facial expressions, and co-speech gestures from a single reference image.

\paratitle{Persona-Consistent Interaction.}
Maintaining identity consistency across interactions is essential for virtual influencers, customer service avatars, and personalized AI companions~\citep{guan2025audcast,yi2023talkshow}. Foundation models increasingly support one-shot or few-shot avatar creation, where a small set of reference images or videos establishes identity priors that persist across generated content. Real3D-Portrait~\citep{ye2024real3dportrait} reconstructs animatable 3D portraits from a single image, while identity-preserving video generation maintains consistent appearance across varying poses and expressions~\citep{shen2023difftalk}. These capabilities enable applications such as virtual influencer marketing, realistic telepresence avatars~\citep{wang2025choreomuse}, and sign language synthesis for accessibility. Cross-lingual voice conversion and lip-sync adaptation further expand digital human applications to multilingual settings~\citep{zhang2021hdtf}.

\subsection{Human-Centric Intelligent Services}

End-user and enterprise settings reward systems that can listen, watch, and speak on human timescales: meetings, education, and accessibility are representative stress tests for robust multimodal input and output.
Assistants and meeting analytics, learning and training, and inclusive design illustrate that breadth without duplicating the technical review earlier in the survey.

\paratitle{Conversational Assistants and Meeting Intelligence.}
Audio LLMs such as Qwen-Audio~\citep{chu2023qwenaudio, chu2024qwen2audio} and SALMONN~\citep{tang2023salmonn} expand conversational AI from text to native audio understanding. By processing speech directly without intermediate ASR transcription, they preserve paralinguistic cues such as emotion, emphasis, and speaker identity~\citep{deshmukh2023pengi,ghosh2024gama}. Omni models such as GPT-4o~\citep{hurst2024gpt} and Qwen-Omni~\citep{xu2025qwen2, xu2025qwen3omni} further integrate real-time audio-visual perception and generation, enabling more natural multimodal dialogue~\citep{tong2025interactiveomni}. In parallel, meeting intelligence systems use audio-visual understanding for speaker diarization, action item extraction, and conference summarization~\citep{nagrani2017voxceleb,chung2018voxceleb2}. Multimodal fusion improves speaker identification by combining vocal and facial cues~\citep{desplanques2020ecapa}, while emotion recognition supports analysis of meeting dynamics in applications such as collaboration analytics and customer-facing services~\citep{ma2024emotion2vec}.

\paratitle{Education and Training.}
Interactive educational content benefits from both audio-visual generation and understanding~\citep{liu2024improved,li2023blip}. AI tutors can adapt presentation styles according to learner engagement inferred from audio-visual signals~\citep{wu2025avf}, while generated explanatory videos with synchronized narration and demonstrations support self-paced learning. Language learning applications further use lip-sync generation~\citep{prajwal2020lip} and audio-driven avatar teachers~\citep{fan2022faceformer} to demonstrate pronunciation and articulatory movements. In high-stakes training domains such as medicine, aviation, and emergency response, audio-visual generation can produce realistic scenarios without expensive physical infrastructure~\citep{gao2023sonicverse,chen2022soundspaces20}. Audio-visual analysis also enables automated feedback on communication, procedural performance, and situational awareness~\citep{radford2023robust}.

\paratitle{Accessibility and Inclusive AI.}
Audio-visual foundation models significantly improve accessibility for users with sensory impairments~\citep{drossos2020clotho,mei2021audiocaptioning}. Audio captioning systems~\citep{mei2021audiocaptioning} generate descriptions of acoustic scenes for deaf and hard-of-hearing users, while audio description generation helps blind and low-vision users understand visual content through synchronized narration~\citep{maaz2024video}. Audio Question Answering~\citep{lipping2022clothoaqa} enables natural-language interaction with sound scenes, and sign language synthesis combined with realistic avatar rendering supports communication for deaf communities~\citep{liu2024emage}. These applications also underscore the importance of inclusive training data and evaluation across diverse user groups and acoustic conditions~\citep{radford2023robust,chen2022wavlm}.

\subsection{Immersive Experience and Metaverse}

Immersive applications (e.g., XR/AR/VR) depend on sound and image staying coherent as the user moves, which ties rendering, spatial audio, and scene understanding more tightly than in flat media.

\paratitle{Spatial Audio and Scene Rendering.}
Immersive experiences require audio that responds naturally to user movement and environmental context~\citep{ChenJSGAIRG20SoundSpaces,chen2022soundspaces20}. Neural acoustic field methods~\citep{luo2022learning} render spatial audio as a function of listener position within reconstructed 3D scenes, enabling 6DoF audio with appropriate attenuation, occlusion, and room acoustics. AV-NeRF~\citep{liang2024avnerf} and related approaches jointly model visual appearance and acoustic propagation, allowing users to navigate virtual environments while hearing spatially consistent sound. These techniques also support virtual concerts and live events, where audiences can experience performances from different virtual viewpoints~\citep{majumder2022few}. However, deployment on head-mounted displays and spatial computing platforms requires low latency, often below 20ms, to avoid perceptual mismatch and motion sickness. Therefore, current research emphasizes efficient inference while preserving audio-visual consistency in dynamic and acoustically complex scenes~\citep{chen2022soundspaces20}.

\paratitle{3D-Aware AV Interaction.}
Audio-Visual Language Maps~\citep{huang2024avlmaps} and related representations enable semantic understanding of 3D environments through combined audio and visual sensing~\citep{ChenJSGAIRG20SoundSpaces,chen2022soundspaces20}. They support natural language queries about spatial scenes, such as locating a ringing phone, and enable more context-aware interaction in virtual environments. Echo-based 3D reconstruction further complements visual sensing by using acoustic reflections to infer room geometry when line-of-sight is limited~\citep{luo2022learning}. Metaverse applications additionally require seamless integration of user-generated content with synthesized environments~\citep{chen2022soundspaces20}. Models that place sound sources at plausible 3D locations with physically consistent propagation can support collaborative creation in shared virtual spaces, although challenges remain in maintaining consistent rendering across devices and network conditions while meeting the low-latency demands of presence and embodiment~\citep{gao2023sonicverse}.

\subsection{Embodied AI and Robotics}

Physical platforms use vision for geometry and affordances while sound often marks goals, off-screen events, and contact; the two modalities rarely substitute for one another in deployment.

\paratitle{Audio-Visual Navigation.}
Embodied agents operating in real-world environments benefit from multimodal perception that combines visual and acoustic information~\citep{ChenJSGAIRG20SoundSpaces,chen2022soundspaces20}. The SoundSpaces platform~\citep{ChenJSGAIRG20SoundSpaces} and follow-up work train agents to locate sound-emitting targets in complex 3D environments by using reverberation patterns, intensity gradients, and visual context~\citep{chen2022soundspaces20}. Semantic audio-visual navigation extends this setting to object-level goals, such as locating a ringing phone through joint reasoning over acoustic and visual evidence~\citep{huang2024avlmaps,yu2022soundadversarial}. For service robots and autonomous systems, acoustic sensing also complements vision in low-light conditions, under occlusion, and beyond the visual field~\citep{gao2023sonicverse}. Vision-Language-Action models~\citep{kimopenvla,zitkovich2023rt,black2024pi_0} further integrate multimodal perception into end-to-end control policies that respond to verbal commands, environmental sounds, and manipulation feedback.

\paratitle{Manipulation with Feedback.}
Acoustic contact sensing provides rich information about manipulation interactions that vision alone often cannot capture~\citep{gao2023sonicverse}. Sounds produced during contact can reveal surface material, slip events, and grasp stability, improving grasping, insertion, and assembly tasks when visual feedback is limited~\citep{team2024octo,li2024cogact}. Human-robot interaction likewise benefits from audio-visual understanding of speech, gesture, and facial expression~\citep{zheng2024tracevla,wen2025tinyvla}, enabling more natural collaboration~\citep{bjorck2025gr00t,shukor2025smolvla}. In safety-critical applications, current research emphasizes real-time reliability, failure detection~\citep{lin2025failsafe}, and sim-to-real transfer under changing acoustic conditions~\citep{gao2023sonicverse,cen2025rynnvla}.

\subsection{Ubiquitous Perception and Security}

Large-scale sensing for safety, industry, and the IoT still hinges on fusing what cameras and microphones can jointly establish about scenes, actors, and anomalies, often at the edge with tight compute budgets.

\paratitle{Smart City and Forensics.}
Audio-visual event detection systems support public safety and urban monitoring~\citep{tian2018audio}. Joint analysis of surveillance video and environmental audio improves detection of anomalous events such as accidents, altercations, and emergencies compared with single-modality systems~\citep{tian2018audio,bao2023cross}. Audio-visual scene understanding can also localize sound sources within visual scenes, supporting incident reconstruction and situational awareness for first responders~\citep{bao2023cross,tian2018audio}. In forensics, audio-visual analysis assists evidence authentication and investigation~\citep{chung2016outoftime,kumar2009robustaudiovisualspeechsynchronydetection}. Speaker recognition based on neural embeddings~\citep{desplanques2020ecapa} enables voice identification, while deepfake detection methods examine the consistency between lip movements and speech to identify manipulated media~\citep{chung2016outoftime}. As generation quality advances, detection methods must continue to improve to preserve trust in digital evidence~\citep{chung2016outoftime}.

\paratitle{Industry and IoT.}
Industrial applications employ audio-visual monitoring for predictive maintenance, quality control, and safety compliance~\citep{gemmeke2017audioset,chen2020vggsound}. Acoustic anomaly detection identifies equipment faults from characteristic sound signatures~\citep{chen2023beats,baevski2020wav2vec}, while visual inspection detects defects in manufactured products. Combining both modalities improves reliability by correlating complementary indicators of system state~\citep{nagrani2021attention,girdhar2023imagebind}. IoT deployments similarly integrate distributed audio-visual sensors for environmental monitoring, smart building management, and agriculture~\citep{yang2024svad,schu2023using}. Because edge deployment must satisfy tight power and compute budgets, efficient architectures are essential~\citep{zeghidour2021soundstream,defossezhigh}. Privacy-preserving pipelines that process data locally or transmit only semantic representations are also increasingly important for continuous sensing in public and private spaces~\citep{wu2022wav2clip,guzhov2022audioclip}.

\section{Open Challenges, Limitations, and Future Directions}
\label{sec:future}

Audio-visual intelligence has progressed from deciding whether two modalities correspond, to generating synchronized videos and sounds, and further to systems that can hear, see, speak, edit, and act. The central difficulty, however, is not merely that audio and vision are different signal types. They are two partially observed views of a dynamic world. Sounds are produced by events, propagate through space, interact with materials and occlusions, and are interpreted by users under conversational, social, and physical context. This makes AVI different from generic multimodal learning: the field must explain \textit{why audio and visual evidence co-occur, and how}.

Therefore, the next stage of AVI should not be framed as a longer checklist of data, robustness, efficiency, and safety issues. These issues remain important, but they are too generic to define the field. A more useful framing is a research agenda for building audio-visual systems that are causal, contextual, controllable, verifiable, and interactive. The preceding sections reviewed perception, generation, and interaction as three pillars of AVI. 
Below we develop 6 major axes (cf. \cref{tab:avi_future_axes}): causal event-source grounding, audio-visual world models, audio-visual context memory, causal audio-visual intervention, verifier and reward ecosystems, and interactive and responsible AVI. \cref{fig:avi_future_stages} frames these axes as a staged roadmap: the field is moving from correspondence, perception, and generation toward interactive systems, and then toward causal-contextual and verifiable agentic AVI.

\begin{table*}[t]
    \centering
    \caption{Six central research axes for future AVI. Each axis connects a concrete limitation of current systems to a structural capability required by audio-visual intelligence.}
    \label{tab:avi_future_axes}
    \benchmarktablestyle
    \begin{tabularx}{\linewidth}{>{\raggedright\arraybackslash}X >{\raggedright\arraybackslash}X >{\raggedright\arraybackslash}X}
        \toprule
        \benchhead \tblhead{Aspect} & \tblhead{Current dominant assumption} & \tblhead{Deeper AVI-specific objective} \\
        \midrule
        Synchronization & Audio and video are aligned if their embeddings, labels, or offsets match. & Model source-level, event-level, and causal alignment under delay, occlusion, off-screen sound, and multi-source mixtures. \\
        World modeling & Audio and vision are paired observations of the same clip. & Treat audio and vision as complementary evidence for geometry, material, dynamics, affordance, and user/social state. \\
        Context and memory & Longer context windows or more tokens improve long-form AVI. & Build selective, hierarchical, and provenance-aware audio-visual memory across streaming, episodic, and semantic levels. \\
        Generation and editing & Prompts specify desired audio-visual content. & Support local, causal, and synchronized interventions over objects, sounds, identities, emotions, space, and time. \\
        Evaluation & FAD/FVD/CLIP/SyncNet-style metrics approximate quality and alignment. & Develop verifier and reward ecosystems for grounding, physical plausibility, audio indispensability, long-horizon coherence, and task utility. \\
        Interaction and deployment & Omni models can extend static perception and generation directly. & Balance real-time response, deliberative reasoning, user intent modeling, privacy, consent, and provenance in embodied and conversational AVI. \\
        \bottomrule
    \end{tabularx}
\end{table*}

\begin{center}
    \centering
    \includegraphics[width=\textwidth]{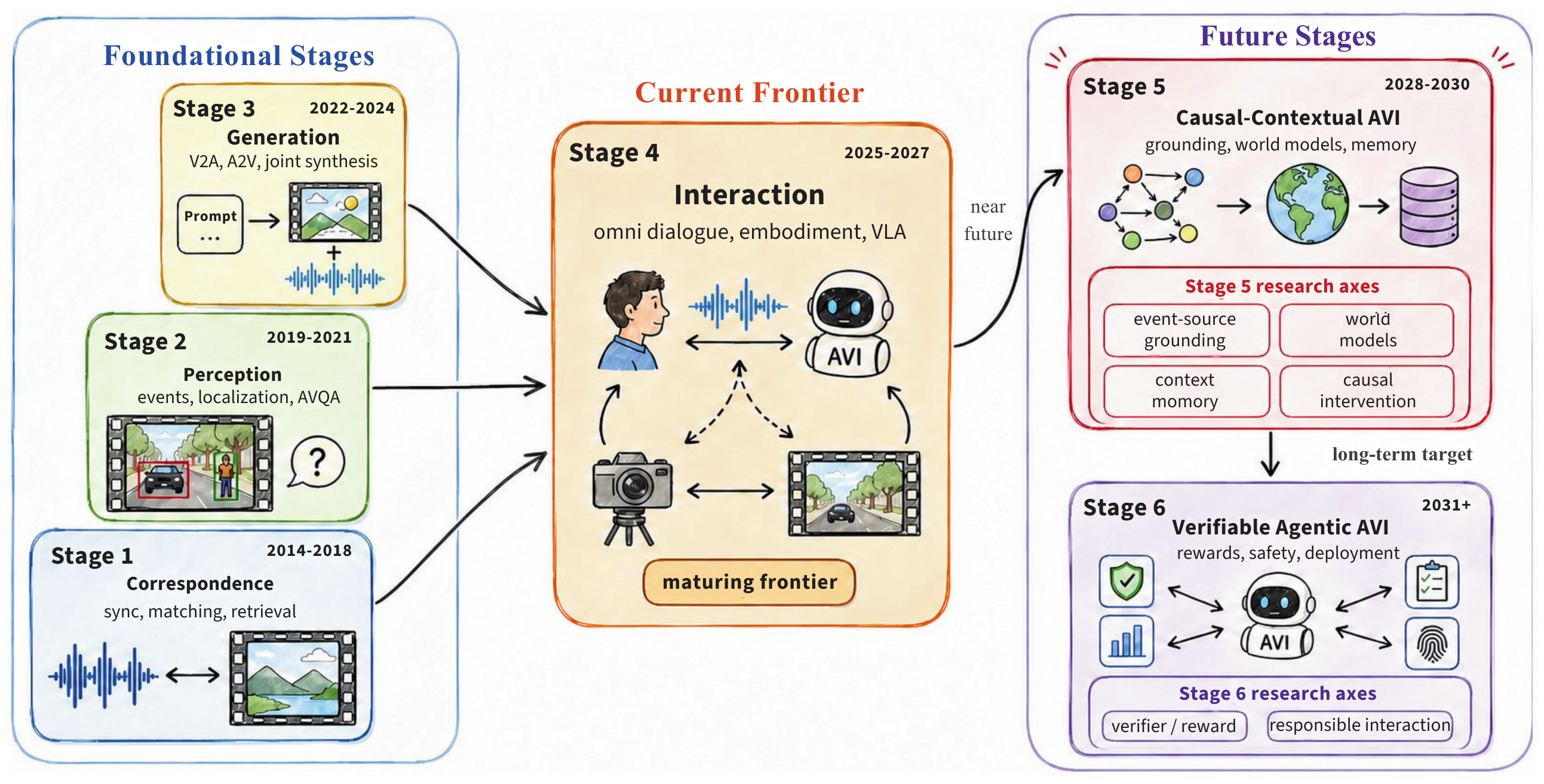}
    \captionof{figure}{Roadmap of AVI development stages. The first three stages consolidate correspondence, perception, and generation; the current frontier centers on interactive omni-modal and embodied systems; the next two stages emphasize causal-contextual AVI and verifiable agentic AVI. The associated research axes connect the discussion below to concrete future capabilities.}
    \label{fig:avi_future_stages}
\end{center}

\subsection{From Temporal Synchronization To Causal Event-Source Grounding}

\begin{wrapfigure}{r}{0.54\textwidth}
    \vspace{-0.6em}
    \centering
    \includegraphics[width=\linewidth]{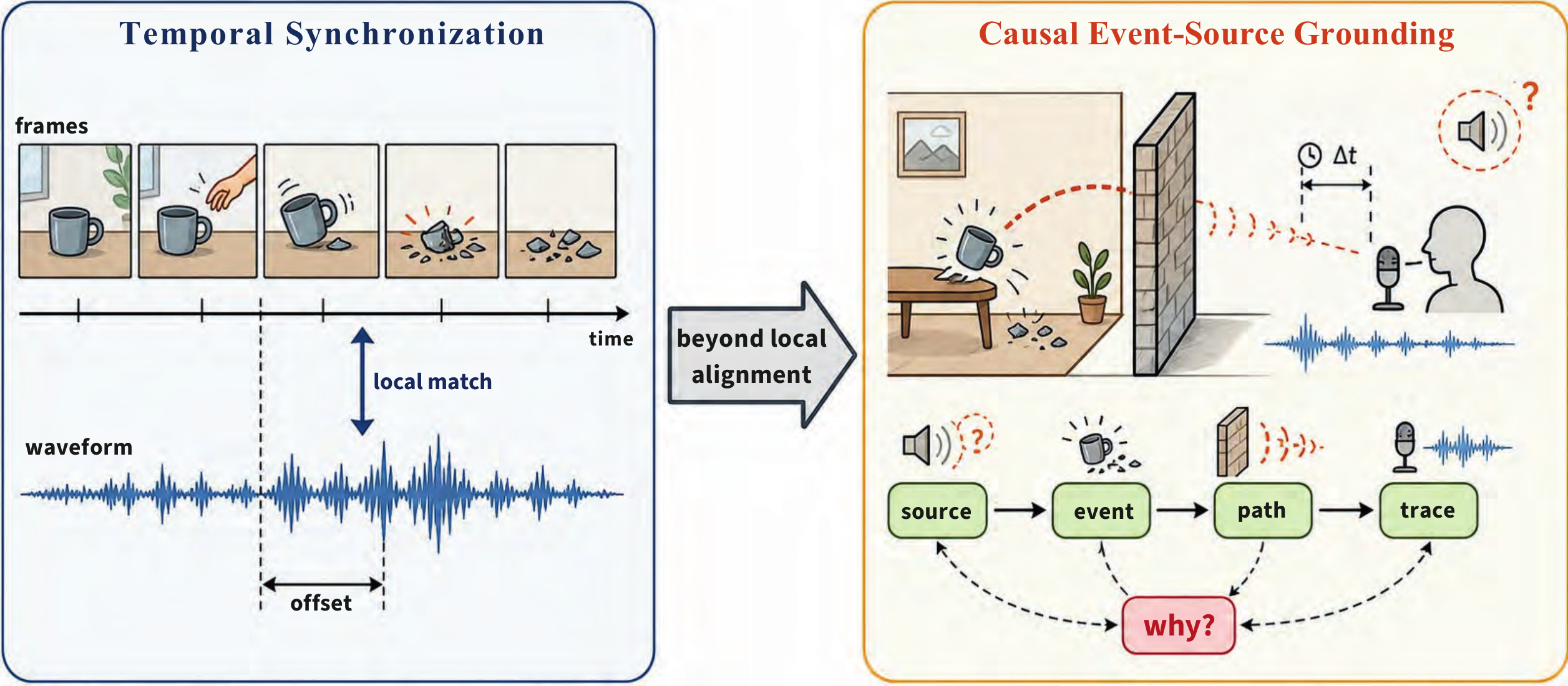}
    \caption{Temporal synchronization checks offsets; causal grounding further infers sources, events, propagation, occlusion, and uncertainty.}
    \label{fig:causal_event_source_grounding}
    \vspace{-0.8em}
\end{wrapfigure}
Synchronization is often treated as a local temporal matching problem: estimate whether the audio and video are aligned, or predict the offset between them. This formulation was effective for early lip-sync and synchronization systems, where mouth motion and speech are tightly coupled~\citep{chung2016outoftime,prajwal2020lip}, and it remains useful for measuring local correspondence in generated media. General AVI, however, requires a richer notion of alignment. A sound is not aligned with a frame simply because the two embeddings are close. It is aligned with a visual event if there is a plausible source, a temporal onset, a propagation path, and a causal relation between what is seen and what is heard. This distinction becomes critical beyond talking heads, especially in V2A, T2AV, AVQA, AV segmentation, and embodied perception. In practice, alignment is still often reduced to labels, offsets, or embedding proximity, which is insufficient when listeners must infer \emph{which} physical event caused \emph{which} sensory trace.

\cref{fig:causal_event_source_grounding} illustrates this shift: synchronization checks local temporal agreement, whereas event-source grounding asks which source produced which sound through which causal path.

\paratitle{Denser Correspondence and Generation.}
Dense audio--visual correspondence and coordinated generation have improved: finer localization and retrieval emerge from dense features~\citep{mark2024denseav}, and CAV-MAE Sync strengthens masked pretraining with explicit synchronization~\citep{araujo2025cavmaesync}. Event-aware soundtrack models such as Diff-Foley~\citep{comunita2024diff}, FoleyCrafter~\citep{zhang2024foleycrafter}, MMAudio~\citep{cheng2025mmaudio}, ThinkSound~\citep{thinksound2025}, HunyuanVideo-Foley~\citep{lin2025hunyuanfoley}, and Kling-Foley~\citep{wang2025kling} move video-to-audio beyond generic ambience, while joint generators including JavisDiT~\citep{liu2025javisdit}, AV-DiT~\citep{wang2024avdit}, UniVerse-1~\citep{wang2025universe}, Harmony~\citep{hu2025harmony}, Ovi~\citep{low2025ovi}, UniAVGen~\citep{zhang2025uniavgen}, and JavisDiT++~\citep{liu2026javisdit++} co-synthesize both modalities rather than tacking sound onto video.

\paratitle{Event-Source Graphs and Counterfactual Training.}
The next step is to treat scenes as \emph{event-source graphs}: nodes for objects, latent sources, events, speakers, and environment; edges for production, temporal order, propagation paths, and uncertainty. Perception could then support questions about provenance and evidence; generation could plan coherent event sequences instead of weakly coupled streams; editing could specify what must follow when an object is removed or replaced. Current practice still confuses co-occurrence with causation; a bark may appear with a dog without knowing whether it is on-screen, delayed, off-screen, or muffled by geometry, and global metrics can stay high while source identity, timing, or propagation is wrong: SyncNet-style scores reward lip motion, not whether a glass impact, rain density, or a distant siren matches the scene. Training should therefore mix contrastive alignment with temporal boundaries, separation, interventions, and reconstruction on counterfactual-style supervision: visible-but-silent sources, audible off-screen events, shifted onsets, material conflicts, and mixtures where only one grounding is correct. The objective is not only to decide whether two modalities match, but to explain the event structure that makes them match.

\subsection{From Paired Clips to Action-Conditioned Audio-Visual World Models}

Most current AVI datasets are built from paired audio-video clips. This form encourages correlation learning: a guitar image co-occurs with guitar sound, a speaking face co-occurs with speech, and a car crash co-occurs with impact noise. Correlation is necessary, but it is not sufficient. Real audio-visual intelligence requires a model of the world that can predict how visual and acoustic observations change when the agent moves, when an object is occluded, when a room changes, or when an action is executed. This is especially important for embodied agents, XR, robotics, navigation, and manipulation, where the system must act rather than merely classify.

\paratitle{Partial Progress in Simulation, Fields, Maps, and Touch.}
Pieces of such structure already exist: SoundSpaces~\citep{ChenJSGAIRG20SoundSpaces} and SoundSpaces~2.0~\citep{chen2022soundspaces20} support audio-visual navigation; Neural Acoustic Fields~\citep{luo2022learning}, AV-NeRF~\citep{liang2024avnerf}, AV-GS~\citep{bhosale2024avgs}, and NeRAF~\citep{cai2024neraf} learn spatial acoustic or joint neural fields; Real Acoustic Fields~\citep{ChenGR0LOR24RAF} brings real impulse responses alongside visuals; Audio-Visual and Multimodal Spatial Language Maps~\citep{huang2024avlmaps,huang2025mslmaps} fuse language with traversable spatial memory; ManiWAV~\citep{liu2024maniwav}, Sound of Simulation~\citep{wang2025soundSimulation}, and Audio-VLA~\citep{wei2025audioVLA} connect contact audio and generative acoustics to manipulation and transfer.

\paratitle{Toward Hybrid Action-Conditioned Latent Dynamics.}
Yet rendering-centric fields, navigation benchmarks that treat sound only as a goal cue, VLAs that bolt audio onto vision--language actions, and pixel-heavy video generators each stop short of a shared predictive abstraction. The agenda is \emph{action-conditioned audio-visual world models} that forecast compact latent state, including sources, materials, acoustics of the space, affordances, intent, and uncertainty, and support queries such as muffling behind a wall, slip versus success sounds during grasping, louder sources after a head turn, or co-evolving evidence if synthesized thunder enters a clip. Hybrid stacks can combine learned latent dynamics with approximate acoustic physics and task-scoped scorers without rendering HD video every planning step.

Evaluation should prioritize navigation gains, manipulation success, spatial reasoning, and counterfactual prediction rather than waveform or pixel reconstruction alone.

\subsection{Audio-Visual Context Memory}

\paratitle{Semantic Span Is Not Raw Token Count.}
Longer multimodal contexts are often addressed by scaling windows or stuffing more tokens, which is the wrong abstraction when a two-second sound matters more than hundreds of redundant frames.
Omni conversational stacks (GPT-4o~\citep{hurst2024gpt}; Qwen2.5-Omni~\citep{xu2025qwen2}; Qwen3-Omni~\citep{xu2025qwen3omni}; Ming-Omni~\citep{ai2025ming}; InteractiveOmni~\citep{tong2025interactiveomni}) illustrate unified interaction; unified representation lines (ImageBind~\citep{girdhar2023imagebind}, LanguageBind~\citep{zhu2023languagebind}, OneLLM~\citep{han2024onellm}, AnyGPT~\citep{anygpt2024}, DenseAV~\citep{mark2024denseav}) add shared embeddings.
Yet flattened token sequences still lose speaker identity, mix concurrent sources, or starve scarce audio evidence.

\begin{wrapfigure}{r}{0.54\textwidth}
    \centering
    \includegraphics[width=\linewidth]{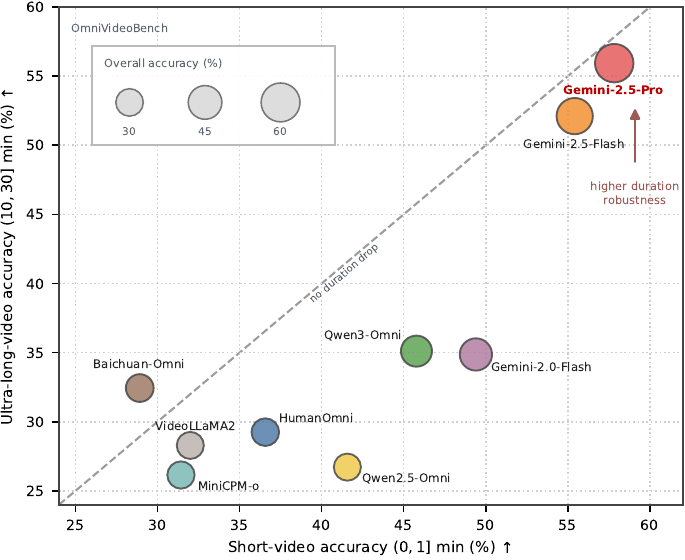}
    \caption{Duration robustness on OmniVideoBench. The x-axis shows short-video accuracy, the y-axis ultra-long-video accuracy, and the bubble area denotes overall accuracy; distance below the diagonal indicates long-context degradation.}
    \label{fig:av_context_memory_performance}
    \vspace{-1em}
\end{wrapfigure}

\cref{fig:av_context_memory_performance} uses OmniVideoBench duration splits~\citep{li2025omnivideobench} to show the same issue empirically: high short-video accuracy does not automatically translate to robust ultra-long-video accuracy.

\paratitle{Layered AV Memory.}
Treat context as layered \emph{audio-visual memory}: sensory buffers preserving recoverable fidelity where needed; event memory tracking onsets, tracks, masks, and positions; semantic summaries tethered to evidence; user/task state for goals and commitments, updated continuously without severing pointers to raw observations.
Specialists (audio events, trackers, diarizers, temporal verifiers, intent models) should exchange structured messages rather than forcing one backbone to attend uniformly. Faithfulness is the bottleneck: brittle symbols, leaky summaries, and opaque embeddings each fail alone, so credible systems retain raw handles, dense retrieval, graphs, and summaries together, and benchmarks probe delayed recall, audio-critical QA, modality conflict, and belief revision.

\subsection{Causal AV Intervention}

\paratitle{Controlled Edits Versus One-Shot Prompts.}
User intent in creation is rarely ``any plausible soundtrack'': edits are local causal requests, such as removing only the guitar stem, rerouting spatialized sirens, or changing emotion without identity, that general prompts mis-specify.
Video-to-audio models (Diff-Foley~\citep{comunita2024diff}, MMAudio~\citep{cheng2025mmaudio}, ThinkSound~\citep{thinksound2025}, HunyuanVideo-Foley~\citep{lin2025hunyuanfoley}) and audio-driven visuals (Wav2Lip~\citep{prajwal2020lip}, DiffTalk~\citep{shen2023difftalk}, EMO~\citep{tian2024emo}, VASA-1~\citep{xu2024vasa}, OmniAvatar~\citep{gan2025omniavatar}, AudCast~\citep{guan2025audcast}) anchor one axis.
Joint generators (JavisDiT~\citep{liu2025javisdit}, Harmony~\citep{hu2025harmony}, Ovi~\citep{low2025ovi}, UniAVGen~\citep{zhang2025uniavgen}) and localized editors (Object-AVEdit~\citep{fu2025objectavedit}, AVI-Edit~\citep{zheng2025aviedit}) anchor another.

\paratitle{Scene Graphs, Disentangled Latents, and Cross-Modal Feedback.}
Residual entanglement persists: prompts couple camera and identity; edits leak unrelated ambience or forget to update counterpart modalities. The path forward is causal editing via explicit or latent \emph{audio-visual scene graphs}: objects, stems, identities, motions, causal links. Operations become graph interventions; generators must freeze what is untouched while propagating edits along dependencies. Practical control needs factorized latents separating stems, speech, ambience, identity, geometry, lighting, and camera so timing, ambience, and emotion edits decouple unless the task demands otherwise. Verifier-guided refinement only helps when feedback sees both modalities, otherwise visuals can sharpen while synchronization silently drifts.

\begin{frontierbox}[\faLightbulbO~Open Problem: Editing Needs Causal Scope]
The hardest part of joint AV editing is deciding the scope of change. An edit should alter every consequence of the requested intervention, but preserve unrelated sources, identities, ambience, and narrative state. This requires models to reason over source-object bindings and downstream effects, not simply regenerate a visually plausible clip.
\end{frontierbox}

\subsection{Verifier and Reward Ecosystems}

\begin{wrapfigure}{r}{0.54\textwidth}
    \vspace{0pt}
    \centering
    \includegraphics[width=\linewidth]{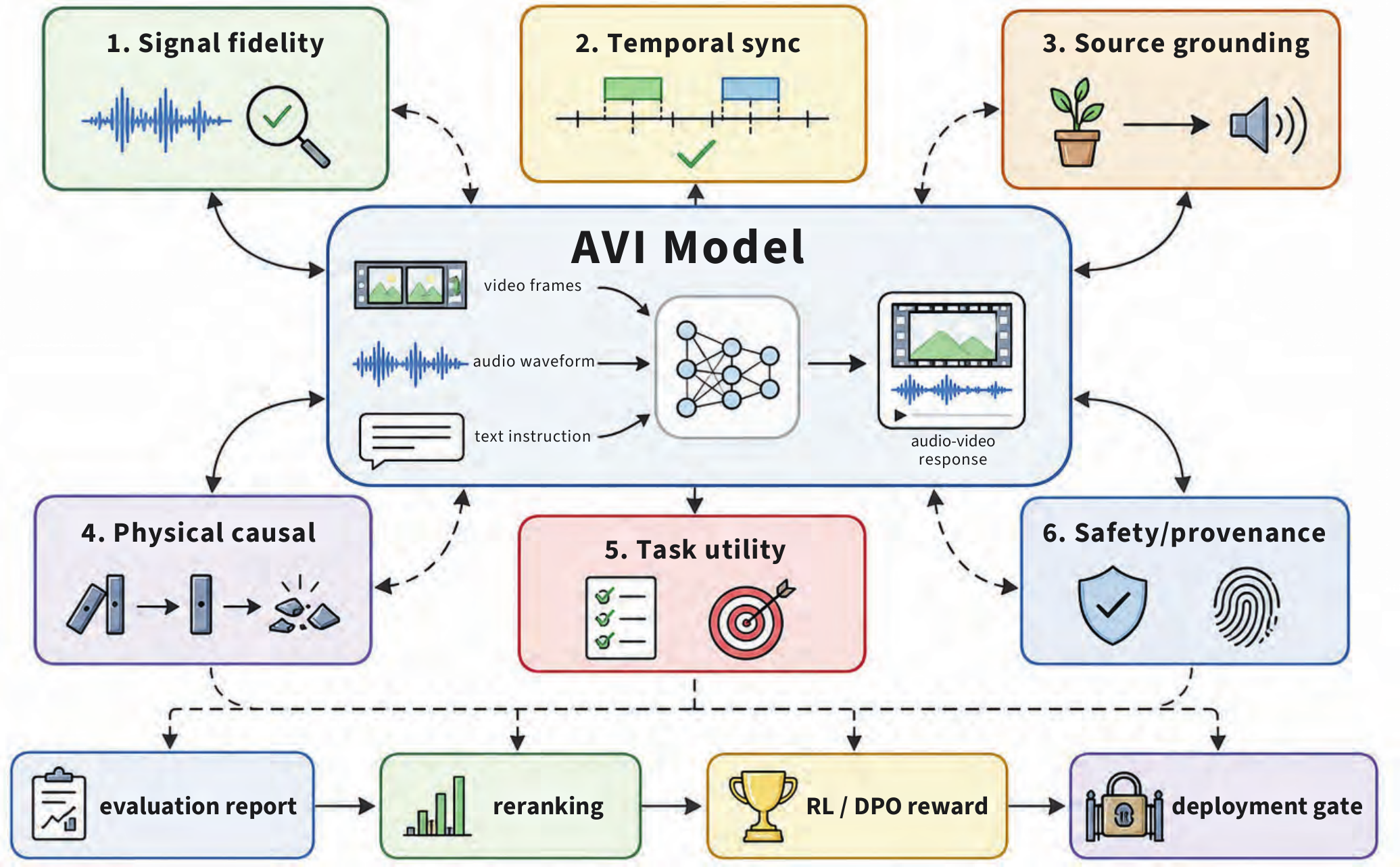}
    \captionof{figure}{Verifier and reward ecosystems decompose AVI correctness into signal, synchronization, grounding, causal/physical coherence, utility, safety, and provenance signals for evaluation and training.}
    \label{fig:verifier_reward_ecosystem}
\end{wrapfigure}

\paratitle{Scalar Metrics Are Not Verifiers.}
\cref{fig:verifier_reward_ecosystem} summarizes the target stack: FID-style audio and video scores, semantic similarity, and lip-sync surrogates slice the problem~\citep{kilgour2019fad,unterthiner2019fvd,chung2016outoftime,girdhar2023imagebind}, yet none certify joint correctness: clip-level audio quality can coexist with wrong sources; prompt alignment may violate physics; lip sync can obscure mis-timed impacts.

Understanding-centric suites such as AVUT~\citep{yang2025avut}, JointAVBench~\citep{chao2025jointavbench}, OmniVideoBench~\citep{li2025omnivideobench}, and AURA~\citep{galougah2025aura} broaden dependency and hallucination tests; generation-oriented benchmarks including JavisBench~\citep{liu2025javisdit}, Verse-Bench~\citep{wang2025universe}, PhyAVBench~\citep{xie2025phyavbench}, and FoleyBench~\citep{foleybench2025} push alignment and acoustics realism. Coverage is improving, but downstream utility ties remain uneven.

\paratitle{Layered Verifiers as Evaluators and Training Rewards.}
Deploy a staged verifier stack: signal fidelity; temporal synchronization; source grounding across visible, occluded, and off-screen cases; causal/physical coherence; instruction or task payoff; plus safety, identity, and provenance. These models double as RL/DPO rerankers: recent preference-tuned audio-video pipelines~\citep{liu2025improving,liu2025flow} hint at the appetite, though rewards must unpack faithfulness across modalities rather than collapsing to a prettier frame.
Treat narrow verifiers like narrow curricula: optimizing SyncNet alone can starve prosody or expression; over-relying on CLIP-like similarity excuses temporal hallucinations. Optimization should therefore be paired with adversarial/counterfactual probes and explicit failure taxonomies rather than leaderboard-only scores alone.

\subsection{Interactive and Responsible AVI}

Interaction is AVI's stringent regime: latency, streaming evidence, interruptions, embodied safety, privacy. Static captioning or offline generation skips these constraints entirely.

\paratitle{Omni Conversation and VLA-Style Action Interfaces.}
Realtime omni pipelines (GPT-4o~\citep{hurst2024gpt}, Qwen-Omni family~\citep{xu2025qwen2,xu2025qwen3omni}, VITA~\citep{fu2025vita}, Mini-Omni~\citep{xie2024miniomni}, Freeze-Omni~\citep{wang2024freezeomni}, Ming-Omni~\citep{ai2025ming}, InteractiveOmni~\citep{tong2025interactiveomni}) and architectures such as X-Streamer~\citep{xie2025x} emphasize streaming audiovisual world modeling; embodied stacks connect foundation perception to manipulation via OpenVLA~\citep{kimopenvla}, RT-2~\citep{zitkovich2023rt}, Octo~\citep{team2024octo}, $\pi_0$~\citep{black2024pi_0}, Audio-VLA~\citep{wei2025audioVLA}, and Sound of Simulation~\citep{wang2025soundSimulation}. Useful policies must schedule perception, reasoning, and actuation over time rather than concatenate encoders around an LLM.

\paratitle{Fast Reactions, Deliberation, User Trust, and Safeguards.}
Couple fast reactive pathways, such as collisions and turn-taking, with verifier-backed deliberation for editing and planning and with uncertainty gates deciding when to ask, defer, or fetch memory. Sensitive user modeling (beliefs, affect, conversational roles) inevitably collides with privacy, consent, impersonation risks, continuous sensing, synthetic faces, and calibrated anthropomorphism. Mitigate by design with provenance, watermarking where appropriate, user-controlled memory partitioning, incidental-data minimization, and consent-aware synthesis of speech and avatars, evaluating on-device, federated, or redacted pipelines explicitly against accuracy and usability constraints.

Overall, the frontier is not solely larger models or datasets but weaving together \emph{causal event-source grounding}, \emph{audio-visual world models}, \emph{audio-visual context memory}, \emph{causal audio-visual intervention}, \emph{verifier and reward ecosystems}, and \emph{interactive and responsible AVI} into systems that jointly explain multimodal evidence, forecast consequences under action, retain faithful memory, intervene with causal control, close the loop with multimodal verifiers, and deploy safely alongside humans.

\section{Conclusion}
\label{sec:conc}

This survey provides a comprehensive examination of audio-visual intelligence (AVI) in the era of foundation models, organizing the landscape around three interconnected pillars: representation-centric methods for signal embedding, generation-centric methods for cross-modal synthesis, and LLM-centric methods that utilize large language models as reasoning engines.
Our review has traced the progression of audio-visual understanding from event recognition to complex spatial reasoning, while the generation survey highlighted the pivotal role of diffusion and autoregressive modeling in achieving synchronized, high-fidelity synthesis. 
A central theme throughout is that cross-modal alignment remains the foundational challenge underlying these diverse tasks. Looking ahead, the field is moving toward unified architectures that handle perception and generation within single "omni" systems, yet this brings new hurdles in temporal synchronization at scale and training stability. 
Furthermore, as AVI extends into embodied robotics and immersive XR, the demands for efficiency, robustness, and safety become paramount. 
Ultimately, as these capabilities mature, addressing ethical risks such as deepfakes and privacy remains essential for the responsible development of AVI. 
We expect this survey to offer a rigorous foundation for researchers navigating this rapidly evolving and consequential frontier of artificial intelligence.

\clearpage
\newpage
\bibliographystyle{assets/plainnat}
\bibliography{paper}

@inproceedings{son2017lip,
  title={Lip reading sentences in the wild},
  author={Son Chung, Joon and Senior, Andrew and Vinyals, Oriol and Zisserman, Andrew},
  booktitle={Proceedings of the IEEE conference on computer vision and pattern recognition},
  pages={6447--6456},
  year={2017}
}

@inproceedings{roth2020ava,
  title={Ava active speaker: An audio-visual dataset for active speaker detection},
  author={Roth, Joseph and Chaudhuri, Sourish and Klejch, Ondrej and Marvin, Radhika and Gallagher, Andrew and Kaver, Liat and Ramaswamy, Sharadh and Stopczynski, Arkadiusz and Schmid, Cordelia and Xi, Zhonghua and others},
  booktitle={ICASSP 2020-2020 IEEE international conference on acoustics, speech and signal processing (ICASSP)},
  pages={4492--4496},
  year={2020},
  organization={IEEE}
}

@inproceedings{li2022invariant,
  title={Invariant grounding for video question answering},
  author={Li, Yicong and Wang, Xiang and Xiao, Junbin and Ji, Wei and Chua, Tat-Seng},
  booktitle={Proceedings of the IEEE/CVF Conference on Computer Vision and Pattern Recognition},
  pages={2928--2937},
  year={2022}
}

@inproceedings{tzinis2020two,
  title={Two-step sound source separation: Training on learned latent targets},
  author={Tzinis, Efthymios and Venkataramani, Shrikant and Wang, Zhepei and Subakan, Cem and Smaragdis, Paris},
  booktitle={ICASSP 2020-2020 IEEE International Conference on Acoustics, Speech and Signal Processing (ICASSP)},
  pages={31--35},
  year={2020},
  organization={IEEE}
}

@inproceedings{su2023pandagpt,
  title={Pandagpt: One model to instruction-follow them all},
  author={Su, Yixuan and Lan, Tian and Li, Huayang and Xu, Jialu and Wang, Yan and Cai, Deng},
  booktitle={Proceedings of the 1st Workshop on Taming Large Language Models: Controllability in the era of Interactive Assistants!},
  pages={11--23},
  year={2023}
}

@article{panagopoulou2023x,
  title={X-instructblip: A framework for aligning x-modal instruction-aware representations to llms and emergent cross-modal reasoning},
  author={Panagopoulou, Artemis and Xue, Le and Yu, Ning and Li, Junnan and Li, Dongxu and Joty, Shafiq and Xu, Ran and Savarese, Silvio and Xiong, Caiming and Niebles, Juan Carlos},
  journal={arXiv preprint arXiv:2311.18799},
  year={2023}
}

@article{lyu2023macaw,
  title={Macaw-llm: Multi-modal language modeling with image, audio, video, and text integration},
  author={Lyu, Chenyang and Wu, Minghao and Wang, Longyue and Huang, Xinting and Liu, Bingshuai and Du, Zefeng and Shi, Shuming and Tu, Zhaopeng},
  journal={arXiv preprint arXiv:2306.09093},
  year={2023}
}

@inproceedings{tang2025empowering,
  title={Empowering llms with pseudo-untrimmed videos for audio-visual temporal understanding},
  author={Tang, Yunlong and Shimada, Daiki and Bi, Jing and Feng, Mingqian and Hua, Hang and Xu, Chenliang},
  booktitle={Proceedings of the AAAI Conference on Artificial Intelligence},
  volume={39},
  pages={7293--7301},
  year={2025}
}

@inproceedings{han2024onellm,
  title={Onellm: One framework to align all modalities with language},
  author={Han, Jiaming and Gong, Kaixiong and Zhang, Yiyuan and Wang, Jiaqi and Zhang, Kaipeng and Lin, Dahua and Qiao, Yu and Gao, Peng and Yue, Xiangyu},
  booktitle={Proceedings of the IEEE/CVF Conference on Computer Vision and Pattern Recognition},
  pages={26584--26595},
  year={2024}
}

@article{yu2024crema,
  title={Crema: Generalizable and efficient video-language reasoning via multimodal modular fusion},
  author={Yu, Shoubin and Yoon, Jaehong and Bansal, Mohit},
  journal={ICLR},
  year={2025}
}

@article{monfort2019moments,
  title={Moments in time dataset: one million videos for event understanding},
  author={Monfort, Mathew and Andonian, Alex and Zhou, Bolei and Ramakrishnan, Kandan and Bargal, Sarah Adel and Yan, Tom and Brown, Lisa and Fan, Quanfu and Gutfreund, Dan and Vondrick, Carl and others},
  journal={IEEE transactions on pattern analysis and machine intelligence},
  volume={42},
  number={2},
  pages={502--508},
  year={2019},
  publisher={IEEE}
}

@article{wang2021audio2head,
  title={Audio2head: Audio-driven one-shot talking-head generation with natural head motion},
  author={Wang, Suzhen and Li, Lincheng and Ding, Yu and Fan, Changjie and Yu, Xin},
  journal={arXiv preprint arXiv:2107.09293},
  year={2021}
}

@inproceedings{zhang2025dialogue,
  title={Dialogue director: Bridging the gap in dialogue visualization for multimodal storytelling},
  author={Zhang, Min and Wang, Zilin and Chen, Liyan and Liu, Kunhong and Lin, Juncong},
  booktitle={2025 IEEE International Conference on Multimedia and Expo (ICME)},
  pages={1--6},
  year={2025},
  organization={IEEE}
}

@article{xie2025x,
  title={X-streamer: Unified human world modeling with audiovisual interaction},
  author={Xie, You and Gu, Tianpei and Li, Zenan and Zhang, Chenxu and Song, Guoxian and Zhao, Xiaochen and Liang, Chao and Jiang, Jianwen and Xu, Hongyi and Luo, Linjie},
  journal={arXiv preprint arXiv:2509.21574},
  year={2025}
}

@article{chen2025blip3o,
  title={Blip3o-next: Next frontier of native image generation},
  author={Chen, Jiuhai and Xue, Le and Xu, Zhiyang and Pan, Xichen and Yang, Shusheng and Qin, Can and Yan, An and Zhou, Honglu and Chen, Zeyuan and Huang, Lifu and others},
  journal={arXiv preprint arXiv:2510.15857},
  year={2025}
}

@article{yin2024survey,
  title={A survey on multimodal large language models},
  author={Yin, Shukang and Fu, Chaoyou and Zhao, Sirui and Li, Ke and Sun, Xing and Xu, Tong and Chen, Enhong},
  journal={National Science Review},
  volume={11},
  number={12},
  pages={nwae403},
  year={2024},
  publisher={Oxford University Press}
}

@article{ma2024foundation,
  title={Foundation models for music: A survey},
  author={Ma, Yinghao and {\O}land, Anders and Ragni, Anton and Del Sette, Bleiz MacSen and Saitis, Charalampos and Donahue, Chris and Lin, Chenghua and Plachouras, Christos and Benetos, Emmanouil and Shatri, Elona and others},
  journal={arXiv preprint arXiv:2408.14340},
  year={2024}
}

@article{kong2020panns,
  title={Panns: Large-scale pretrained audio neural networks for audio pattern recognition},
  author={Kong, Qiuqiang and Cao, Yin and Iqbal, Turab and Wang, Yuxuan and Wang, Wenwu and Plumbley, Mark D},
  journal={IEEE/ACM Transactions on Audio, Speech, and Language Processing},
  volume={28},
  pages={2880--2894},
  year={2020},
  publisher={IEEE}
}

@article{mccallum2022supervised,
  title={Supervised and Unsupervised Learning of Audio Representations for Music Understanding},
  author={McCallum, Matthew C and Korzeniowski, Filip and Oramas, Sergio and Gouyon, Fabien and Ehmann, Andreas F},
  journal={Ismir 2022 Hybrid Conference},
  year={2022}
}

@inproceedings{baevski2022data2vec,
  title={Data2vec: A general framework for self-supervised learning in speech, vision and language},
  author={Baevski, Alexei and Hsu, Wei-Ning and Xu, Qiantong and Babu, Arun and Gu, Jiatao and Auli, Michael},
  booktitle={International conference on machine learning},
  pages={1298--1312},
  year={2022},
  organization={PMLR}
}

@inproceedings{li2024mert,
  title={MERT: Acoustic Music Understanding Model with Large-Scale Self-supervised Training},
  author={Li, Yizhi and Yuan, Ruibin and Zhang, Ge and Ma, Yinghao and Chen, Xingran and Yin, Hanzhi and Lin, Chenghua and Ragni, Anton and Benetos, Emmanouil and Gyenge, Norbert and others},
  booktitle={International Conference on Learning Representations},
  year={2024}
}

@article{borsos2023audiolm,
  title={Audiolm: a language modeling approach to audio generation},
  author={Borsos, Zal{\'a}n and Marinier, Rapha{\"e}l and Vincent, Damien and Kharitonov, Eugene and Pietquin, Olivier and Sharifi, Matt and Roblek, Dominik and Teboul, Olivier and Grangier, David and Tagliasacchi, Marco and others},
  journal={IEEE/ACM transactions on audio, speech, and language processing},
  volume={31},
  pages={2523--2533},
  year={2023},
  publisher={IEEE}
}

@article{agostinelli2023musiclm,
  title={Musiclm: Generating music from text},
  author={Agostinelli, Andrea and Denk, Timo I and Borsos, Zal{\'a}n and Engel, Jesse and Verzetti, Mauro and Caillon, Antoine and Huang, Qingqing and Jansen, Aren and Roberts, Adam and Tagliasacchi, Marco and others},
  journal={arXiv preprint arXiv:2301.11325},
  year={2023}
}

@article{kreuk2022audiogen,
  title={Audiogen: Textually guided audio generation},
  author={Kreuk, Felix and Synnaeve, Gabriel and Polyak, Adam and Singer, Uriel and D{\'e}fossez, Alexandre and Copet, Jade and Parikh, Devi and Taigman, Yaniv and Adi, Yossi},
  journal={arXiv preprint arXiv:2209.15352},
  year={2022}
}

@inproceedings{ye2025codec,
  title={Codec does matter: Exploring the semantic shortcoming of codec for audio language model},
  author={Ye, Zhen and Sun, Peiwen and Lei, Jiahe and Lin, Hongzhan and Tan, Xu and Dai, Zheqi and Kong, Qiuqiang and Chen, Jianyi and Pan, Jiahao and Liu, Qifeng and others},
  booktitle={Proceedings of the AAAI Conference on Artificial Intelligence},
  pages={25697--25705},
  year={2025}
}

@inproceedings{jiwavtokenizer,
  title={WavTokenizer: an Efficient Acoustic Discrete Codec Tokenizer for Audio Language Modeling},
  author={Ji, Shengpeng and Jiang, Ziyue and Wang, Wen and Chen, Yifu and Fang, Minghui and Zuo, Jialong and Yang, Qian and Cheng, Xize and Wang, Zehan and Li, Ruiqi and others},
  booktitle={The Thirteenth International Conference on Learning Representations},
  year={2025}
}

@inproceedings{arandjelovic2017look,
  title={Look, listen and learn},
  author={Arandjelovic, Relja and Zisserman, Andrew},
  booktitle={Proceedings of the IEEE international conference on computer vision},
  pages={609--617},
  year={2017}
}

@article{korbar2018cooperative,
  title={Cooperative learning of audio and video models from self-supervised synchronization},
  author={Korbar, Bruno and Tran, Du and Torresani, Lorenzo},
  journal={Advances in Neural Information Processing Systems},
  volume={31},
  year={2018}
}

@article{alwassel2020self,
  title={Self-supervised learning by cross-modal audio-video clustering},
  author={Alwassel, Humam and Mahajan, Dhruv and Korbar, Bruno and Torresani, Lorenzo and Ghanem, Bernard and Tran, Du},
  journal={Advances in neural information processing systems},
  volume={33},
  pages={9758--9770},
  year={2020}
}

@inproceedings{morgado2021audio,
  title={Audio-visual instance discrimination with cross-modal agreement},
  author={Morgado, Pedro and Vasconcelos, Nuno and Misra, Ishan},
  booktitle={Proceedings of the IEEE/CVF conference on computer vision and pattern recognition},
  pages={12475--12486},
  year={2021}
}

@article{akbari2021vatt,
  title={Vatt: Transformers for multimodal self-supervised learning from raw video, audio and text},
  author={Akbari, Hassan and Yuan, Liangzhe and Qian, Rui and Chuang, Wei-Hong and Chang, Shih-Fu and Cui, Yin and Gong, Boqing},
  journal={Advances in neural information processing systems},
  volume={34},
  pages={24206--24221},
  year={2021}
}

@inproceedings{zhang2024es3,
  title={Es3: Evolving self-supervised learning of robust audio-visual speech representations},
  author={Zhang, Yuanhang and Yang, Shuang and Shan, Shiguang and Chen, Xilin},
  booktitle={Proceedings of the IEEE/CVF Conference on Computer Vision and Pattern Recognition},
  pages={27069--27079},
  year={2024}
}

@inproceedings{kim2024equiav,
  title={EquiAV: leveraging equivariance for audio-visual contrastive learning},
  author={Kim, Jongsuk and Lee, Hyeongkeun and Rho, Kyeongha and Kim, Junmo and Chung, Joon Son},
  booktitle={Proceedings of the 41st International Conference on Machine Learning},
  pages={24327--24341},
  year={2024}
}

@inproceedings{sarkar2023self,
  title={Self-supervised audio-visual representation learning with relaxed cross-modal synchronicity},
  author={Sarkar, Pritam and Etemad, Ali},
  booktitle={Proceedings of the AAAI Conference on Artificial Intelligence},
  pages={9723--9732},
  year={2023}
}

@article{nagrani2021attention,
  title={Attention bottlenecks for multimodal fusion},
  author={Nagrani, Arsha and Yang, Shan and Arnab, Anurag and Jansen, Aren and Schmid, Cordelia and Sun, Chen},
  journal={Advances in neural information processing systems},
  volume={34},
  pages={14200--14213},
  year={2021}
}

@inproceedings{hao2024improving,
  title={Improving audio-visual segmentation with bidirectional generation},
  author={Hao, Dawei and Mao, Yuxin and He, Bowen and Han, Xiaodong and Dai, Yuchao and Zhong, Yiran},
  booktitle={Proceedings of the AAAI conference on artificial intelligence},
  pages={2067--2075},
  year={2024}
}

@article{shi2022learning,
  title={Learning audio-visual speech representation by masked multimodal cluster prediction},
  author={Shi, Bowen and Hsu, Wei-Ning and Lakhotia, Kushal and Mohamed, Abdelrahman},
  journal={arXiv preprint arXiv:2201.02184},
  year={2022}
}

@article{ashraf2023hybrid,
  title={A hybrid cnn and rnn variant model for music classification},
  author={Ashraf, Mohsin and Abid, Fazeel and Din, Ikram Ud and Rasheed, Jawad and Yesiltepe, Mirsat and Yeo, Sook Fern and Ersoy, Merve T},
  journal={Applied Sciences},
  volume={13},
  number={3},
  pages={1476},
  year={2023},
  publisher={MDPI}
}

@article{huang2022mulan,
  title={Mulan: A joint embedding of music audio and natural language},
  author={Huang, Qingqing and Jansen, Aren and Lee, Joonseok and Ganti, Ravi and Li, Judith Yue and Ellis, Daniel PW},
  journal={arXiv preprint arXiv:2208.12415},
  year={2022}
}

@inproceedings{OmniAVS,
  title={Towards omnimodal expressions and reasoning in referring audio-visual segmentation},
  author={Ying, Kaining and Ding, Henghui and Jie, Guangquan and Jiang, Yu-Gang},
  booktitle={Proceedings of the IEEE/CVF International Conference on Computer Vision},
  pages={22575--22585},
  year={2025}
}

@article{xing2025echoink,
  title={Echoink-r1: Exploring audio-visual reasoning in multimodal llms via reinforcement learning},
  author={Xing, Zhenghao and Hu, Xiaowei and Fu, Chi-Wing and Wang, Wenhai and Dai, Jifeng and Heng, Pheng-Ann},
  journal={arXiv preprint arXiv:2505.04623},
  year={2025}
}

@article{li2024omnibench,
  title={Omnibench: Towards the future of universal omni-language models},
  author={Li, Yizhi and Zhang, Ge and Ma, Yinghao and Yuan, Ruibin and Zhu, Kang and Guo, Hangyu and Liang, Yiming and Liu, Jiaheng and Wang, Zekun and Yang, Jian and others},
  journal={arXiv preprint arXiv:2409.15272},
  year={2024}
}

@article{li2025omnivideobench,
  title={OmniVideoBench: Towards Audio-Visual Understanding Evaluation for Omni MLLMs},
  author={Li, Caorui and Chen, Yu and Ji, Yiyan and Xu, Jin and Cui, Zhenyu and Li, Shihao and Zhang, Yuanxing and Tang, Jiafu and Song, Zhenghao and Zhang, Dingling and others},
  journal={arXiv preprint arXiv:2510.10689},
  year={2025}
}

@article{hong2025musicinfuser,
  title={MusicInfuser: Making Video Diffusion Listen and Dance},
  author={Hong, Susung and Kemelmacher-Shlizerman, Ira and Curless, Brian and Seitz, Steven M},
  journal={arXiv preprint arXiv:2503.14505},
  year={2025}
}

@inproceedings{wang2025choreomuse,
  title={Choreomuse: Robust music-to-dance video generation with style transfer and beat-adherent motion},
  author={Wang, Xuanchen and Wang, Heng and Cai, Weidong},
  booktitle={Proceedings of the 33rd ACM International Conference on Multimedia},
  pages={7912--7921},
  year={2025}
}

@article{nagrani2017voxceleb,
  title={Voxceleb: a large-scale speaker identification dataset},
  author={Nagrani, Arsha and Chung, Joon Son and Zisserman, Andrew},
  journal={arXiv preprint arXiv:1706.08612},
  year={2017}
}

@inproceedings{guan2025audcast,
  title={AudCast: Audio-Driven Human Video Generation by Cascaded Diffusion Transformers},
  author={Guan, Jiazhi and Wang, Kaisiyuan and Xu, Zhiliang and Yang, Quanwei and Sun, Yasheng and He, Shengyi and Liang, Borong and Cao, Yukang and Li, Yingying and Feng, Haocheng and others},
  booktitle={Proceedings of the Computer Vision and Pattern Recognition Conference},
  pages={10678--10689},
  year={2025}
}

@article{tong2025interactiveomni,
  title={Interactiveomni: A unified omni-modal model for audio-visual multi-turn dialogue},
  author={Tong, Wenwen and Guo, Hewei and Ran, Dongchuan and Chen, Jiangnan and Lu, Jiefan and Wang, Kaibin and Li, Keqiang and Zhu, Xiaoxu and Li, Jiakui and Li, Kehan and others},
  journal={arXiv preprint arXiv:2510.13747},
  year={2025}
}

@article{park2024let,
  title={Let's Go Real Talk: Spoken Dialogue Model for Face-to-Face Conversation},
  author={Park, Se Jin and Kim, Chae Won and Rha, Hyeongseop and Kim, Minsu and Hong, Joanna and Yeo, Jeong Hun and Ro, Yong Man},
  journal={arXiv preprint arXiv:2406.07867},
  year={2024}
}

@inproceedings{fu2025video,
  title={Video-mme: The first-ever comprehensive evaluation benchmark of multi-modal llms in video analysis},
  author={Fu, Chaoyou and Dai, Yuhan and Luo, Yongdong and Li, Lei and Ren, Shuhuai and Zhang, Renrui and Wang, Zihan and Zhou, Chenyu and Shen, Yunhang and Zhang, Mengdan and others},
  booktitle={Proceedings of the Computer Vision and Pattern Recognition Conference},
  pages={24108--24118},
  year={2025}
}

@article{wang2026avibench,
  title={AVIBench: Toward Human-like Audio-Visual Intelligence of Omni-MLLMs},
  author={Wang, Yaoting and Zhang, Ziyi and Tu, Wenming and Xu, Shaoxuan and Du, Wenjie and Liang, Cheng and Wang, Weijun and Li, Yuanchao and Li, Guangyao and Fei, Hao and Li, Yuanchun and Ding, Henghui and Liu, Yunxin},
  journal={Preprint},
  year={2026}
}

@inproceedings{le2023music,
  title={Music-driven group choreography},
  author={Le, Nhat and Pham, Thang and Do, Tuong and Tjiputra, Erman and Tran, Quang D and Nguyen, Anh},
  booktitle={Proceedings of the IEEE/CVF Conference on Computer Vision and Pattern Recognition},
  pages={8673--8682},
  year={2023}
}

@inproceedings{shen2023difftalk,
  title={Difftalk: Crafting diffusion models for generalized audio-driven portraits animation},
  author={Shen, Shuai and Zhao, Wenliang and Meng, Zibin and Li, Wanhua and Zhu, Zheng and Zhou, Jie and Lu, Jiwen},
  booktitle={Proceedings of the IEEE/CVF conference on computer vision and pattern recognition},
  pages={1982--1991},
  year={2023}
}

@inproceedings{zhang2023sadtalker,
  title={SadTalker: Learning Realistic 3D Motion Coefficients for Stylized Audio-Driven Single Image Talking Face Animation},
  author={Zhang, Wenxuan and Cun, Xiaodong and Wang, Xuan and Zhang, Yong and Shen, Xi and Guo, Yu and Shan, Ying and Wang, Fei},
  booktitle={Proceedings of the IEEE/CVF Conference on Computer Vision and Pattern Recognition},
  pages={8652--8661},
  year={2023}
}

@inproceedings{zhu2021let,
  title={Let's play music: Audio-driven performance video generation},
  author={Zhu, Hao and Li, Yi and Zhu, Feixia and Zheng, Aihua and He, Ran},
  booktitle={2020 25th International Conference on Pattern Recognition (ICPR)},
  pages={3574--3581},
  year={2021},
  organization={IEEE}
}

@inproceedings{ruan2023mm,
  title={Mm-diffusion: Learning multi-modal diffusion models for joint audio and video generation},
  author={Ruan, Ludan and Ma, Yiyang and Yang, Huan and He, Huiguo and Liu, Bei and Fu, Jianlong and Yuan, Nicholas Jing and Jin, Qin and Guo, Baining},
  booktitle={Proceedings of the IEEE/CVF Conference on Computer Vision and Pattern Recognition},
  pages={10219--10228},
  year={2023}
}

@inproceedings{yariv2024diverse,
  title={Diverse and aligned audio-to-video generation via text-to-video model adaptation},
  author={Yariv, Guy and Gat, Itai and Benaim, Sagie and Wolf, Lior and Schwartz, Idan and Adi, Yossi},
  booktitle={Proceedings of the AAAI Conference on Artificial Intelligence},
  pages={6639--6647},
  year={2024}
}

@article{zhou2025daily,
  title={Daily-Omni: Towards Audio-Visual Reasoning with Temporal Alignment across Modalities},
  author={Zhou, Ziwei and Wang, Rui and Wu, Zuxuan},
  journal={arXiv preprint arXiv:2505.17862},
  year={2025}
}

@article{jung2025avcd,
  title={AVCD: Mitigating Hallucinations in Audio-Visual Large Language Models through Contrastive Decoding},
  author={Jung, Chaeyoung and Jang, Youngjoon and Chung, Joon Son},
  journal={arXiv preprint arXiv:2505.20862},
  year={2025}
}

@inproceedings{zhao2022beat,
  title={Beat Transformer: Demixed Beat and Downbeat Tracking with Dilated Self-Attention},
  author={Zhao, Jingwei and Xia, Gus and Wang, Ye},
  booktitle={Ismir 2022 Hybrid Conference},
  year={2022}
}

@article{luo2023music,
  title={Music source separation with band-split RNN},
  author={Luo, Yi and Yu, Jianwei},
  journal={IEEE/ACM Transactions on Audio, Speech, and Language Processing},
  volume={31},
  pages={1893--1901},
  year={2023},
  publisher={IEEE}
}

@inproceedings{rouard2023hybrid,
  title={Hybrid transformers for music source separation},
  author={Rouard, Simon and Massa, Francisco and D{\'e}fossez, Alexandre},
  booktitle={ICASSP 2023-2023 IEEE International Conference on Acoustics, Speech and Signal Processing (ICASSP)},
  pages={1--5},
  year={2023},
  organization={IEEE}
}

@inproceedings{liu2024audio,
  title={Audio prompt tuning for universal sound separation},
  author={Liu, Yuzhuo and Liu, Xubo and Zhao, Yan and Wang, Yuanyuan and Xia, Rui and Tain, Pingchuan and Wang, Yuxuan},
  booktitle={ICASSP 2024-2024 IEEE International Conference on Acoustics, Speech and Signal Processing (ICASSP)},
  pages={1446--1450},
  year={2024},
  organization={IEEE}
}

@inproceedings{jiang2025dual,
  title={Dual-path mamba: Short and long-term bidirectional selective structured state space models for speech separation},
  author={Jiang, Xilin and Han, Cong and Mesgarani, Nima},
  booktitle={ICASSP 2025-2025 IEEE International Conference on Acoustics, Speech and Signal Processing (ICASSP)},
  pages={1--5},
  year={2025},
  organization={IEEE}
}

@inproceedings{wang2023tf,
  title={TF-GridNet: Making time-frequency domain models great again for monaural speaker separation},
  author={Wang, Zhong-Qiu and Cornell, Samuele and Choi, Shukjae and Lee, Younglo and Kim, Byeong-Yeol and Watanabe, Shinji},
  booktitle={ICASSP 2023-2023 IEEE international conference on acoustics, speech and signal processing (ICASSP)},
  pages={1--5},
  year={2023},
  organization={IEEE}
}

@inproceedings{schu2023using,
  title={On using the UA-Speech and TORGO databases to validate automatic dysarthric speech classification approaches},
  author={Schu, Guilherme and Janbakhshi, Parvaneh and Kodrasi, Ina},
  booktitle={ICASSP 2023-2023 IEEE International Conference on Acoustics, Speech and Signal Processing (ICASSP)},
  pages={1--5},
  year={2023},
  organization={IEEE}
}

@inproceedings{shao2024fine,
  title={Fine-tune the pretrained ATST model for sound event detection},
  author={Shao, Nian and Li, Xian and Li, Xiaofei},
  booktitle={ICASSP 2024-2024 IEEE International Conference on Acoustics, Speech and Signal Processing (ICASSP)},
  pages={911--915},
  year={2024},
  organization={IEEE}
}

@inproceedings{yang2024svad,
  title={Svad: A robust, low-power, and light-weight voice activity detection with spiking neural networks},
  author={Yang, Qu and Liu, Qianhui and Li, Nan and Ge, Meng and Song, Zeyang and Li, Haizhou},
  booktitle={ICASSP 2024-2024 IEEE International Conference on Acoustics, Speech and Signal Processing (ICASSP)},
  pages={221--225},
  year={2024},
  organization={IEEE}
}

@inproceedings{martin2023training,
  title={Training sound event detection with soft labels from crowdsourced annotations},
  author={Mart{\'\i}n-Morat{\'o}, Irene and Harju, Manu and Ahokas, Paul and Mesaros, Annamaria},
  booktitle={ICASSP 2023-2023 IEEE International Conference on Acoustics, Speech and Signal Processing (ICASSP)},
  pages={1--5},
  year={2023},
  organization={IEEE}
}

@article{chang2023speechprompt,
  title={Speechprompt v2: Prompt tuning for speech classification tasks},
  author={Chang, Kai-Wei and Wang, Yu-Kai and Shen, Hua and Kang, Iu-thing and Tseng, Wei-Cheng and Li, Shang-Wen and Lee, Hung-yi},
  journal={arXiv preprint arXiv:2303.00733},
  year={2023}
}

@inproceedings{wu2025avf,
  title={AVF-MAE++: Scaling Affective Video Facial Masked Autoencoders via Efficient Audio-Visual Self-Supervised Learning},
  author={Wu, Xuecheng and Sun, Heli and Wang, Yifan and Nie, Jiayu and Zhang, Jie and Wang, Yabing and Xue, Junxiao and He, Liang},
  booktitle={Proceedings of the Computer Vision and Pattern Recognition Conference},
  pages={9142--9153},
  year={2025}
}

@inproceedings{bao2023cross,
  title={Cross-modal label contrastive learning for unsupervised audio-visual event localization},
  author={Bao, Peijun and Yang, Wenhan and Ng, Boon Poh and Er, Meng Hwa and Kot, Alex C},
  booktitle={Proceedings of the AAAI Conference on Artificial Intelligence},
  pages={215--222},
  year={2023}
}

@inproceedings{huang2023mavil,
  title={MAViL: masked audio-video learners},
  author={Huang, Po-Yao and Sharma, Vasu and Xu, Hu and Ryali, Chaitanya and Fan, Haoqi and Li, Yanghao and Li, Shang-Wen and Ghosh, Gargi and Malik, Jitendra and Feichtenhofer, Christoph},
  booktitle={Proceedings of the 37th International Conference on Neural Information Processing Systems},
  pages={20371--20393},
  year={2023}
}

@inproceedings{tsiamas2025sequential,
  title={Sequential contrastive audio-visual learning},
  author={Tsiamas, Ioannis and Pascual, Santiago and Yeh, Chunghsin and Serr{\`a}, Joan},
  booktitle={ICASSP 2025-2025 IEEE International Conference on Acoustics, Speech and Signal Processing (ICASSP)},
  pages={1--5},
  year={2025},
  organization={IEEE}
}

@inproceedings{girdhar2023imagebind,
  title={Imagebind: One embedding space to bind them all},
  author={Girdhar, Rohit and El-Nouby, Alaaeldin and Liu, Zhuang and Singh, Mannat and Alwala, Kalyan Vasudev and Joulin, Armand and Misra, Ishan},
  booktitle={Proceedings of the IEEE/CVF conference on computer vision and pattern recognition},
  pages={15180--15190},
  year={2023}
}

@inproceedings{wu2022wav2clip,
  title={Wav2clip: Learning robust audio representations from clip},
  author={Wu, Ho-Hsiang and Seetharaman, Prem and Kumar, Kundan and Bello, Juan Pablo},
  booktitle={ICASSP 2022-2022 IEEE International Conference on Acoustics, Speech and Signal Processing (ICASSP)},
  pages={4563--4567},
  year={2022},
  organization={IEEE}
}

@inproceedings{copet2023simple,
  title={Simple and controllable music generation},
  author={Copet, Jade and Kreuk, Felix and Gat, Itai and Remez, Tal and Kant, David and Synnaeve, Gabriel and Adi, Yossi and D{\'e}fossez, Alexandre},
  booktitle={Proceedings of the 37th International Conference on Neural Information Processing Systems},
  pages={47704--47720},
  year={2023}
}

@article{zhu2025muq,
  title={Muq: Self-supervised music representation learning with mel residual vector quantization},
  author={Zhu, Haina and Zhou, Yizhi and Chen, Hangting and Yu, Jianwei and Ma, Ziyang and Gu, Rongzhi and Luo, Yi and Tan, Wei and Chen, Xie},
  journal={arXiv preprint arXiv:2501.01108},
  year={2025}
}

@article{zeghidour2021soundstream,
  title={Soundstream: An end-to-end neural audio codec},
  author={Zeghidour, Neil and Luebs, Alejandro and Omran, Ahmed and Skoglund, Jan and Tagliasacchi, Marco},
  journal={IEEE/ACM Transactions on Audio, Speech, and Language Processing},
  volume={30},
  pages={495--507},
  year={2021},
  publisher={IEEE}
}

@article{defossezhigh,
  title={High Fidelity Neural Audio Compression},
  author={D{\'e}fossez, Alexandre and Copet, Jade and Synnaeve, Gabriel and Adi, Yossi},
  journal={Transactions on Machine Learning Research},
  year={2022}
}

@inproceedings{van2017neural,
  title={Neural discrete representation learning},
  author={van den Oord, Aaron and Vinyals, Oriol and Kavukcuoglu, Koray},
  booktitle={Proceedings of the 31st International Conference on Neural Information Processing Systems},
  pages={6309--6318},
  year={2017}
}

@inproceedings{chen2023beats,
  title={BEATs: Audio Pre-Training with Acoustic Tokenizers},
  author={Chen, Sanyuan and Wu, Yu and Wang, Chengyi and Liu, Shujie and Tompkins, Daniel and Chen, Zhuo and Che, Wanxiang and Yu, Xiangzhan and Wei, Furu},
  booktitle={International Conference on Machine Learning},
  pages={5178--5193},
  year={2023},
  organization={PMLR}
}

@article{huang2022masked,
  title={Masked autoencoders that listen},
  author={Huang, Po-Yao and Xu, Hu and Li, Juncheng and Baevski, Alexei and Auli, Michael and Galuba, Wojciech and Metze, Florian and Feichtenhofer, Christoph},
  journal={Advances in Neural Information Processing Systems},
  volume={35},
  pages={28708--28720},
  year={2022}
}

@inproceedings{ms-coco-2014,
  author       = {Tsung{-}Yi Lin and
                  Michael Maire and
                  Serge J. Belongie and
                  James Hays and
                  Pietro Perona and
                  Deva Ramanan and
                  Piotr Doll{\'{a}}r and
                  C. Lawrence Zitnick},
  title        = {Microsoft {COCO:} Common Objects in Context},
  booktitle    = {Computer Vision - {ECCV} 2014 - 13th European Conference, Zurich,
                  Switzerland, September 6-12, 2014, Proceedings, Part {V}},
  volume       = {8693},
  pages        = {740--755},
  year         = {2014}
}

@article{ImageNet,
  author       = {Olga Russakovsky and
                  Jia Deng and
                  Hao Su and
                  Jonathan Krause and
                  Sanjeev Satheesh and
                  Sean Ma and
                  Zhiheng Huang and
                  Andrej Karpathy and
                  Aditya Khosla and
                  Michael S. Bernstein and
                  Alexander C. Berg and
                  Li Fei{-}Fei},
  title        = {ImageNet Large Scale Visual Recognition Challenge},
  journal      = {Int. J. Comput. Vis.},
  volume       = {115},
  number       = {3},
  pages        = {211--252},
  year         = {2015}
}

@article{hurst2024gpt,
  title={Gpt-4o system card},
  author={Hurst, Aaron and Lerer, Adam and Goucher, Adam P and Perelman, Adam and Ramesh, Aditya and Clark, Aidan and Ostrow, AJ and Welihinda, Akila and Hayes, Alan and Radford, Alec and others},
  journal={arXiv preprint arXiv:2410.21276},
  year={2024}
}

@article{achiam2023gpt,
  title={Gpt-4 technical report},
  author={Achiam, Josh and Adler, Steven and Agarwal, Sandhini and Ahmad, Lama and Akkaya, Ilge and Aleman, Florencia Leoni and Almeida, Diogo and Altenschmidt, Janko and Altman, Sam and Anadkat, Shyamal and others},
  journal={arXiv preprint arXiv:2303.08774},
  year={2023}
}

@article{touvron2023llama,
  title={Llama: Open and efficient foundation language models},
  author={Touvron, Hugo and Lavril, Thibaut and Izacard, Gautier and Martinet, Xavier and Lachaux, Marie-Anne and Lacroix, Timoth{\'e}e and Rozi{\`e}re, Baptiste and Goyal, Naman and Hambro, Eric and Azhar, Faisal and others},
  journal={arXiv preprint arXiv:2302.13971},
  year={2023}
}

@article{yang2025qwen3,
  title={Qwen3 technical report},
  author={Yang, An and Li, Anfeng and Yang, Baosong and Zhang, Beichen and Hui, Binyuan and Zheng, Bo and Yu, Bowen and Gao, Chang and Huang, Chengen and Lv, Chenxu and others},
  journal={arXiv preprint arXiv:2505.09388},
  year={2025}
}

@article{guo2025deepseek,
  title={Deepseek-r1: Incentivizing reasoning capability in llms via reinforcement learning},
  author={Guo, Daya and Yang, Dejian and Zhang, Haowei and Song, Junxiao and Zhang, Ruoyu and Xu, Runxin and Zhu, Qihao and Ma, Shirong and Wang, Peiyi and Bi, Xiao and others},
  journal={arXiv preprint arXiv:2501.12948},
  year={2025}
}

@inproceedings{radford2021learning,
  title={Learning transferable visual models from natural language supervision},
  author={Radford, Alec and Kim, Jong Wook and Hallacy, Chris and Ramesh, Aditya and Goh, Gabriel and Agarwal, Sandhini and Sastry, Girish and Askell, Amanda and Mishkin, Pamela and Clark, Jack and others},
  booktitle={International conference on machine learning},
  pages={8748--8763},
  year={2021},
  organization={PmLR}
}

@inproceedings{zhuminigpt,
  title={MiniGPT-4: Enhancing Vision-Language Understanding with Advanced Large Language Models},
  author={Zhu, Deyao and Chen, Jun and Shen, Xiaoqian and Li, Xiang and Elhoseiny, Mohamed},
  booktitle={The Twelfth International Conference on Learning Representations},
  year = {2024}
}

@article{zhang2024llava,
  title={Llava-video: Video instruction tuning with synthetic data},
  author={Zhang, Yuanhan and Wu, Jinming and Li, Wei and Li, Bo and Ma, Zejun and Liu, Ziwei and Li, Chunyuan},
  journal={arXiv preprint arXiv:2410.02713},
  year={2024}
}

@article{meinardus2024chrono,
  title={Chrono: A simple blueprint for representing time in mllms},
  author={Meinardus, Boris and Rodriguez, Hector and Batra, Anil and Rohrbach, Anna and Rohrbach, Marcus},
  journal={arXiv preprint arXiv:2406.18113},
  year={2024}
}

@article{jung2026fastav,
  title={FastAV: Efficient Token Pruning for Audio-Visual Large Language Model Inference},
  author={Jung, Chaeyoung and Jang, Youngjoon and Lee, Seungwoo and Chung, Joon Son},
  journal={arXiv preprint arXiv:2601.13143},
  year={2026}
}

@article{bai2025qwen3,
  title={Qwen3-vl technical report},
  author={Bai, Shuai and Cai, Yuxuan and Chen, Ruizhe and Chen, Keqin and Chen, Xionghui and Cheng, Zesen and Deng, Lianghao and Ding, Wei and Gao, Chang and Ge, Chunjiang and others},
  journal={arXiv preprint arXiv:2511.21631},
  year={2025}
}

@inproceedings{liu2024improved,
  title={Improved baselines with visual instruction tuning},
  author={Liu, Haotian and Li, Chunyuan and Li, Yuheng and Lee, Yong Jae},
  booktitle={Proceedings of the IEEE/CVF conference on computer vision and pattern recognition},
  pages={26296--26306},
  year={2024}
}

@article{bai2025qwen2,
  title={Qwen2. 5-vl technical report},
  author={Bai, Shuai and Chen, Keqin and Liu, Xuejing and Wang, Jialin and Ge, Wenbin and Song, Sibo and Dang, Kai and Wang, Peng and Wang, Shijie and Tang, Jun and others},
  journal={arXiv preprint arXiv:2502.13923},
  year={2025}
}

@article{liu2023visual,
  title={Visual instruction tuning},
  author={Liu, Haotian and Li, Chunyuan and Wu, Qingyang and Lee, Yong Jae},
  journal={Advances in neural information processing systems},
  volume={36},
  pages={34892--34916},
  year={2023}
}

@inproceedings{zhai2023sigmoid,
  title={Sigmoid loss for language image pre-training},
  author={Zhai, Xiaohua and Mustafa, Basil and Kolesnikov, Alexander and Beyer, Lucas},
  booktitle={Proceedings of the IEEE/CVF international conference on computer vision},
  pages={11975--11986},
  year={2023}
}

@article{alayrac2022flamingo,
  title={Flamingo: a visual language model for few-shot learning},
  author={Alayrac, Jean-Baptiste and Donahue, Jeff and Luc, Pauline and Miech, Antoine and Barr, Iain and Hasson, Yana and Lenc, Karel and Mensch, Arthur and Millican, Katherine and Reynolds, Malcolm and others},
  journal={Advances in neural information processing systems},
  volume={35},
  pages={23716--23736},
  year={2022}
}

@inproceedings{li2023blip,
  title={Blip-2: Bootstrapping language-image pre-training with frozen image encoders and large language models},
  author={Li, Junnan and Li, Dongxu and Savarese, Silvio and Hoi, Steven},
  booktitle={International conference on machine learning},
  pages={19730--19742},
  year={2023},
  organization={PMLR}
}

@inproceedings{zhang2023video,
  title={Video-LLaMA: An Instruction-tuned Audio-Visual Language Model for Video Understanding},
  author={Zhang, Hang and Li, Xin and Bing, Lidong},
  booktitle={Proceedings of the 2023 Conference on Empirical Methods in Natural Language Processing: System Demonstrations},
  pages={543--553},
  year={2023}
}

@inproceedings{goyal2017making,
  title={Making the v in vqa matter: Elevating the role of image understanding in visual question answering},
  author={Goyal, Yash and Khot, Tejas and Summers-Stay, Douglas and Batra, Dhruv and Parikh, Devi},
  booktitle={Proceedings of the IEEE conference on computer vision and pattern recognition},
  pages={6904--6913},
  year={2017}
}

@article{chen2015microsoft,
  title={Microsoft coco captions: Data collection and evaluation server},
  author={Chen, Xinlei and Fang, Hao and Lin, Tsung-Yi and Vedantam, Ramakrishna and Gupta, Saurabh and Doll{\'a}r, Piotr and Zitnick, C Lawrence},
  journal={arXiv preprint arXiv:1504.00325},
  year={2015}
}

@inproceedings{maaz2024video,
  title={Video-chatgpt: Towards detailed video understanding via large vision and language models},
  author={Maaz, Muhammad and Rasheed, Hanoona and Khan, Salman and Khan, Fahad},
  booktitle={Proceedings of the 62nd Annual Meeting of the Association for Computational Linguistics (Volume 1: Long Papers)},
  pages={12585--12602},
  year={2024}
}

@inproceedings{lin2024video,
  title={Video-llava: Learning united visual representation by alignment before projection},
  author={Lin, Bin and Ye, Yang and Zhu, Bin and Cui, Jiaxi and Ning, Munan and Jin, Peng and Yuan, Li},
  booktitle={Proceedings of the 2024 Conference on Empirical Methods in Natural Language Processing},
  pages={5971--5984},
  year={2024}
}

@inproceedings{li2024llama,
  title={Llama-vid: An image is worth 2 tokens in large language models},
  author={Li, Yanwei and Wang, Chengyao and Jia, Jiaya},
  booktitle={European Conference on Computer Vision},
  pages={323--340},
  year={2024},
  organization={Springer}
}

@inproceedings{xiao2021next,
  title={Next-qa: Next phase of question-answering to explaining temporal actions},
  author={Xiao, Junbin and Shang, Xindi and Yao, Angela and Chua, Tat-Seng},
  booktitle={Proceedings of the IEEE/CVF conference on computer vision and pattern recognition},
  pages={9777--9786},
  year={2021}
}

@inproceedings{li2024mvbench,
  title={Mvbench: A comprehensive multi-modal video understanding benchmark},
  author={Li, Kunchang and Wang, Yali and He, Yinan and Li, Yizhuo and Wang, Yi and Liu, Yi and Wang, Zun and Xu, Jilan and Chen, Guo and Luo, Ping and others},
  booktitle={Proceedings of the IEEE/CVF Conference on Computer Vision and Pattern Recognition},
  pages={22195--22206},
  year={2024}
}

@article{xu2025qwen2,
  title={Qwen2. 5-omni technical report},
  author={Xu, Jin and Guo, Zhifang and He, Jinzheng and Hu, Hangrui and He, Ting and Bai, Shuai and Chen, Keqin and Wang, Jialin and Fan, Yang and Dang, Kai and others},
  journal={arXiv preprint arXiv:2503.20215},
  year={2025}
}

@article{zhang2024long,
  title={Long context transfer from language to vision},
  author={Zhang, Peiyuan and Zhang, Kaichen and Li, Bo and Zeng, Guangtao and Yang, Jingkang and Zhang, Yuanhan and Wang, Ziyue and Tan, Haoran and Li, Chunyuan and Liu, Ziwei},
  journal={arXiv preprint arXiv:2406.16852},
  year={2024}
}

@article{shen2024longvu,
  title={Longvu: Spatiotemporal adaptive compression for long video-language understanding},
  author={Shen, Xiaoqian and Xiong, Yunyang and Zhao, Changsheng and Wu, Lemeng and Chen, Jun and Zhu, Chenchen and Liu, Zechun and Xiao, Fanyi and Varadarajan, Balakrishnan and Bordes, Florian and others},
  journal={arXiv preprint arXiv:2410.17434},
  year={2024}
}

@inproceedings{chen2024image,
  title={An image is worth 1/2 tokens after layer 2: Plug-and-play inference acceleration for large vision-language models},
  author={Chen, Liang and Zhao, Haozhe and Liu, Tianyu and Bai, Shuai and Lin, Junyang and Zhou, Chang and Chang, Baobao},
  booktitle={European Conference on Computer Vision},
  pages={19--35},
  year={2024},
  organization={Springer}
}

@article{li2024llava,
  title={Llava-onevision: Easy visual task transfer},
  author={Li, Bo and Zhang, Yuanhan and Guo, Dong and Zhang, Renrui and Li, Feng and Zhang, Hao and Zhang, Kaichen and Zhang, Peiyuan and Li, Yanwei and Liu, Ziwei and others},
  journal={arXiv preprint arXiv:2408.03326},
  year={2024}
}

@article{an2025llava,
  title={Llava-onevision-1.5: Fully open framework for democratized multimodal training},
  author={An, Xiang and Xie, Yin and Yang, Kaicheng and Zhang, Wenkang and Zhao, Xiuwei and Cheng, Zheng and Wang, Yirui and Xu, Songcen and Chen, Changrui and Wu, Chunsheng and others},
  journal={arXiv preprint arXiv:2509.23661},
  year={2025}
}

@article{chen2024expanding,
  title={Expanding performance boundaries of open-source multimodal models with model, data, and test-time scaling},
  author={Chen, Zhe and Wang, Weiyun and Cao, Yue and Liu, Yangzhou and Gao, Zhangwei and Cui, Erfei and Zhu, Jinguo and Ye, Shenglong and Tian, Hao and Liu, Zhaoyang and others},
  journal={arXiv preprint arXiv:2412.05271},
  year={2024}
}

@article{sun2025visual,
  title={Visual Intention Grounding for Egocentric Assistants},
  author={Sun, Pengzhan and Xiao, Junbin and Tse, Tze Ho Elden and Li, Yicong and Akula, Arjun and Yao, Angela},
  journal={ICCV},
  year={2025}
}

@article{wang2025internvl3,
  title={Internvl3. 5: Advancing open-source multimodal models in versatility, reasoning, and efficiency},
  author={Wang, Weiyun and Gao, Zhangwei and Gu, Lixin and Pu, Hengjun and Cui, Long and Wei, Xingguang and Liu, Zhaoyang and Jing, Linglin and Ye, Shenglong and Shao, Jie and others},
  journal={arXiv preprint arXiv:2508.18265},
  year={2025}
}

@inproceedings{zhou2025egotextvqa,
  title={Egotextvqa: Towards egocentric scene-text aware video question answering},
  author={Zhou, Sheng and Xiao, Junbin and Li, Qingyun and Li, Yicong and Yang, Xun and Guo, Dan and Wang, Meng and Chua, Tat-Seng and Yao, Angela},
  booktitle={Proceedings of the Computer Vision and Pattern Recognition Conference},
  pages={3363--3373},
  year={2025}
}

@inproceedings{huang2024audiogpt,
  title={Audiogpt: Understanding and generating speech, music, sound, and talking head},
  author={Huang, Rongjie and Li, Mingze and Yang, Dongchao and Shi, Jiatong and Chang, Xuankai and Ye, Zhenhui and Wu, Yuning and Hong, Zhiqing and Huang, Jiawei and Liu, Jinglin and others},
  booktitle={Proceedings of the AAAI Conference on Artificial Intelligence},
  volume={38},
  pages={23802--23804},
  year={2024}
}

@article{brown2020language,
  title={Language models are few-shot learners},
  author={Brown, Tom and Mann, Benjamin and Ryder, Nick and Subbiah, Melanie and Kaplan, Jared D and Dhariwal, Prafulla and Neelakantan, Arvind and Shyam, Pranav and Sastry, Girish and Askell, Amanda and others},
  journal={Advances in neural information processing systems},
  volume={33},
  pages={1877--1901},
  year={2020}
}

@article{baevski2020wav2vec,
  title={wav2vec 2.0: A framework for self-supervised learning of speech representations},
  author={Baevski, Alexei and Zhou, Yuhao and Mohamed, Abdelrahman and Auli, Michael},
  journal={Advances in neural information processing systems},
  volume={33},
  pages={12449--12460},
  year={2020}
}

@inproceedings{guzhov2022audioclip,
  title={Audioclip: Extending clip to image, text and audio},
  author={Guzhov, Andrey and Raue, Federico and Hees, J{\"o}rn and Dengel, Andreas},
  booktitle={ICASSP 2022-2022 IEEE International Conference on Acoustics, Speech and Signal Processing (ICASSP)},
  pages={976--980},
  year={2022},
  organization={IEEE}
}

@article{hsu2021hubert,
  title={Hubert: Self-supervised speech representation learning by masked prediction of hidden units},
  author={Hsu, Wei-Ning and Bolte, Benjamin and Tsai, Yao-Hung Hubert and Lakhotia, Kushal and Salakhutdinov, Ruslan and Mohamed, Abdelrahman},
  journal={IEEE/ACM transactions on audio, speech, and language processing},
  volume={29},
  pages={3451--3460},
  year={2021},
  publisher={IEEE}
}

@inproceedings{ao2022speecht5,
  title={Speecht5: Unified-modal encoder-decoder pre-training for spoken language processing},
  author={Ao, Junyi and Wang, Rui and Zhou, Long and Wang, Chengyi and Ren, Shuo and Wu, Yu and Liu, Shujie and Ko, Tom and Li, Qing and Zhang, Yu and others},
  booktitle={Proceedings of the 60th annual meeting of the association for computational linguistics (volume 1: Long papers)},
  pages={5723--5738},
  year={2022}
}

@inproceedings{radford2023robust,
  title={Robust speech recognition via large-scale weak supervision},
  author={Radford, Alec and Kim, Jong Wook and Xu, Tao and Brockman, Greg and McLeavey, Christine and Sutskever, Ilya},
  booktitle={International conference on machine learning},
  pages={28492--28518},
  year={2023},
  organization={PMLR}
}

@article{deshmukh2023pengi,
  title={Pengi: An audio language model for audio tasks},
  author={Deshmukh, Soham and Elizalde, Benjamin and Singh, Rita and Wang, Huaming},
  journal={Advances in Neural Information Processing Systems},
  volume={36},
  pages={18090--18108},
  year={2023}
}

@article{ghosh2024gama,
  title={Gama: A large audio-language model with advanced audio understanding and complex reasoning abilities},
  author={Ghosh, Sreyan and Kumar, Sonal and Seth, Ashish and Evuru, Chandra Kiran Reddy and Tyagi, Utkarsh and Sakshi, S and Nieto, Oriol and Duraiswami, Ramani and Manocha, Dinesh},
  journal={arXiv preprint arXiv:2406.11768},
  year={2024}
}

@article{shu2023llasm,
  title={Llasm: Large language and speech model},
  author={Shu, Yu and Dong, Siwei and Chen, Guangyao and Huang, Wenhao and Zhang, Ruihua and Shi, Daochen and Xiang, Qiqi and Shi, Yemin},
  journal={arXiv preprint arXiv:2308.15930},
  year={2023}
}

@article{cheng2024videollama,
  title={Videollama 2: Advancing spatial-temporal modeling and audio understanding in video-llms},
  author={Cheng, Zesen and Leng, Sicong and Zhang, Hang and Xin, Yifei and Li, Xin and Chen, Guanzheng and Zhu, Yongxin and Zhang, Wenqi and Luo, Ziyang and Zhao, Deli and others},
  journal={arXiv preprint arXiv:2406.07476},
  year={2024}
}

@inproceedings{wu2024next,
  title={Next-GPT: Any-to-any Multimodal LLM},
  author={Wu, Shengqiong and Fei, Hao and Qu, Leigang and Ji, Wei and Chua, Tat-Seng},
  booktitle={Forty-first International Conference on Machine Learning},
  year={2024}
}

@inproceedings{zitkovich2023rt,
  title={Rt-2: Vision-language-action models transfer web knowledge to robotic control},
  author={Zitkovich, Brianna and Yu, Tianhe and Xu, Sichun and Xu, Peng and Xiao, Ted and Xia, Fei and Wu, Jialin and Wohlhart, Paul and Welker, Stefan and Wahid, Ayzaan and others},
  booktitle={Conference on Robot Learning},
  pages={2165--2183},
  year={2023},
  organization={PMLR}
}

@inproceedings{RT-Trajectory,
  title={RT-Trajectory: Robotic Task Generalization via Hindsight Trajectory Sketches},
  author={Gu, Jiayuan and Kirmani, Sean and Wohlhart, Paul and Lu, Yao and Arenas, Montserrat Gonzalez and Rao, Kanishka and Yu, Wenhao and Fu, Chuyuan and Gopalakrishnan, Keerthana and Xu, Zhuo and others},
  booktitle={The Twelfth International Conference on Learning Representations},
  year={2023}
}

@inproceedings{belkhale2024rt,
  title={RT-H: Action Hierarchies using Language},
  author={Belkhale, Suneel and Ding, Tianli and Xiao, Ted and Sermanet, Pierre and Vuong, Quan and Tompson, Jonathan and Chebotar, Yevgen and Dwibedi, Debidatta and Sadigh, Dorsa},
  booktitle={Robotics: Science and Systems},
  year={2024}
}

@article{oquab2024dinov2,
  title={DINOv2: Learning Robust Visual Features without Supervision},
  author={Oquab, Maxime and Darcet, Timoth{\'e}e and Moutakanni, Th{\'e}o and Vo, Huy and Szafraniec, Marc and Khalidov, Vasil and Fernandez, Pierre and Haziza, Daniel and Massa, Francisco and El-Nouby, Alaaeldin and others},
  journal={Transactions on Machine Learning Research Journal},
  pages={1--31},
  year={2024}
}

@inproceedings{kimopenvla,
  title={OpenVLA: An Open-Source Vision-Language-Action Model},
  author={Kim, Moo Jin and Pertsch, Karl and Karamcheti, Siddharth and Xiao, Ted and Balakrishna, Ashwin and Nair, Suraj and Rafailov, Rafael and Foster, Ethan P and Sanketi, Pannag R and Vuong, Quan and others},
  booktitle={8th Annual Conference on Robot Learning},
  year={2024}
}

@article{black2024pi_0,
  title={$\pi_0$: A Vision-Language-Action Flow Model for General Robot Control},
  author={Black, Kevin and Brown, Noah and Driess, Danny and Esmail, Adnan and Equi, Michael and Finn, Chelsea and Fusai, Niccolo and Groom, Lachy and Hausman, Karol and Ichter, Brian and others},
  journal={arXiv preprint arXiv:2410.24164},
  year={2024}
}

@article{team2024octo,
  title={Octo: An open-source generalist robot policy},
  author={Team, Octo Model and Ghosh, Dibya and Walke, Homer and Pertsch, Karl and Black, Kevin and Mees, Oier and Dasari, Sudeep and Hejna, Joey and Kreiman, Tobias and Xu, Charles and others},
  journal={arXiv preprint arXiv:2405.12213},
  year={2024}
}

@article{wen2025tinyvla,
  title={Tinyvla: Towards fast, data-efficient vision-language-action models for robotic manipulation},
  author={Wen, Junjie and Zhu, Yichen and Li, Jinming and Zhu, Minjie and Tang, Zhibin and Wu, Kun and Xu, Zhiyuan and Liu, Ning and Cheng, Ran and Shen, Chaomin and others},
  journal={IEEE Robotics and Automation Letters},
  year={2025},
  publisher={IEEE}
}

@article{shukor2025smolvla,
  title={Smolvla: A vision-language-action model for affordable and efficient robotics},
  author={Shukor, Mustafa and Aubakirova, Dana and Capuano, Francesco and Kooijmans, Pepijn and Palma, Steven and Zouitine, Adil and Aractingi, Michel and Pascal, Caroline and Russi, Martino and Marafioti, Andres and others},
  journal={arXiv preprint arXiv:2506.01844},
  year={2025}
}

@article{bjorck2025gr00t,
  title={Gr00t n1: An open foundation model for generalist humanoid robots},
  author={Bjorck, Johan and Casta{\~n}eda, Fernando and Cherniadev, Nikita and Da, Xingye and Ding, Runyu and Fan, Linxi and Fang, Yu and Fox, Dieter and Hu, Fengyuan and Huang, Spencer and others},
  journal={arXiv preprint arXiv:2503.14734},
  year={2025}
}

@article{team2025gemini,
  title={Gemini robotics: Bringing ai into the physical world},
  author={Team, Gemini Robotics and Abeyruwan, Saminda and Ainslie, Joshua and Alayrac, Jean-Baptiste and Arenas, Montserrat Gonzalez and Armstrong, Travis and Balakrishna, Ashwin and Baruch, Robert and Bauza, Maria and Blokzijl, Michiel and others},
  journal={arXiv preprint arXiv:2503.20020},
  year={2025}
}

@article{zheng2024tracevla,
  title={Tracevla: Visual trace prompting enhances spatial-temporal awareness for generalist robotic policies},
  author={Zheng, Ruijie and Liang, Yongyuan and Huang, Shuaiyi and Gao, Jianfeng and Daum{\'e} III, Hal and Kolobov, Andrey and Huang, Furong and Yang, Jianwei},
  journal={arXiv preprint arXiv:2412.10345},
  year={2024}
}

@article{li2024cogact,
  title={Cogact: A foundational vision-language-action model for synergizing cognition and action in robotic manipulation},
  author={Li, Qixiu and Liang, Yaobo and Wang, Zeyu and Luo, Lin and Chen, Xi and Liao, Mozheng and Wei, Fangyun and Deng, Yu and Xu, Sicheng and Zhang, Yizhong and others},
  journal={arXiv preprint arXiv:2411.19650},
  year={2024}
}

@article{li2025cogvla,
  title={Cogvla: Cognition-aligned vision-language-action model via instruction-driven routing \& sparsification},
  author={Li, Wei and Zhang, Renshan and Shao, Rui and He, Jie and Nie, Liqiang},
  journal={arXiv preprint arXiv:2508.21046},
  year={2025}
}

@article{lin2025failsafe,
  title={Failsafe: Reasoning and recovery from failures in vision-language-action models},
  author={Lin, Zijun and Duan, Jiafei and Fang, Haoquan and Fox, Dieter and Krishna, Ranjay and Tan, Cheston and Wen, Bihan},
  journal={arXiv preprint arXiv:2510.01642},
  year={2025}
}

@article{cen2025rynnvla,
  title={RynnVLA-002: A Unified Vision-Language-Action and World Model},
  author={Cen, Jun and Huang, Siteng and Yuan, Yuqian and Yuan, Hangjie and Yu, Chaohui and Jiang, Yuming and Guo, Jiayan and Li, Kehan and Luo, Hao and Wang, Fan and others},
  journal={arXiv preprint arXiv:2511.17502},
  year={2025}
}

@article{deng2025graspvla,
  title={Graspvla: a grasping foundation model pre-trained on billion-scale synthetic action data},
  author={Deng, Shengliang and Yan, Mi and Wei, Songlin and Ma, Haixin and Yang, Yuxin and Chen, Jiayi and Zhang, Zhiqi and Yang, Taoyu and Zhang, Xuheng and Zhang, Wenhao and others},
  journal={arXiv preprint arXiv:2505.03233},
  year={2025}
}

@inproceedings{yu2019activitynet,
  title={Activitynet-qa: A dataset for understanding complex web videos via question answering},
  author={Yu, Zhou and Xu, Dejing and Yu, Jun and Yu, Ting and Zhao, Zhou and Zhuang, Yueting and Tao, Dacheng},
  booktitle={Proceedings of the AAAI Conference on Artificial Intelligence},
  volume={33},
  pages={9127--9134},
  year={2019}
}

@inproceedings{antol2015vqa,
  title={Vqa: Visual question answering},
  author={Antol, Stanislaw and Agrawal, Aishwarya and Lu, Jiasen and Mitchell, Margaret and Batra, Dhruv and Zitnick, C Lawrence and Parikh, Devi},
  booktitle={Proceedings of the IEEE international conference on computer vision},
  pages={2425--2433},
  year={2015}
}

@inproceedings{he2016deep,
  title={Deep residual learning for image recognition},
  author={He, Kaiming and Zhang, Xiangyu and Ren, Shaoqing and Sun, Jian},
  booktitle={Proceedings of the IEEE conference on computer vision and pattern recognition},
  pages={770--778},
  year={2016}
}

@article{dosovitskiy2020image,
  title={An image is worth 16x16 words: Transformers for image recognition at scale},
  author={Dosovitskiy, Alexey},
  journal={arXiv preprint arXiv:2010.11929},
  year={2020}
}

@article{yu2022coca,
  title={Coca: Contrastive captioners are image-text foundation models},
  author={Yu, Jiahui and Wang, Zirui and Vasudevan, Vijay and Yeung, Legg and Seyedhosseini, Mojtaba and Wu, Yonghui},
  journal={arXiv preprint arXiv:2205.01917},
  year={2022}
}

@article{chen2022pali,
  title={Pali: A jointly-scaled multilingual language-image model},
  author={Chen, Xi and Wang, Xiao and Changpinyo, Soravit and Piergiovanni, Anthony J and Padlewski, Piotr and Salz, Daniel and Goodman, Sebastian and Grycner, Adam and Mustafa, Basil and Beyer, Lucas and others},
  journal={arXiv preprint arXiv:2209.06794},
  year={2022}
}

@article{wang2021simvlm,
  title={Simvlm: Simple visual language model pretraining with weak supervision},
  author={Wang, Zirui and Yu, Jiahui and Yu, Adams Wei and Dai, Zihang and Tsvetkov, Yulia and Cao, Yuan},
  journal={arXiv preprint arXiv:2108.10904},
  year={2021}
}

@inproceedings{xu2015show,
  title={Show, attend and tell: Neural image caption generation with visual attention},
  author={Xu, Kelvin and Ba, Jimmy and Kiros, Ryan and Cho, Kyunghyun and Courville, Aaron and Salakhudinov, Ruslan and Zemel, Rich and Bengio, Yoshua},
  booktitle={International conference on machine learning},
  pages={2048--2057},
  year={2015},
  organization={PMLR}
}

@inproceedings{anderson2018bottom,
  title={Bottom-up and top-down attention for image captioning and visual question answering},
  author={Anderson, Peter and He, Xiaodong and Buehler, Chris and Teney, Damien and Johnson, Mark and Gould, Stephen and Zhang, Lei},
  booktitle={Proceedings of the IEEE conference on computer vision and pattern recognition},
  pages={6077--6086},
  year={2018}
}

@inproceedings{li2022blip,
  title={Blip: Bootstrapping language-image pre-training for unified vision-language understanding and generation},
  author={Li, Junnan and Li, Dongxu and Xiong, Caiming and Hoi, Steven},
  booktitle={International conference on machine learning},
  pages={12888--12900},
  year={2022},
  organization={PMLR}
}

@inproceedings{zhong2022video,
  title={Video question answering: Datasets, algorithms and challenges},
  author={Zhong, Yaoyao and Ji, Wei and Xiao, Junbin and Li, Yicong and Deng, Weihong and Chua, Tat-Seng},
  booktitle={Proceedings of the 2022 conference on empirical methods in natural language processing},
  pages={6439--6455},
  year={2022}
}

@misc{bai2023qwenvlversatilevisionlanguagemodel,
      title={Qwen-VL: A Versatile Vision-Language Model for Understanding, Localization, Text Reading, and Beyond}, 
      author={Jinze Bai and Shuai Bai and Shusheng Yang and Shijie Wang and Sinan Tan and Peng Wang and Junyang Lin and Chang Zhou and Jingren Zhou},
      year={2023},
      eprint={2308.12966},
      archivePrefix={arXiv},
      primaryClass={cs.CV},
      url={https://arxiv.org/abs/2308.12966}, 
}

@inproceedings{vedantam2015cider,
  title={Cider: Consensus-based image description evaluation},
  author={Vedantam, Ramakrishna and Lawrence Zitnick, C and Parikh, Devi},
  booktitle={Proceedings of the IEEE conference on computer vision and pattern recognition},
  pages={4566--4575},
  year={2015}
}

@article{bai2024hallucination,
  title={Hallucination of multimodal large language models: A survey},
  author={Bai, Zechen and Wang, Pichao and Xiao, Tianjun and He, Tong and Han, Zongbo and Zhang, Zheng and Shou, Mike Zheng},
  journal={arXiv preprint arXiv:2404.18930},
  year={2024}
}

@inproceedings{chen2024internvl,
  title={Internvl: Scaling up vision foundation models and aligning for generic visual-linguistic tasks},
  author={Chen, Zhe and Wu, Jiannan and Wang, Wenhai and Su, Weijie and Chen, Guo and Xing, Sen and Zhong, Muyan and Zhang, Qinglong and Zhu, Xizhou and Lu, Lewei and others},
  booktitle={Proceedings of the IEEE/CVF conference on computer vision and pattern recognition},
  pages={24185--24198},
  year={2024}
}

@inproceedings{mao2016generation,
  title={Generation and comprehension of unambiguous object descriptions},
  author={Mao, Junhua and Huang, Jonathan and Toshev, Alexander and Camburu, Oana and Yuille, Alan L and Murphy, Kevin},
  booktitle={Proceedings of the IEEE conference on computer vision and pattern recognition},
  pages={11--20},
  year={2016}
}

@inproceedings{yu2018mattnet,
  title={Mattnet: Modular attention network for referring expression comprehension},
  author={Yu, Licheng and Lin, Zhe and Shen, Xiaohui and Yang, Jimei and Lu, Xin and Bansal, Mohit and Berg, Tamara L},
  booktitle={Proceedings of the IEEE conference on computer vision and pattern recognition},
  pages={1307--1315},
  year={2018}
}

@inproceedings{kamath2021mdetr,
  title={Mdetr-modulated detection for end-to-end multi-modal understanding},
  author={Kamath, Aishwarya and Singh, Mannat and LeCun, Yann and Synnaeve, Gabriel and Misra, Ishan and Carion, Nicolas},
  booktitle={Proceedings of the IEEE/CVF international conference on computer vision},
  pages={1780--1790},
  year={2021}
}

@inproceedings{li2022grounded,
  title={Grounded language-image pre-training},
  author={Li, Liunian Harold and Zhang, Pengchuan and Zhang, Haotian and Yang, Jianwei and Li, Chunyuan and Zhong, Yiwu and Wang, Lijuan and Yuan, Lu and Zhang, Lei and Hwang, Jenq-Neng and others},
  booktitle={Proceedings of the IEEE/CVF conference on computer vision and pattern recognition},
  pages={10965--10975},
  year={2022}
}

@article{chen2023minigpt,
  title={Minigpt-v2: large language model as a unified interface for vision-language multi-task learning},
  author={Chen, Jun and Zhu, Deyao and Shen, Xiaoqian and Li, Xiang and Liu, Zechun and Zhang, Pengchuan and Krishnamoorthi, Raghuraman and Chandra, Vikas and Xiong, Yunyang and Elhoseiny, Mohamed},
  journal={arXiv preprint arXiv:2310.09478},
  year={2023}
}

@inproceedings{liu2024grounding,
  title={Grounding dino: Marrying dino with grounded pre-training for open-set object detection},
  author={Liu, Shilong and Zeng, Zhaoyang and Ren, Tianhe and Li, Feng and Zhang, Hao and Yang, Jie and Jiang, Qing and Li, Chunyuan and Yang, Jianwei and Su, Hang and others},
  booktitle={European conference on computer vision},
  pages={38--55},
  year={2024},
  organization={Springer}
}

@article{wang2024cogvlm,
  title={Cogvlm: Visual expert for pretrained language models},
  author={Wang, Weihan and Lv, Qingsong and Yu, Wenmeng and Hong, Wenyi and Qi, Ji and Wang, Yan and Ji, Junhui and Yang, Zhuoyi and Zhao, Lei and XiXuan, Song and others},
  journal={Advances in Neural Information Processing Systems},
  volume={37},
  pages={121475--121499},
  year={2024}
}

@article{lei2021detecting,
  title={Detecting moments and highlights in videos via natural language queries},
  author={Lei, Jie and Berg, Tamara L and Bansal, Mohit},
  journal={Advances in Neural Information Processing Systems},
  volume={34},
  pages={11846--11858},
  year={2021}
}

@inproceedings{yang2023vid2seq,
  title={Vid2seq: Large-scale pretraining of a visual language model for dense video captioning},
  author={Yang, Antoine and Nagrani, Arsha and Seo, Paul Hongsuck and Miech, Antoine and Pont-Tuset, Jordi and Laptev, Ivan and Sivic, Josef and Schmid, Cordelia},
  booktitle={Proceedings of the IEEE/CVF conference on computer vision and pattern recognition},
  pages={10714--10726},
  year={2023}
}

@inproceedings{ren2024timechat,
  title={Timechat: A time-sensitive multimodal large language model for long video understanding},
  author={Ren, Shuhuai and Yao, Linli and Li, Shicheng and Sun, Xu and Hou, Lu},
  booktitle={Proceedings of the IEEE/CVF Conference on Computer Vision and Pattern Recognition},
  pages={14313--14323},
  year={2024}
}

@inproceedings{huang2024vtimellm,
  title={Vtimellm: Empower llm to grasp video moments},
  author={Huang, Bin and Wang, Xin and Chen, Hong and Song, Zihan and Zhu, Wenwu},
  booktitle={Proceedings of the IEEE/CVF Conference on Computer Vision and Pattern Recognition},
  pages={14271--14280},
  year={2024}
}

@article{liu2025time,
  title={Time-R1: Towards Comprehensive Temporal Reasoning in LLMs},
  author={Liu, Zijia and Han, Peixuan and Yu, Haofei and Li, Haoru and You, Jiaxuan},
  journal={arXiv preprint arXiv:2505.13508},
  year={2025}
}

@article{li2025videochat,
  title={Videochat-r1: Enhancing spatio-temporal perception via reinforcement fine-tuning},
  author={Li, Xinhao and Yan, Ziang and Meng, Desen and Dong, Lu and Zeng, Xiangyu and He, Yinan and Wang, Yali and Qiao, Yu and Wang, Yi and Wang, Limin},
  journal={arXiv preprint arXiv:2504.06958},
  year={2025}
}

@InProceedings{Jung_2025_CVPR,
    author    = {Jung, Minjoon and Xiao, Junbin and Zhang, Byoung-Tak and Yao, Angela},
    title     = {On the Consistency of Video Large Language Models in Temporal Comprehension},
    booktitle = {Proceedings of the IEEE/CVF Conference on Computer Vision and Pattern Recognition (CVPR)},
    month     = {June},
    year      = {2025},
    pages     = {13713-13722}
}

@misc{jung2025egoexoconexploringviewinvariantvideo,
      title={EgoExo-Con: Exploring View-Invariant Video Temporal Understanding}, 
      author={Minjoon Jung and Junbin Xiao and Junghyun Kim and Byoung-Tak Zhang and Angela Yao},
      year={2025},
      eprint={2510.26113},
      archivePrefix={arXiv},
      primaryClass={cs.CV},
      url={https://arxiv.org/abs/2510.26113}, 
}

@article{yan2025videochat,
  title={Videochat-r1. 5: Visual test-time scaling to reinforce multimodal reasoning by iterative perception},
  author={Yan, Ziang and Li, Xinhao and He, Yinan and Yue, Zhengrong and Zeng, Xiangyu and Wang, Yali and Qiao, Yu and Wang, Limin and Wang, Yi},
  journal={arXiv preprint arXiv:2509.21100},
  year={2025}
}

@article{zeng2024timesuite,
  title={Timesuite: Improving mllms for long video understanding via grounded tuning},
  author={Zeng, Xiangyu and Li, Kunchang and Wang, Chenting and Li, Xinhao and Jiang, Tianxiang and Yan, Ziang and Li, Songze and Shi, Yansong and Yue, Zhengrong and Wang, Yi and others},
  journal={arXiv preprint arXiv:2410.19702},
  year={2024}
}

@article{di2025streaming,
  title={Streaming video question-answering with in-context video kv-cache retrieval},
  author={Di, Shangzhe and Yu, Zhelun and Zhang, Guanghao and Li, Haoyuan and Zhong, Tao and Cheng, Hao and Li, Bolin and He, Wanggui and Shu, Fangxun and Jiang, Hao},
  journal={ICLR},
  year={2025}
}

@article{yang2025streammem,
  title={Streammem: Query-agnostic kv cache memory for streaming video understanding},
  author={Yang, Yanlai and Zhao, Zhuokai and Shukla, Satya Narayan and Singh, Aashu and Mishra, Shlok Kumar and Zhang, Lizhu and Ren, Mengye},
  journal={arXiv preprint arXiv:2508.15717},
  year={2025}
}

@inproceedings{xiao2024can,
  title={Can i trust your answer? visually grounded video question answering},
  author={Xiao, Junbin and Yao, Angela and Li, Yicong and Chua, Tat-Seng},
  booktitle={Proceedings of the IEEE/CVF Conference on Computer Vision and Pattern Recognition},
  pages={13204--13214},
  year={2024}
}

@article{xiao2025egoblind,
  title={EgoBlind: Towards Egocentric Visual Assistance for the Blind People},
  author={Xiao, Junbin and Huang, Nanxin and Qiu, Hao and Tao, Zhulin and Yang, Xun and Hong, Richang and Wang, Meng and Yao, Angela},
  journal={arXiv preprint arXiv:2503.08221},
  year={2025}
}

@article{xiao2025videoqa,
  title={VideoQA in the Era of LLMs: An Empirical Study},
  author={Xiao, Junbin and Huang, Nanxin and Qin, Hangyu and Li, Dongyang and Li, Yicong and Zhu, Fengbin and Tao, Zhulin and Yu, Jianxing and Lin, Liang and Chua, Tat-Seng and others},
  journal={International Journal of Computer Vision},
  volume={133},
  number={7},
  pages={3970--3993},
  year={2025},
  publisher={Springer US New York}
}

@inproceedings{lei2018tvqa,
  title={Tvqa: Localized, compositional video question answering},
  author={Lei, Jie and Yu, Licheng and Bansal, Mohit and Berg, Tamara},
  booktitle={Proceedings of the 2018 conference on empirical methods in natural language processing},
  pages={1369--1379},
  year={2018}
}

@inproceedings{yang2022avqa,
  title={Avqa: A dataset for audio-visual question answering on videos},
  author={Yang, Pinci and Wang, Xin and Duan, Xuguang and Chen, Hong and Hou, Runze and Jin, Cong and Zhu, Wenwu},
  booktitle={Proceedings of the 30th ACM international conference on multimedia},
  pages={3480--3491},
  year={2022}
}

@inproceedings{ye2024cat,
  title={Cat: Enhancing multimodal large language model to answer questions in dynamic audio-visual scenarios},
  author={Ye, Qilang and Yu, Zitong and Shao, Rui and Xie, Xinyu and Torr, Philip and Cao, Xiaochun},
  booktitle={European Conference on Computer Vision},
  pages={146--164},
  year={2024},
  organization={Springer}
}

@InProceedings{Shu_2025_ICCV,
    author    = {Shu, Fangxun and Zhang, Lei and Jiang, Hao and Xie, Cihang},
    title     = {Audio-Visual LLM for Video Understanding},
    booktitle = {Proceedings of the IEEE/CVF International Conference on Computer Vision (ICCV) Workshops},
    month     = {October},
    year      = {2025},
    pages     = {4246-4255}
}

@inproceedings{kim2025question,
  title={Question-Aware Gaussian Experts for Audio-Visual Question Answering},
  author={Kim, Hongyeob and Jung, Inyoung and Suh, Dayoon and Zhang, Youjia and Lee, Sangmin and Hong, Sungeun},
  booktitle={Proceedings of the Computer Vision and Pattern Recognition Conference},
  pages={13681--13690},
  year={2025}
}

@article{zhang2025av,
  title={AV-Master: Dual-Path Comprehensive Perception Makes Better Audio-Visual Question Answering},
  author={Zhang, Jiayu and Ye, Qilang and Ye, Shuo and Lin, Xun and Song, Zihan and Yu, Zitong},
  journal={arXiv preprint arXiv:2510.18346},
  year={2025}
}

@inproceedings{wu2025avqacl,
  title={AVQACL: A Novel Benchmark for Audio-Visual Question Answering Continual Learning},
  author={Wu, Kaixuan and Li, Xinde and Li, Xinling and Hu, Chuanfei and Wu, Guoliang},
  booktitle={Proceedings of the Computer Vision and Pattern Recognition Conference},
  pages={3252--3261},
  year={2025}
}

@inproceedings{riahi2025valor32k,
  title={Valor32k-AVQA v2. 0: Open-Ended Audio-Visual Question Answering Dataset and Benchmark},
  author={Riahi, Ines and Radman, Abduljalil and Guo, Zixin and Hedjam, Rachid and Laaksonen, Jorma},
  booktitle={Proceedings of the 33rd ACM International Conference on Multimedia},
  pages={13097--13103},
  year={2025}
}

@article{ma2025fortisavqa,
  title={Fortisavqa and maven: a benchmark dataset and debiasing framework for robust multimodal reasoning},
  author={Ma, Jie and Gao, Zhitao and Chai, Qi and Liu, Jun and Wang, Pinghui and Tao, Jing and Su, Zhou},
  journal={arXiv preprint arXiv:2504.00487},
  year={2025}
}

@inproceedings{zellers2019recognition,
  title={From recognition to cognition: Visual commonsense reasoning},
  author={Zellers, Rowan and Bisk, Yonatan and Farhadi, Ali and Choi, Yejin},
  booktitle={Proceedings of the IEEE/CVF conference on computer vision and pattern recognition},
  pages={6720--6731},
  year={2019}
}

@inproceedings{johnson2017clevr,
  title={Clevr: A diagnostic dataset for compositional language and elementary visual reasoning},
  author={Johnson, Justin and Hariharan, Bharath and Van Der Maaten, Laurens and Fei-Fei, Li and Lawrence Zitnick, C and Girshick, Ross},
  booktitle={Proceedings of the IEEE conference on computer vision and pattern recognition},
  pages={2901--2910},
  year={2017}
}

@inproceedings{niu2021counterfactual,
  title={Counterfactual vqa: A cause-effect look at language bias},
  author={Niu, Yulei and Tang, Kaihua and Zhang, Hanwang and Lu, Zhiwu and Hua, Xian-Sheng and Wen, Ji-Rong},
  booktitle={Proceedings of the IEEE/CVF conference on computer vision and pattern recognition},
  pages={12700--12710},
  year={2021}
}

@inproceedings{hudson2019gqa,
  title={Gqa: A new dataset for real-world visual reasoning and compositional question answering},
  author={Hudson, Drew A and Manning, Christopher D},
  booktitle={Proceedings of the IEEE/CVF conference on computer vision and pattern recognition},
  pages={6700--6709},
  year={2019}
}

@inproceedings{jang2017tgif,
  title={Tgif-qa: Toward spatio-temporal reasoning in visual question answering},
  author={Jang, Yunseok and Song, Yale and Yu, Youngjae and Kim, Youngjin and Kim, Gunhee},
  booktitle={Proceedings of the IEEE conference on computer vision and pattern recognition},
  pages={2758--2766},
  year={2017}
}

@inproceedings{ji2020action,
  title={Action genome: Actions as compositions of spatio-temporal scene graphs},
  author={Ji, Jingwei and Krishna, Ranjay and Fei-Fei, Li and Niebles, Juan Carlos},
  booktitle={Proceedings of the IEEE/CVF conference on computer vision and pattern recognition},
  pages={10236--10247},
  year={2020}
}

@inproceedings{shi2019explainable,
  title={Explainable and explicit visual reasoning over scene graphs},
  author={Shi, Jiaxin and Zhang, Hanwang and Li, Juanzi},
  booktitle={Proceedings of the IEEE/CVF conference on computer vision and pattern recognition},
  pages={8376--8384},
  year={2019}
}

@inproceedings{li2019relation,
  title={Relation-aware graph attention network for visual question answering},
  author={Li, Linjie and Gan, Zhe and Cheng, Yu and Liu, Jingjing},
  booktitle={Proceedings of the IEEE/CVF international conference on computer vision},
  pages={10313--10322},
  year={2019}
}

@article{hildebrandt2020scene,
  title={Scene graph reasoning for visual question answering},
  author={Hildebrandt, Marcel and Li, Hang and Koner, Rajat and Tresp, Volker and G{\"u}nnemann, Stephan},
  journal={arXiv preprint arXiv:2007.01072},
  year={2020}
}

@article{subramanian2023modular,
  title={Modular visual question answering via code generation},
  author={Subramanian, Sanjay and Narasimhan, Medhini and Khangaonkar, Kushal and Yang, Kevin and Nagrani, Arsha and Schmid, Cordelia and Zeng, Andy and Darrell, Trevor and Klein, Dan},
  journal={arXiv preprint arXiv:2306.05392},
  year={2023}
}

@inproceedings{suris2023vipergpt,
  title={Vipergpt: Visual inference via python execution for reasoning},
  author={Sur{\'\i}s, D{\'\i}dac and Menon, Sachit and Vondrick, Carl},
  booktitle={Proceedings of the IEEE/CVF international conference on computer vision},
  pages={11888--11898},
  year={2023}
}

@inproceedings{andreas2016neural,
  title={Neural module networks},
  author={Andreas, Jacob and Rohrbach, Marcus and Darrell, Trevor and Klein, Dan},
  booktitle={Proceedings of the IEEE conference on computer vision and pattern recognition},
  pages={39--48},
  year={2016}
}

@inproceedings{chen2021meta,
  title={Meta module network for compositional visual reasoning},
  author={Chen, Wenhu and Gan, Zhe and Li, Linjie and Cheng, Yu and Wang, William and Liu, Jingjing},
  booktitle={Proceedings of the IEEE/CVF Winter Conference on Applications of Computer Vision},
  pages={655--664},
  year={2021}
}

@article{yi2018neural,
  title={Neural-symbolic vqa: Disentangling reasoning from vision and language understanding},
  author={Yi, Kexin and Wu, Jiajun and Gan, Chuang and Torralba, Antonio and Kohli, Pushmeet and Tenenbaum, Josh},
  journal={Advances in neural information processing systems},
  volume={31},
  year={2018}
}

@inproceedings{girdharcater,
  title={CATER: A diagnostic dataset for Compositional Actions \& TEmporal Reasoning},
  author={Girdhar, Rohit and Ramanan, Deva},
  booktitle={International Conference on Learning Representations},
  year={2020}
}

@article{yi2019clevrer,
  title={Clevrer: Collision events for video representation and reasoning},
  author={Yi, Kexin and Gan, Chuang and Li, Yunzhu and Kohli, Pushmeet and Wu, Jiajun and Torralba, Antonio and Tenenbaum, Joshua B},
  journal={arXiv preprint arXiv:1910.01442},
  year={2019}
}

@article{qin2025chain,
  title={Chain-of-Visual-Thought: Teaching VLMs to See and Think Better with Continuous Visual Tokens},
  author={Qin, Yiming and Wei, Bomin and Ge, Jiaxin and Kallidromitis, Konstantinos and Fu, Stephanie and Darrell, Trevor and Wang, XuDong},
  journal={arXiv preprint arXiv:2511.19418},
  year={2025}
}

@inproceedings{han2025videoespresso,
  title={Videoespresso: A large-scale chain-of-thought dataset for fine-grained video reasoning via core frame selection},
  author={Han, Songhao and Huang, Wei and Shi, Hairong and Zhuo, Le and Su, Xiu and Zhang, Shifeng and Zhou, Xu and Qi, Xiaojuan and Liao, Yue and Liu, Si},
  booktitle={Proceedings of the Computer Vision and Pattern Recognition Conference},
  pages={26181--26191},
  year={2025}
}

@article{tan2025reason,
  title={Reason-RFT: Reinforcement Fine-Tuning for Visual Reasoning of Vision Language Models},
  author={Tan, Huajie and Ji, Yuheng and Hao, Xiaoshuai and Lin, Minglan and Wang, Pengwei and Wang, Zhongyuan and Zhang, Shanghang},
  journal={arXiv preprint arXiv:2503.20752},
  year={2025}
}

@article{ni2025point,
  title={Point-rft: Improving multimodal reasoning with visually grounded reinforcement finetuning},
  author={Ni, Minheng and Yang, Zhengyuan and Li, Linjie and Lin, Chung-Ching and Lin, Kevin and Zuo, Wangmeng and Wang, Lijuan},
  journal={arXiv preprint arXiv:2505.19702},
  year={2025}
}

@article{wang2024charxiv,
  title={Charxiv: Charting gaps in realistic chart understanding in multimodal llms},
  author={Wang, Zirui and Xia, Mengzhou and He, Luxi and Chen, Howard and Liu, Yitao and Zhu, Richard and Liang, Kaiqu and Wu, Xindi and Liu, Haotian and Malladi, Sadhika and others},
  journal={Advances in Neural Information Processing Systems},
  volume={37},
  pages={113569--113697},
  year={2024}
}

@article{xia2025visionary,
  title={Visionary-r1: Mitigating shortcuts in visual reasoning with reinforcement learning},
  author={Xia, Jiaer and Zang, Yuhang and Gao, Peng and Li, Sharon and Zhou, Kaiyang},
  journal={arXiv preprint arXiv:2505.14677},
  year={2025}
}

@misc{claude35,
    title={Claude 3.5 Sonnet Model Card Addendum},
    author={{Anthropic}},
    year={2024},
    url={https://www-cdn.anthropic.com/fed9cc193a14b84131812372d8d5857f8f304c52/Model_Card_Claude_3_Addendum.pdf}
}

@article{wu2025vtool,
  title={VTool-R1: VLMs Learn to Think with Images via Reinforcement Learning on Multimodal Tool Use},
  author={Wu, Mingyuan and Yang, Jingcheng and Jiang, Jize and Li, Meitang and Yan, Kaizhuo and Yu, Hanchao and Zhang, Minjia and Zhai, Chengxiang and Nahrstedt, Klara},
  journal={arXiv preprint arXiv:2505.19255},
  year={2025}
}

@article{feng2025video,
  title={Video-r1: Reinforcing video reasoning in mllms},
  author={Feng, Kaituo and Gong, Kaixiong and Li, Bohao and Guo, Zonghao and Wang, Yibing and Peng, Tianshuo and Wu, Junfei and Zhang, Xiaoying and Wang, Benyou and Yue, Xiangyu},
  journal={arXiv preprint arXiv:2503.21776},
  year={2025}
}

@article{wang2025time,
  title={Time-R1: Post-Training Large Vision Language Model for Temporal Video Grounding},
  author={Wang, Ye and Wang, Ziheng and Xu, Boshen and Du, Yang and Lin, Kejun and Xiao, Zihan and Yue, Zihao and Ju, Jianzhong and Zhang, Liang and Yang, Dingyi and others},
  journal={arXiv preprint arXiv:2503.13377},
  year={2025}
}

@article{wang2025videorft,
  title={VideoRFT: Incentivizing Video Reasoning Capability in MLLMs via Reinforced Fine-Tuning},
  author={Wang, Qi and Yu, Yanrui and Yuan, Ye and Mao, Rui and Zhou, Tianfei},
  journal={arXiv preprint arXiv:2505.12434},
  year={2025}
}

@inproceedings{voas2025temporallyRealSync,
  title={Temporally Streaming Audio-Visual Synchronization for Real-World Videos},
  author={Voas, Jordan and Tseng, Wei-Cheng and Berry, Layne and Hu, Xixi and Peng, Puyuan and Stuedemann, James and Harwath, David},
  booktitle={2025 IEEE/CVF Winter Conference on Applications of Computer Vision (WACV)},
  pages={1--9},
  year={2025}
}

@article{afouras2018lrs3,
  title={{LRS3-TED}: a large-scale dataset for visual speech recognition},
  author={Afouras, Triantafyllos and Chung, Joon Son and Zisserman, Andrew},
  journal={arXiv preprint arXiv:1809.00496},
  year={2018}
}

@inproceedings{halperin2019dynamictemporallips,
  title={Dynamic temporal alignment of speech to lips},
  author={Halperin, Tavi and Ephrat, Ariel and Peleg, Shmuel},
  booktitle={ICASSP 2019-2019 IEEE International Conference on Acoustics, Speech and Signal Processing (ICASSP)},
  pages={3980--3984},
  year={2019}
}

@inproceedings{gao2021visualvoice,
  title={Visualvoice: Audio-visual speech separation with cross-modal consistency},
  author={Gao, Ruohan and Grauman, Kristen},
  booktitle={2021 IEEE/CVF Conference on Computer Vision and Pattern Recognition (CVPR)},
  pages={15490--15500},
  year={2021}
}

@article{hershey1999audiovision,
  title={Audio vision: Using audio-visual synchrony to locate sounds},
  author={Hershey, John and Movellan, Javier},
  journal={Advances in neural information processing systems},
  volume={12},
  year={1999}
}

@article{chen2021audiovisualsynchronisation,
  title={Audio-visual synchronisation in the wild},
  author={Chen, Honglie and Xie, Weidi and Afouras, Triantafyllos and Nagrani, Arsha and Vedaldi, Andrea and Zisserman, Andrew},
  journal={arXiv preprint arXiv:2112.04432},
  year={2021}
}

@inproceedings{chen2020vggsound,
  title={Vggsound: A large-scale audio-visual dataset},
  author={Chen, Honglie and Xie, Weidi and Vedaldi, Andrea and Zisserman, Andrew},
  booktitle={ICASSP 2020-2020 IEEE International Conference on Acoustics, Speech and Signal Processing (ICASSP)},
  pages={721--725},
  year={2020}
}

@inproceedings{khosravan2019attentionAudioVisual,
  title={On Attention Modules for Audio-Visual Synchronization.},
  author={Khosravan, Naji and Ardeshir, Shervin and Puri, Rohit},
  booktitle={CVPR workshops},
  pages={25--28},
  year={2019}
}

@inproceedings{gemmeke2017audioset,
  title={Audio set: An ontology and human-labeled dataset for audio events},
  author={Gemmeke, Jort F and Ellis, Daniel PW and Freedman, Dylan and Jansen, Aren and Lawrence, Wade and Moore, R Channing and Plakal, Manoj and Ritter, Marvin},
  booktitle={2017 IEEE international conference on acoustics, speech and signal processing (ICASSP)},
  pages={776--780},
  year={2017}
}

@inproceedings{ebenezer2021detectionaudiovideosynchronization,
  title={Detection of audio-video synchronization errors via event detection},
  author={Ebenezer, Joshua P and Wu, Yongjun and Wei, Hai and Sethuraman, Sriram and Liu, Zongyi},
  booktitle={ICASSP 2021-2021 IEEE International Conference on Acoustics, Speech and Signal Processing (ICASSP)},
  pages={4345--4349},
  year={2021}
}

@inproceedings{kumar2009robustaudiovisualspeechsynchronydetection,
  title={Robust audio-visual speech synchrony detection by generalized bimodal linear prediction.},
  author={Kumar, Kshitiz and Navratil, Jiri and Marcheret, Etienne and Libal, Vit and Potamianos, Gerasimos},
  booktitle={INTERSPEECH},
  pages={2251--2254},
  year={2009}
}

@article{park2024interpretableSyncNet,
  title={Interpretable Convolutional SyncNet},
  author={Park, Sungjoon and Yun, Jaesub and Lee, Donggeon and Park, Minsik},
  journal={arXiv preprint arXiv:2409.00971},
  year={2024}
}

@article{raina2022syncnet,
  title={SyncNet: Correlating Objective for Time Delay Estimation in Audio Signals},
  author={Raina, Akshay and Arora, Vipul},
  journal={arXiv preprint arXiv:2203.14639},
  year={2022}
}

@inproceedings{chung2016outoftime,
  title={Out of time: automated lip sync in the wild},
  author={Chung, Joon Son and Zisserman, Andrew},
  booktitle={Asian conference on computer vision},
  pages={251--263},
  year={2016}
}

@inproceedings{gupta2023modeformer,
  title={Modeformer: Modality-preserving embedding for audio-video synchronization using transformers},
  author={Gupta, Akash and Tripathi, Rohun and Jang, Wondong},
  booktitle={ICASSP 2023-2023 IEEE International Conference on Acoustics, Speech and Signal Processing (ICASSP)},
  pages={1--5},
  year={2023}
}

@article{kadandale2022vocalist,
  title={Vocalist: An audio-visual synchronisation model for lips and voices},
  author={Kadandale, Venkatesh S and Montesinos, Juan F and Haro, Gloria},
  journal={arXiv preprint arXiv:2204.02090},
  year={2022}
}

@inproceedings{kim2021endtoendlipsynchronisation,
  title={End-to-end lip synchronisation based on pattern classification},
  author={Kim, You Jin and Heo, Hee Soo and Chung, Soo-Whan and Lee, Bong-Jin},
  booktitle={2021 IEEE Spoken Language Technology Workshop (SLT)},
  pages={598--605},
  year={2021}
}

@inproceedings{chung2019perfectmatch,
  title={Perfect match: Improved cross-modal embeddings for audio-visual synchronisation},
  author={Chung, Soo-Whan and Chung, Joon Son and Kang, Hong-Goo},
  booktitle={ICASSP 2019-2019 IEEE International Conference on Acoustics, Speech and Signal Processing (ICASSP)},
  pages={3965--3969},
  year={2019}
}

@inproceedings{fernandez2024divas,
  title={DiVAS: Video and Audio Synchronization with Dynamic Frame Rates},
  author={Fernandez-Labrador, Clara and Ak{\c{c}}ay, Mertcan and Abecassis, Eitan and Massich, Joan and Schroers, Christopher},
  booktitle={Proceedings of the IEEE/CVF Conference on Computer Vision and Pattern Recognition},
  pages={26846--26854},
  year={2024}
}

@article{aytar2016soundnet,
  title={Soundnet: Learning sound representations from unlabeled video},
  author={Aytar, Yusuf and Vondrick, Carl and Torralba, Antonio},
  journal={Advances in neural information processing systems},
  volume={29},
  year={2016}
}

@inproceedings{owens2016ambient,
  title={Ambient sound provides supervision for visual learning},
  author={Owens, Andrew and Wu, Jiajun and McDermott, Josh H and Freeman, William T and Torralba, Antonio},
  booktitle={European conference on computer vision},
  pages={801--816},
  year={2016}
}

@inproceedings{arandjelovic2018objects,
  title={Objects that sound},
  author={Arandjelovic, Relja and Zisserman, Andrew},
  booktitle={Proceedings of the European conference on computer vision (ECCV)},
  pages={435--451},
  year={2018}
}

@inproceedings{owens2018audio,
  title={Audio-visual scene analysis with self-supervised multisensory features},
  author={Owens, Andrew and Efros, Alexei A},
  booktitle={Proceedings of the European conference on computer vision (ECCV)},
  pages={631--648},
  year={2018}
}

@article{rouditchenko2020avlnet,
  title={Avlnet: Learning audio-visual language representations from instructional videos},
  author={Rouditchenko, Andrew and Boggust, Angie and Harwath, David and Chen, Brian and Joshi, Dhiraj and Thomas, Samuel and Audhkhasi, Kartik and Kuehne, Hilde and Panda, Rameswar and Feris, Rogerio and others},
  journal={arXiv preprint arXiv:2006.09199},
  year={2020}
}

@inproceedings{sun2023learning,
  title={Learning audio-visual source localization via false negative aware contrastive learning},
  author={Sun, Weixuan and Zhang, Jiayi and Wang, Jianyuan and Liu, Zheyuan and Zhong, Yiran and Feng, Tianpeng and Guo, Yandong and Zhang, Yanhao and Barnes, Nick},
  booktitle={Proceedings of the IEEE/CVF conference on computer vision and pattern recognition},
  pages={6420--6429},
  year={2023}
}

@inproceedings{morgado2021robust,
  title={Robust audio-visual instance discrimination},
  author={Morgado, Pedro and Misra, Ishan and Vasconcelos, Nuno},
  booktitle={Proceedings of the IEEE/CVF Conference on Computer Vision and Pattern Recognition},
  pages={12934--12945},
  year={2021}
}

@inproceedings{recasens2021broaden,
  title={Broaden your views for self-supervised video learning},
  author={Recasens, Adria and Luc, Pauline and Alayrac, Jean-Baptiste and Wang, Luyu and Strub, Florian and Tallec, Corentin and Malinowski, Mateusz and P{\u{a}}tr{\u{a}}ucean, Viorica and Altch{\'e}, Florent and Valko, Michal and others},
  booktitle={Proceedings of the IEEE/CVF international conference on computer vision},
  pages={1255--1265},
  year={2021}
}

@article{wang2021multimodalaudio,
  title={Multimodal self-supervised learning of general audio representations},
  author={Wang, Luyu and Luc, Pauline and Recasens, Adria and Alayrac, Jean-Baptiste and Oord, Aaron van den},
  journal={arXiv preprint arXiv:2104.12807},
  year={2021}
}

@article{zeng2021contrastivegloballocal,
  title={Contrastive learning of global and local video representations},
  author={Zeng, Zhaoyang and McDuff, Daniel and Song, Yale and others},
  journal={Advances in Neural Information Processing Systems},
  volume={34},
  pages={7025--7040},
  year={2021}
}

@article{gong2022CAVMAE,
  title={Contrastive audio-visual masked autoencoder},
  author={Gong, Yuan and Rouditchenko, Andrew and Liu, Alexander H and Harwath, David and Karlinsky, Leonid and Kuehne, Hilde and Glass, James},
  journal={arXiv preprint arXiv:2210.07839},
  year={2022}
}

@inproceedings{georgescu2023AVMAE,
  title={Audiovisual masked autoencoders},
  author={Georgescu, Mariana-Iuliana and Fonseca, Eduardo and Ionescu, Radu Tudor and Lucic, Mario and Schmid, Cordelia and Arnab, Anurag},
  booktitle={Proceedings of the IEEE/CVF International Conference on Computer Vision},
  pages={16144--16154},
  year={2023}
}

@inproceedings{yi2023pcavmae,
    author = {Li, Yi and Plamen P. Angelov},
    title = {Explainable audio-visual representation learning via prototypical contrastive masked autoencoder},
    booktitle = {NeurIPS 2024 Workshop: Self-Supervised Learning-Theory and Practice},
    year = {2024}
}

@inproceedings{guo2024crossmae,
  title={Crossmae: Cross-modality masked autoencoders for region-aware audio-visual pre-training},
  author={Guo, Yuxin and Sun, Siyang and Ma, Shuailei and Zheng, Kecheng and Bao, Xiaoyi and Ma, Shijie and Zou, Wei and Zheng, Yun},
  booktitle={Proceedings of the IEEE/CVF Conference on Computer Vision and Pattern Recognition},
  pages={26721--26731},
  year={2024}
}

@inproceedings{lin2024avsiam,
  author       = {Yan{-}Bo Lin and
                  Gedas Bertasius},
  title        = {Siamese Vision Transformers are Scalable Audio-Visual Learners},
  booktitle    = {Computer Vision - {ECCV} 2024 - 18th European Conference, Milan, Italy,
                  September 29-October 4, 2024, Proceedings, Part {XIV}},
  series       = {Lecture Notes in Computer Science},
  volume       = {15072},
  pages        = {303--321},
  year         = {2024}
}

@inproceedings{SuLS24VAB,
  author       = {Kun Su and
                  Xiulong Liu and
                  Eli Shlizerman},
  title        = {From Vision to Audio and Beyond: {A} Unified Model for Audio-Visual
                  Representation and Generation},
  booktitle    = {Forty-first International Conference on Machine Learning, {ICML} 2024,
                  Vienna, Austria, July 21-27, 2024},
  year         = {2024}
}

@inproceedings{araujo2025cavmaesync,
  author       = {Edson Araujo and
                  Andrew Rouditchenko and
                  Yuan Gong and
                  Saurabhchand Bhati and
                  Samuel Thomas and
                  Brian Kingsbury and
                  Leonid Karlinsky and
                  Rog{\'{e}}rio Feris and
                  James R. Glass and
                  Hilde Kuehne},
  title        = {{CAV-MAE} Sync: Improving Contrastive Audio-Visual Mask Autoencoders
                  via Fine-Grained Alignment},
  booktitle    = {{IEEE/CVF} Conference on Computer Vision and Pattern Recognition,
                  {CVPR} 2025, Nashville, TN, USA, June 11-15, 2025},
  pages        = {18794--18803},
  year         = {2025}
}

@inproceedings{zhu2023languagebind,
  author       = {Bin Zhu and
                  Bin Lin and
                  Munan Ning and
                  Yang Yan and
                  Jiaxi Cui and
                  Hongfa Wang and
                  Yatian Pang and
                  Wenhao Jiang and
                  Junwu Zhang and
                  Zongwei Li and
                  Caiwan Zhang and
                  Zhifeng Li and
                  Wei Liu and
                  Li Yuan},
  title        = {LanguageBind: Extending Video-Language Pretraining to N-modality by
                  Language-based Semantic Alignment},
  booktitle    = {The Twelfth International Conference on Learning Representations,
                  {ICLR} 2024, Vienna, Austria, May 7-11, 2024},
  year         = {2024}
}

@inproceedings{mark2024denseav,
  author       = {Mark Hamilton and
                  Andrew Zisserman and
                  John R. Hershey and
                  William T. Freeman},
  title        = {Separating the "Chirp" from the "Chat": Self-supervised Visual Grounding
                  of Sound and Language},
  booktitle    = {{IEEE/CVF} Conference on Computer Vision and Pattern Recognition,
                  {CVPR} 2024, Seattle, WA, USA, June 16-22, 2024},
  pages        = {13117--13127},
  year         = {2024}
}

@inproceedings{CaronTMJMBJ21dino,
  author       = {Mathilde Caron and
                  Hugo Touvron and
                  Ishan Misra and
                  Herv{\'{e}} J{\'{e}}gou and
                  Julien Mairal and
                  Piotr Bojanowski and
                  Armand Joulin},
  title        = {Emerging Properties in Self-Supervised Vision Transformers},
  booktitle    = {2021 {IEEE/CVF} International Conference on Computer Vision, {ICCV}
                  2021, Montreal, QC, Canada, October 10-17, 2021},
  pages        = {9630--9640},
  year         = {2021}
}

@inproceedings{xu2016msrvtt,
  author       = {Jun Xu and
                  Tao Mei and
                  Ting Yao and
                  Yong Rui},
  title        = {{MSR-VTT:} {A} Large Video Description Dataset for Bridging Video
                  and Language},
  booktitle    = {2016 {IEEE} Conference on Computer Vision and Pattern Recognition,
                  {CVPR} 2016, Las Vegas, NV, USA, June 27-30, 2016},
  pages        = {5288--5296},
  year         = {2016}
}

@inproceedings{zhou2018YouCook2,
  author       = {Luowei Zhou and
                  Chenliang Xu and
                  Jason J. Corso},
  title        = {Towards Automatic Learning of Procedures From Web Instructional Videos},
  booktitle    = {Proceedings of the Thirty-Second {AAAI} Conference on Artificial Intelligence,
                  (AAAI-18), the 30th innovative Applications of Artificial Intelligence
                  (IAAI-18), and the 8th {AAAI} Symposium on Educational Advances in
                  Artificial Intelligence (EAAI-18), New Orleans, Louisiana, USA, February
                  2-7, 2018},
  pages        = {7590--7598},
  year         = {2018}
}

@inproceedings{miech2019howto100m,
  author       = {Antoine Miech and
                  Dimitri Zhukov and
                  Jean{-}Baptiste Alayrac and
                  Makarand Tapaswi and
                  Ivan Laptev and
                  Josef Sivic},
  title        = {HowTo100M: Learning a Text-Video Embedding by Watching Hundred Million
                  Narrated Video Clips},
  booktitle    = {2019 {IEEE/CVF} International Conference on Computer Vision, {ICCV}
                  2019, Seoul, Korea (South), October 27 - November 2, 2019},
  pages        = {2630--2640},
  year         = {2019}
}

@inproceedings{hayes2022mugen,
  author       = {Thomas Hayes and
                  Songyang Zhang and
                  Xi Yin and
                  Guan Pang and
                  Sasha Sheng and
                  Harry Yang and
                  Songwei Ge and
                  Qiyuan Hu and
                  Devi Parikh},
  title        = {{MUGEN:} {A} Playground for Video-Audio-Text Multimodal Understanding
                  and GENeration},
  booktitle    = {Computer Vision - {ECCV} 2022 - 17th European Conference, Tel Aviv,
                  Israel, October 23-27, 2022, Proceedings, Part {VIII}},
  pages        = {431--449},
  year         = {2022}
}

@inproceedings{IshikawaNMSKA25,
  author       = {Yuchi Ishikawa and
                  Shota Nakada and
                  Hokuto Munakata and
                  Kazuhiro Saito and
                  Tatsuya Komatsu and
                  Yoshimitsu Aoki},
  title        = {Language-Guided Contrastive Audio-Visual Masked Autoencoder with Automatically
                  Generated Audio-Visual-Text Triplets from Videos},
  booktitle    = {26th Annual Conference of the International Speech Communication Association,
                  Interspeech 2025, Rotterdam, The Netherlands, 17-21 August 2025},
  year         = {2025}
}

@misc{cheng2025vset10m,
    author = {Xize Cheng and Ziang Zhang and Minghui Fang and Zehan Wang and Ruofan Hu and Siqi Zheng and Rongjie Huang and Tao Jin and Zhou Zhao},
    title = {AVSET-10M: An Open Large-Scale Audio-Visual Dataset with High Correspondence},
    note = {Hugging Face dataset: https://huggingface.co/datasets/avset10m/avset10m},
    year = {2025}
}

@inproceedings{ChenJSGAIRG20SoundSpaces,
  author       = {Changan Chen and
                  Unnat Jain and
                  Carl Schissler and
                  Sebasti{\`{a}} Vicenc Amengual Gar{\'{\i}} and
                  Ziad Al{-}Halah and
                  Vamsi Krishna Ithapu and
                  Philip W. Robinson and
                  Kristen Grauman},
  title        = {SoundSpaces: Audio-Visual Navigation in 3D Environments},
  booktitle    = {Computer Vision - {ECCV} 2020 - 16th European Conference, Glasgow,
                  UK, August 23-28, 2020, Proceedings, Part {VI}},
  volume       = {12351},
  pages        = {17--36},
  year         = {2020}
}

@article{chen2022soundspaces20,
  title={Soundspaces 2.0: A simulation platform for visual-acoustic learning},
  author={Chen, Changan and Schissler, Carl and Garg, Sanchit and Kobernik, Philip and Clegg, Alexander and Calamia, Paul and Batra, Dhruv and Robinson, Philip and Grauman, Kristen},
  journal={Advances in Neural Information Processing Systems},
  volume={35},
  pages={8896--8911},
  year={2022}
}

@inproceedings{ChenAG21SAVi,
  author       = {Changan Chen and
                  Ziad Al{-}Halah and
                  Kristen Grauman},
  title        = {Semantic Audio-Visual Navigation},
  booktitle    = {{IEEE} Conference on Computer Vision and Pattern Recognition, {CVPR}
                  2021, virtual, June 19-25, 2021},
  pages        = {15516--15525},
  year         = {2021}
}

@article{gao2023sonicverse,
  title={Sonicverse: A multisensory simulation platform for embodied household agents that see and hear},
  author={Gao, Ruohan and Li, Hao and Dharan, Gokul and Wang, Zhuzhu and Li, Chengshu and Xia, Fei and Savarese, Silvio and Fei-Fei, Li and Wu, Jiajun},
  journal={arXiv preprint arXiv:2306.00923},
  year={2023}
}

@article{li2025audioDynamicModality,
  title={Audio-Guided Dynamic Modality Fusion with Stereo-Aware Attention for Audio-Visual Navigation},
  author={Li, Jia and Yu, Yinfeng and Wang, Liejun and Sun, Fuchun and Zheng, Wendong},
  journal={arXiv preprint arXiv:2509.16924},
  year={2025}
}

@inproceedings{GanZ0GT20,
  author       = {Chuang Gan and
                  Yiwei Zhang and
                  Jiajun Wu and
                  Boqing Gong and
                  Joshua B. Tenenbaum},
  title        = {Look, Listen, and Act: Towards Audio-Visual Embodied Navigation},
  booktitle    = {2020 {IEEE} International Conference on Robotics and Automation, {ICRA}
                  2020, Paris, France, May 31 - August 31, 2020},
  pages        = {9701--9707},
  year         = {2020}
}

@inproceedings{ChenMAGRG21AVWaN,
  author       = {Changan Chen and
                  Sagnik Majumder and
                  Ziad Al{-}Halah and
                  Ruohan Gao and
                  Santhosh Kumar Ramakrishnan and
                  Kristen Grauman},
  title        = {Learning to Set Waypoints for Audio-Visual Navigation},
  booktitle    = {9th International Conference on Learning Representations, {ICLR} 2021,
                  Virtual Event, Austria, May 3-7, 2021},
  year         = {2021}
}

@inproceedings{ChenGR0LOR24RAF,
  author       = {Ziyang Chen and
                  Israel D. Gebru and
                  Christian Richardt and
                  Anurag Kumar and
                  William Laney and
                  Andrew Owens and
                  Alexander Richard},
  title        = {Real Acoustic Fields: An Audio-Visual Room Acoustics Dataset and Benchmark},
  booktitle    = {{IEEE/CVF} Conference on Computer Vision and Pattern Recognition,
                  {CVPR} 2024, Seattle, WA, USA, June 16-22, 2024},
  pages        = {21886--21896},
  year         = {2024}
}

@inproceedings{ChenWLLY23,
  author       = {Jinyu Chen and
                  Wenguan Wang and
                  Si Liu and
                  Hongsheng Li and
                  Yi Yang},
  title        = {Omnidirectional Information Gathering for Knowledge Transfer-based
                  Audio-Visual Navigation},
  booktitle    = {{IEEE/CVF} International Conference on Computer Vision, {ICCV} 2023,
                  Paris, France, October 1-6, 2023},
  pages        = {10959--10969},
  year         = {2023}
}

@article{li2023transformermemory,
  title={Transformer memory for interactive visual navigation in cluttered environments},
  author={Li, Weiyuan and Hong, Ruoxin and Shen, Jiwei and Yuan, Liang and Lu, Yue},
  journal={IEEE Robotics and Automation Letters},
  volume={8},
  number={3},
  pages={1731--1738},
  year={2023}
}

@inproceedings{yang2024rila,
  title={Rila: Reflective and imaginative language agent for zero-shot semantic audio-visual navigation},
  author={Yang, Zeyuan and Liu, Jiageng and Chen, Peihao and Cherian, Anoop and Marks, Tim K and Le Roux, Jonathan and Gan, Chuang},
  booktitle={Proceedings of the IEEE/CVF Conference on Computer Vision and Pattern Recognition},
  pages={16251--16261},
  year={2024}
}

@article{yu2022soundadversarial,
  title={Sound adversarial audio-visual navigation},
  author={Yu, Yinfeng and Huang, Wenbing and Sun, Fuchun and Chen, Changan and Wang, Yikai and Liu, Xiaohong},
  journal={arXiv preprint arXiv:2202.10910},
  year={2022}
}

@inproceedings{liu2024caven,
  title={Caven: An embodied conversational agent for efficient audio-visual navigation in noisy environments},
  author={Liu, Xiulong and Paul, Sudipta and Chatterjee, Moitreya and Cherian, Anoop},
  booktitle={Proceedings of the AAAI conference on artificial intelligence},
  volume={38},
  pages={3765--3773},
  year={2024}
}

@article{paul2022avlen,
  title={Avlen: Audio-visual-language embodied navigation in 3d environments},
  author={Paul, Sudipta and Roy-Chowdhury, Amit and Cherian, Anoop},
  journal={Advances in Neural Information Processing Systems},
  volume={35},
  pages={6236--6249},
  year={2022}
}

@inproceedings{zhao2025audioAntiBacktracking,
  title={Audio-Visual Navigation with Anti-Backtracking},
  author={Zhao, Zhenghao and Tang, Hao and Yan, Yan},
  booktitle={International Conference on Pattern Recognition},
  pages={358--372},
  year={2025}
}

@inproceedings{shi2025towardsNoisyEnvironments,
  title={Towards Audio-Visual Navigation in Noisy Environments: A Large-Scale Benchmark Dataset and an Architecture Considering Multiple Sound-Sources},
  author={Shi, Zhanbo and Zhang, Lin and Li, Linfei and Shen, Ying},
  booktitle={Proceedings of the AAAI Conference on Artificial Intelligence},
  volume={39},
  pages={14673--14680},
  year={2025}
}

@article{younes2022dynamical,
  title={Dynamical Audio-Visual Navigation: Catching Unheard Moving Sound Sources in Unmapped 3D Environments},
  author={Younes, Abdelrahman},
  journal={arXiv preprint arXiv:2201.04279},
  year={2022}
}

@inproceedings{wang2025boostingChannelAttention,
  title={Boosting Audio-Visual Navigation Performance with Channel Attention in 3D Environments},
  author={Wang, Yi and Zhu, Meiling and Yu, Yinfeng},
  booktitle={Chinese Intelligent Automation Conference},
  pages={198--206},
  year={2025},
  organization={Springer}
}

@article{li2022seehearfeel,
  title={See, hear, and feel: Smart sensory fusion for robotic manipulation},
  author={Li, Hao and Zhang, Yizhi and Zhu, Junzhe and Wang, Shaoxiong and Lee, Michelle A and Xu, Huazhe and Adelson, Edward and Fei-Fei, Li and Gao, Ruohan and Wu, Jiajun},
  journal={arXiv preprint arXiv:2212.03858},
  year={2022}
}

@article{huang2025unifiedForceful,
  title={Unified Multimodal Diffusion Forcing for Forceful Manipulation},
  author={Huang, Zixuan and Hou, Huaidian and Berenson, Dmitry},
  journal={arXiv preprint arXiv:2511.04812},
  year={2025}
}

@article{zhao2025vlas,
  title={Vlas: Vision-language-action model with speech instructions for customized robot manipulation},
  author={Zhao, Wei and Ding, Pengxiang and Zhang, Min and Gong, Zhefei and Bai, Shuanghao and Zhao, Han and Wang, Donglin},
  journal={arXiv preprint arXiv:2502.13508},
  year={2025}
}

@article{wei2025audioVLA,
  title={Audio-VLA: Adding Contact Audio Perception to Vision-Language-Action Model for Robotic Manipulation},
  author={Wei, Xiangyi and Zhang, Haotian and Cao, Xinyi and Xie, Siyu and Ge, Weifeng and Li, Yang and Wang, Changbo},
  journal={arXiv preprint arXiv:2511.09958},
  year={2025}
}

@article{wang2024allinone,
  title={All robots in one: A new standard and unified dataset for versatile, general-purpose embodied agents},
  author={Wang, Zhiqiang and Zheng, Hao and Nie, Yunshuang and Xu, Wenjun and Wang, Qingwei and Ye, Hua and Li, Zhe and Zhang, Kaidong and Cheng, Xuewen and Dong, Wanxi and others},
  journal={arXiv preprint arXiv:2408.10899},
  year={2024}
}

@article{jiang2025kaiwu,
  title={Kaiwu: A Multimodal Manipulation Dataset and Framework for Robot Learning and Human-Robot Interaction},
  author={Jiang, Shuo and Li, Haonan and Ren, Ruochen and Zhou, Yanmin and Wang, Zhipeng and He, Bin},
  journal={arXiv preprint arXiv:2503.05231},
  year={2025}
}

@article{liu2024maniwav,
  title={Maniwav: Learning robot manipulation from in-the-wild audio-visual data},
  author={Liu, Zeyi and Chi, Cheng and Cousineau, Eric and Kuppuswamy, Naveen and Burchfiel, Benjamin and Song, Shuran},
  journal={arXiv preprint arXiv:2406.19464},
  year={2024}
}

@inproceedings{wang2022audiovisualgrounding,
  title={Audio-visual grounding referring expression for robotic manipulation},
  author={Wang, Yefei and Wang, Kaili and Wang, Yi and Guo, Di and Liu, Huaping and Sun, Fuchun},
  booktitle={2022 International Conference on Robotics and Automation (ICRA)},
  pages={9258--9264},
  year={2022}
}

@article{zhang2025learningAudioWorld,
  title={Learning Robot Manipulation from Audio World Models},
  author={Zhang, Fan and Gienger, Michael},
  journal={arXiv preprint arXiv:2512.08405},
  year={2025}
}

@article{yi2024visualauditory,
  title={Visual-auditory Extrinsic Contact Estimation},
  author={Yi, Xili and Lee, Jayjun and Fazeli, Nima},
  journal={arXiv preprint arXiv:2409.14608},
  year={2024}
}

@inproceedings{wang2025soundSimulation,
  title={The Sound of Simulation: Learning Multimodal Sim-to-Real Robot Policies with Generative Audio},
  author={Wang, Renhao and Geng, Haoran and Li, Tingle and Wu, Philipp and Wang, Feishi and Anumanchipalli, Gopala and Darrell, Trevor and Li, Boyi and Abbeel, Pieter and Malik, Jitendra and others},
  booktitle={9th Annual Conference on Robot Learning},
  year={2025}
}

@inproceedings{mejia2024hearingtouch,
  title={Hearing touch: Audio-visual pretraining for contact-rich manipulation},
  author={Mejia, Jared and Dean, Victoria and Hellebrekers, Tess and Gupta, Abhinav},
  booktitle={2024 IEEE International Conference on Robotics and Automation (ICRA)},
  pages={6912--6919},
  year={2024}
}

@article{schneider2019wav2vec,
  title={wav2vec: Unsupervised pre-training for speech recognition},
  author={Schneider, Steffen and Baevski, Alexei and Collobert, Ronan and Auli, Michael},
  journal={arXiv preprint arXiv:1904.05862},
  year={2019}
}

@article{chen2022wavlm,
  title={Wavlm: Large-scale self-supervised pre-training for full stack speech processing},
  author={Chen, Sanyuan and Wang, Chengyi and Chen, Zhengyang and Wu, Yu and Liu, Shujie and Chen, Zhuo and Li, Jinyu and Kanda, Naoyuki and Yoshioka, Takuya and Xiao, Xiong and others},
  journal={IEEE Journal of Selected Topics in Signal Processing},
  volume={16},
  number={6},
  pages={1505--1518},
  year={2022}
}

@inproceedings{chen2022htsat,
  title={Hts-at: A hierarchical token-semantic audio transformer for sound classification and detection},
  author={Chen, Ke and Du, Xingjian and Zhu, Bilei and Ma, Zejun and Berg-Kirkpatrick, Taylor and Dubnov, Shlomo},
  booktitle={ICASSP 2022-2022 IEEE International Conference on Acoustics, Speech and Signal Processing (ICASSP)},
  pages={646--650},
  year={2022}
}




\end{document}